%% file: bmvc_review.tex
\documentclass{bmvc2k}

\usepackage{times}
\usepackage{epsfig}
\usepackage{graphicx}
\usepackage{amsmath}
\usepackage{amssymb}
\usepackage{comment}

\usepackage{subfigure}
\usepackage{rotating}
\usepackage{xcolor}

\usepackage{multirow}

\title{LUCES: A Dataset for Near-Field \\ Point Light Source Photometric Stereo}

\addauthor{Roberto Mecca}{rmecca@crl.toshiba.co.uk}{1}
\addauthor{Fotios Logothetis}{flogothetis@crl.toshiba.co.uk}{1}
\addauthor{Ignas Budvytis}{ib255@cam.ac.uk}{2}
\addauthor{Roberto Cipolla}{rc10001@cam.ac.uk}{2}
\addinstitution{
Cambridge Research Laboratory,\\
Toshiba Europe, Cambridge, UK\\
Cambridge, UK
}
\addinstitution{
University of Cambridge\\
 Cambridge, UK
}

\runninghead{R. Mecca et. al }{LUCES: Near-field Photometric Stereo Dataset}

\def\etal{\emph{et al}\bmvaOneDot}

\begin{document}

\maketitle

\begin{abstract}

Three-dimensional reconstruction of objects from shading information is a challenging task in computer vision. As most of the approaches facing the Photometric Stereo problem use simplified far-field assumptions, real-world scenarios have essentially more complex physical effects that need to be handled for accurately reconstructing the 3D shape. An increasing number of methods have been proposed to address the problem when point light sources are assumed to be nearby the target object. 

To understand the capability of the approaches dealing with this near-field scenario, the literature till now has used synthetically rendered photometric images or minimal and very customised real-world data. In order to fill the gap in evaluating near-field photometric stereo methods, we introduce LUCES the first real-world 'dataset for near-fieLd point light soUrCe photomEtric Stereo' of 14 objects of different materials. 52 LEDs have been used to lit each object positioned 10 to 30 centimeters away from the camera. Together with the raw images, in order to evaluate the 3D reconstructions, the dataset includes both normal and depth maps for comparing different features of the retrieved 3D geometry. Furthermore, we evaluate the performance of the latest near-field Photometric Stereo algorithms on the proposed dataset to assess the state-of-the-art method with respect to actual close range effects and object materials. 

\end{abstract}

\input{sections/introduction}

\input{sections/relatedwork}
\input{sections/capture}

\input{sections/calibration}

\input{sections/Experiments}

\input{sections/conclusion}
\newpage

{\small
\bibliography{egbib}
}

\clearpage

\appendix

\section{Appendix}

This appendix provides supplementary material for the main publication. Section~\ref{sec:pssetup} provides additional details of the Photometric Stereo camera and LED setup. Section~\ref{sec:gt} provides a discussion on the inherent errors introduced in shape estimation from normals. Section~\ref{sec:rec} provides a complete qualitative comparison of all 5 methods evaluated in the main publication.

\input{sections/leds.tex}

\input{sections/gt.tex}
\input{sections/reconstructions.tex}
\end{document}

%% file: sections/introduction.tex
\section{Introduction}
\label{sec:intro}

Since the introduction of the Photometric Stereo problem (PS) by Woodham in the early '80s \cite{Woodham1980}, a wide variety of approaches tackled the very same problem of reconstructing 3D geometry of an object under varying illumination from the same view point. Despite the very simplified assumption in \cite{Woodham1980} to make the PS problem solvable as an (over-determined) linear system, similar simplifications are often still considered nowadays to make the problem applicable to real-world scenarios. Nonetheless, diffuse material assumption was relaxed in 
\cite{Ikeuchi81,Wolff1994relative, Shi2018}, camera perspective viewing was modelled in \cite{Tankus2005, mecca2014perspective}, and robust optimisation methods were employed by \cite{HarrisonJ12, Ikehata2012Robust} to increase robustness to outliers.  Light calibration assumption was also relaxed by \cite{QueauLD15, Papadhimitri2014}.

\begin{figure*}[t]
\centering
 \includegraphics[trim={0 0cm 0 0cm},clip,width=0.99\textwidth]{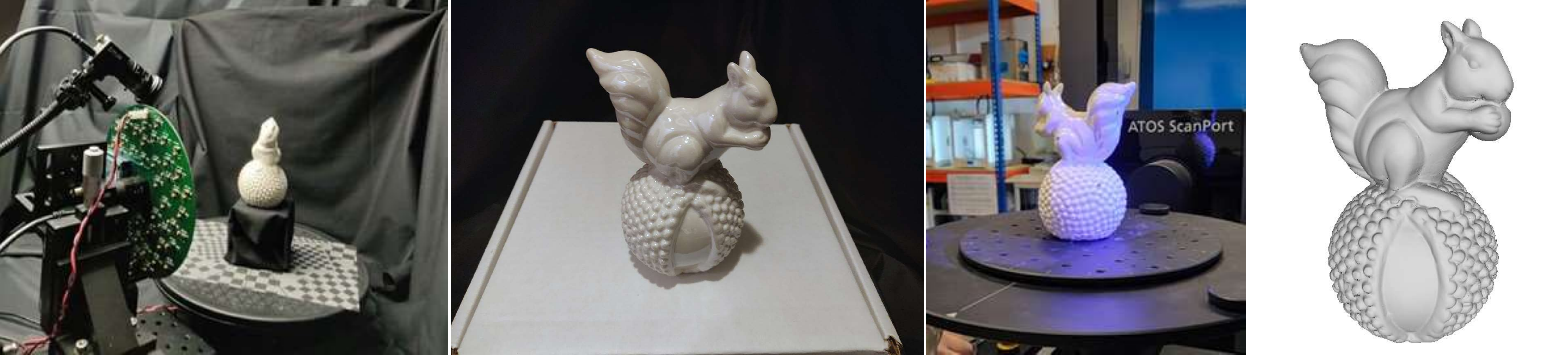}
 \caption{From left to right:(1) the stage of our Photometric Stereo setup (2) a top view of a sample object (Squirrel), (3) acquisition with the GOM scanner (4) the 3D scanned mesh.}
 \label{fig:setup}
\end{figure*}

One of the most challenging aims of more recent PS methods is realistic illumination modelling, as uniform directional lighting is hard to achieve in practice. For this purpose, several methods have proposed using point light sources instead of directional ones \cite{Iwahori1990point, Clark1992, Mecca2014near, NieS16, queau2018led, logothetis2017semi, logothetis2020cnn}. As LED illumination based technology has spread widely, point light source has become by far the more adopted alternative to directional lighting. However, point light sources require non-linear modeling of light propagation and attenuation, but they are a more realistic assumption than directional lights for near-field photometric imaging acquisitions. Note that proximity of the camera and lights to the object are very favorable in order to capture detailed geometry and minimise the ambient light interference. For example, near-field photometric stereo has been used in practice with handheld acquisition devices \cite{Higo2009} and 
in endoscope-like inspections \cite{CollinsB12}. Whereas, the far-field assumptions do not allow to combine PS with multi-view for volumetric reconstruction \cite{logothetis2019differential}.

However, despite the increased contribution from the computer vision community to tackle the near-field PS problem, the evaluation of such methods has relied on synthetic \cite{logothetis2020cnn} or very minimal real-world datasets~\cite{queau2018led,santo2020deep}. 
The lack of shared data has prevented detailed and fair comparisons across the different methods. The aim of this work is to provide a comprehensive near-field PS benchmark with a variety of objects having different materials 
in order to evaluate several algorithms and understand their strengths and weaknesses. For this purpose, ground truth normal map and depth are provided for each object.

Our contribution is as follows:
\begin{itemize}
    \item introducing the first near-field PS dataset of 14 real objects having a wide variety of materials; 
    \item evaluating most relevant algorithms for the near-field PS problem and establish the actual state-of-the-art method.
\end{itemize}

The dataset (including all images, light and camera calibration parameters and ground truth meshes) and the evaluation of the methods are available for download at: \\ \href{http://www.robertomecca.com/psdataset.html}{http://www.robertomecca.com/psdataset.html}.

%% file: sections/relatedwork.tex
\section{Related Work}
\label{sec:relatedwork}

\begin{figure*}
\includegraphics[trim={0.0cm 47cm 0 0},clip,width=\textwidth]{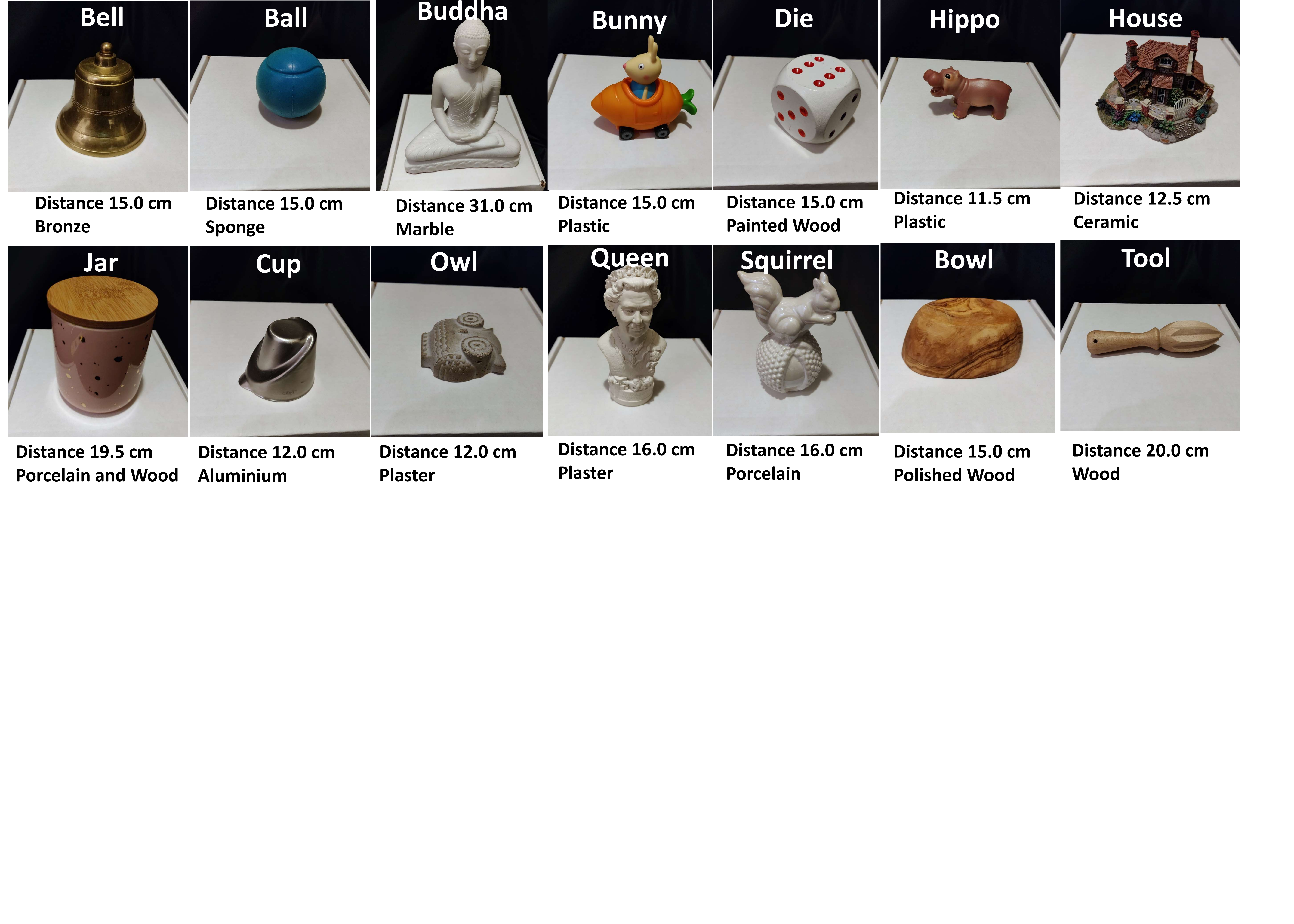}
\caption{Top view of the objects captured for this dataset. Below every object the acquisition distance between the object and the camera, and the material of the object are reported.}
    \label{tab:objects}
\end{figure*}

A number of approaches for the PS problem has been proposed since it was first introduced \cite{Woodham1980}. We refer to some fairly recent surveys \cite{AckermannG15, HerbortW11} to cover the initial evolution of the PS methodologies. Here we discuss more contemporary algorithms as this work focuses on evaluating their performances on the proposed dataset.

\subsection{PS datasets} Across the years, a number of custom real-world PS datasets have been created to suit the purposes of the proposed approaches. Alldrin \etal \cite{Alldrin2008} proposed a dataset consisting of 3 objects lit by roughly a hundred distant light directions. The light calibration in terms of positioning and intensity has been performed by using respectively a mirror sphere and a diffuse sphere. Xiong \etal \cite{XiongCBGJZ15} have proposed a dataset of 7 objects using 20 directional lights calibrated with two chrome spheres. As the approach was mostly modeling PS images with Lamberitan irradiance equations, the material of the objects was quite diffuse. A limited number of PS data has been released by Qu\'eau \etal to prove the working principle of an edge preserving method \cite{queau2015edge} and a multi-spectral PS approach \cite{Queau2016colored}.

Although initially designed for evaluating multi-view approaches, the datasets released by Aan{\ae}s \etal \cite{AanaesDP12, AanaesJVTD16} are useful for evaluating PS approaches as they also contain images under varying illumination.

As most of the methods aimed at tackling the PS problem deal with the far-field setting, recently Shi \etal \cite{Shi2018} introduced the first dataset in this category, namely DiLiGenT aimed at evaluating reconstruction methods over a wide variety of materials for 10 different objects. This work also contains a well discussed taxonomy for non-Lambertian and uncalibrated PS approaches. Their setup consists of 96 LEDs placed several meters away from the objects to approximate directional illumination and the camera
(with a 50mm lens) was placed at 1.5m from the object. Such distance between the object and the camera/lights system does not provide to this dataset the near-field light variation studied in many recent approaches. 

\paragraph{Near-field datasets:} There are very limited, proper near-field labeled data including a single object from \cite{queau2018led} and 3 simple objects from  \cite{santo2020deep}.

\subsection{Near-field PS} 

The near-field setting is intrinsically more complicated to model than the far-field one as it requires handling not only different type of BRDFs \cite{Chandraker2012, Ngan} but also anisotropic light propagation \cite{NieS16}, inconsistent light intensity among the set of LEDs \cite{queau2017semi,logothetis2017semi} and finally the uncalibrated case \cite{Papadhimitri2014}. Given this wide variety of difficulties, most of the proposed near-field PS methods have presented custom PS data.

With the aim to tackle all these issues simultaneously for the near-field PS, latest approaches have been exploiting deep learning capability training their networks with synthetically rendered data and data driven rendered data \cite{Matusik2003}. In particular, Logothetis \etal \cite{logothetis2020cnn} used a per-pixel training strategy that allows to render unlimited data without carrying any training dataset. The sim-to-real gap is then filled by augmenting the data with physical effects such as noise, ambient light, interreflections, etc. Finally, the far to near field compensation is performed by integrating the normal field to compute the depth. By doing so iterativelly, the method converges to an estimate of the 3D geometry. Santo \etal \cite{santo2020deep} have recently introduced a near-field PS method where the near-field compensation is computed after computing the far-field normals map from PSFCN \cite{Chen2018flexible}. The surface optimisation is performed through a differentiable renderer which fuses the normal predictions and the lighting model to re-project to the original images. This step limits the evaluation of the method to small images due to very high requirements of GPU RAM (around 20GB for 0.5Mpx images). Furthermore, despite the near-field setting, the camera viewing is assumed orthographic. 


%% file: sections/capture.tex
\section{Data Capture}
\label{sec:capsetup}

This section gives an overview of the data capture and calibration procedure.

\subsection{Photometric Stereo Data Capture}

\paragraph{The Photometric Stereo setup.} Our setup (see Figure \ref{fig:setup}, left) consists of the following main components:
\begin{itemize}
\setlength\itemsep{0.1mm}
    \item RGB camera FLIR BFS-U3-32S4C-C with 8mm lens
    \item 52 LED Golden Dragon OSRAM
    \item variable voltage for adjustable LED power
    \item Arduino Mega 2560
\end{itemize}

A custom printed circuit board (PCB) has been designed to host 52 bright LED controlled with by an Arduino Mega. The configuration of the LEDs was planar around the camera. 
A set of 52 images was captured per object. The camera parameters (aperture and shutter speed) and LED voltage were adjusted to achieve the best object exposure, which is very critical for specular objects. In particular, ISO sensitivity was set to zero and the exposure time has been changed depending on the shininess of the object (between 9 and 500 ms).  We also changed the power of the LEDs for particularly specular objects to avoid over saturated images. We used the maximum color-depth possible for the camera which was 12-bit. All camera prepossessing was turned off during the acquisition, including white-balance and analog gain.

Several optomechanical tools have been used for holding the camera and the PCB jointly. A manual XYZ translation stage with differential adjusters has been used to positioning the camera accurately through the printed circuit board.

In order to limit interreflections and  ambient light, the walls surrounding the setup have been covered with black, polyurethane-coated nylon fabric. 


\paragraph{Camera Intrinsics.} This is performed using 100 checkerboard images and the OpenCV calibration toolbox. Fourth degree radial distortion is estimated 
and this is used to rectify all the images. The calibration re-projection error was $0.42$px. The RAW data (before demosaicing and rectification) will also be made available.

%% file: sections/calibration.tex
\begin{figure*}[t]
\centering
 \includegraphics[height=0.14\textwidth,trim={0cm 0cm 0cm  0cm},clip]{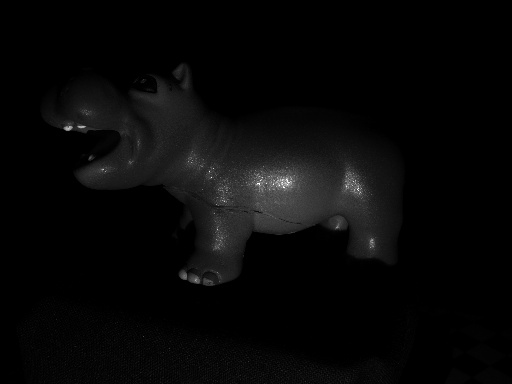}
  \includegraphics[height=0.14\textwidth,trim={0cm 0cm 0cm  0cm},clip]{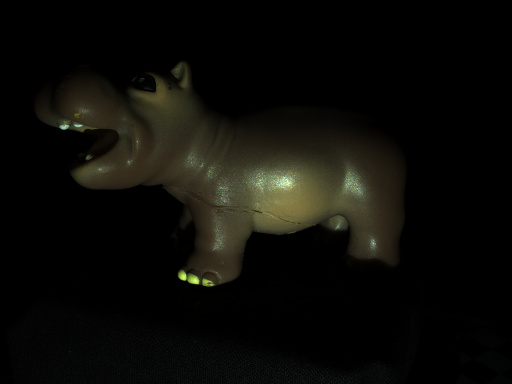}
 \includegraphics[height=0.14\textwidth,trim={1.5cm 0.9cm 1.5cm  0.9cm},clip]{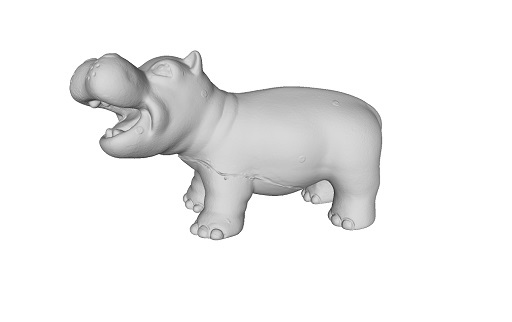}
  \includegraphics[height=0.14\textwidth,trim={0cm 0cm 0cm  0cm},clip]{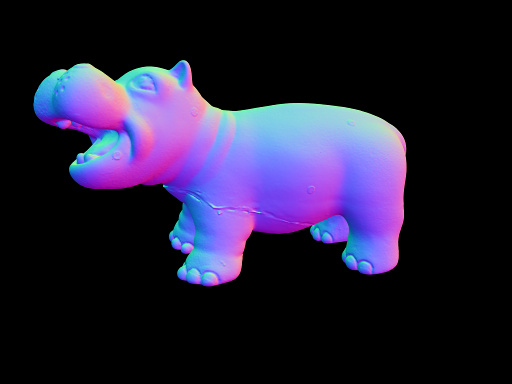}
 \includegraphics[height=0.14\textwidth,trim={0cm 0cm 0cm  0cm},clip]{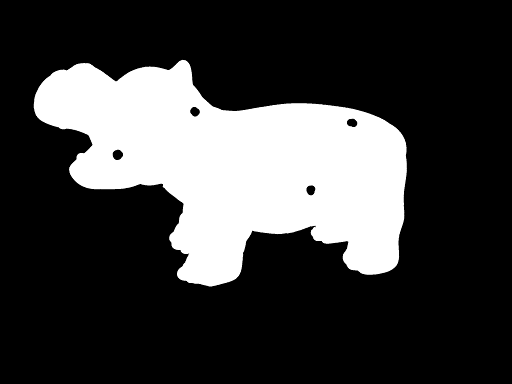}
 \caption{Demonstration of the processing steps performed per object. Firstly, compensation for radial distortion and demosaicing is performed on raw images to get RBG ones (left). Laser-scanned ground truth meshes are aligned with RGB images and ground truth normal maps are rendered (middle). Segmentation masks are generated (removing the pixels corresponding to markers)(right). 
 }
 \label{fig:steps}
\end{figure*}

\paragraph{Near Lighting Model.}%

 The lighting model is the anisotropic point light sources \cite{Mecca2014near} which is used for all SOTA methods (\cite{logothetis2017semi, queau2018led, logothetis2020cnn, santo2020deep} evaluated in Section~\ref{sec:exp}). This model assumes that a light source $m$, has a position $\mathbf{P}_m \in \mathbf{R}^3$, principal direction $\mathbf{D}_m \in \mathbb{R}^3$, RGB brightness $\mathbf{\phi}_m \in \mathbb{R}^3$ and angular dissipation factor $\mu _m \in \mathbb{R}$. Therefore, a point $\mathbf{X} \in \mathbb{R}^3$ has a lighting vector $\mathbf{L}_m (\mathbf{X}) =\mathbf{P}_m-\mathbf{X}$ and assuming $\hat{\mathbf{L}}_m=\frac{\mathbf{L}_m}{||\mathbf{L}_m||}$ as the normalised light direction, we consider the following light attenuation:
\begin{equation}
\label{eq:attenuation}
a_m(\mathbf{X})= \frac{( \hat{\mathbf{L}}_m (\mathbf{X}) \cdot \hat{\mathbf{D}}_m)^{\mu _m}}{||\mathbf{L}_m (\mathbf{X})||^2}.
\end{equation}
\paragraph{Light Calibration.} The aim here is not only to estimate the point light position $\mathbf{P}_m$ \cite{SantoWLSM20}, but also the other LED parameters $\mathbf{D}_m$, $\mathbf{\phi}_m$ and $\mu _m$. Instead of employing methods that aim at estimating these parameters while reconstructing the geometry \cite{logothetis2017semi, queau2017semi, queau2018led}, we developed a custom method that accurately estimates $\mathbf{P}_m$, $\mathbf{D}_m$, $\mathbf{\phi}_m$ and $\mu _m$ from PS images of a purely diffuse reflectance plane. To do so, we used a plane with 99\% nominal reflectance in UV-VIS-NIR wavelength range (350 - 1600nm). To have an initial estimate of $\mathbf{\phi}_m$, we measured the LED brightness with a LuxMeter.  

For every object, the calibration plane was captured twice, at different distances, in order to get data redundancy and produce a more accurate calibration. Thus, the Lambertian calibration object with albedo $\rho$ and surface normal $\mathbf{N}$, should satisfy the resulting image irradiance equation:
\begin{equation}
\label{eq:irad}
\mathbf{I}_m= \mathbf{\phi}_m a_m \rho {\hat{\mathbf{L}}}_m \cdot\hat{\mathbf{N}}.
\end{equation}

 The irradiance Equation~\ref{eq:irad} was implemented into a differentiable renderer (using Keras of Tensorflow v2.0) with the LED parameters being the model weights thus allowing refinement from a reasonable initial estimate. The parameters were initialised as follows: $\mathbf{\phi}_m$ from the LuxMeter, $\mathbf{D}_m=[0,0,1]$, $\mu_m=0.5$, $\mathbf{P}_m$ from the schematic of the printed circuit board of the LEDs and $\rho =1$. We used $L_1$ loss function for 30 epochs and converged to around 0.005 error i.e 0.5\% of the maximum image intensity.  The complete calibration parameters are included in the dataset. 

\subsection{3D Ground Truth Capture}

\paragraph{3D capturing device.} 
3D ground-truth has been acquired with the optical 3D scanner GOM ATOS Core 80/135 
with a reported accuracy of 0.03mm (see Figure \ref{fig:setup}). The GOM scanner uses a stereo camera set-up and more than a dozen scans were performed and fused per object.  In order to keep the geometry of the object consistent with the PS data, no spray coating has been used to ease the acquisition. Indeed, coating material can fill up those regions of the objects that are prone to interreflection and that are noticeably harder to reconstruct. Instead, markers were used for some objects.

 \paragraph{Alignment.} The laser scans of the objects were aligned and merged using MeshLab \cite{CignoniCCDGR08}. Some manual removal of noisy regions was performed and finally screened Poisson reconstruction \cite{kazhdan2013screened} was used in order to obtain full continuous surfaces (which are both useful for rendering normal maps and for mutual information alignment). As expected, not all parts of the surfaces of all objects have the same amount of noise, especially the metallic objects (Bell, Cup). 
Meshes were aligned with the photometric stereo images following the same procedure as in \cite{diligentshi2016benchmark}. This involved manual initialisation and then refinement using the mutual information registration filter of MeshLab. This was performed repeatedly until the projection of the mesh on the images was visually `pixel perfect' (using the semi-transparent overlay). Using the aligned meshes, ground truth normal maps were rendered (using Blender). In addition, manual segmentation was performed to remove regions where the GT was unreliable (markers on the objects, holes etc), those masks are provided in the dataset. 
Furthermore, the dataset contains meshes that have been interpolated in the marker/hole regions. The steps per object are summarised in Figure~\ref{fig:steps}.

\subsection{Dataset Overview}

For each of the 14 object, 52 PS images have been acquired using the BayerRG16 RAW format. The total amount of PS images amounts then to 728. For all objects, rectified RGB PS images will be released (by compensating for the radial distortion). We note that color balancing was not performed on the images as this will distort the saturated pixels (which is an important feature for CNN-based PS methods \cite{logothetis2020px, ikehata2018cnn}). Instead, RGB light source brightness are provided along with the rest of point light source parameters. 
Both normal map and depth ground truth will be provided in order to evaluate the accuracy of near-field PS methods with either cases. 


%% file: sections/Experiments.tex
\section{Experiments}
\label{sec:exp}

In this section, we evaluate four competing near-field methods namely \cite{logothetis2017semi,queau2018led, logothetis2020cnn, santo2020deep}. In addition, we also evaluate with \cite{ikehata2018cnn}, the best performing far-field method (on the far-field benchmark \cite{Shi2018}) to demonstrate the need for a near-field method.

\paragraph{Evaluation hyper-parameters.} \cite{logothetis2017semi, queau2018led} and \cite{santo2020deep} have publicly available code whereas for \cite{logothetis2020cnn} the code has been provided. Indeed, \cite{logothetis2020cnn} has the disadvantage that the light configurations has to be known at train time therefore specific light positions had to be assigned to the networks to be trained for the dataset.  \cite{logothetis2017semi} performs best with a priori initialisation of the specularity parameter $c$ (0 is fully specular, 1 fully Lambertian) we used 0.1 for the Cup, 0.2 for the Bell, 0.25 for the Bawl and Tool, 0.5 for the Ball, Die, Hippo, Jar and Squirrel, 0.75 for the Bunny and 1 for the rest. For \cite{queau2018led}, we used the Cauchy estimator with 0.5 on the respective hyper-parameter. For both  \cite{logothetis2017semi} and \cite{queau2018led} we disabled the lighting calibration parameter. For all methods, we evaluated on full resolution images (2048x1536) except for \cite{santo2020deep}, which is severely limited by GPU RAM so we had to subsample to (512x384) which was the maximum we could fit on 24GB Nvidia Titan RTX. All other approaches are CPU RAM limited but `only' require around 120GB. The computation time was varied from  around 15 minutes (the fastest was \cite{logothetis2017semi} on the Bowl) to  around two hours (the slowest was \cite{logothetis2020cnn} on the Jar). For all of the methods, the initialisation was a flat plane at the mean depth computed exactly using the GT depth map.

\begin{table}[t]
\setlength{\tabcolsep}{3.0pt} 
\begin{center}
\resizebox{1.0\columnwidth}{!}{%
\hspace{-0.0125\columnwidth}
\begin{tabular}{ | c | c c c c c c c c c c c c c c c c|}
 \hline
Method & Error & Bell & Ball & Buddha & Bunny & Die & Hippo & House & Cup & Owl & Jar & Queen & Squirrel & Bowl & Tool & Average \\ \hline
\multirow{3}{*}{\vspace{0.4cm} L17-\cite{logothetis2017semi}} & MAE & 28.25 & \textbf{9.77} & 11.5 & 20.15 & 11.95 & 15.42 & 29.69 & 30.76 & 13.77 & 10.56 & 13.05 & 15.93 & 12.5 & \textbf{15.1} & 17.03 \\ 
 & MZE & 4.45 & 0.81 & 4.67 & 7.51 & 4.58 & 3.19 & 6.99 & 2.67 & 3.64 & 6.56 & \textbf{1.89} & 1.82 & 4.37 & \textbf{3.25} & 4.02 \\  \hline
 \multirow{3}{*}{\vspace{0.4cm} Q18-\cite{queau2018led}} & MAE & 25.8 & 12.12 & 14.07 & 13.73 & 13.77 & 18.51 & 30.63 & 37.63 & 14.74 & 15.66 & 13.16 & \textbf{14.06} & 11.19 & 16.12 & 17.94 \\ 
 & MZE & 12.03 & 2.5 & 9.28 & 7.06 & 5.91 & 6.8 & 8.02 & 4.83 & 5.83 & 16.87 & 6.92 & 2.55 & 6.48 & 6.69 & 7.27 \\   \hline
\multirow{3}{*}{\vspace{0.4cm} S20-\cite{santo2020deep}} & MAE & \textbf{9.5} & 25.42 & 19.17 & 12.5 & \textbf{5.23} & 23.12 & \textbf{28.02} & \textbf{14.22} & 13.08 & 9.27 & 16.62 & 14.07 & 12.44 & 17.42 & 15.72 \\
 & MZE & 1.9 & 5.5 & 5.53 & 6.02 & \textbf{2.76} & 7.04 & \textbf{6.15} & \textbf{1.62} & 3.75 & 6.09 & 3.91 & 2.81 & 5.22 & 4.68 & 4.5 \\ \hline
 \multirow{3}{*}{\vspace{0.4cm} L20-\cite{logothetis2020cnn}} & MAE & 14.74 & 12.43 & \textbf{10.73} & \textbf{8.15} & 6.55 & \textbf{7.75} & 30.03 & 23.35 & \textbf{12.39} & \textbf{8.6} & \textbf{10.96} & 15.12 & \textbf{8.78} & 17.05 & \textbf{13.33} \\
 & MZE &  \textbf{1.53}	& 	\textbf{0.67}	& 	\textbf{3.27}	& 	\textbf{2.49}	& 	4.44	& 	\textbf{1.82}	& 	9.14	& 	2.04	& 	\textbf{3.44}	& 	\textbf{3.86}	& 	1.94	& \textbf{1.01}	& 	\textbf{2.80}	& 	5.90	& 	\textbf{3.17} \\ \hline \hline
\multirow{3}{*}{\vspace{0.4cm} I18-\cite{ikehata2018cnn}} & MAE & 23.55 & 44.29 & 35.29 & 36 & 41.52 & 44.9 & 49.05 & 35.78 & 40.27 & 40.66 & 32.89 & 41.09 & 28.04 & 31.71 & 37.5 \\
 & MZE & 5.93 & 6.59 & 10.92 & 6.88 & 7.83 & 7.59 & 8.98 & 3.17 & 8.67 & 15.54 & 8.08 & 5.8 & 6.69 & 12.45 & 8.22 \\ \hline \hline
\multirow{2}{*}{GT} & Diff-MAE & 2.5 & 2.69 & 2.69 & 2.93 & 2.49 & 3.2 & 9.19 & 2.85 & 4.3 & 1.79 & 4.22 & 3.26 & 2.27 & 2.34 & 3.34 \\
& Int\cite{queau2015edge}-MZE & 0.08 & 	0.22 & 	3.28 & 	2.30&  0.56& 1.28& 	7.43 & 	0.02	& 	3.51	&  0.12	& 	3.25 & 	1.12 & 	0.12 & 	0.13 & 	1.67 \\ \hline
\end{tabular}
} 
\end{center}
\caption{Complete evaluation of five methods on all objects. Mean angular error MAE (degrees) and mean depth error MZE (mm) are reported. The last two lines contain the error obtained after differentiation and integration of GT depth and normals respectively.}
\label{tab:mainresults}
\end{table}

Finally, we also evaluated the far-field method \cite{ikehata2018cnn}. The assumed lighting direction was set the average one for each light and numerical integration was used on the output normal map to be able to compare surfaces. It is worth to mention that our 52 lights is within the range of lights the model in \cite{ikehata2018cnn} is trained for.

\begin{figure}[t]
\centering
 \includegraphics[width=\textwidth]{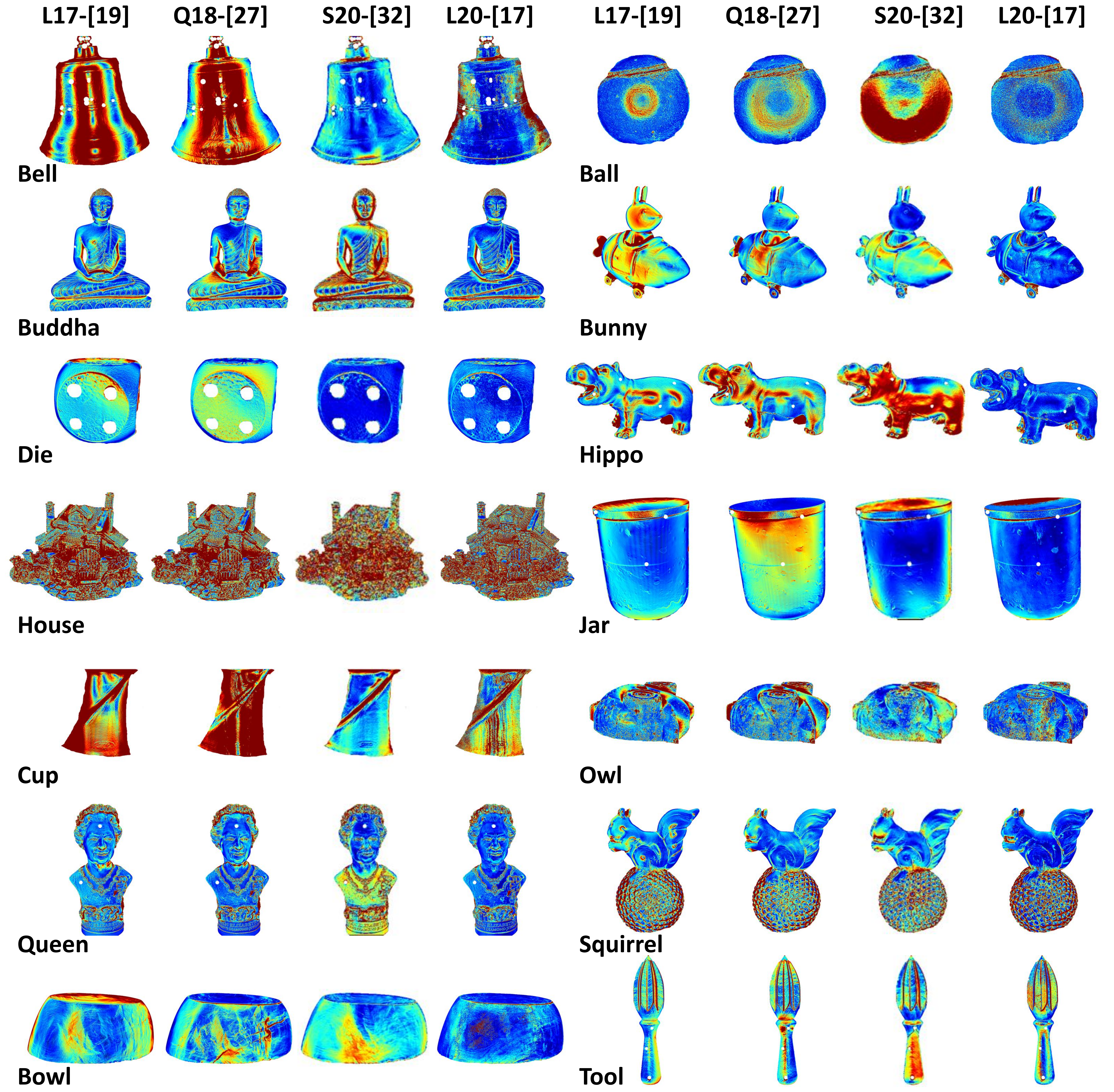}
  \includegraphics[width=0.2\textwidth,trim={0cm 1cm 0cm  2cm},clip]{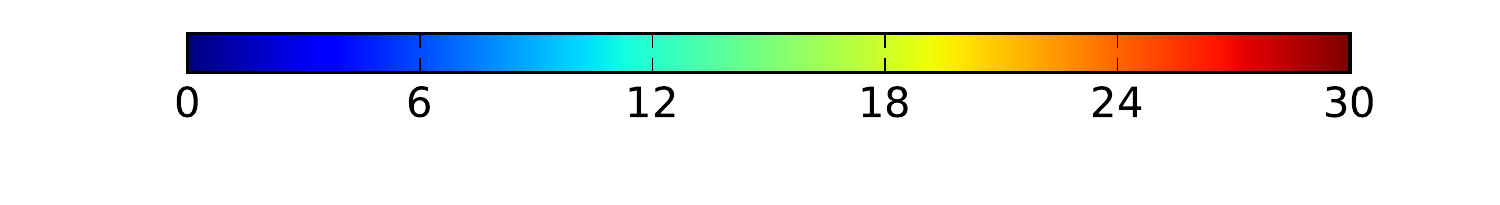}
 \caption{Normal error map comparison for all objects and all near-field methods.}
 \label{fig:maenormals}
\end{figure}

\paragraph{Evaluation metrics.} As it is the standard in PS literature, we first evaluate the competing approaches using the angular error on normal maps. We note that \cite{logothetis2017semi} and \cite{queau2018led} output surfaces as dense depth maps, therefore the normals have been estimated using first order (forward and downward) finite differences. The other 3 methods output both surfaces and normals. It is very important to mention, that normal evaluation has two major limitations. Firstly, for real data, there can be regions where the ground truth normal uncertainty is non-negligible. This is inevitable due to capturing surfaces with a laser scanner that only provides very dense point clouds. Even micro-meter accuracy on the surface can generate a few degress uncertainty of normals in regions of complicated geometry. The second important issue with evaluating on normals is that even on synthetic data,  ground truth normals are not fully consistent with the ground truth depth \cite{QueauDA18, zhu2020least}. This is inevitable due to the fact that for any non-trivial object, the projection operation generates a depth map that is discontinuous and non-differentiable for a significant portion of the pixels. In fact, to quantify this discontinuity measure, we compared the ground truth normals with the normals that are obtained with differentiation of the ground truth depth and indeed observed a $3.3^o$ error on average over the whole dataset (varying from $1.8^o$ on the Jar to $9.2^o$ on the House, as shown on the penultimate row of Table~\ref{tab:mainresults}). Conversely, numerical integration (using \cite{queau2015edge}) of GT normals has an average error of $1.67$mm with repsect to the ground truth.

%% file: sections/conclusion.tex
\section{Results}
\label{sec:dis}

\begin{figure*}[t]
\centering
 \includegraphics[width=\textwidth]{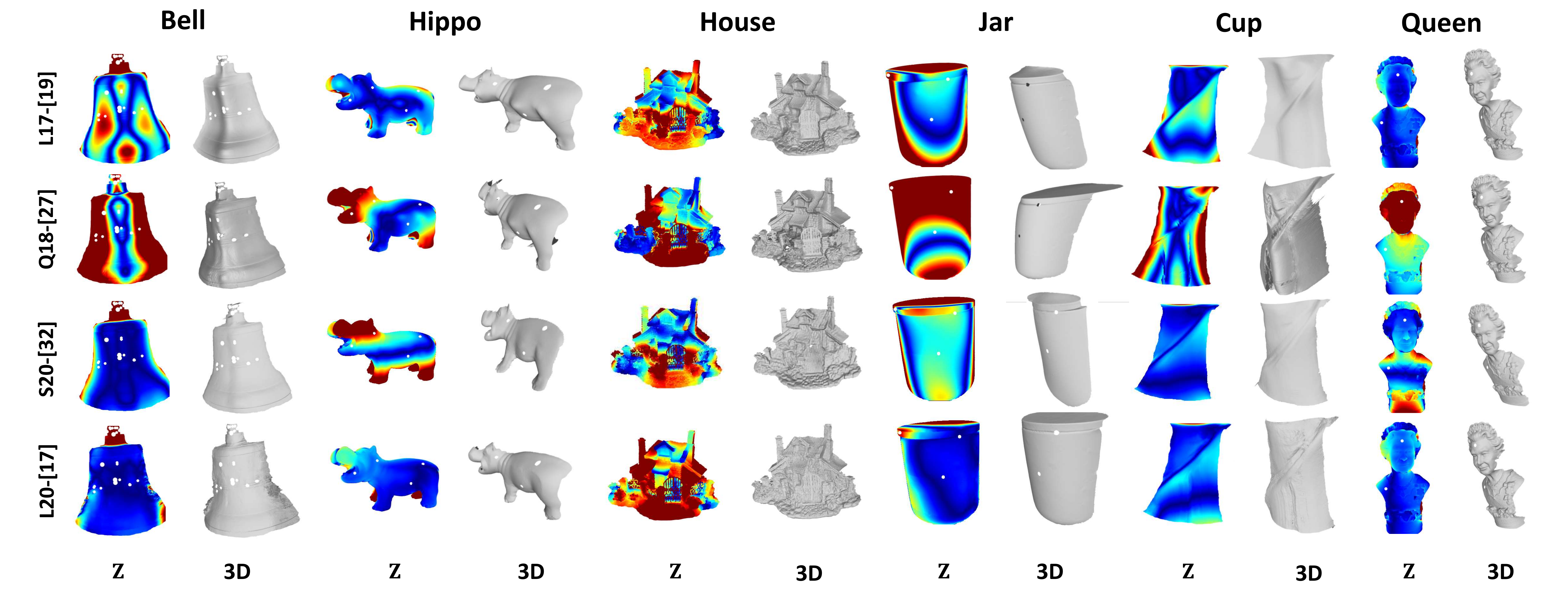}
  \includegraphics[width=0.2\textwidth,trim={0cm 1cm 0cm  1cm},clip]{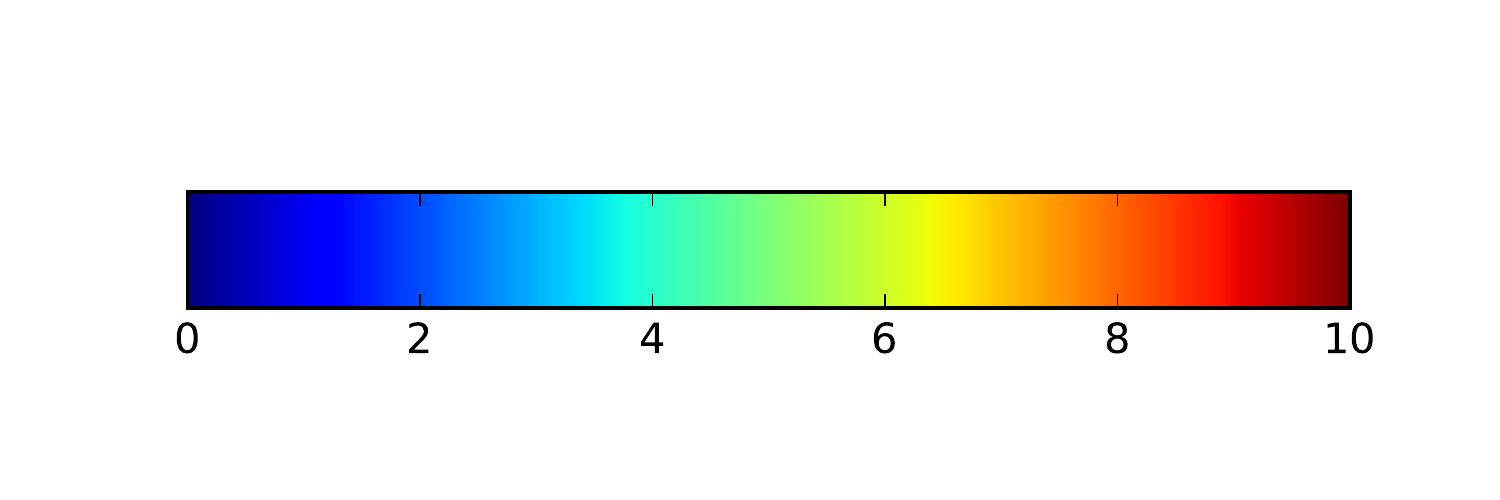}
 \caption{Output surface comparison for 6 objects and all methods. This is shown qualitatively through the 3D meshes and well as  depth $Z$ error maps (errors in mm).}  
 \label{fig:sixobjects}
\end{figure*}

In this section, we analyse the performance of \cite{ikehata2018cnn, logothetis2017semi, queau2018led, logothetis2020cnn, santo2020deep} which is shown quantitatively at Table \ref{tab:mainresults} and qualitatively at Figures~\ref{fig:maenormals} and \ref{fig:sixobjects}.  

We first of all observe that the far-field method \cite{ikehata2018cnn} fails to produce accurate results as expected. The two classical optimisation methods \cite{logothetis2017semi,queau2018led} are outperformed by the deep learning approaches \cite{santo2020deep} and \cite{logothetis2020cnn} on the normal error metric for most scenarios although \cite{logothetis2017semi} is not significantly worse for a few objects and indeed achieves the minimum depth error on the Queen and Tool objects. 
Despite the fact that \cite{logothetis2020cnn} is the best overall performer, we emphasise that it requires knowledge of the setup at train time. 
\cite{santo2020deep} achieves the best performance on object with specular and metallic materials (Die \& Cup both normals \&  depth, Bell normals only), because of the use of a patch-based network that extract the most information of the metallic object. It also achieves best MAE on the Bell. The orthographic camera assumption of \cite{santo2020deep} in terms of error translates to a growing inaccuracy towards the external part of the reconstruction (see Bell, Cup and Jar in Figure \ref{fig:maenormals}).

We also notice that the normal predictions are more noisy as opposed to  depth prediction. This could be due to actual noisy estimates of the normals from ground truth meshes which is inevitable for any laser scanner (see in particular the Ball in Figure \ref{fig:maenormals}). As the ground truth depth is more reliable, it is a better evaluation metric compared to the `ground truth' normals.  See Figure \ref{fig:sixobjects} for depth evaluation.  

By looking at Figure \ref{fig:maenormals} it can be seen that even the best methods perform poorly for recovering the geometry of very oblique regions. This is observable for the Jar, Owl and Cup in Figure \ref{fig:maenormals}. Therefore these represent quite hard regions to retrieve.

An interesting observation is that for both CNN-based methods \cite{logothetis2020cnn,santo2020deep}, the material's specularity does not seem be a significant factor of performance. Indeed, convex regions (where self reflections are negligible) are consistently recovered correctly regardless of the material: diffuse head of Queen, bronze Bell, plastic Hippo, wooden Bowl; with the only exception being the aluminium Cup. This is a clear advantage of CNN methods against the classical ones that require diffuse or mostly diffuse materials.


We observe that the hardest regions are the ones containing high frequency details (sharp boundaries) such as House, bottom part of the Squirrel, details of the Queen etc. 

\section{Conclusion}
\label{sec:conclusion}

In this work we proposed the first dataset for the near-field PS problem. Differently from the far-field assumption, when the target object is close to the camera/light setup several non-linear physical effects as anisotropic light propagation, light attenuation and perspective viewing geometry occur. Current PS datasets, mostly consider scenarios where such effects are negligible as they provide directional light coordinates and use quite long-focus lenses giving orthographic viewing geometry.

Recent research trends on 3D reconstruction using PS have shown an increasing interest to deal under near-field settings. However, the lack of a dataset for this topic has prevented to fairly compare different approaches. For this reason, we benchmark the recent near-field PS approaches and analyse their performance over our dataset which include objects with a wide variety of materials. We also provide a discussion about appropriate ways of evaluation (depth vs normals). In addition, as we noticed that most of the error is expectedly concentrated on the edges and discontinuity regions we conclude that future research has to improve the interpretation of the PS imaging data in these specific areas and possibly exploiting networks with edge detection capability to better deal with interreflctions.

Finally, it is worth investigating the possibility of using completely raw image data without demosicing or radial distortion compensation. This requires incorporating the radial distortion into the image irradiance equation and treating the images as pure intensity and ignoring the potential of recovering colours. The advantage of skipping these two pre-processing steps is the potential of eliminating some image artefacts, especially around image edges, which currently achieve the least accuracy.

%% file: sections/leds.tex
\section{Photometric Stereo Setup}
\label{sec:pssetup}
This section gives additional details on the PS setup.

\begin{figure}[h]
\centering
\includegraphics[width=0.49\textwidth,trim={0cm 0cm 0cm  0cm},clip]{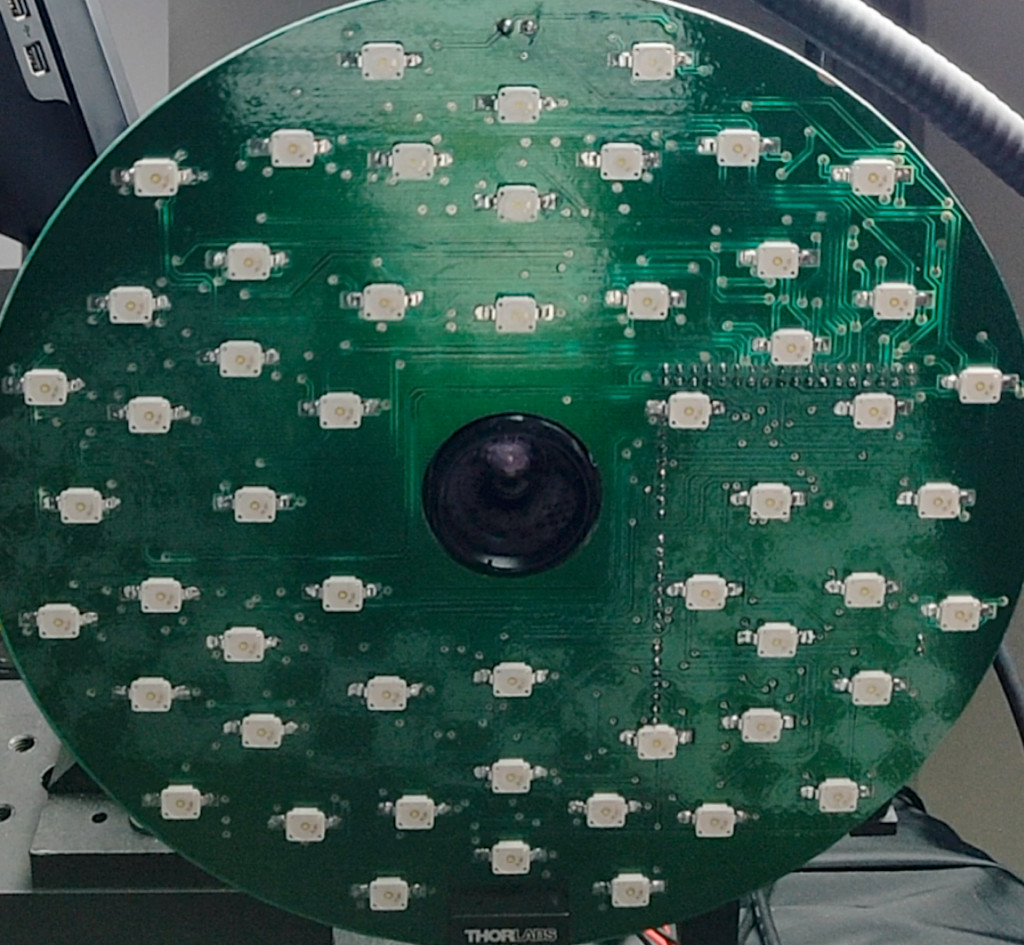}
\caption{Close-up of the setup used for acquiring Photometric Stereo images.}
\label{fig:ledsetup}
\end{figure}
\noindent
It consists of the following main components (see Figure \ref{fig:ledsetup}):
\begin{itemize}
\setlength\itemsep{0.1mm}
    \item RGB camera FLIR BFS-U3-32S4C-C with 8mm lens
    \item 52 LED Golden Dragon OSRAM
    \item variable voltage for adjustable LED power
    \item Arduino Mega 2560
\end{itemize}

The Arduino Mega 2560 controls the LEDs that are turned on and off individually during the Photometric Stereo image capture. The PCB has been designed to accommodate a specific 8mm lens that allows a reasonable wide field of view on a target object placed few centimeters away from the camera. The type of OSRAM LEDs (LW W5SN) are capable of emitting 5600 Kelvin white light up to 191 lumen. The $120^o$ of viewing angle of the LEDs allows a complete lighting of the scene. The LEDs have been distributed on 6 different circumferences (all centered in the camera centre) of radii 35, 45, 55, 65, 75 and 85 mm. Depending on the circumference, the LEDs have been positioned at variable angle of $30^{o}$ and $60^{o}$. 

In order to avoid out of focus areas, the aperture of the lens is kept to the minimum (f/11). Therefore, variable time of exposures (among objects) have been used to prevent saturations or too dark regions. Note that a separate set of calibration images (see main paper Section 3.1) was captured for each different configuration.

The complete LED parameters (positions, brightness, directions, angular dissipation, see main paper Section 3.1) are included in the dataset.

%% file: sections/gt.tex
\section{Ground Truth Meshes}
\label{sec:gt}

\begin{figure*}[t]
\begin{tabular}{c c c c c}
\includegraphics[height=0.19\textwidth,trim={10cm 0cm 10cm  0cm},clip]{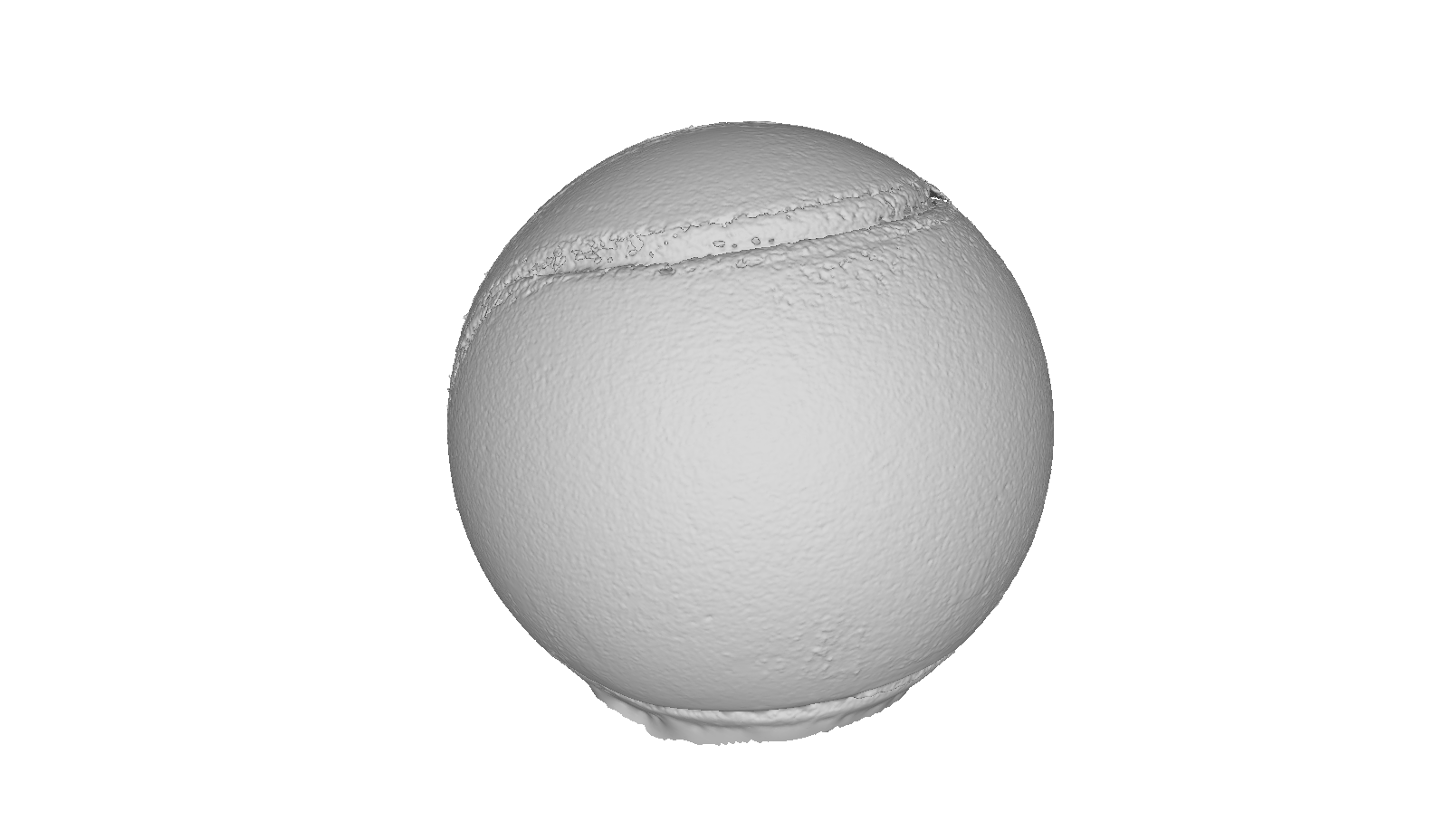} &
\includegraphics[height=0.19\textwidth,trim={10cm 0cm 9cm  0cm},clip]{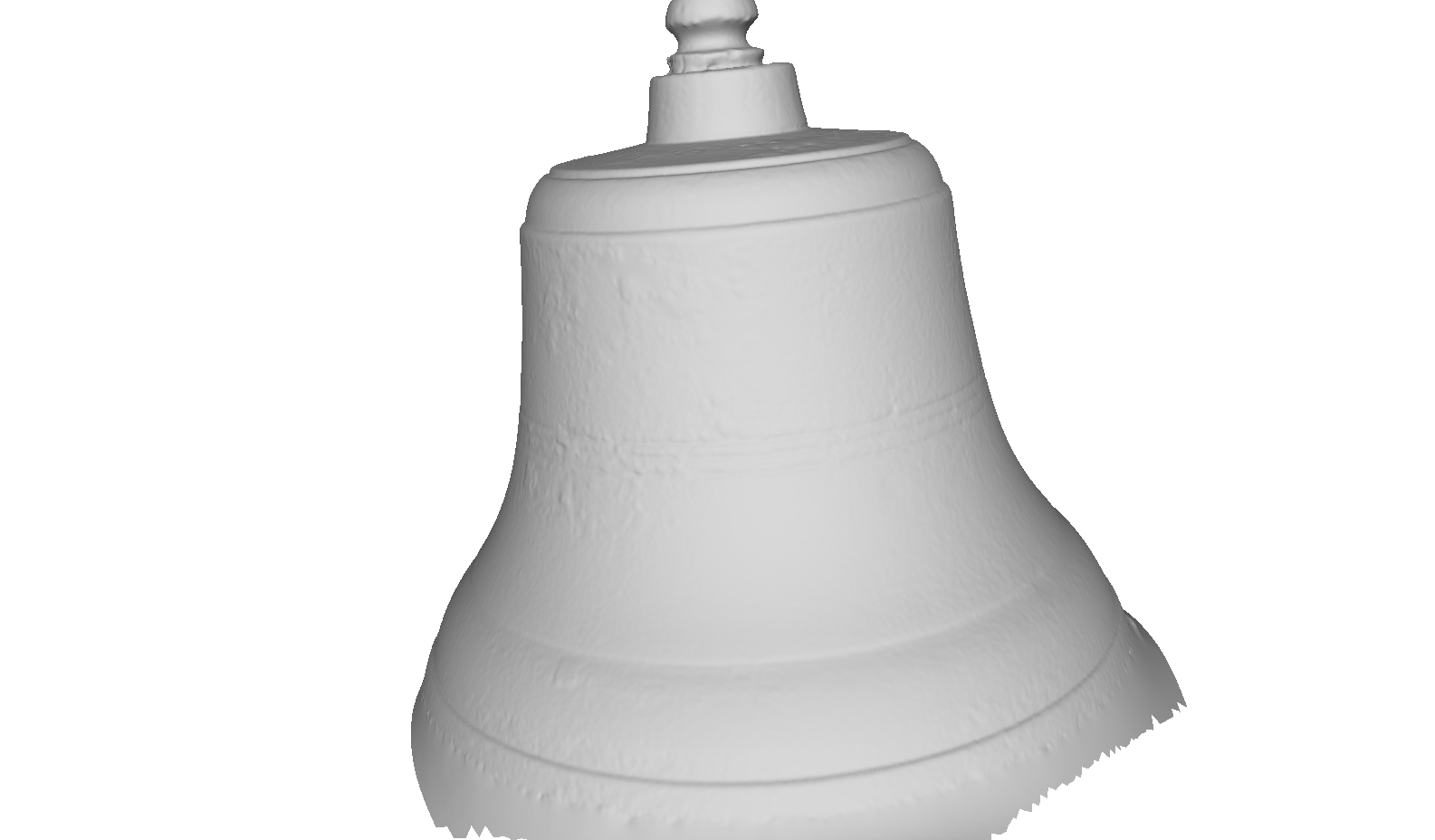} &
\includegraphics[height=0.19\textwidth,trim={11cm 0cm 10cm  0cm},clip]{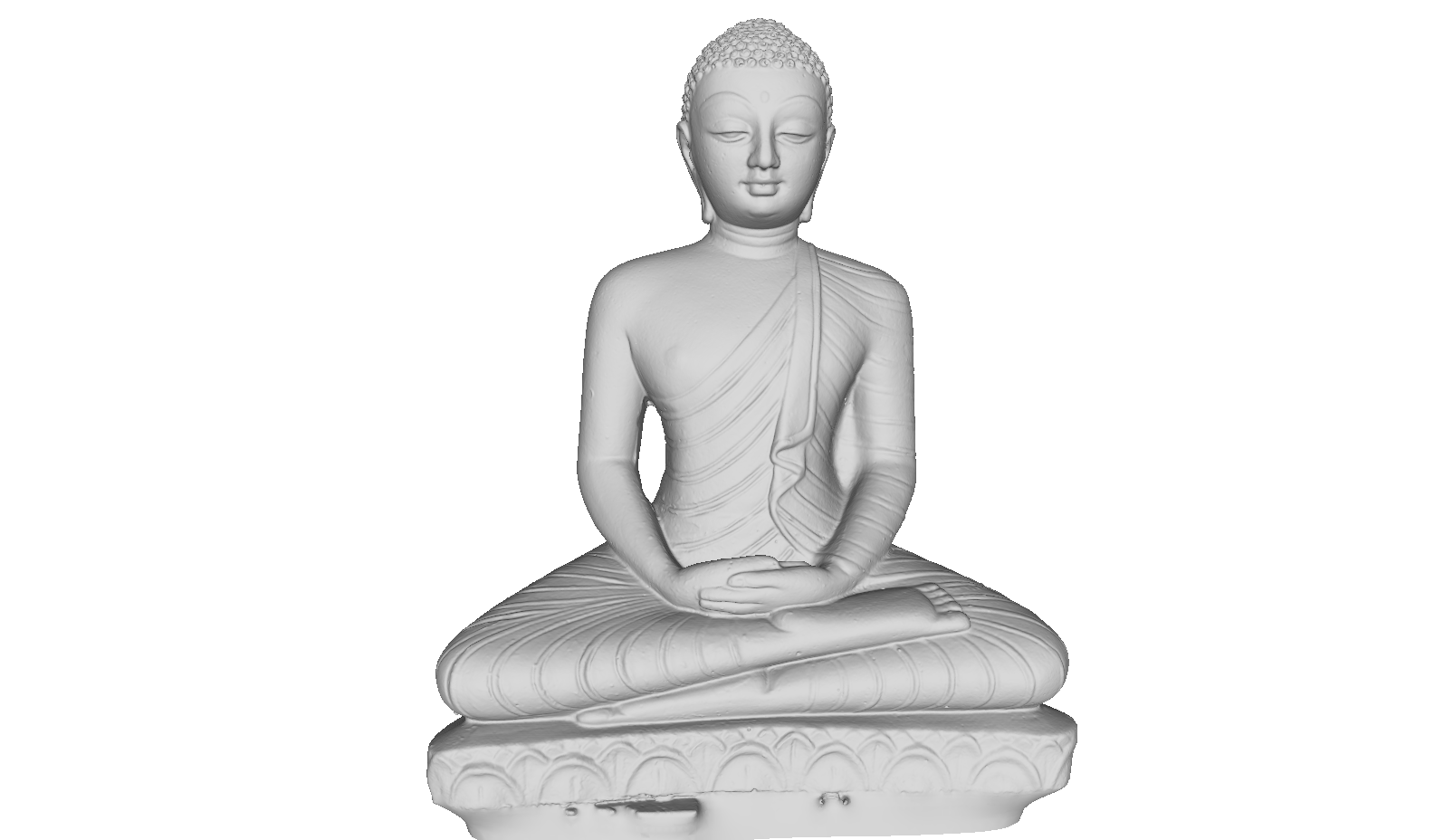} &
\includegraphics[height=0.19\textwidth,trim={15cm 0cm 9cm  0cm},clip]{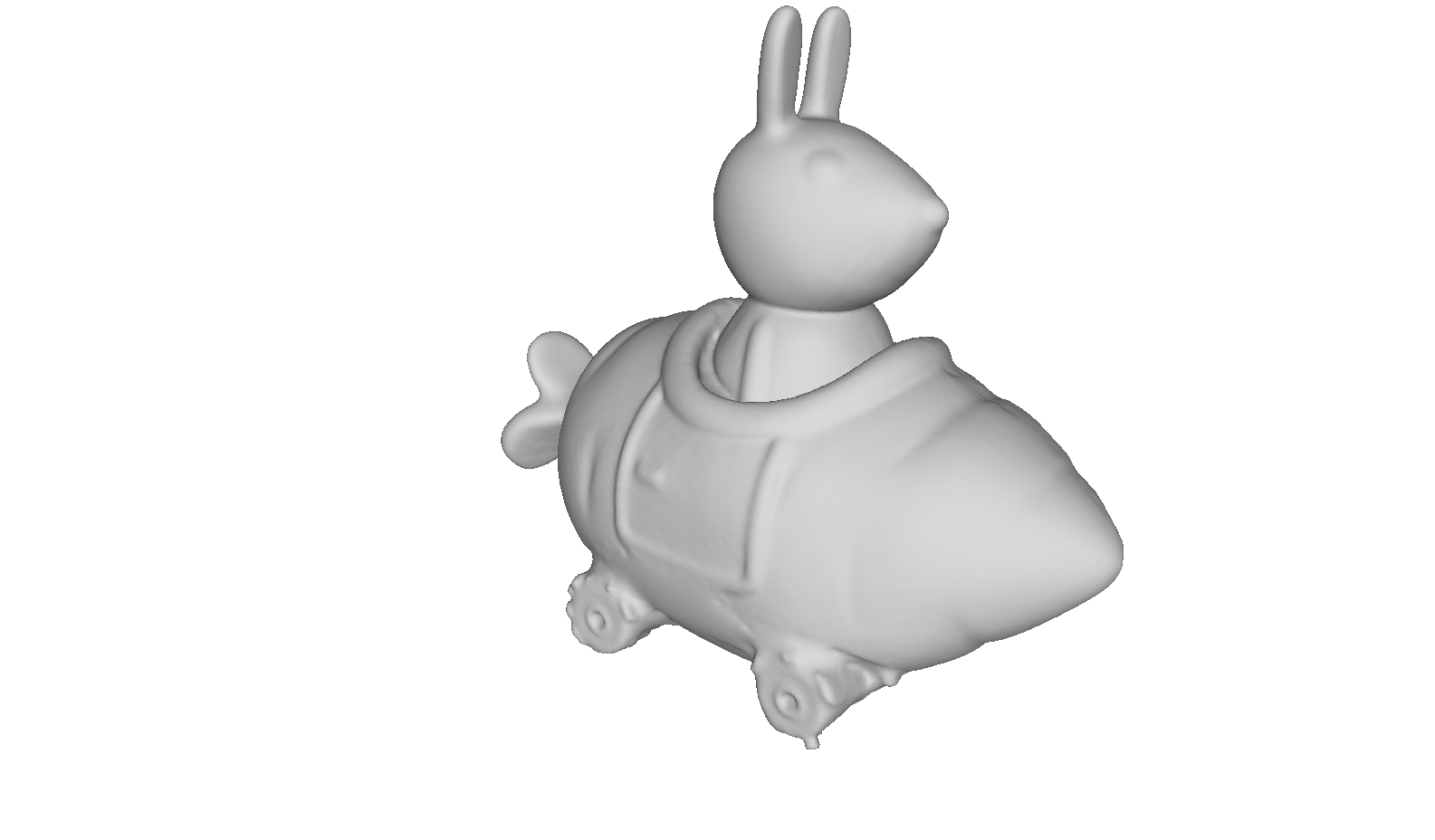} &
\includegraphics[height=0.19\textwidth,trim={11cm 0cm 11cm  0cm},clip]{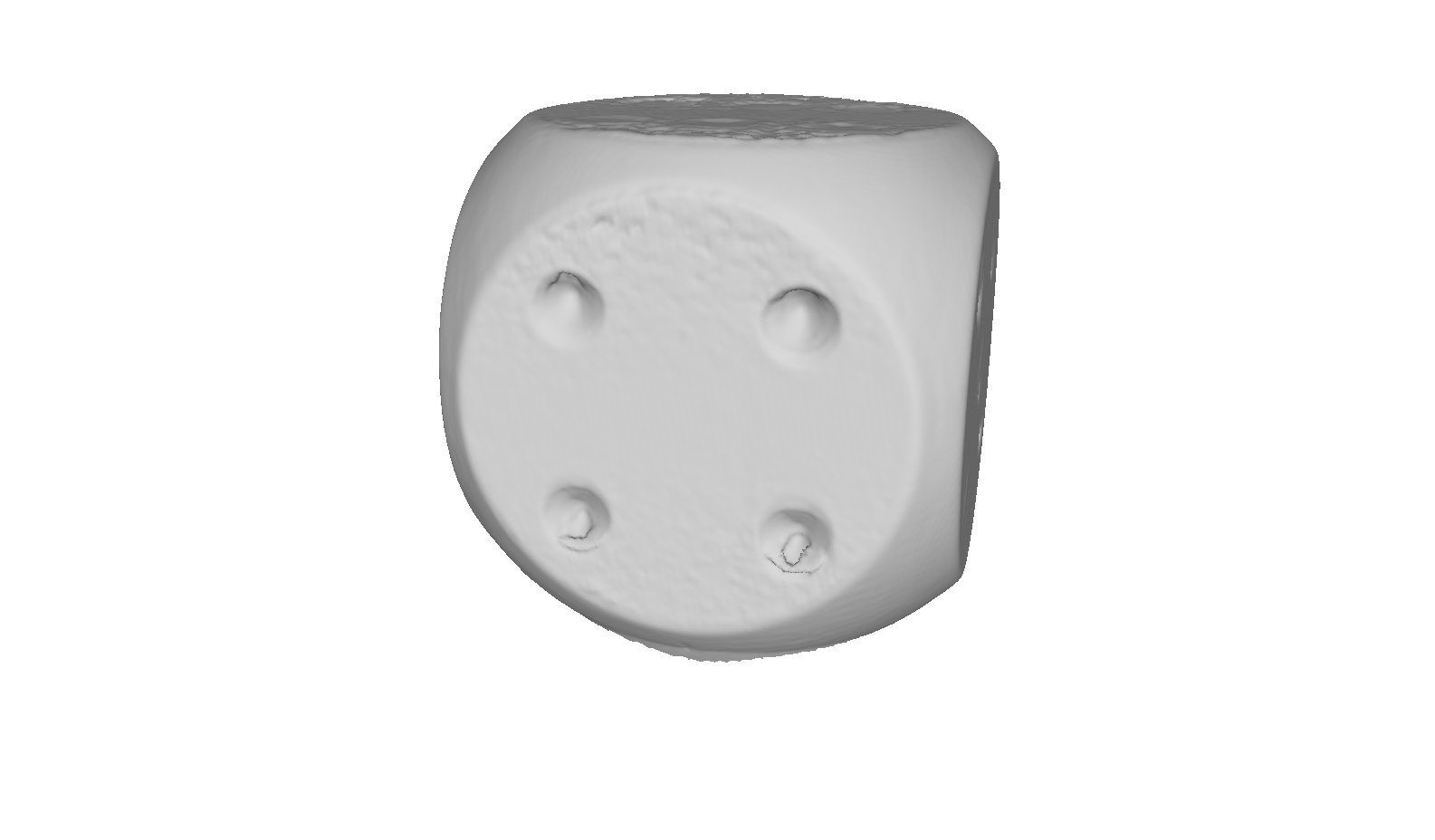} \\
Ball 2M & Bell 1.1M & Buddha 3.4M & Bunny 260K & Die 420K  \\
\includegraphics[height=0.17\textwidth,trim={7cm 0cm 11cm  0cm},clip]{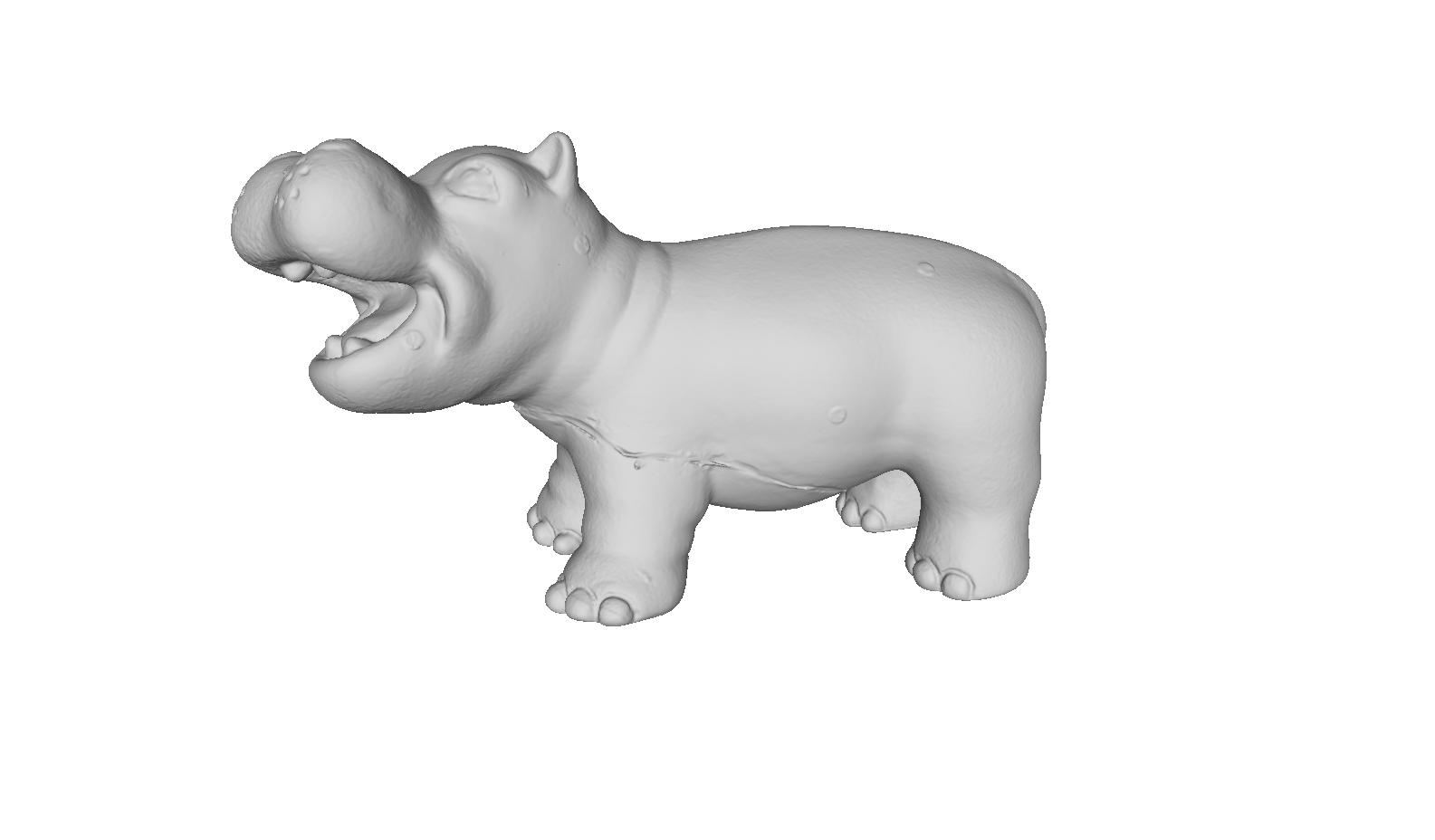} &
\includegraphics[height=0.17\textwidth,trim={7cm 0cm 8cm  0cm},clip]{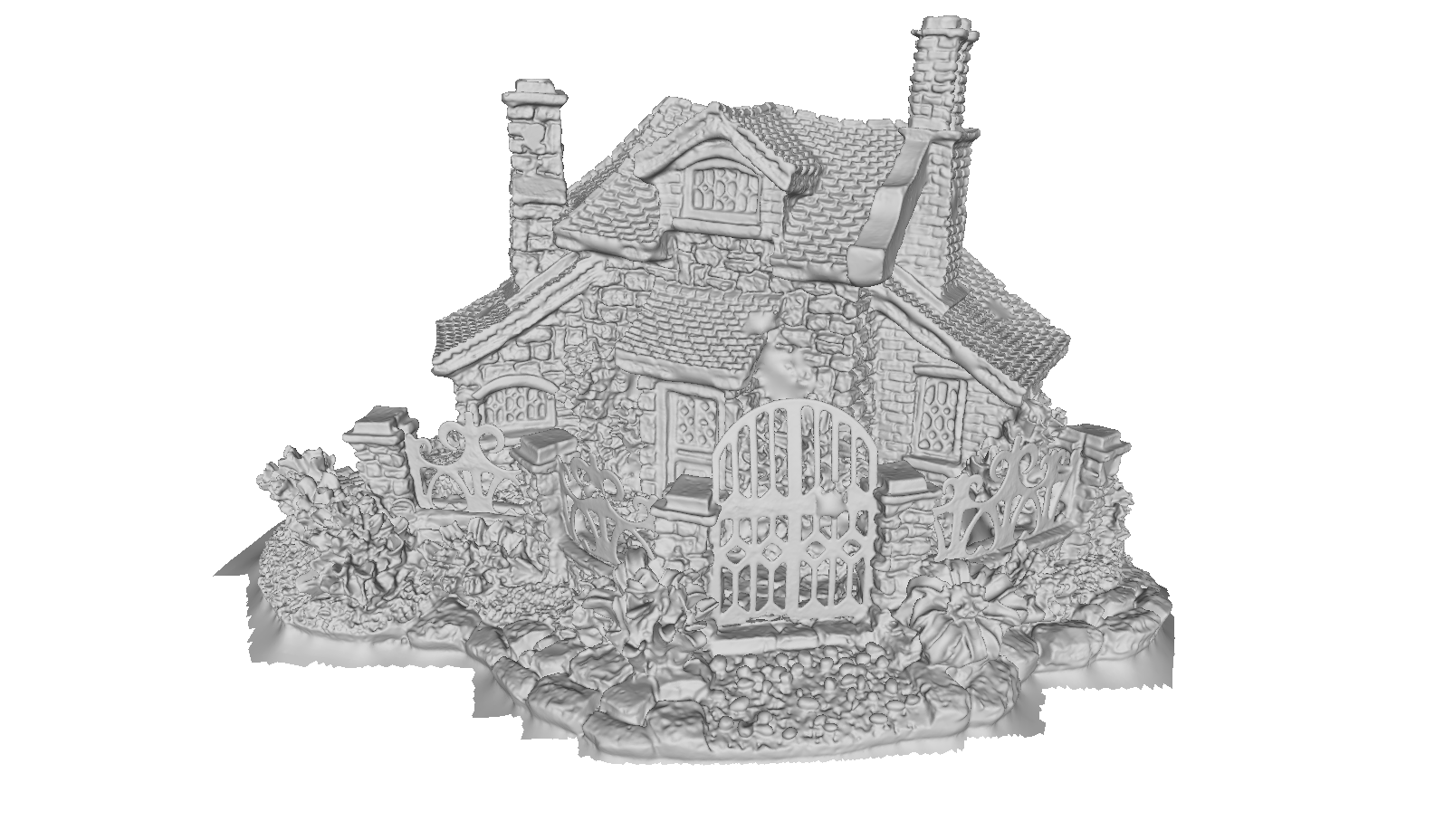} &
\includegraphics[height=0.17\textwidth,trim={11cm 0cm 14cm  0cm},clip]{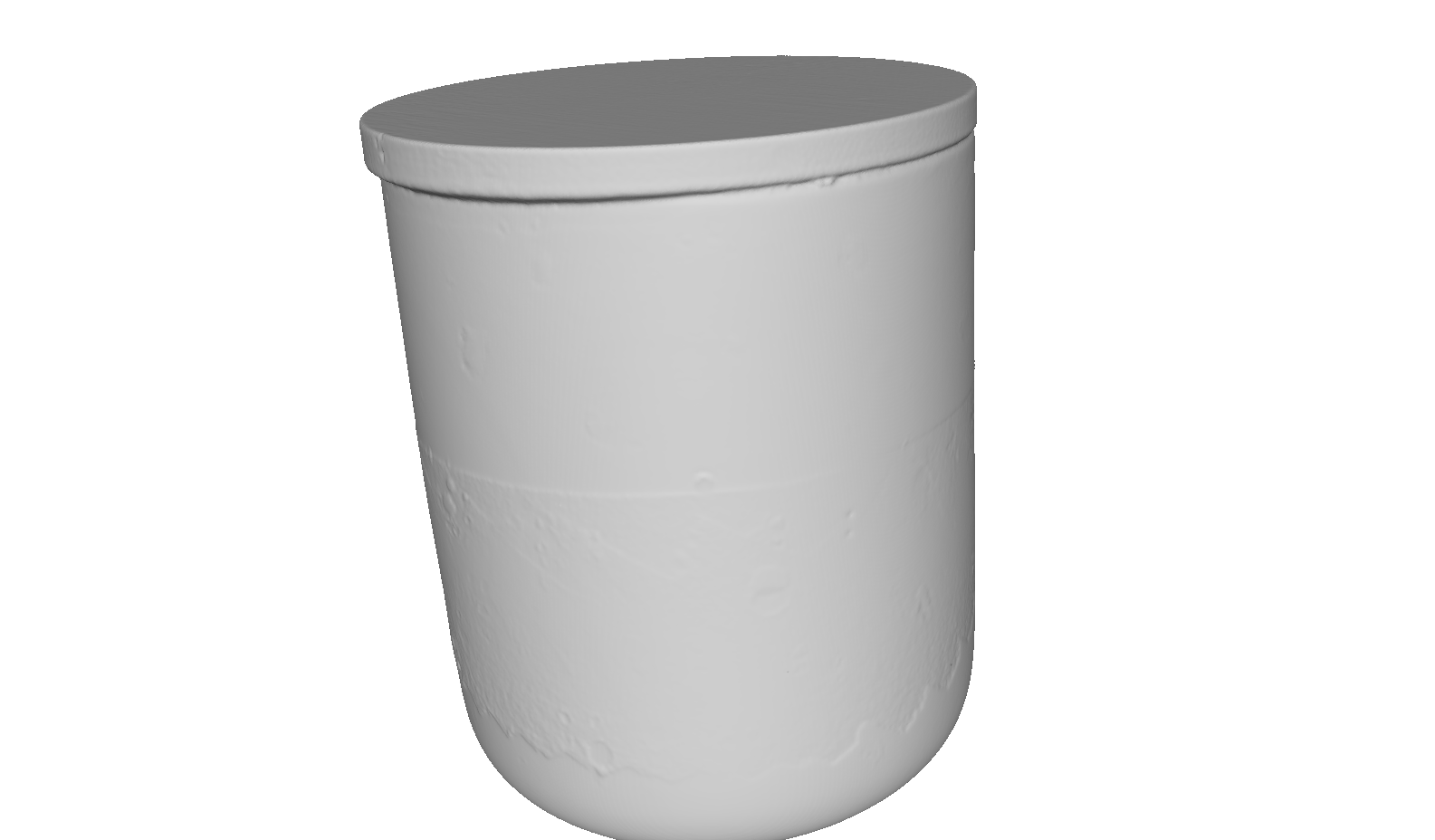}&
\includegraphics[height=0.17\textwidth,trim={12cm 0cm 14cm  0cm},clip]{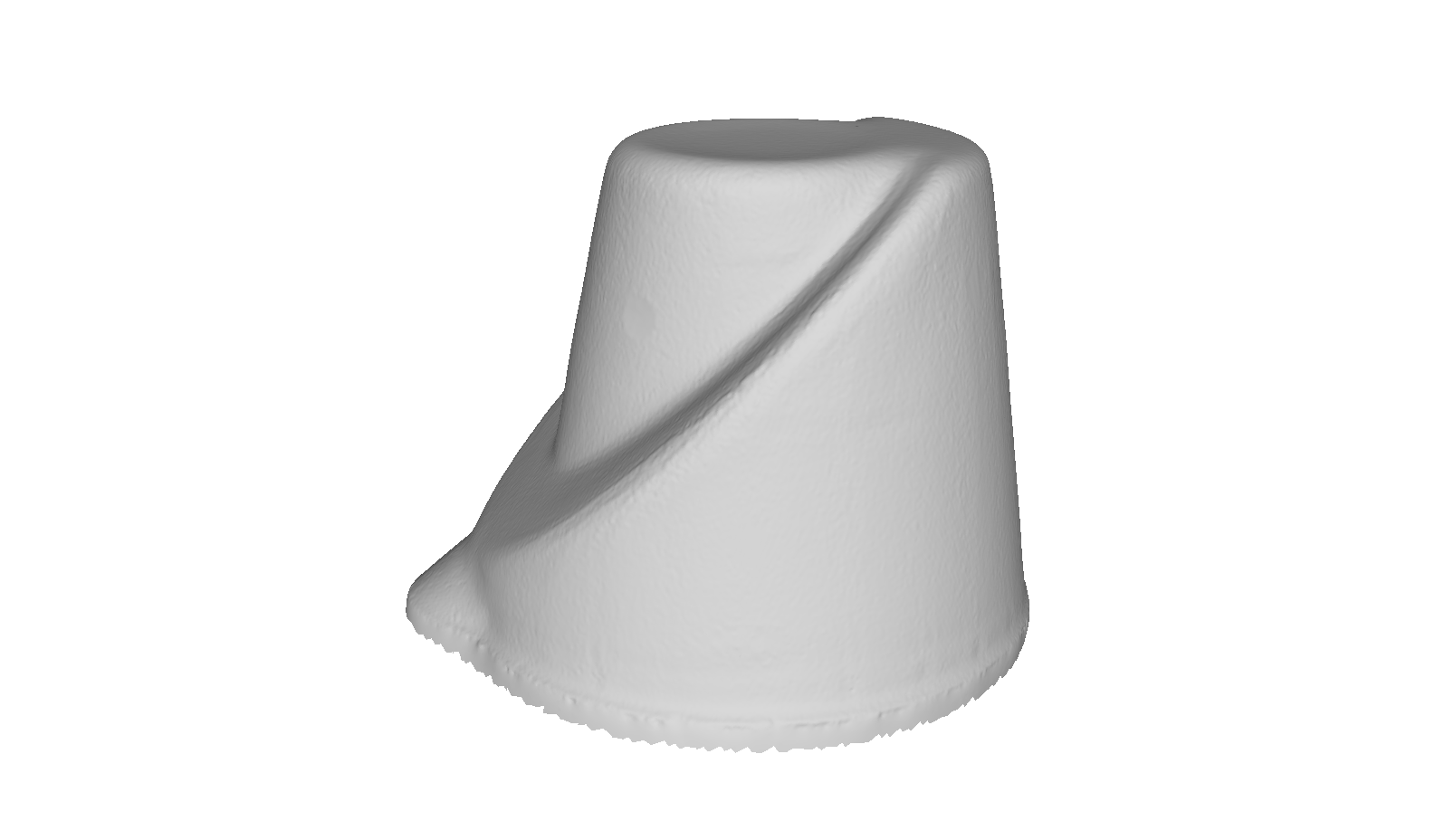} &
\includegraphics[height=0.17\textwidth,trim={10cm 0cm 10cm  0cm},clip]{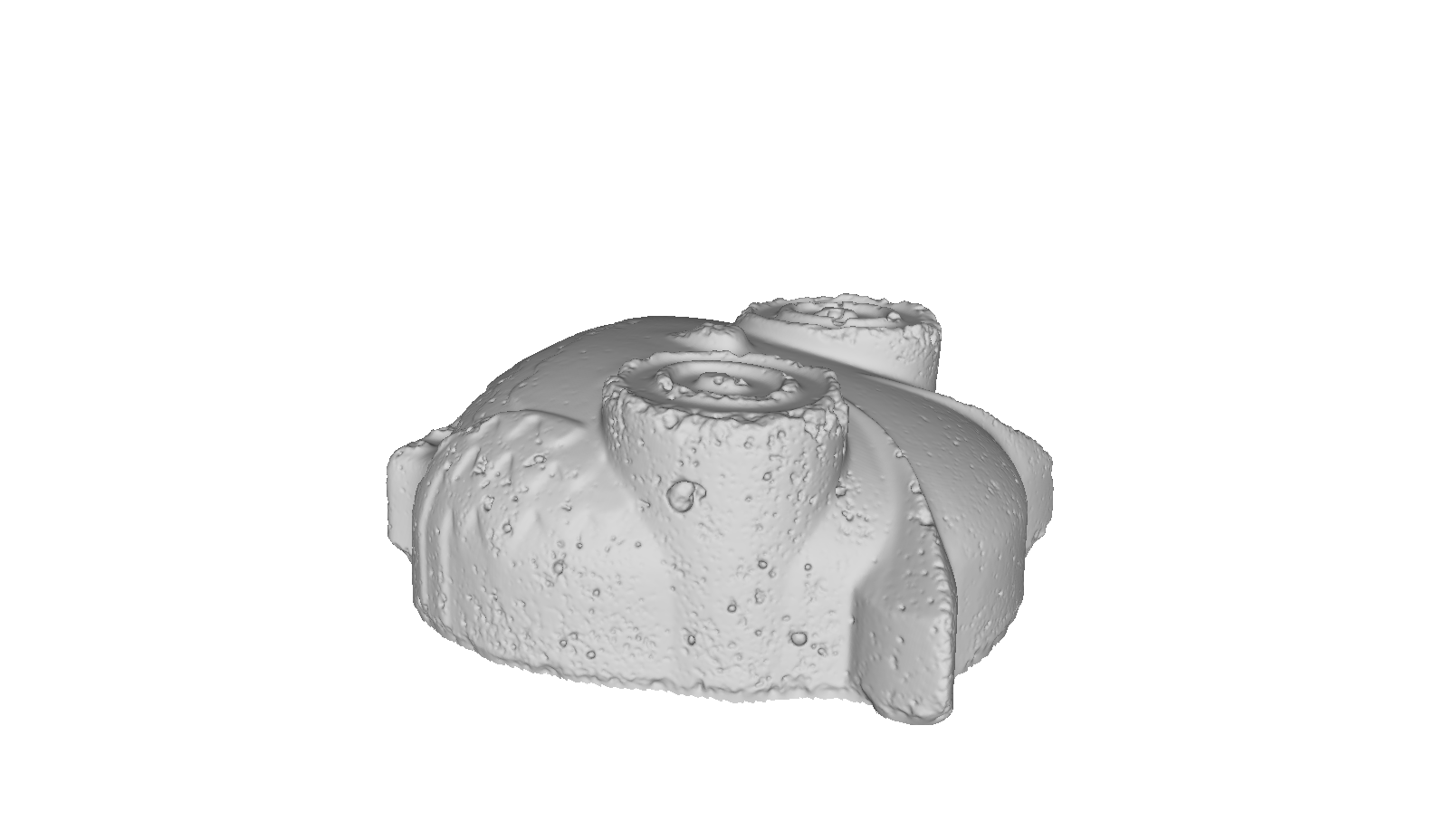} \\
Hippo 2.2M & House 6.1M & Jar 1.5M & Cup 1M & Owl  930K
\end{tabular}
\begin{tabular}{c c c c}
\includegraphics[height=0.21\textwidth,trim={15cm 0cm 15cm  0cm},clip]{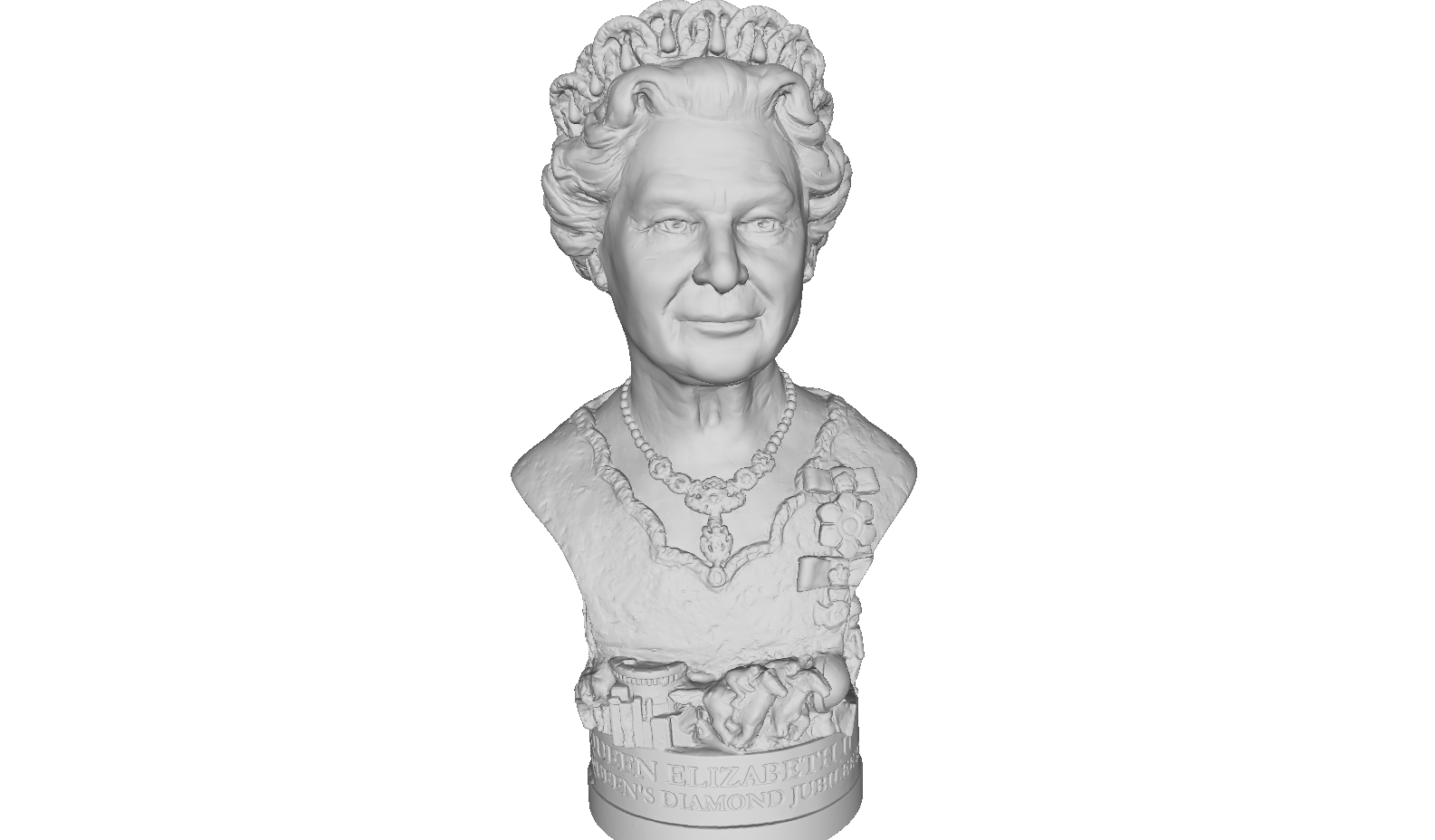} &
\includegraphics[height=0.21\textwidth,trim={13cm 0cm 10cm  0cm},clip]{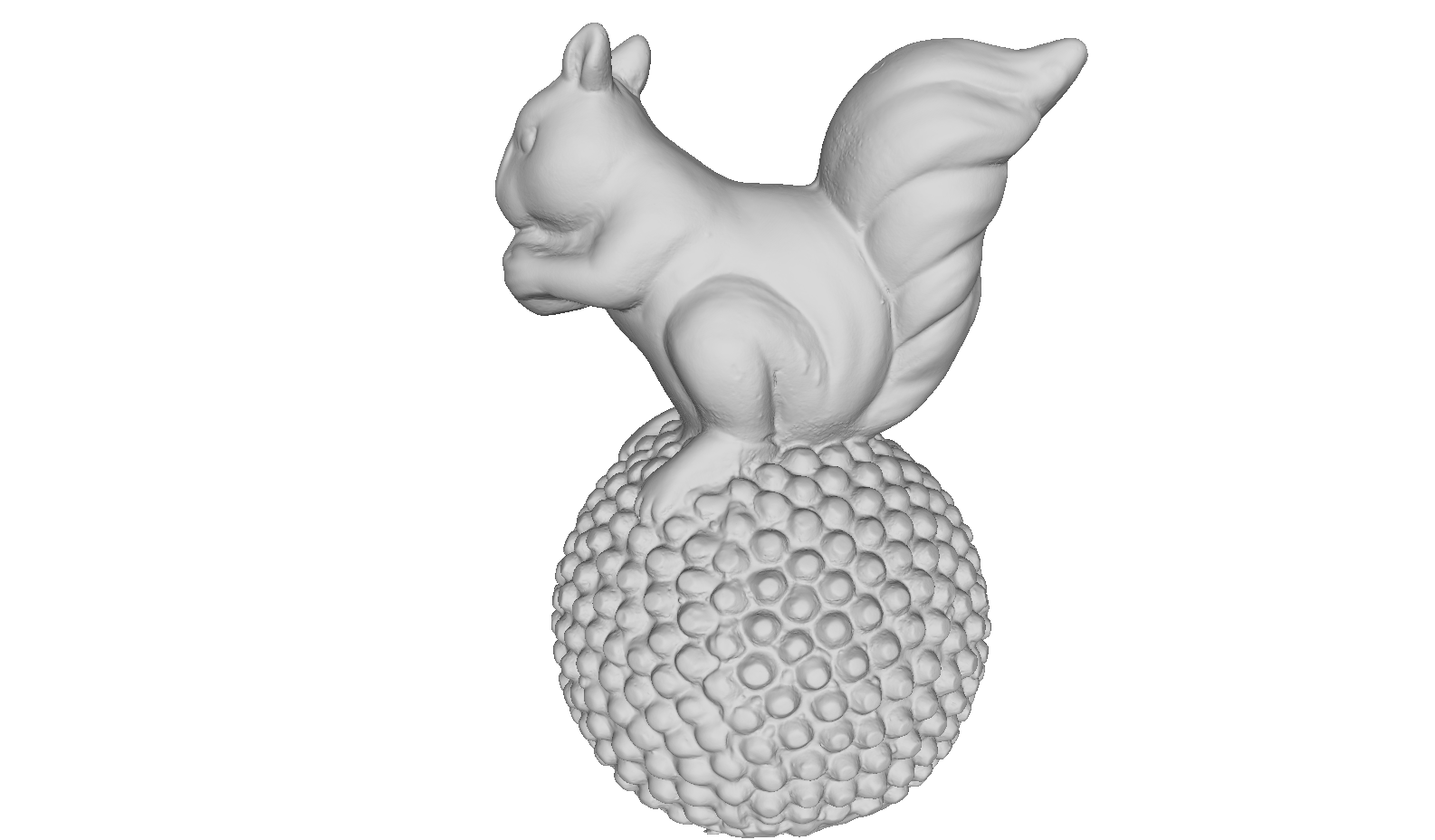} &
\includegraphics[height=0.21\textwidth,trim={9cm 0cm 10cm  0cm},clip]{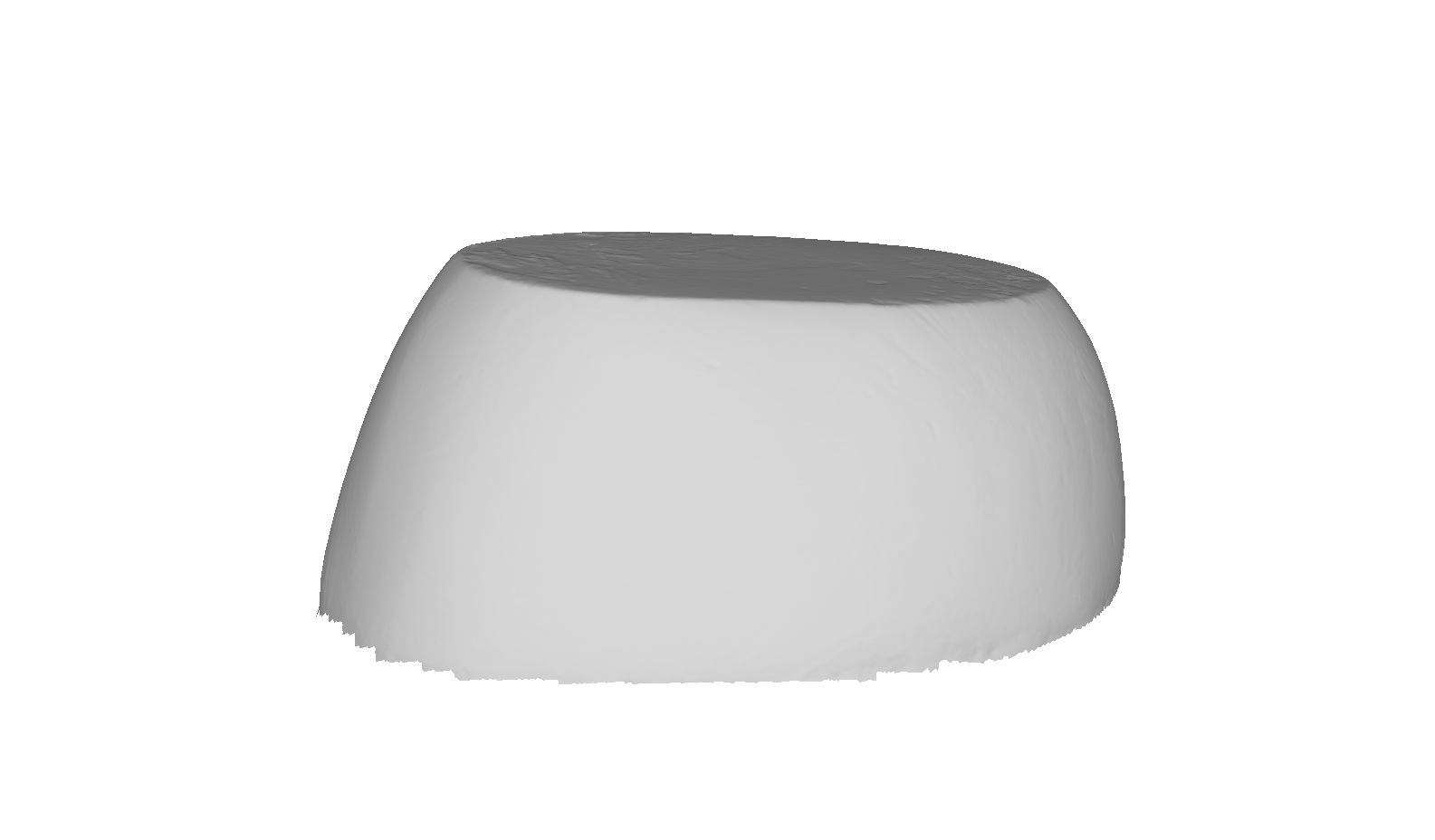} &
\includegraphics[height=0.21\textwidth,trim={13cm 0cm 13cm  0cm},clip]{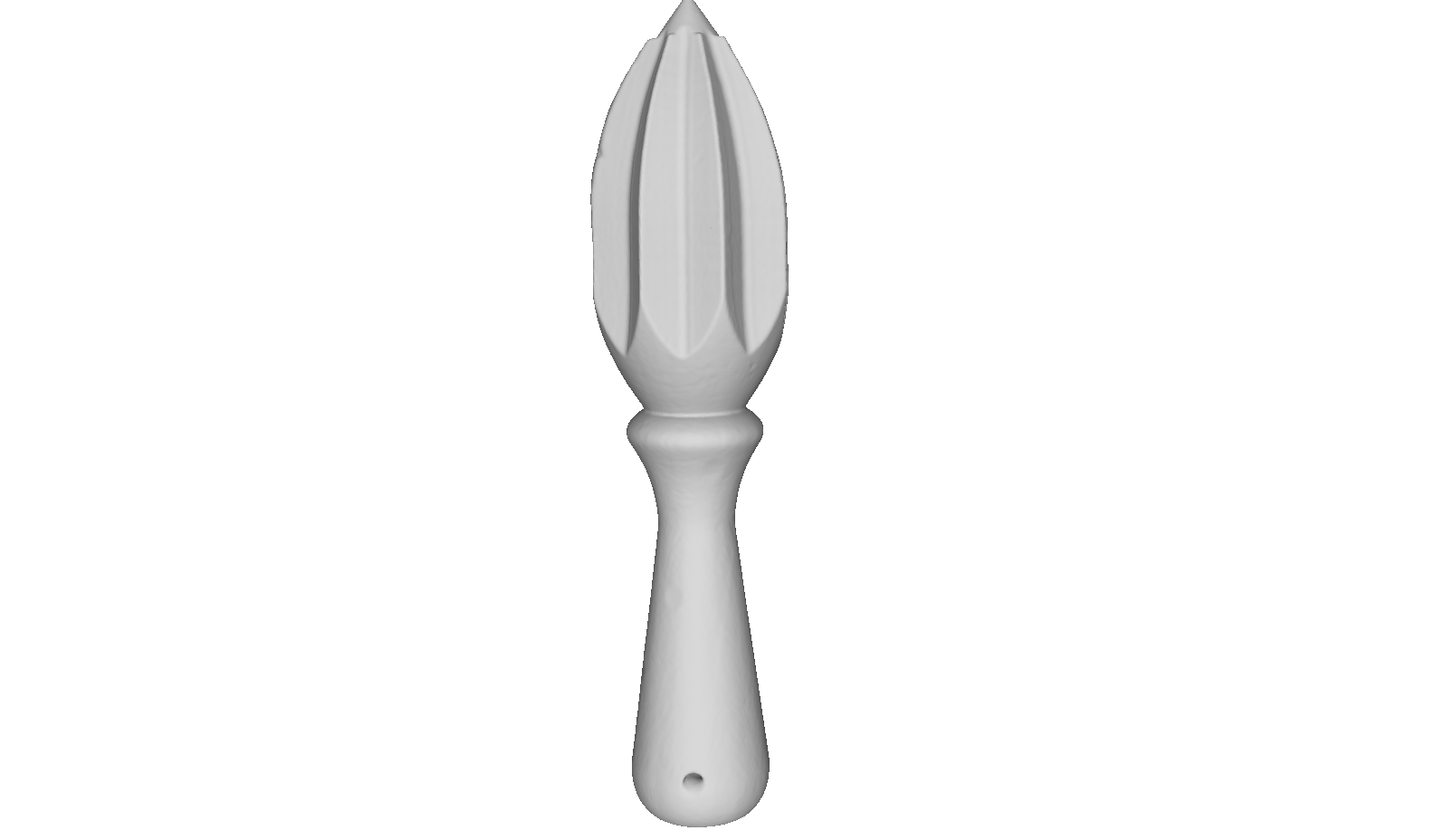} \\
   Queen 4.9M & Squirrel 5.5M & Bowl 750K & Tool 350K\\
\end{tabular}
\caption{Laser-scanned meshes and their respective number of triangles. }
\label{fig:gtmeshes}
\end{figure*}

This section provides supplementary information about the laser-scanned meshes which were used as ground truth for the evaluation of the competing PS approaches. Note that because of the different sizes of the objects and the characteristics of the surfaces, the number of triangles in each object mesh varies from 260K to 6.1M as shown in Figure~\ref{fig:gtmeshes}. As the scans can be of questionable quality at some regions due to visibility and/or specular material (i.e. numbers on the Die), manual segmentation was performed on the image domain to only evaluate on the reliable regions.

\begin{figure*}[t]
\includegraphics[width=0.18\textwidth,trim={2.0cm 0cm 2.0cm  0.0cm},clip]{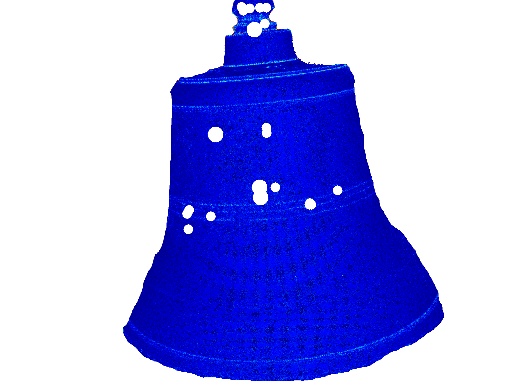} 
\includegraphics[width=0.18\textwidth,trim={2.0cm 0cm 2.0cm  0.0cm},clip]{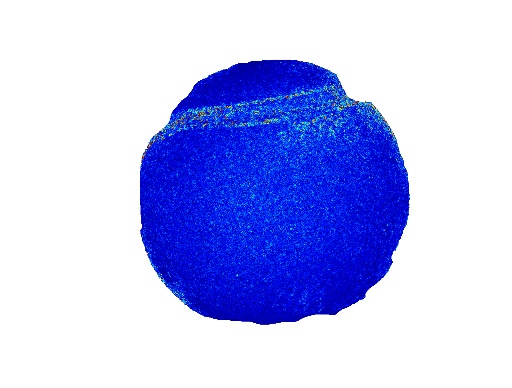} 
\includegraphics[width=0.18\textwidth,trim={2.0cm 0cm 2.0cm  0.0cm},clip]{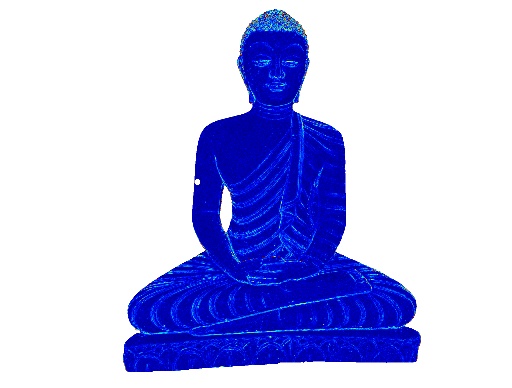} 
\includegraphics[width=0.18\textwidth,trim={2.0cm 0cm 2.0cm  0.0cm},clip]{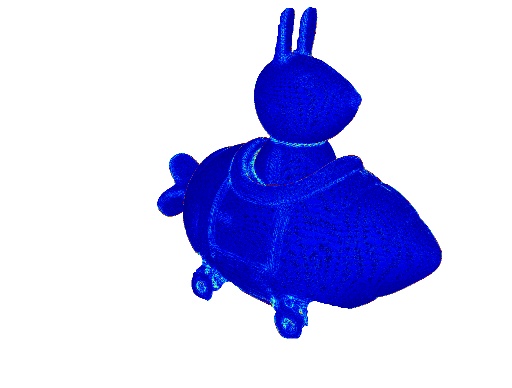} 
\includegraphics[width=0.18\textwidth,trim={2.0cm 0cm 2.0cm  0.0cm},clip]{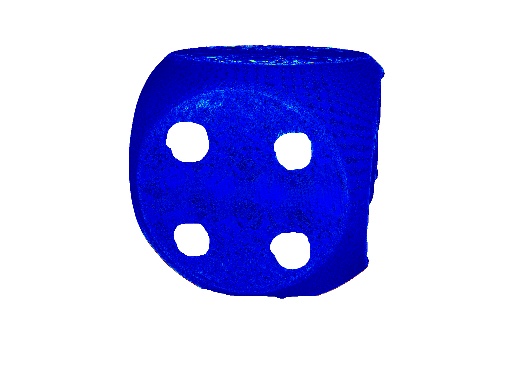} 
\includegraphics[height=0.18\textwidth,trim={1.0cm 0cm 1.0cm  0.0cm},clip]{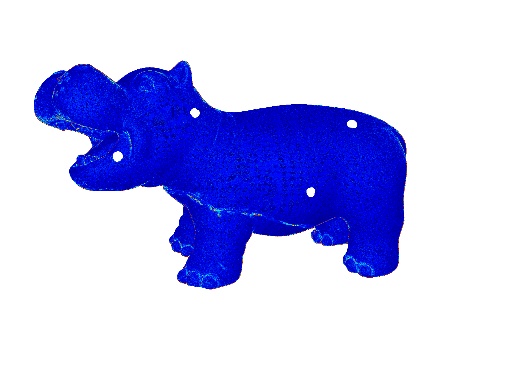} 
\includegraphics[height=0.18\textwidth,trim={1.0cm 0cm 1.0cm  0.0cm},clip]{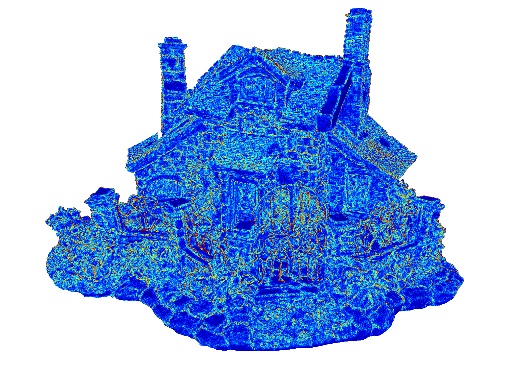} 
\includegraphics[height=0.18\textwidth,trim={5.0cm 0cm 5.0cm  0.0cm},clip]{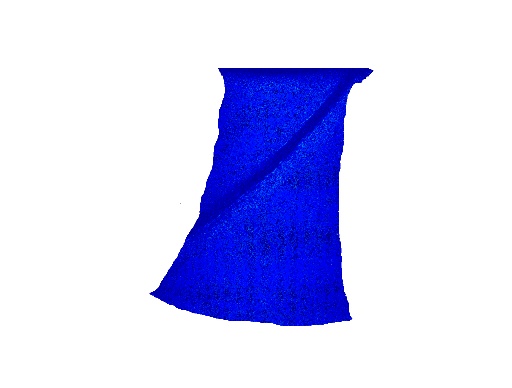} 
\includegraphics[height=0.18\textwidth,trim={2.0cm 0cm 2.0cm  0.0cm},clip]{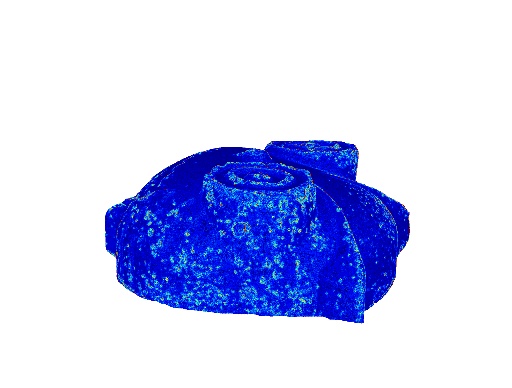} 
\includegraphics[height=0.18\textwidth,trim={2.0cm 0cm 2.0cm  0.0cm},clip]{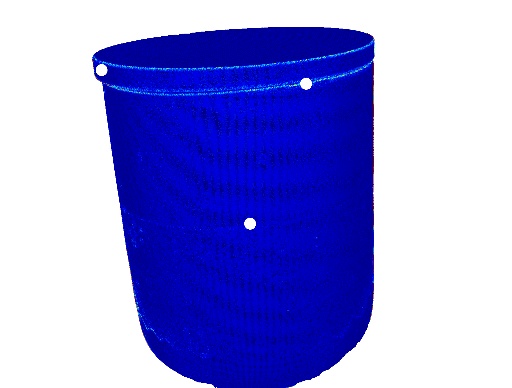} 
\includegraphics[width=0.18\textwidth,trim={2.0cm 0cm 2.0cm  0.0cm},clip]{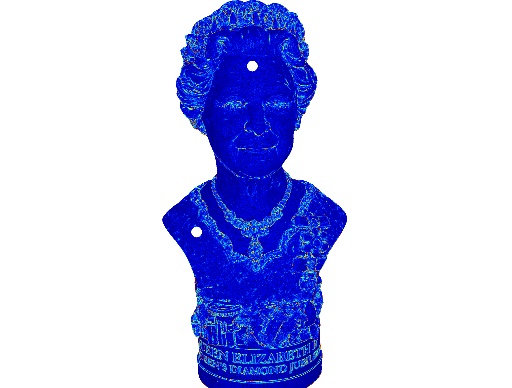} 
\includegraphics[width=0.18\textwidth,trim={2.0cm 0cm 2.0cm  0.0cm},clip]{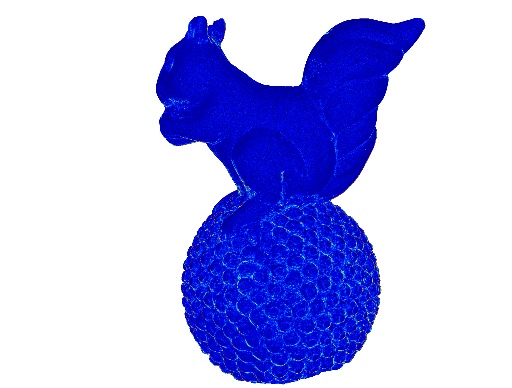} 
\includegraphics[width=0.18\textwidth,trim={2.0cm 0cm 2.0cm  0.0cm},clip]{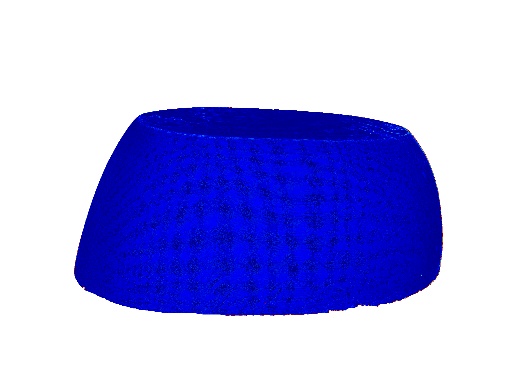} 
\includegraphics[width=0.18\textwidth,trim={2.0cm 0cm 2.0cm  0.0cm},clip]{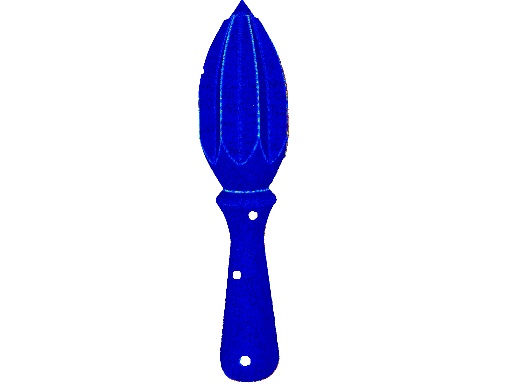} 
\includegraphics[width=0.2\textwidth,trim={1.5cm 1cm 1.0cm  1.0cm},clip]{./images/colorbar_degrees.pdf} \\
Error maps between differentiated normals and ground truth normals. \\
\includegraphics[width=0.18\textwidth,trim={2.0cm 0cm 2.0cm  0.0cm},clip]{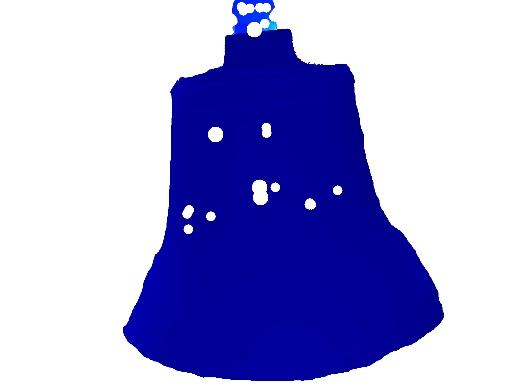} 
\includegraphics[width=0.18\textwidth,trim={2.0cm 0cm 2.0cm  0.0cm},clip]{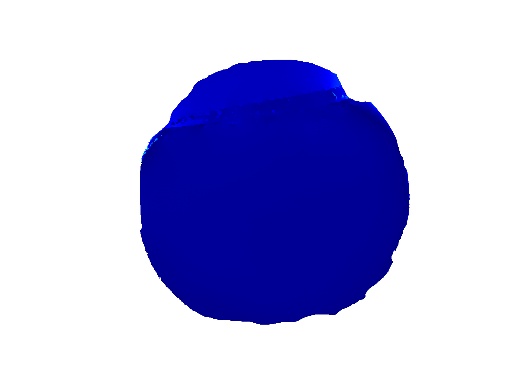} 
\includegraphics[width=0.18\textwidth,trim={2.0cm 0cm 2.0cm  0.0cm},clip]{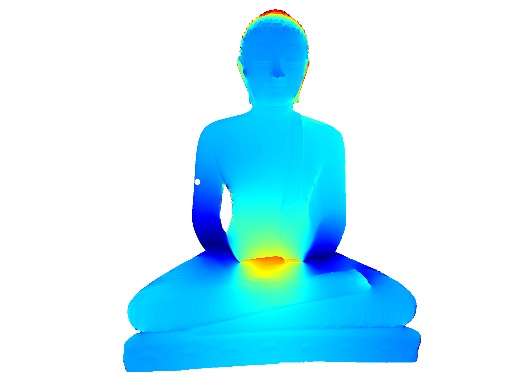} 
\includegraphics[width=0.18\textwidth,trim={2.0cm 0cm 2.0cm  0.0cm},clip]{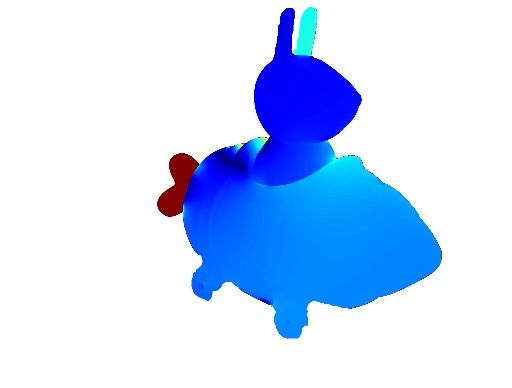} 
\includegraphics[width=0.18\textwidth,trim={2.0cm 0cm 2.0cm  0.0cm},clip]{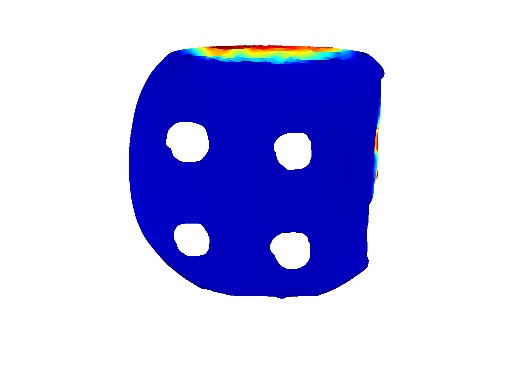} 
\includegraphics[height=0.18\textwidth,trim={1.0cm 0cm 1.0cm  0.0cm},clip]{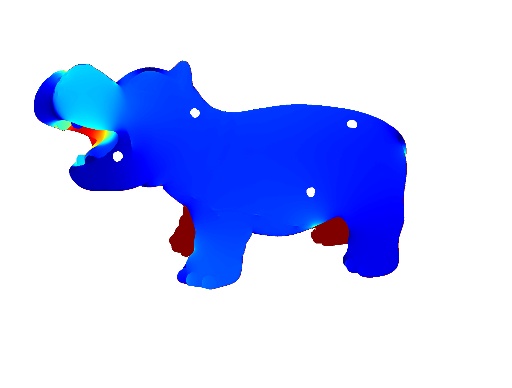} 
\includegraphics[height=0.18\textwidth,trim={1.0cm 0cm 1.0cm  0.0cm},clip]{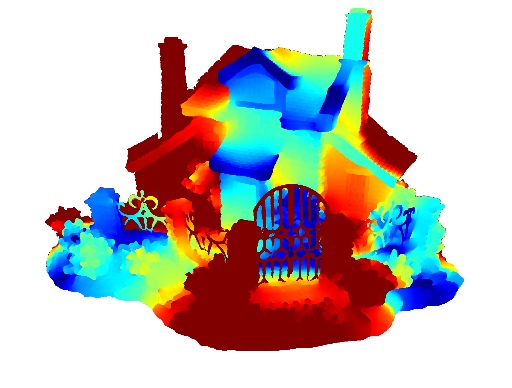} 
\includegraphics[height=0.18\textwidth,trim={5.0cm 0cm 5.0cm  0.0cm},clip]{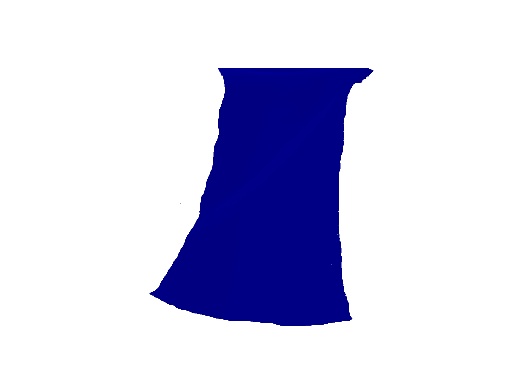} 
\includegraphics[height=0.18\textwidth,trim={2.0cm 0cm 2.0cm  0.0cm},clip]{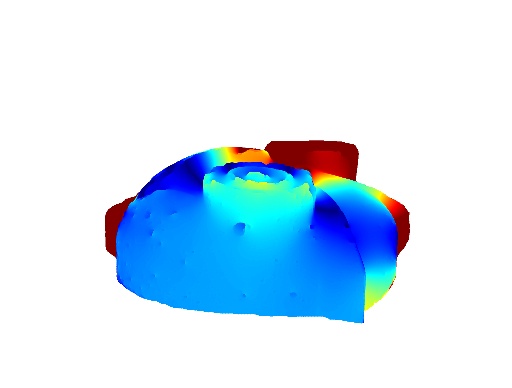} 
\includegraphics[height=0.18\textwidth,trim={2.0cm 0cm 2.0cm  0.0cm},clip]{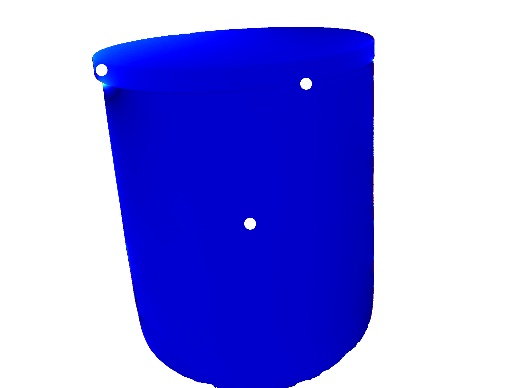} 
\includegraphics[width=0.18\textwidth,trim={2.0cm 0cm 2.0cm  0.0cm},clip]{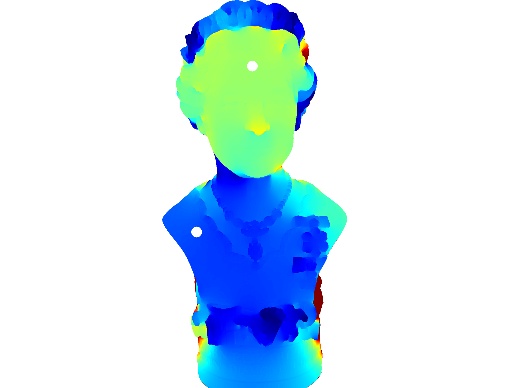} 
\includegraphics[width=0.18\textwidth,trim={2.0cm 0cm 2.0cm  0.0cm},clip]{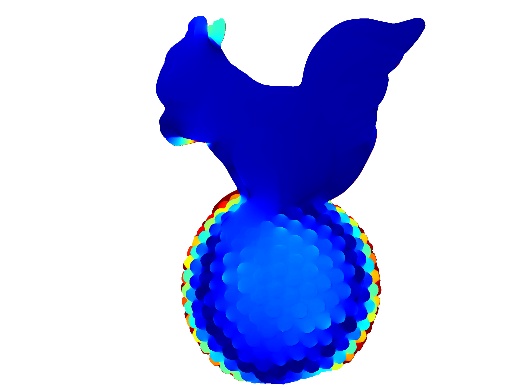} 
\includegraphics[width=0.18\textwidth,trim={2.0cm 0cm 2.0cm  0.0cm},clip]{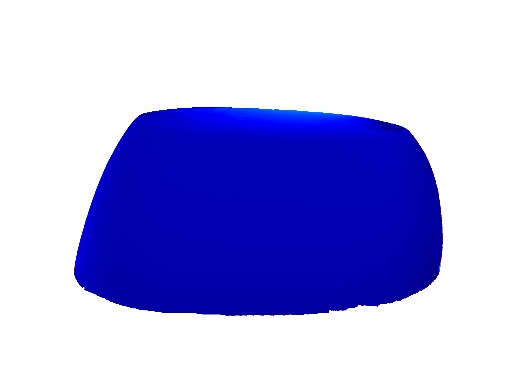} 
\includegraphics[width=0.18\textwidth,trim={2.0cm 0cm 2.0cm  0.0cm},clip]{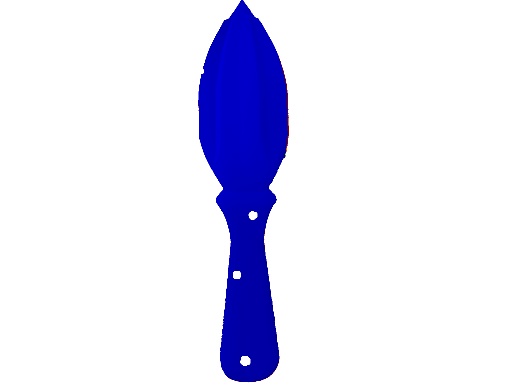} 
\includegraphics[width=0.2\textwidth,trim={0.0cm 1cm 0.0cm  1.0cm},clip]{./images/colorbar_mm.pdf} \\
Error maps between the integration of ground normals and the ground truth depth.
\caption{ Differentiation/Integration errors.} 
 \label{fig:gtnorm}
\end{figure*}

\paragraph{Ground truth discontinuity and non-differentiability.} Over the years, the majority of PS approaches has assumed that the surface can be described as a continuous and differentiable depth map. It has been acknowledged before~\cite{MeccaQLC2016} that this assumption is violated in practice and some care has been taken to include some robustness to it (e.g. $L_1$ loss in~\cite{logothetis2017semi} and Cauchy estimator by~\cite{queau2018led}). However, we believe that the extent of this issue is underestimated and thus we attempt to quantify it here by computing the following two metrics. Firstly, we compute a normal map through numerical differentiation of the ground truth depth map~\cite{queau2015edge} 
and compare it with the ground truth normal map. This is shown in Table 1 of the main submission and illustrated visually here in Figure~\ref{fig:gtnorm}. The mean per-pixel angular error of computed and ground truth normals is on average $3.34^o$ with the maximum value being $9.19^o$ on the complex geometry house object. We emphasise that this effect is completely independent of the actual uncertainty of the `ground truth' meshes and it is solely caused by the projection operation (and so the effect would be identical in synthetic data). Note, the ground truth normal map is computed by rendering (i.e. projection, discretisation and occlusion) of the surface normals into the image plane which is quite different than numerical differentiation of the ground truth depth map. Indeed, in Figure~\ref{fig:gtnorm}, the error is concentrated on boundaries.

In addition, we compute the average per-pixel error between the ground truth depth and the depth obtained by numerical integration (using~\cite{queau2015edge}) of the ground truth normals (i.e. pseudo-depth). The observed error is not-negligible ($1.67$mm on average) and it propagates outwards from occlusion boundaries (as the numerical integration preserves the actual mean depth).

Finally, we note that the two error metrics explained above are likely to be close to the theoretical minimum (for normals and depth respectively) achievable by any approach that is reliant of the differentiable surface assumption. As these error bounds can be non-negligible, we motivate future research that avoids reliance of surface differentiability (e.g. direct depth regression).

%% file: sections/reconstructions.tex
\section{Reconstructions}
\label{sec:rec}

\begin{figure*}[t]
\begin{tabular}{c c c c c c}
~ & L17 & Q18 & I18 & S20 & L20 \\
\begin{sideways} {Bell-3D Shape} \end{sideways} &
\includegraphics[width=0.15\textwidth,trim={10cm 0cm 10cm  0cm},clip]{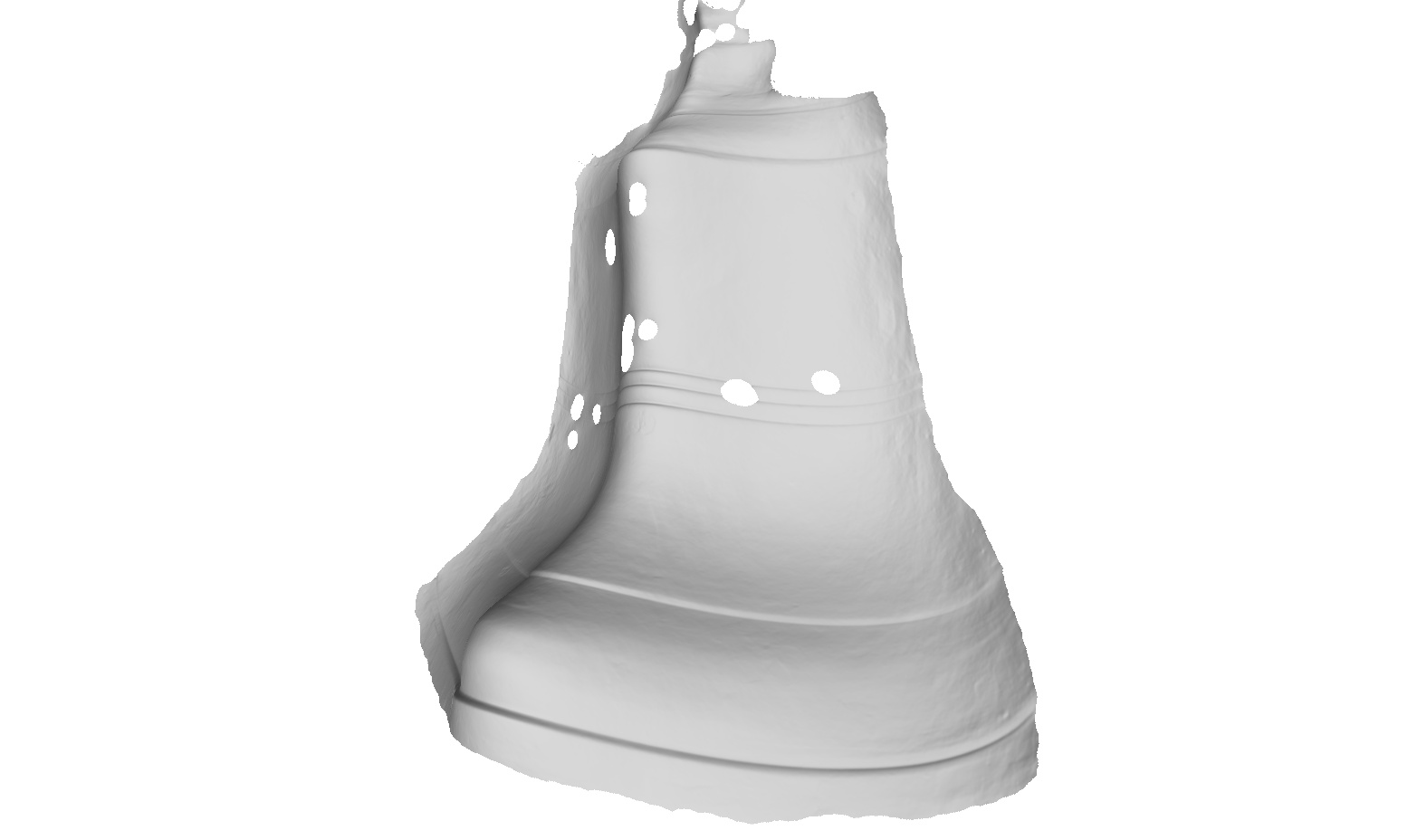} &
\includegraphics[width=0.15\textwidth,trim={10cm 0cm 10cm  0cm},clip]{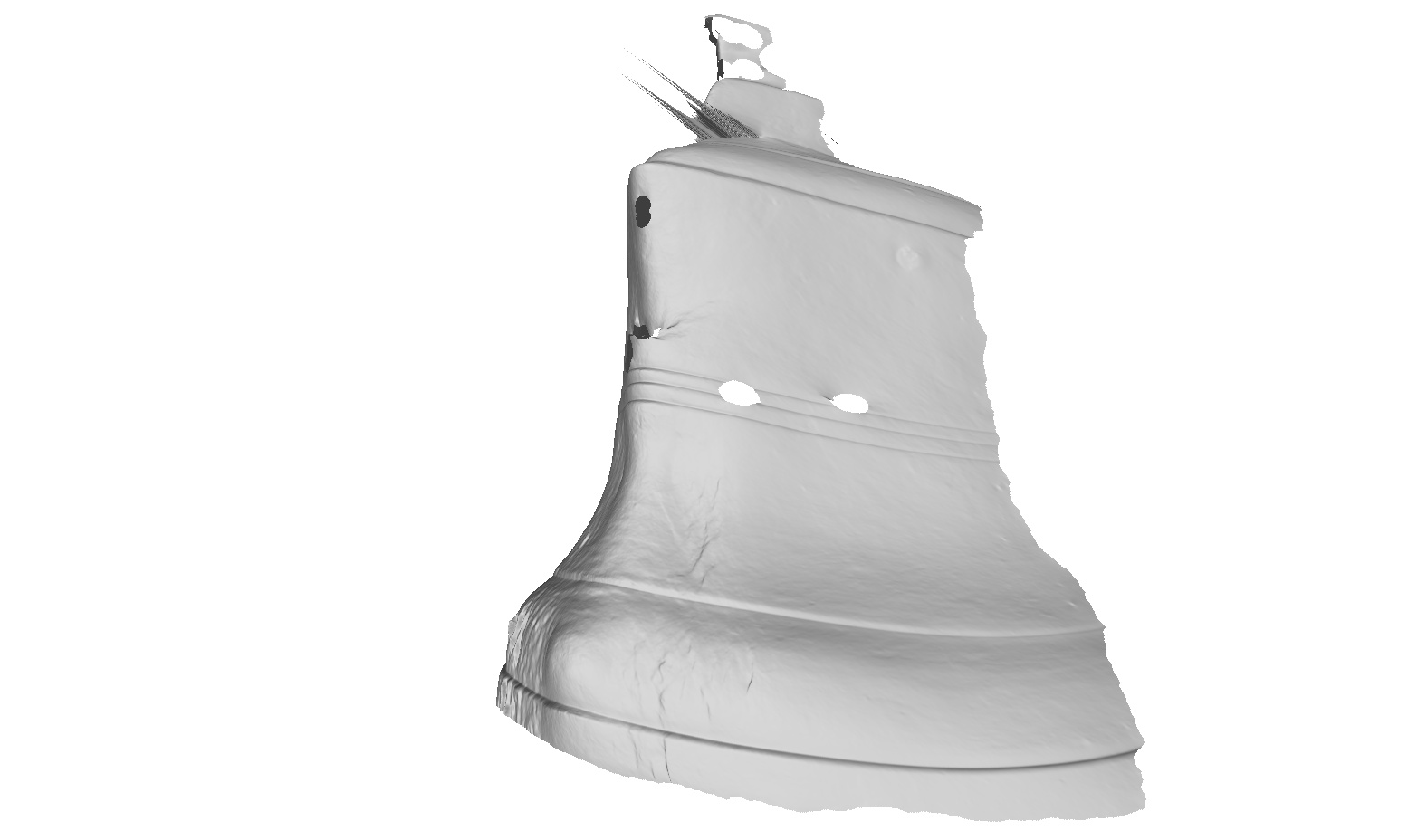} &
\includegraphics[width=0.15\textwidth,trim={10cm 0cm 10cm  0cm},clip]{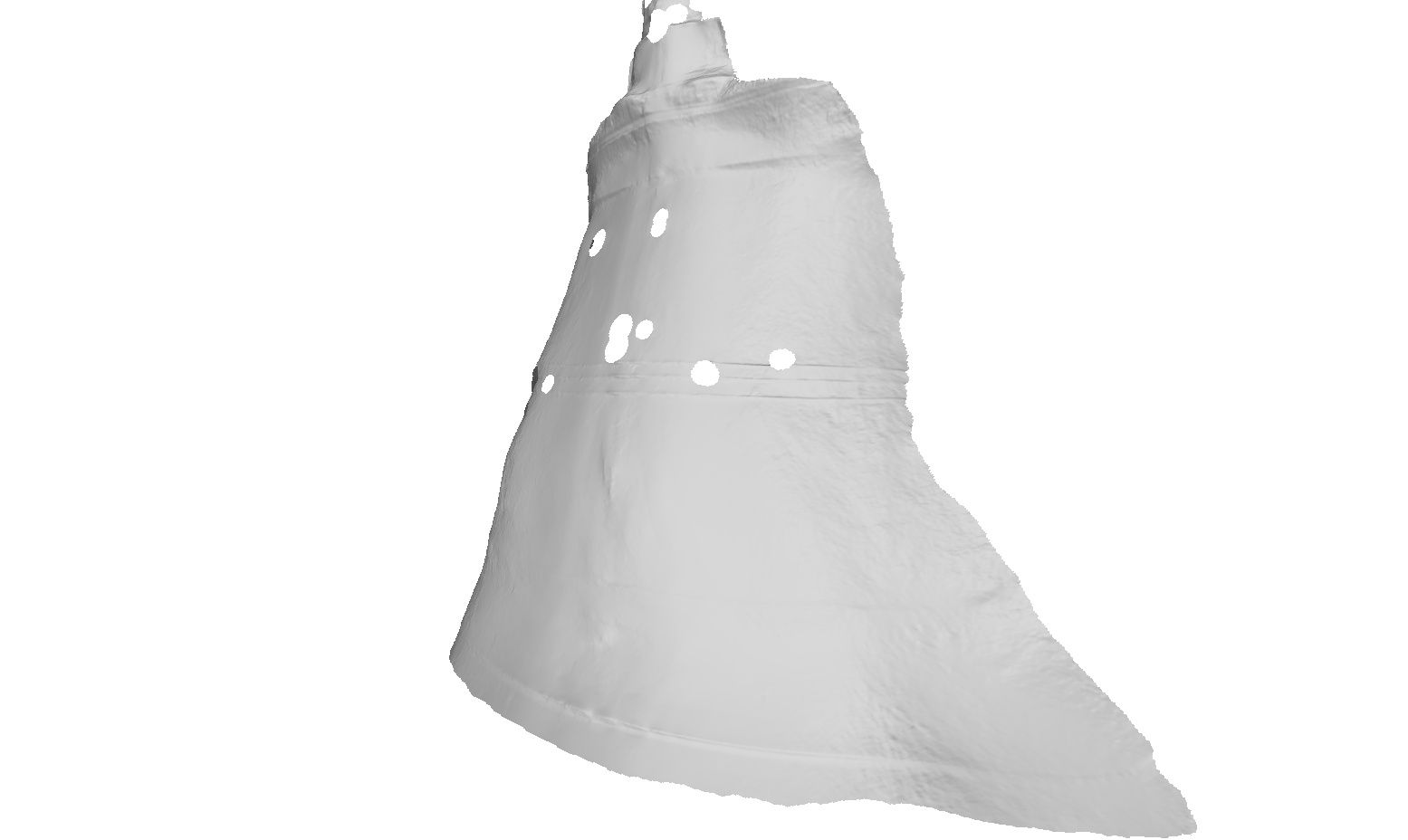} &
\includegraphics[width=0.15\textwidth,trim={10cm 0cm 10cm  0cm},clip]{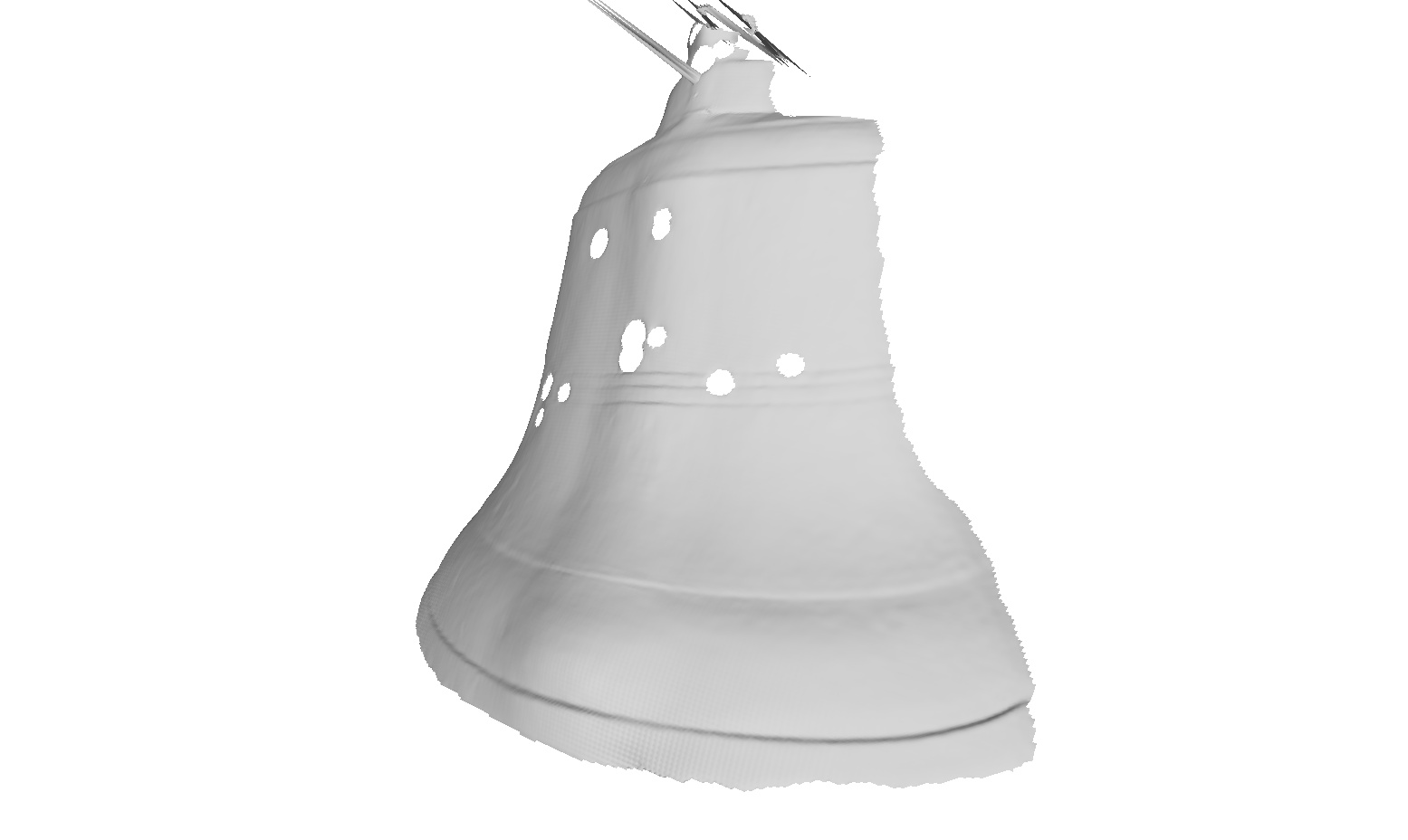} &
\includegraphics[width=0.15\textwidth,trim={10cm 0cm 10cm  0cm},clip]{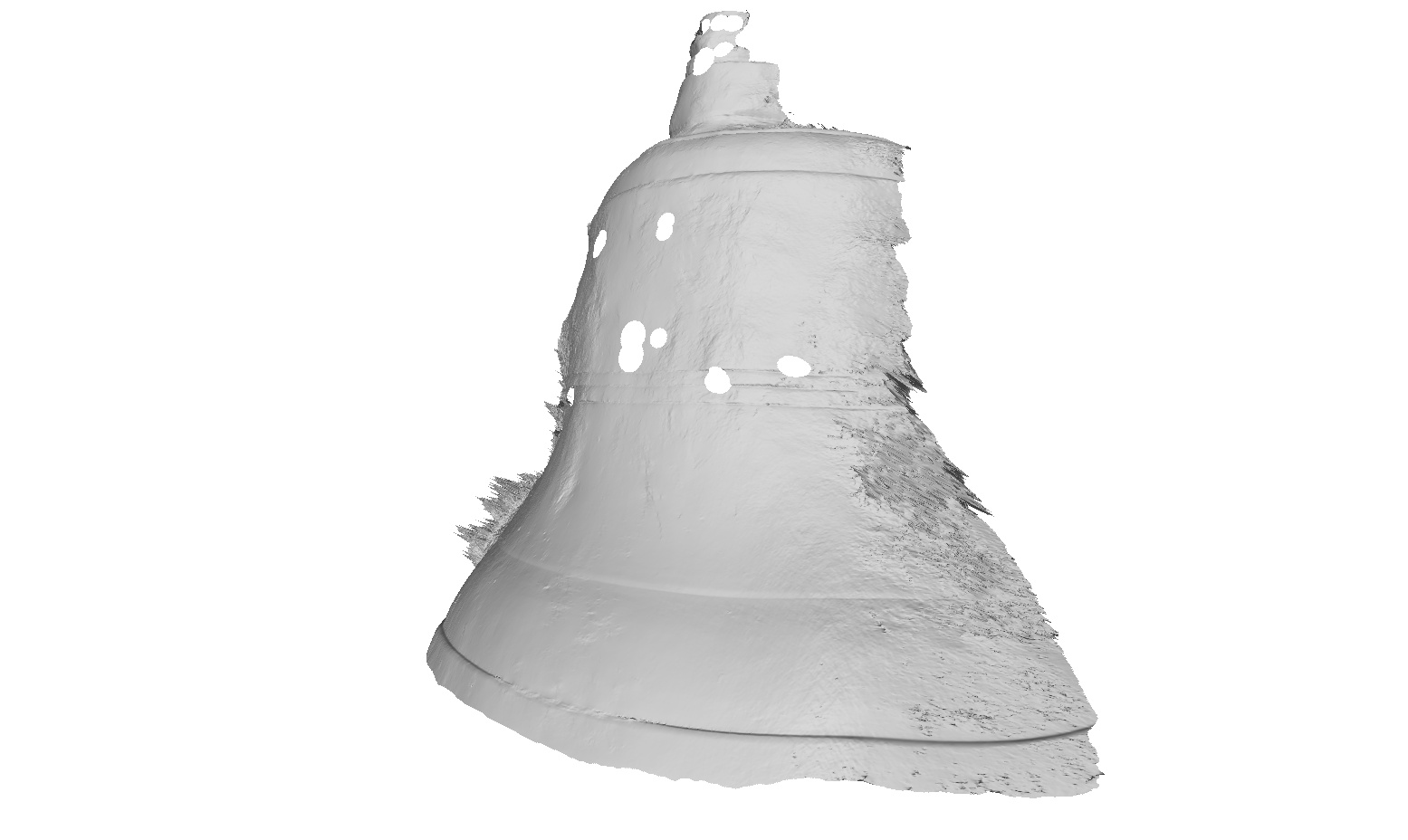} \\
\begin{sideways} {$Z$ Error (mm)} \end{sideways} &
\includegraphics[width=0.15\textwidth,trim={2cm 0cm 2cm  0cm},clip]{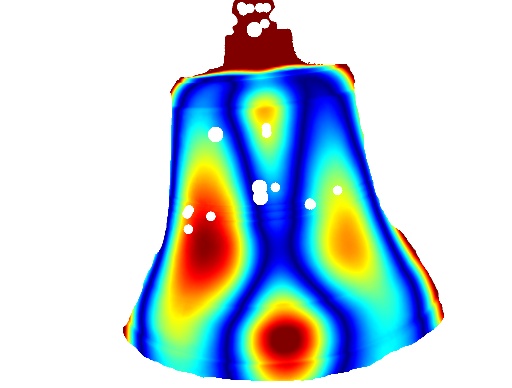} &
\includegraphics[width=0.15\textwidth,trim={2cm 0cm 2cm  0cm},clip]{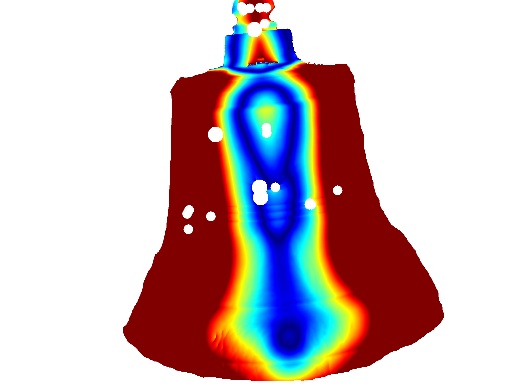} &
\includegraphics[width=0.15\textwidth,trim={2cm 0cm 2cm  0cm},clip]{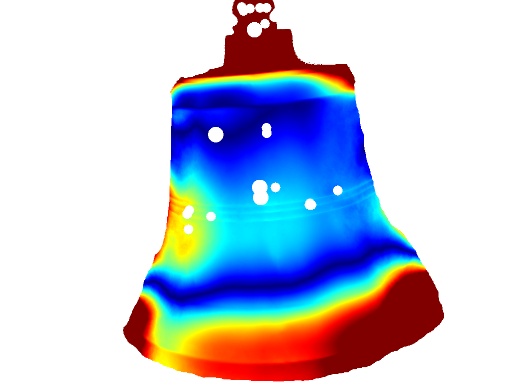} &
\includegraphics[width=0.15\textwidth,trim={2cm 0cm 2cm  0cm},clip]{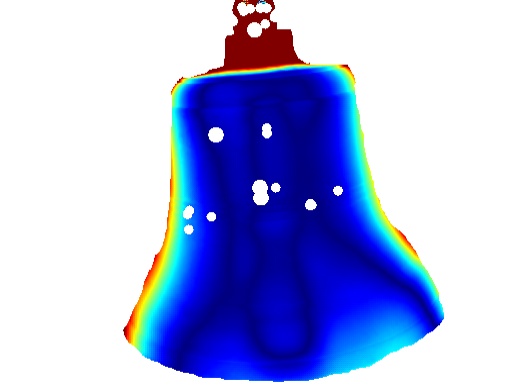} &
\includegraphics[width=0.15\textwidth,trim={2cm 0cm 2cm  0cm},clip]{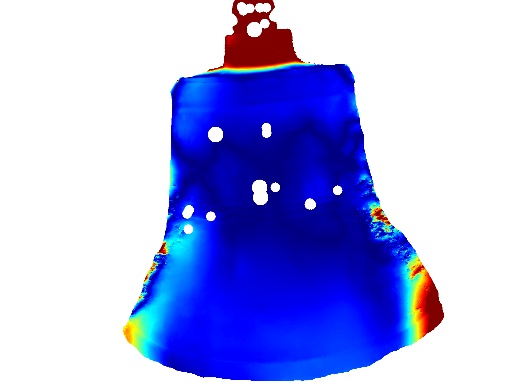} \\
%
%
\begin{sideways} {Ball-3D Shape} \end{sideways} &
\includegraphics[width=0.15\textwidth,trim={12cm 0cm 10cm  0cm},clip]{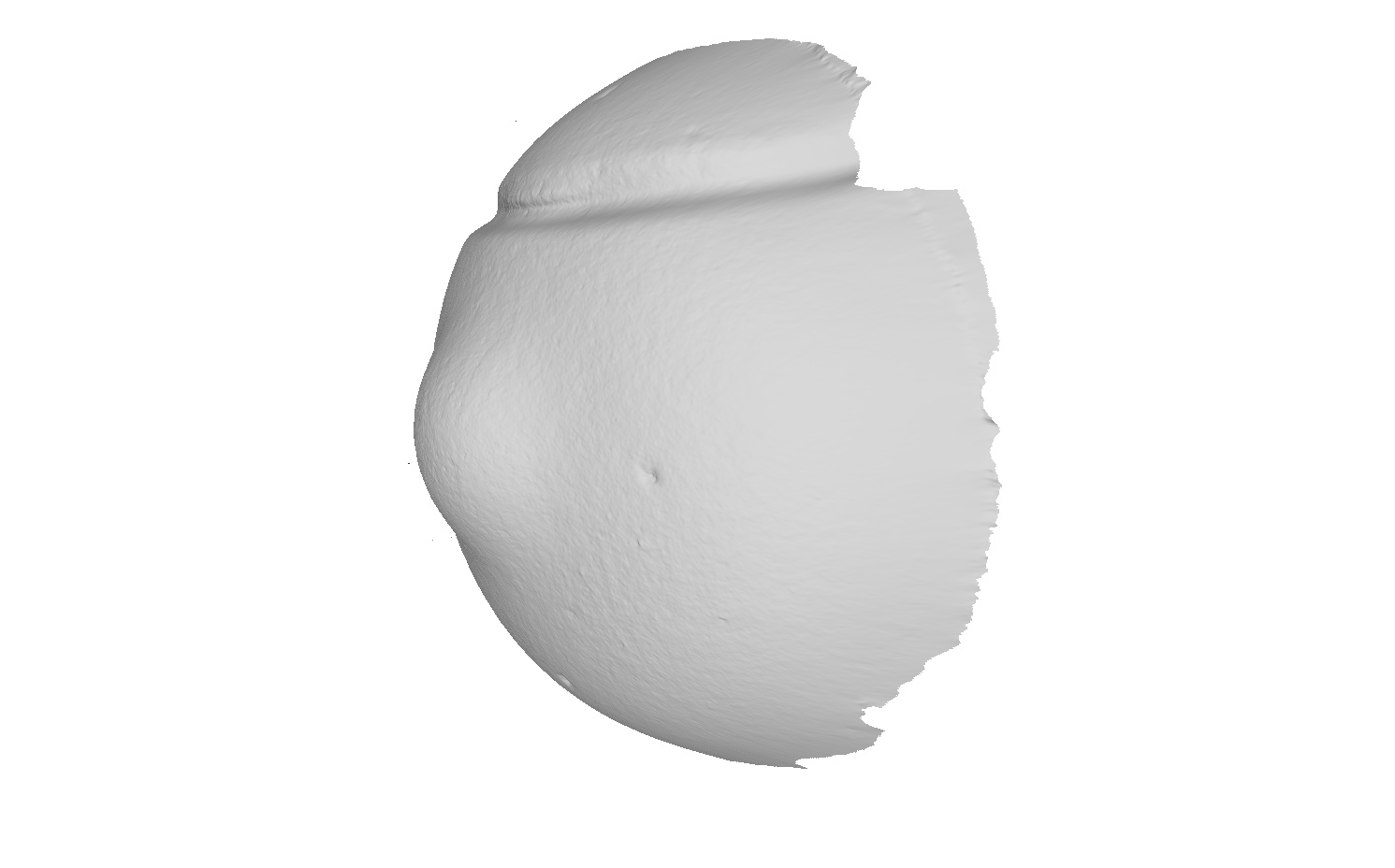} &
\includegraphics[width=0.15\textwidth,trim={12cm 0cm 10cm  0cm},clip]{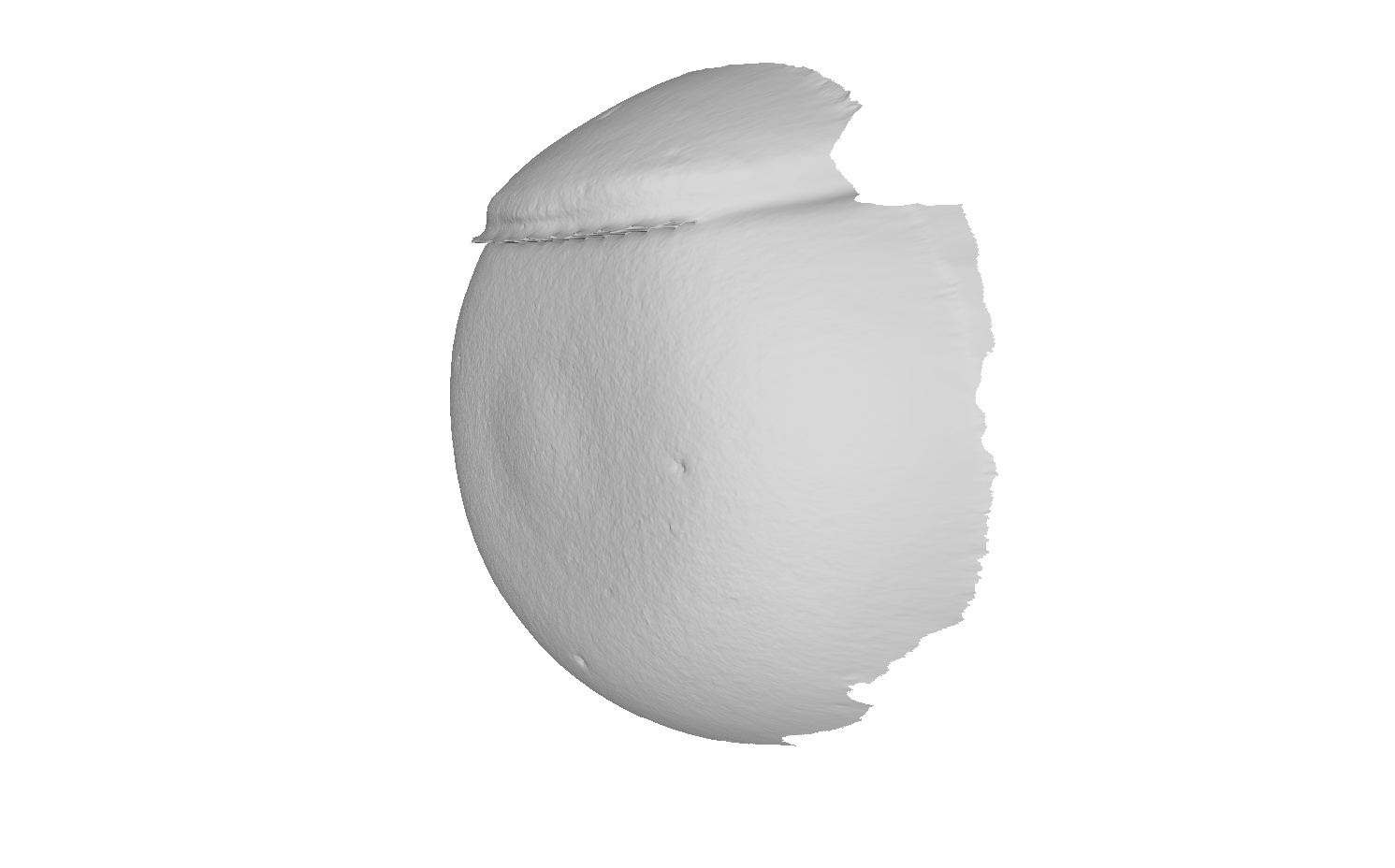} &
\includegraphics[width=0.15\textwidth,trim={12cm 0cm 10cm  0cm},clip]{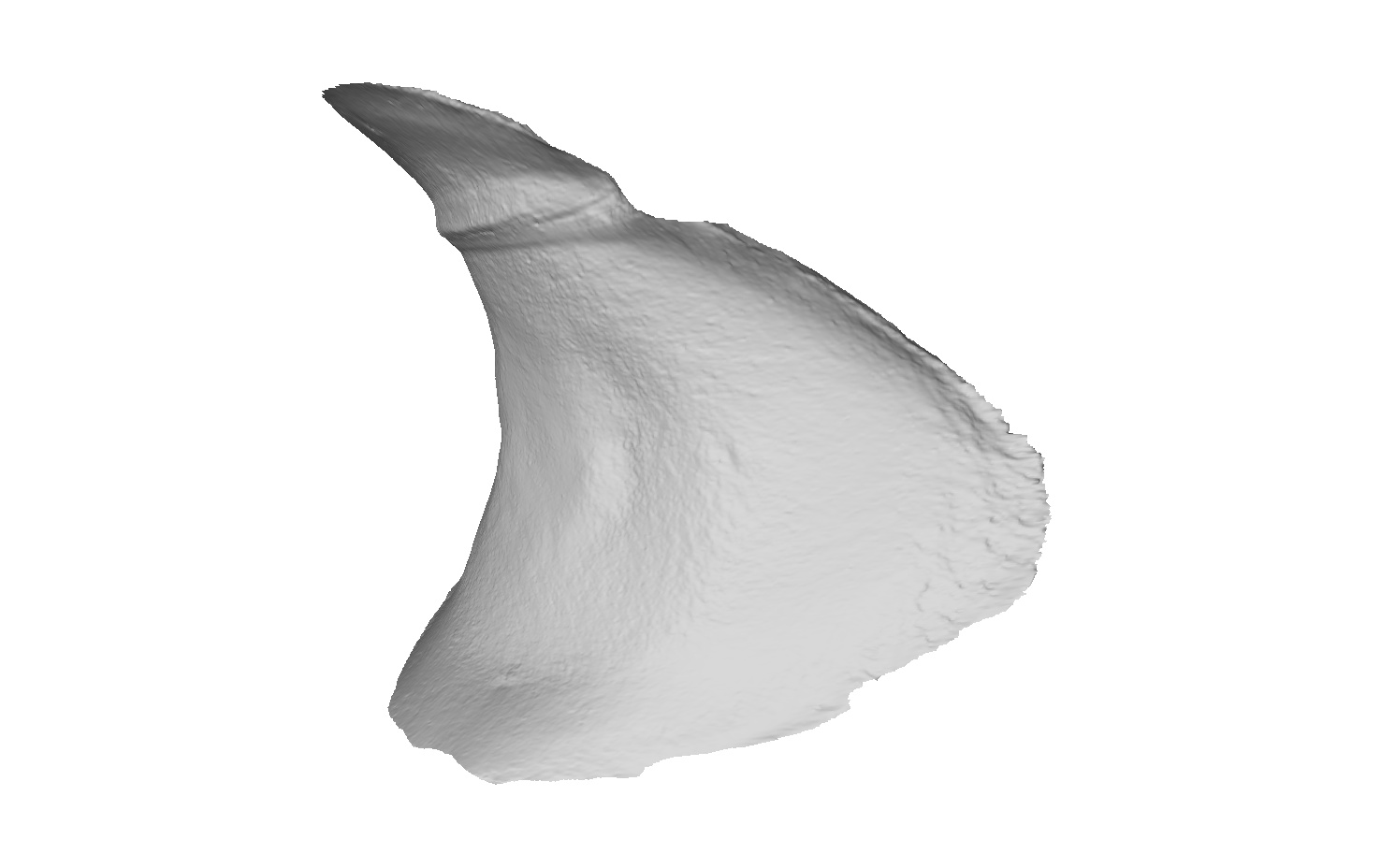} &
\includegraphics[width=0.15\textwidth,trim={12cm 0cm 10cm  0cm},clip]{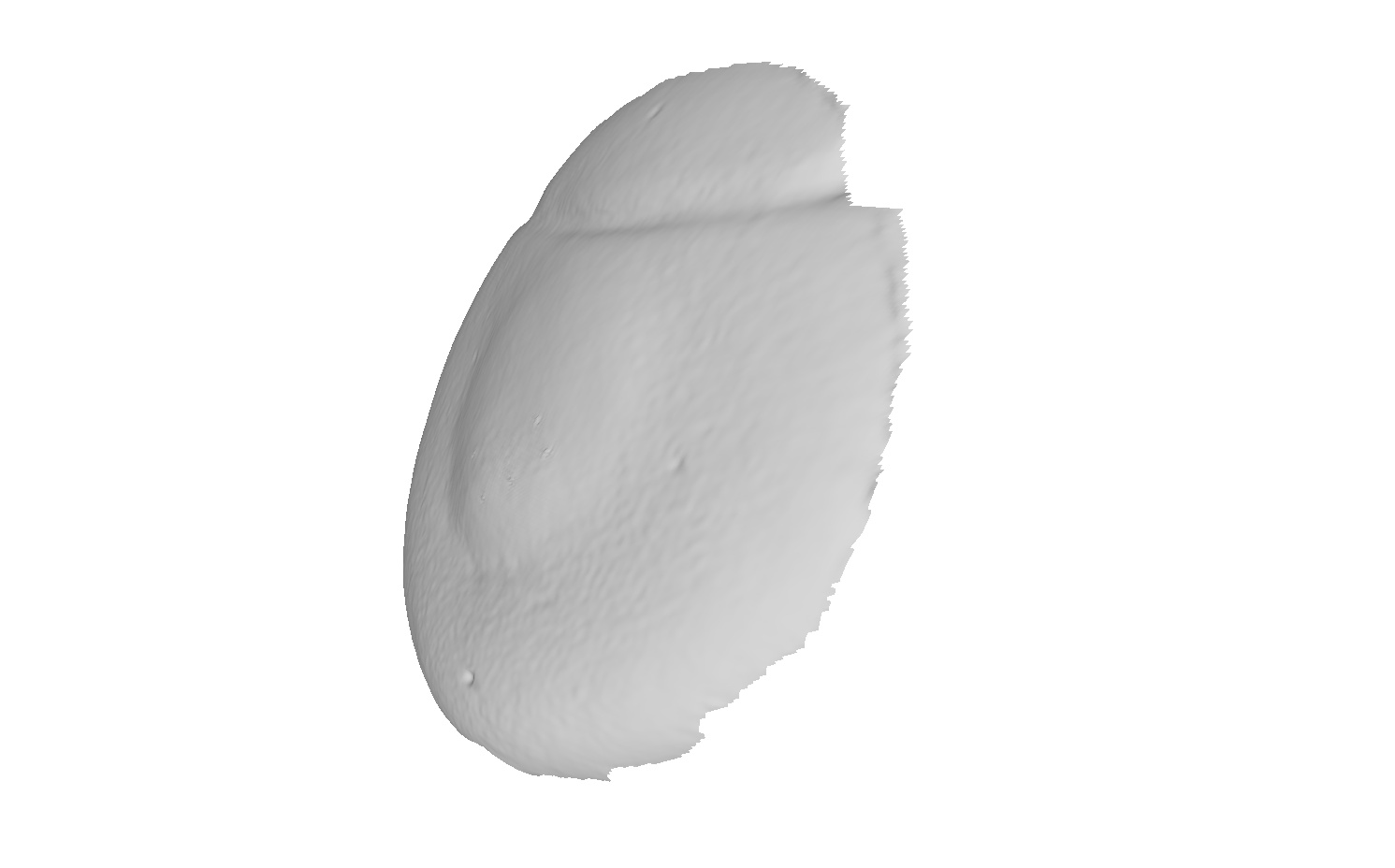} &
\includegraphics[width=0.15\textwidth,trim={12cm 0cm 10cm  0cm},clip]{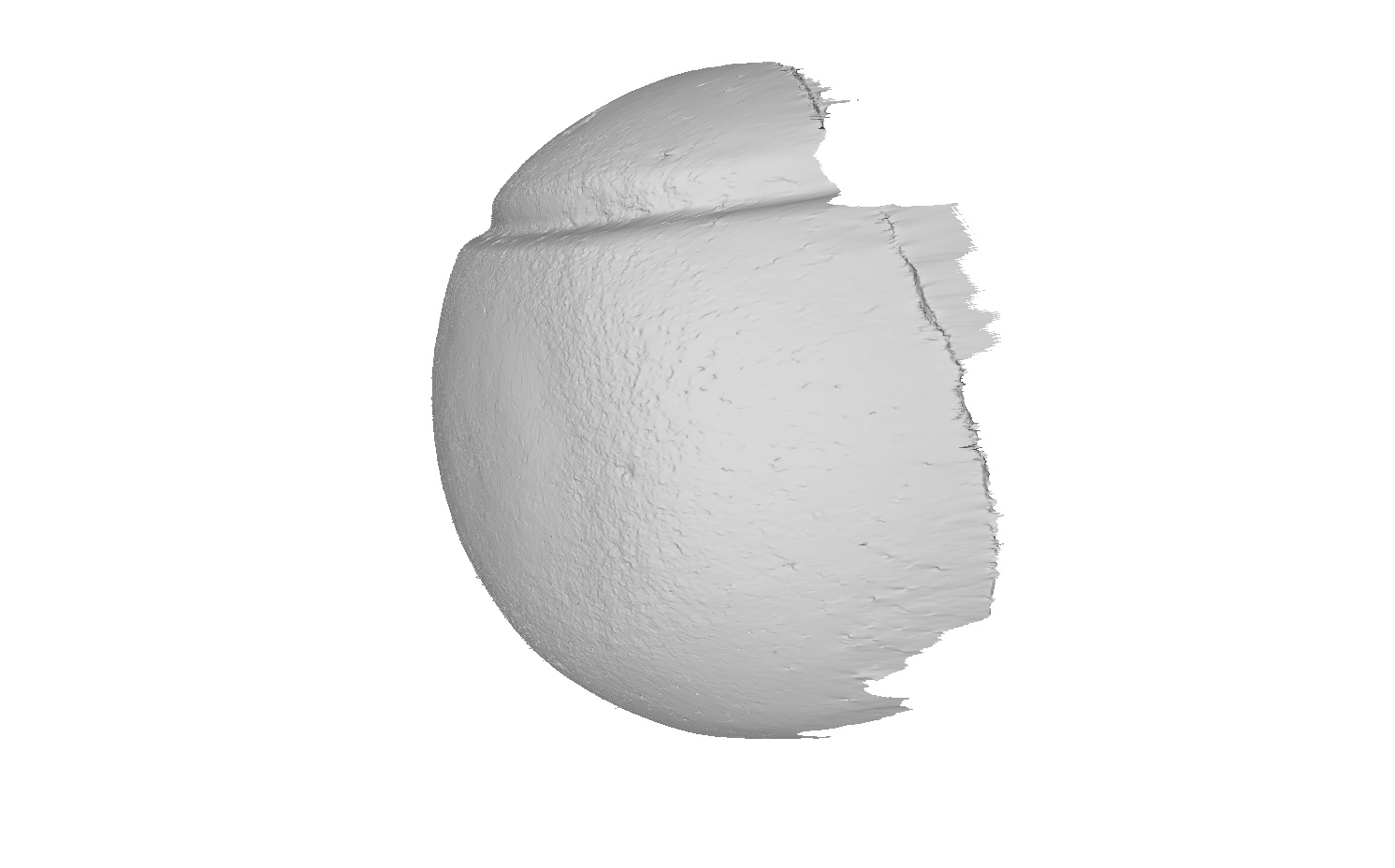} \\
\begin{sideways} {$Z$ Error (mm)} \end{sideways} &
\includegraphics[height=0.15\textwidth,trim={4cm 0cm 4cm  0cm},clip]{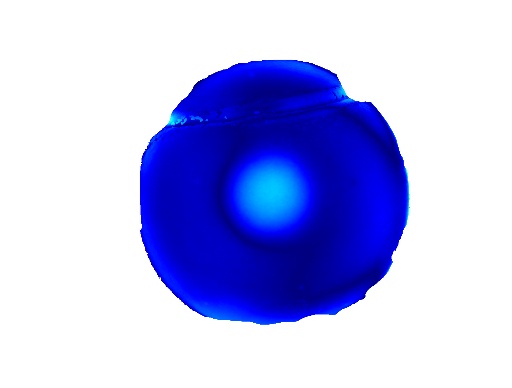} &
\includegraphics[height=0.15\textwidth,trim={4cm 0cm 4cm  0cm},clip]{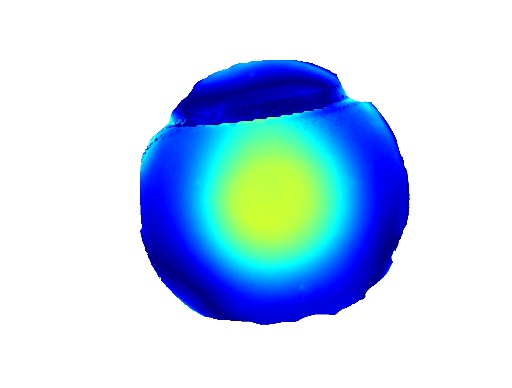} &
\includegraphics[height=0.15\textwidth,trim={4cm 0cm 4cm  0cm},clip]{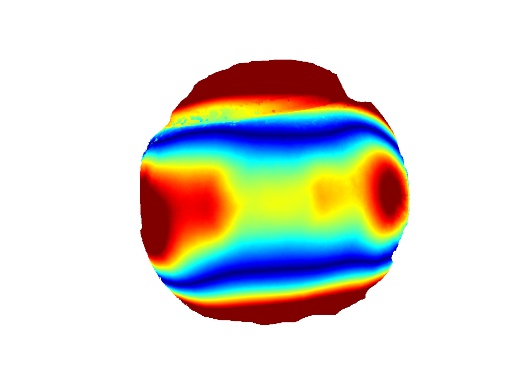} &
\includegraphics[height=0.15\textwidth,trim={4cm 0cm 4cm  0cm},clip]{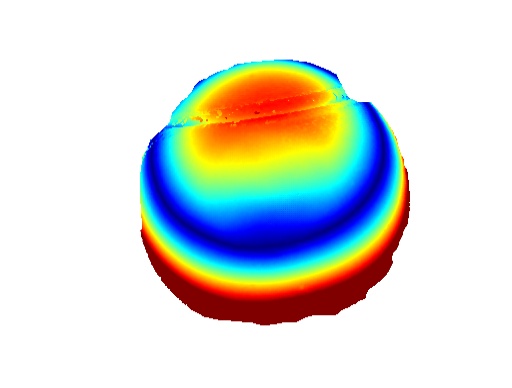} &
\includegraphics[height=0.15\textwidth,trim={4cm 0cm 4cm  0cm},clip]{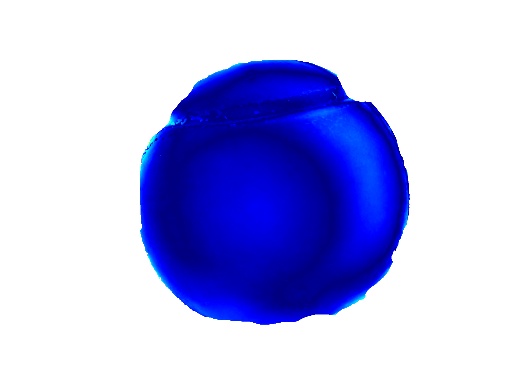} \\
%
\begin{sideways} {Buddha-3D Shape} \end{sideways} &
\includegraphics[width=0.15\textwidth,trim={10cm 0cm 10cm  0cm},clip]{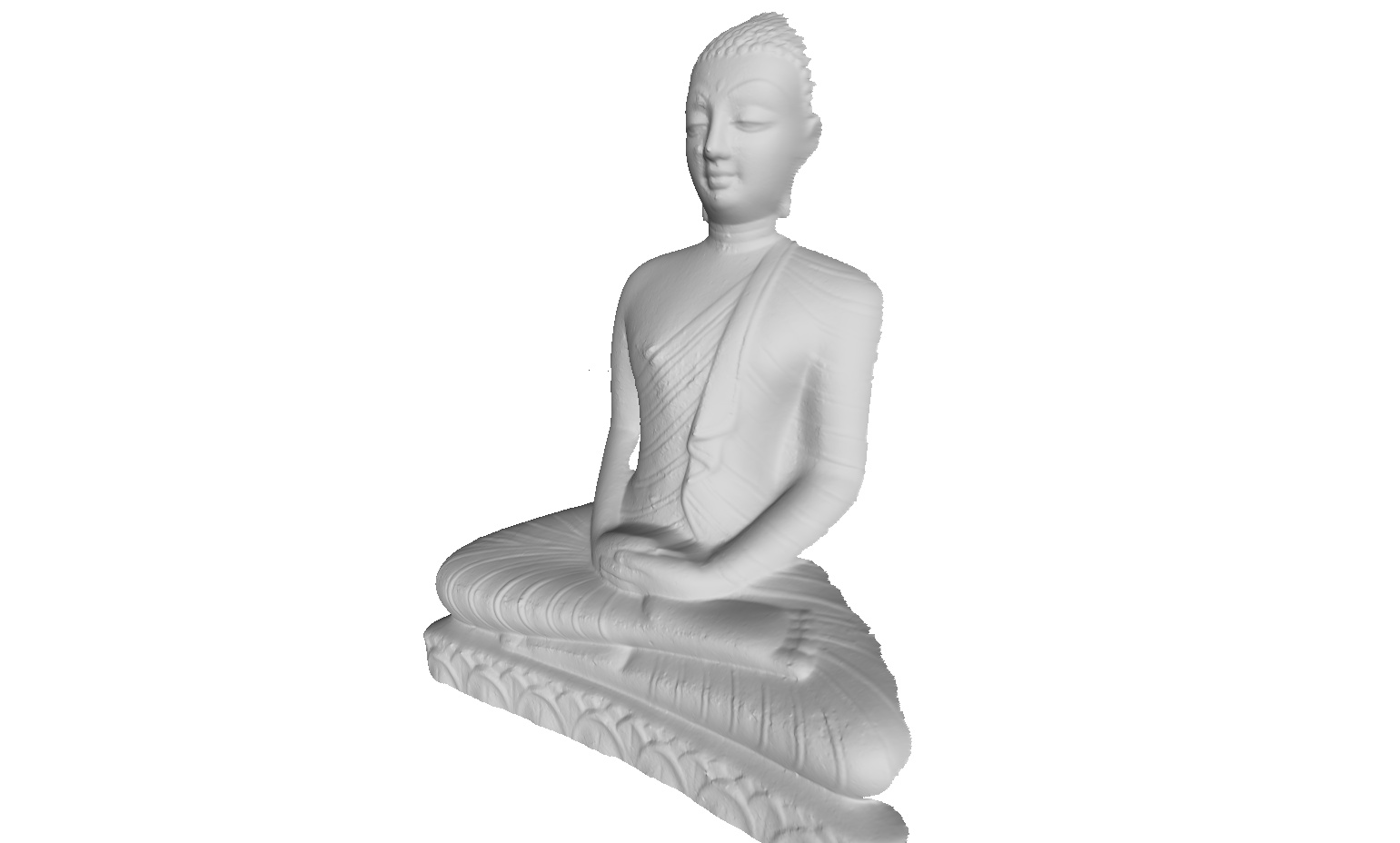} &
\includegraphics[width=0.15\textwidth,trim={10cm 0cm 10cm  0cm},clip]{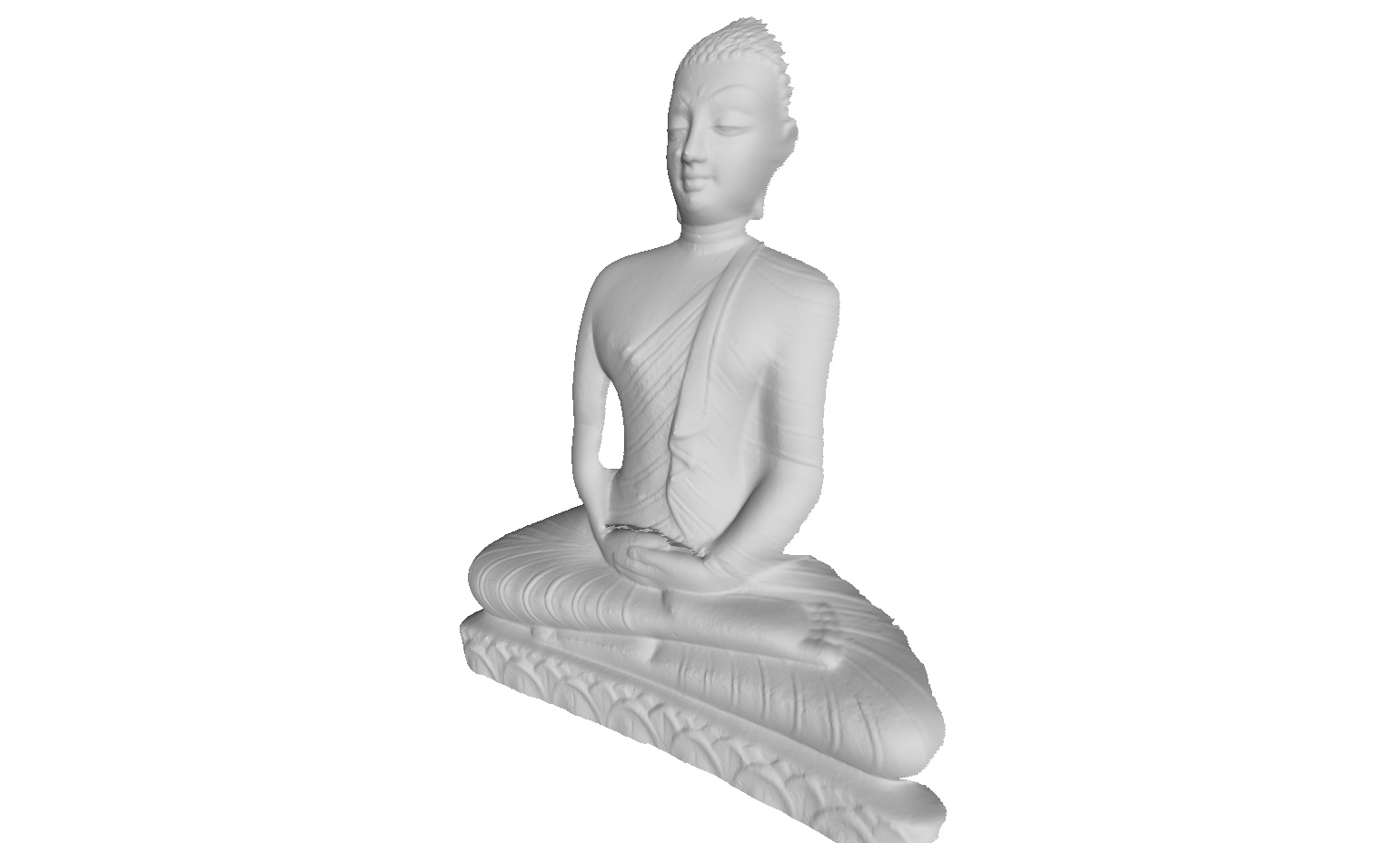} &
\includegraphics[width=0.15\textwidth,trim={10cm 0cm 10cm  0cm},clip]{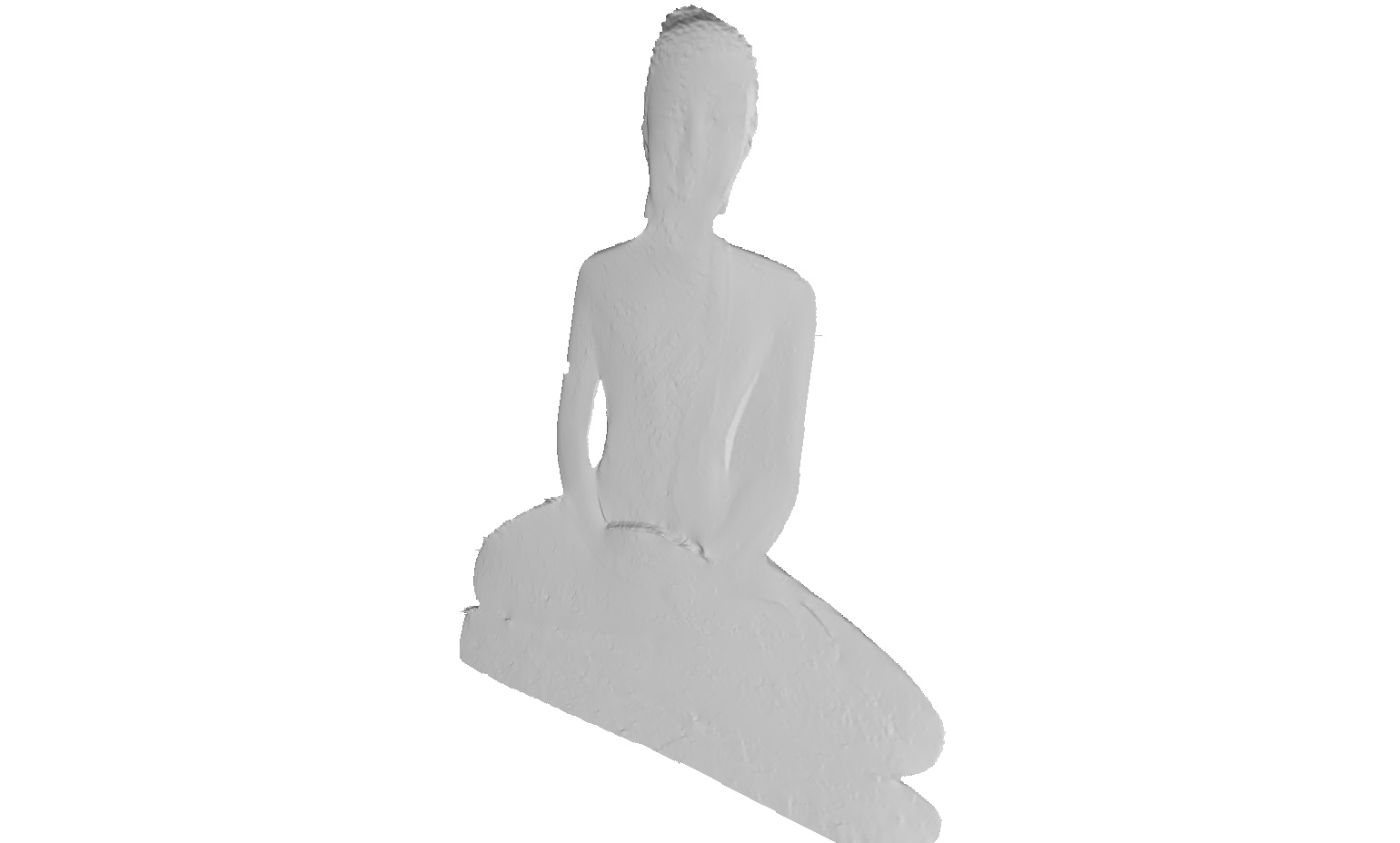} &
\includegraphics[width=0.15\textwidth,trim={10cm 0cm 10cm  0cm},clip]{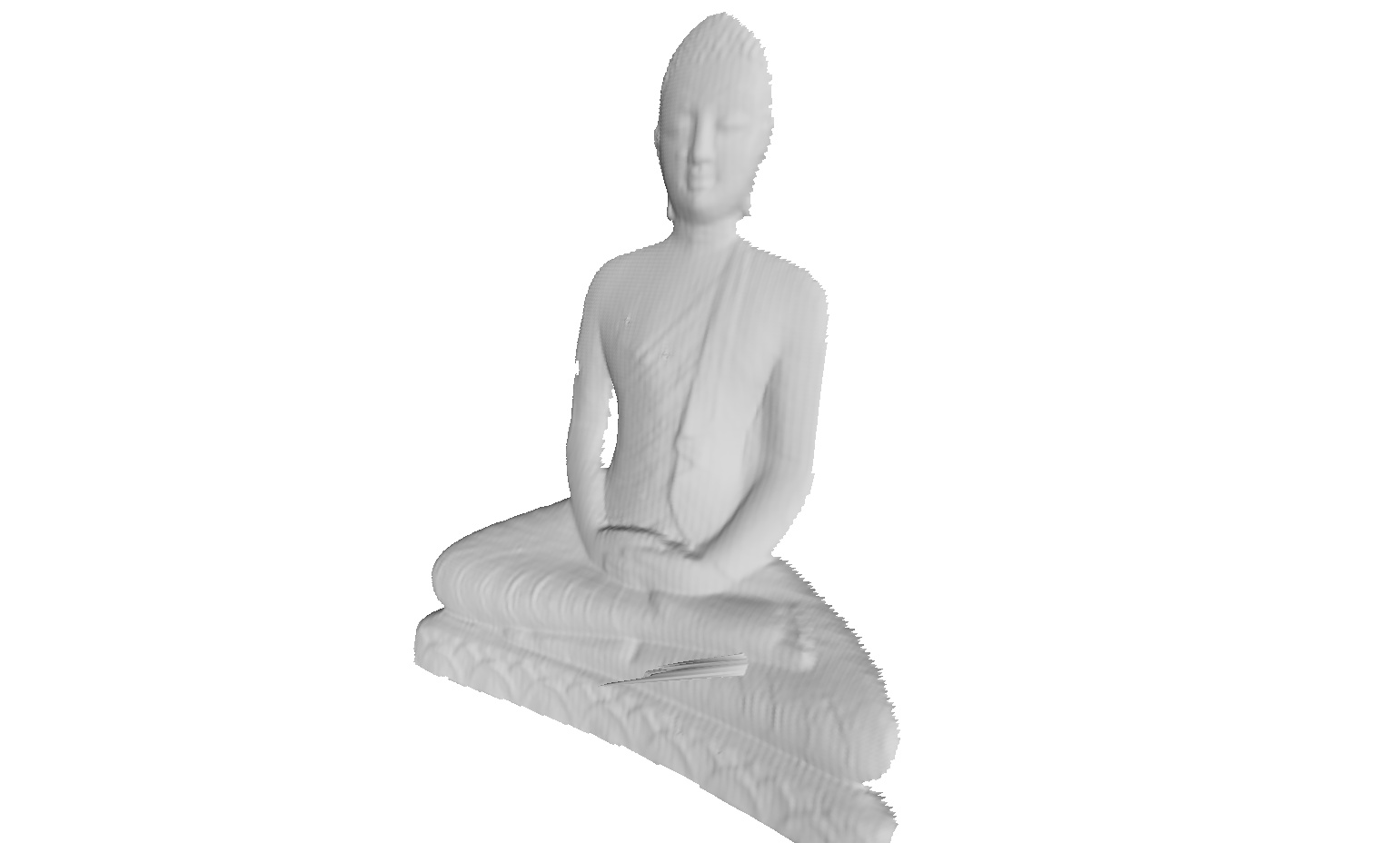} &
\includegraphics[width=0.15\textwidth,trim={10cm 0cm 10cm  0cm},clip]{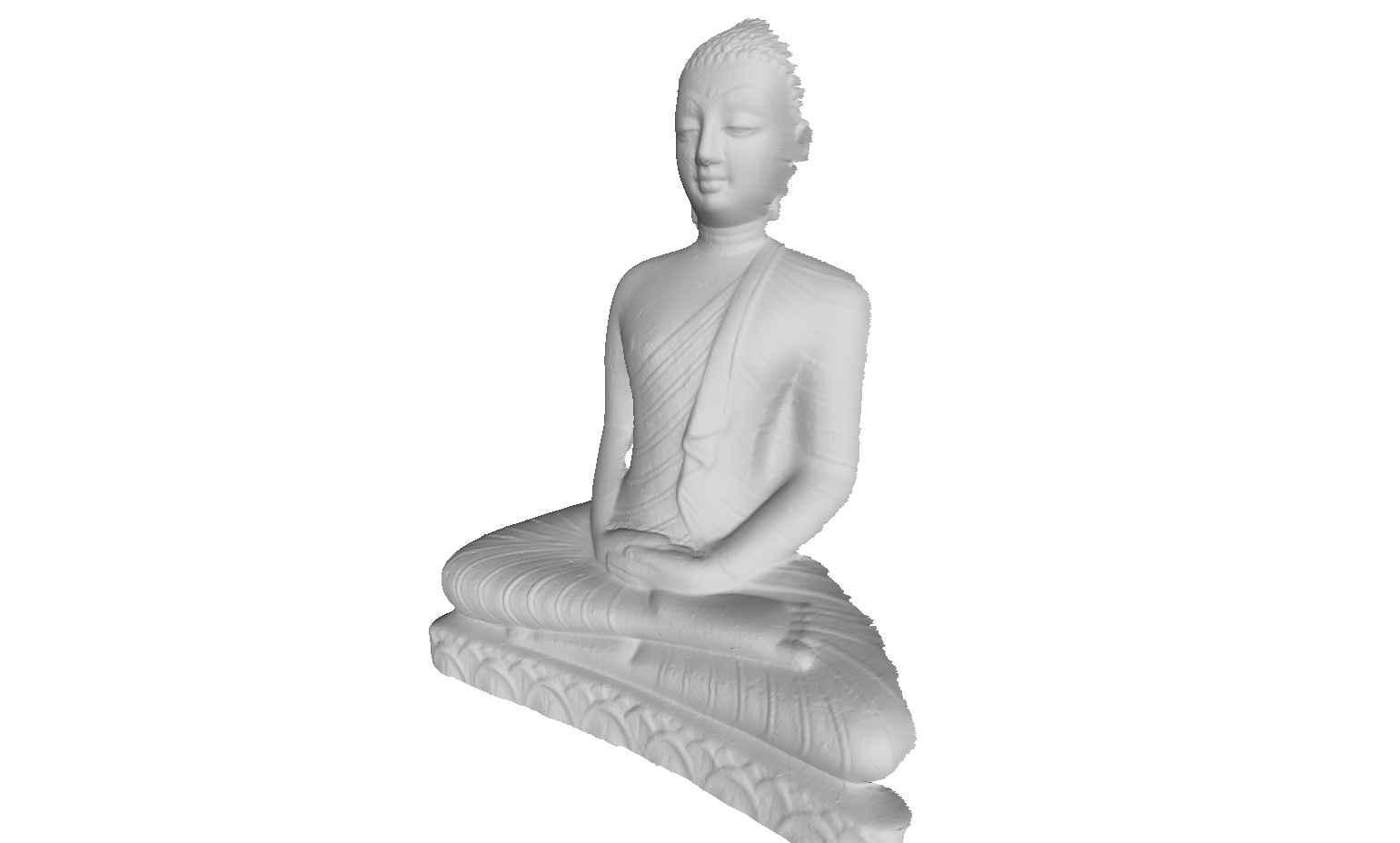} \\

\begin{sideways} {$Z$ Error (mm)} \end{sideways} &
\includegraphics[width=0.15\textwidth,trim={2cm 0cm 2cm  0cm},clip]{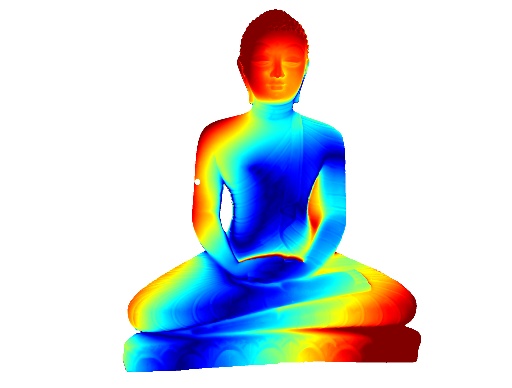} &
\includegraphics[width=0.15\textwidth,trim={2cm 0cm 2cm  0cm},clip]{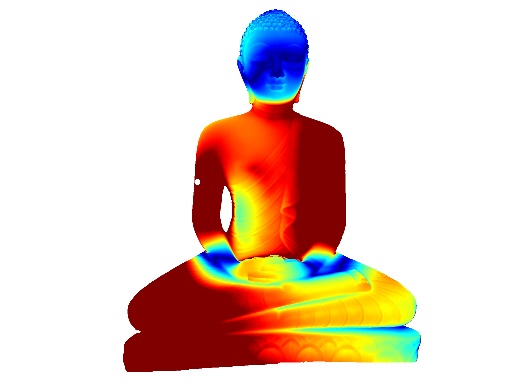} &
\includegraphics[width=0.15\textwidth,trim={2cm 0cm 2cm  0cm},clip]{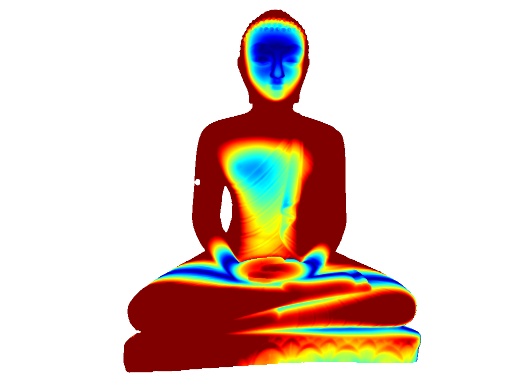} &
\includegraphics[width=0.15\textwidth,trim={2cm 0cm 2cm  0cm},clip]{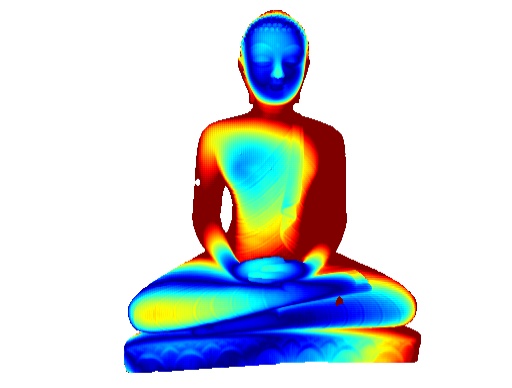} &
\includegraphics[width=0.15\textwidth,trim={2cm 0cm 2cm  0cm},clip]{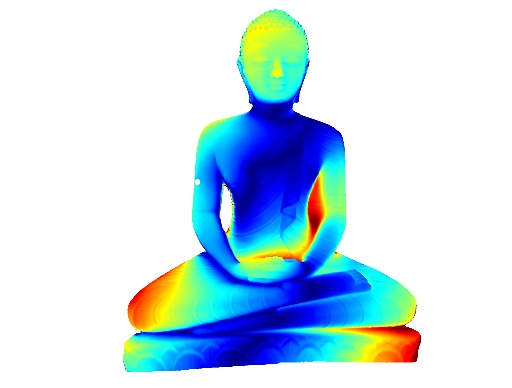} \\
%

\begin{sideways} {Bunny-3D Shape} \end{sideways} &
\includegraphics[width=0.15\textwidth,trim={10cm 0cm 10cm  0cm},clip]{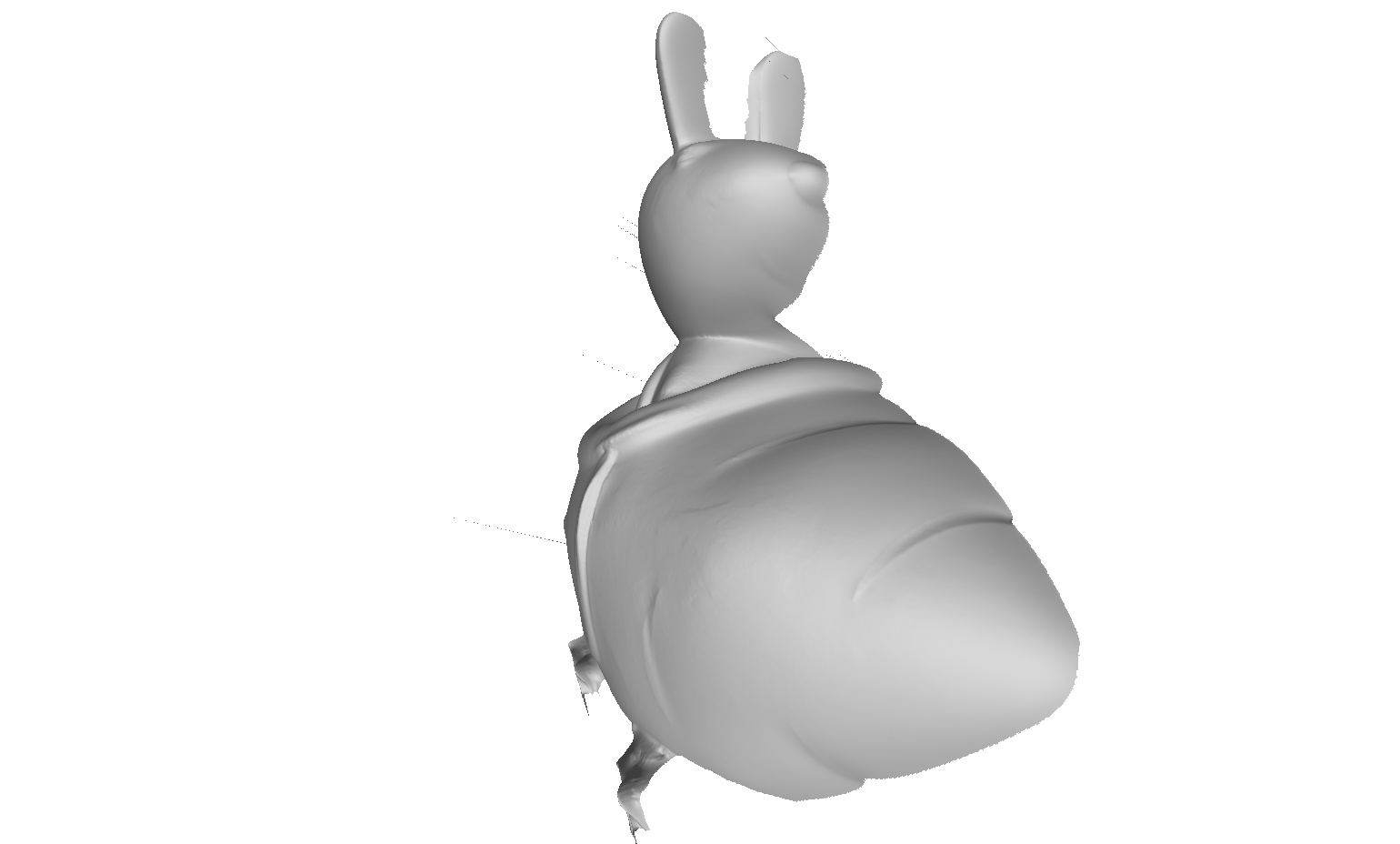} &
\includegraphics[width=0.15\textwidth,trim={10cm 0cm 10cm  0cm},clip]{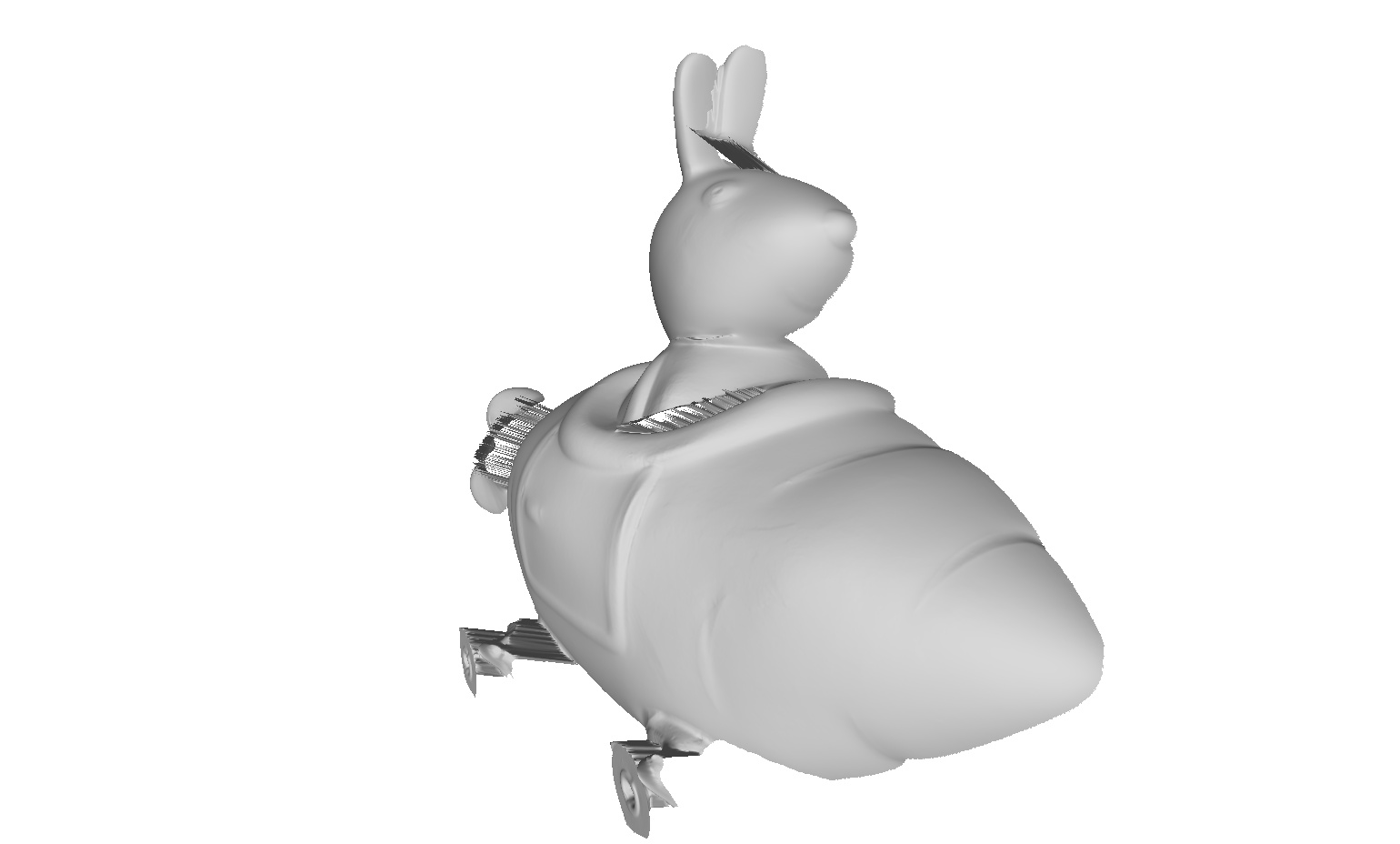} &
\includegraphics[width=0.15\textwidth,trim={10cm 0cm 10cm  0cm},clip]{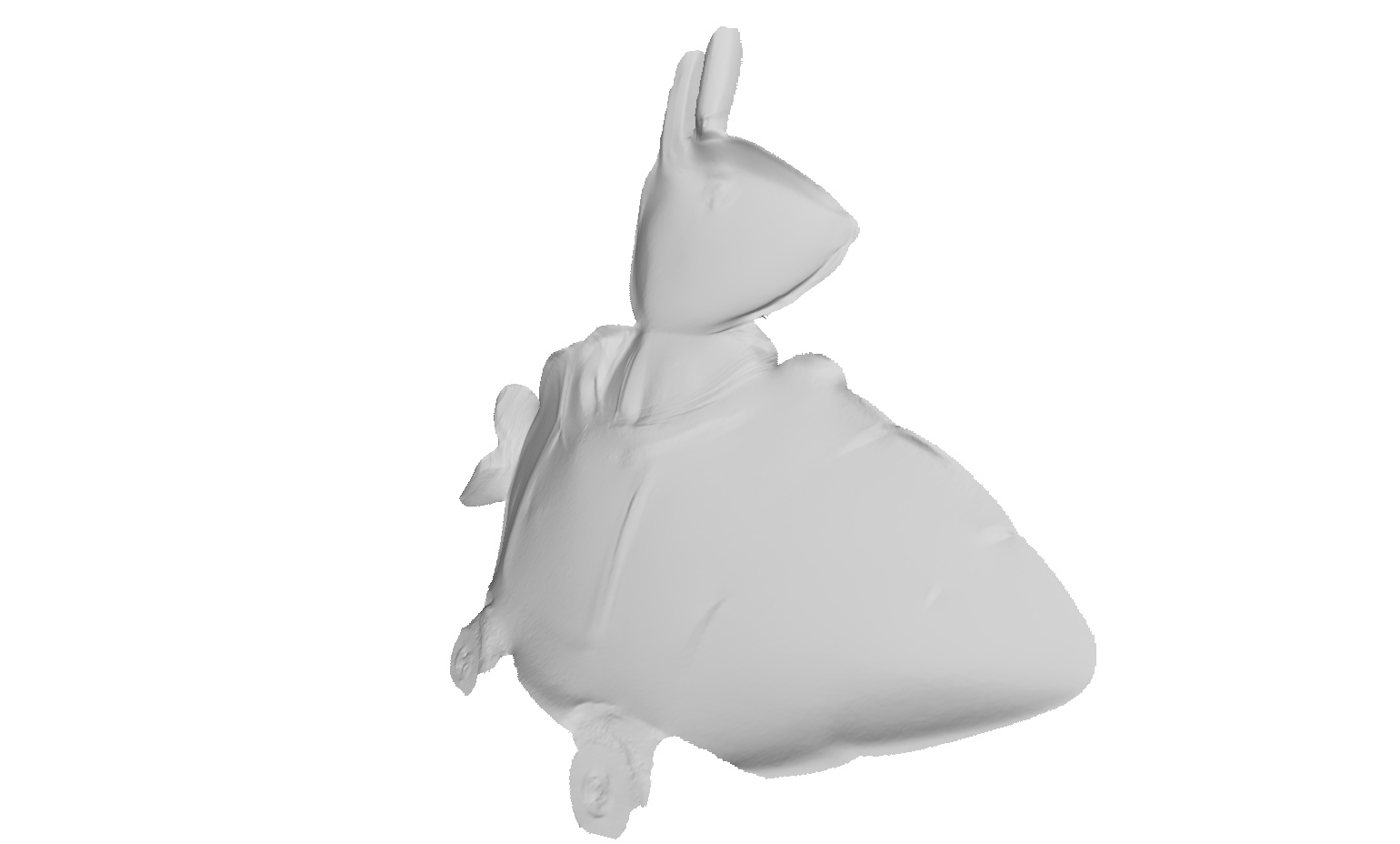} &
\includegraphics[width=0.15\textwidth,trim={10cm 0cm 10cm  0cm},clip]{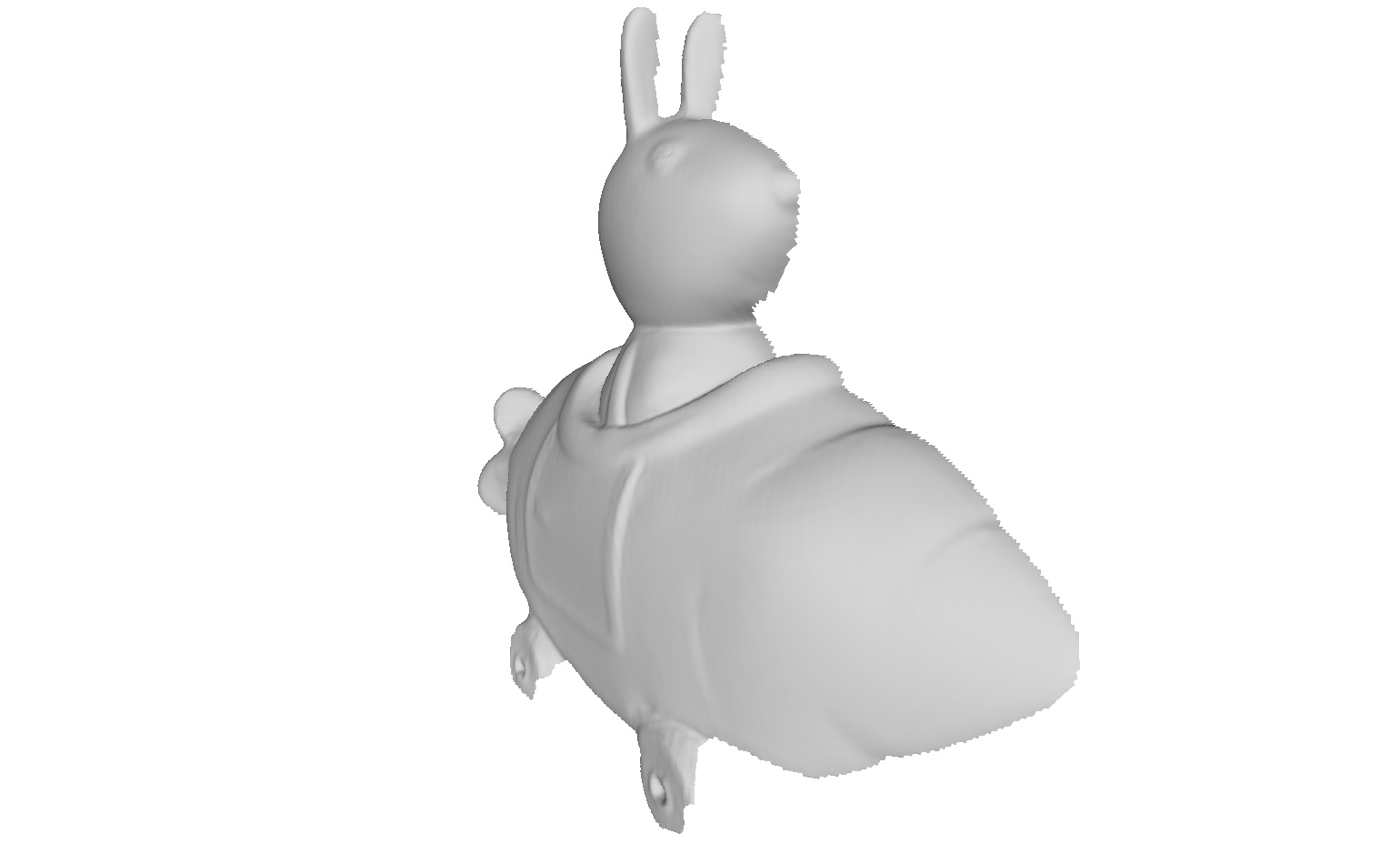} &
\includegraphics[width=0.15\textwidth,trim={10cm 0cm 10cm  0cm},clip]{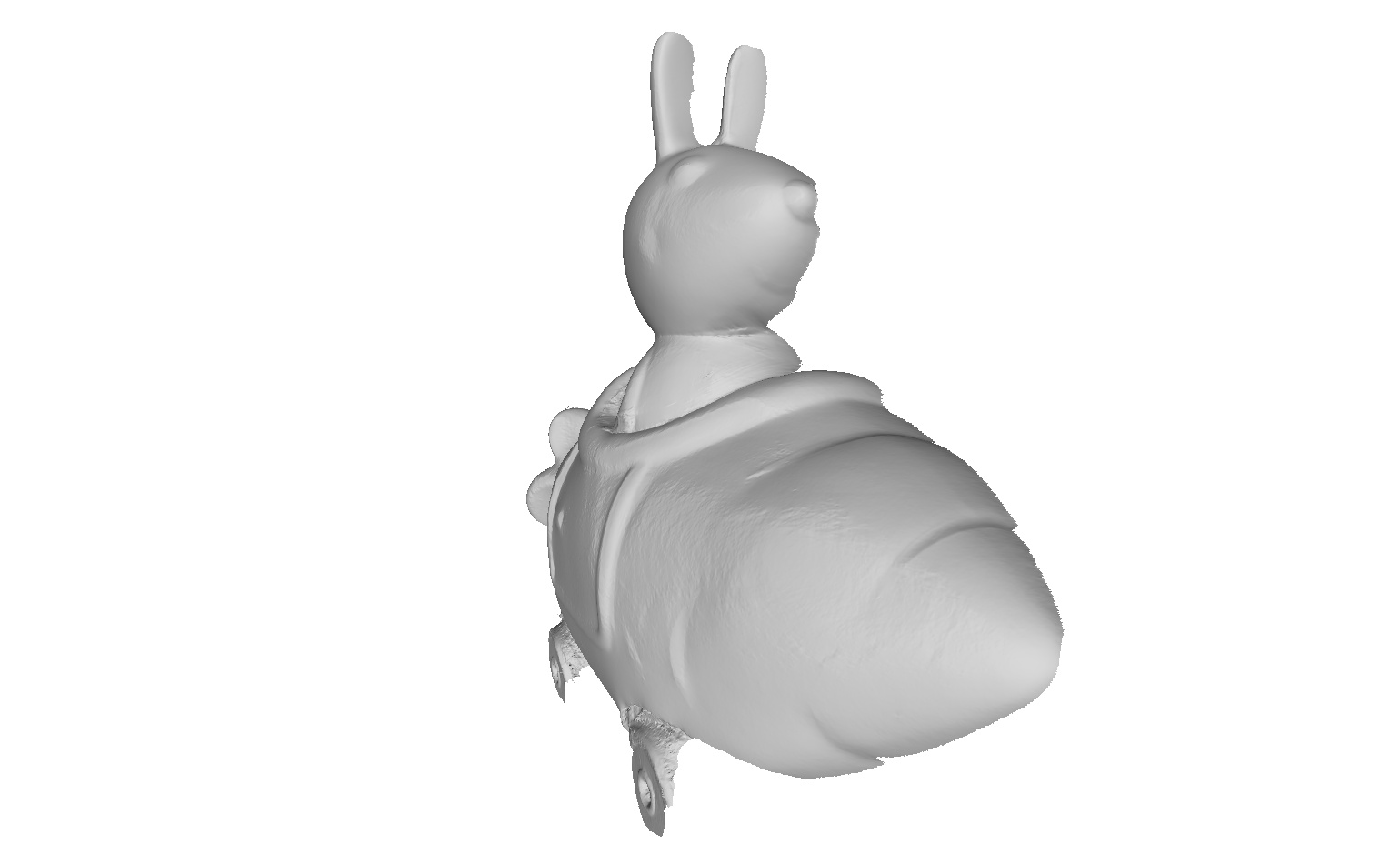} \\

\begin{sideways} {$Z$ Error (mm)} \end{sideways} &
\includegraphics[width=0.15\textwidth,trim={2cm 0cm 2cm  0cm},clip]{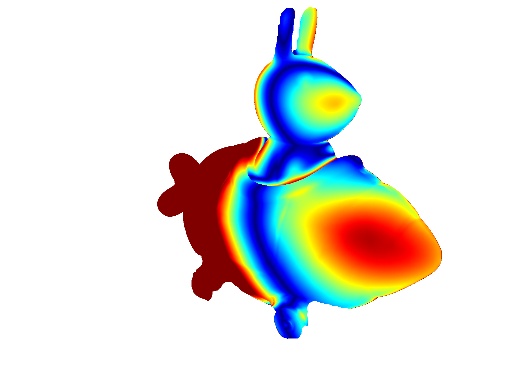} &
\includegraphics[width=0.15\textwidth,trim={2cm 0cm 2cm  0cm},clip]{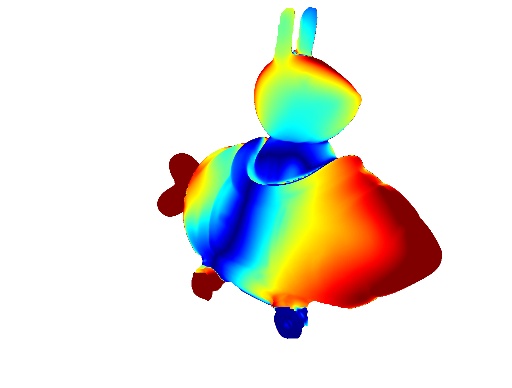} &
\includegraphics[width=0.15\textwidth,trim={2cm 0cm 2cm  0cm},clip]{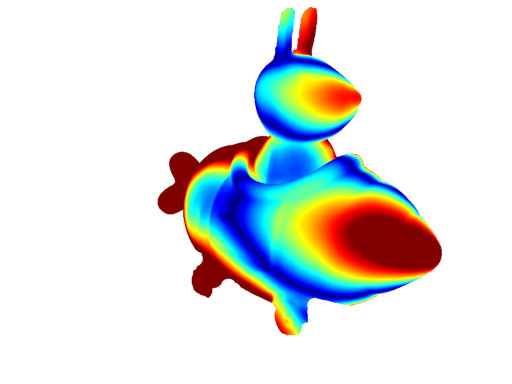} &
\includegraphics[width=0.15\textwidth,trim={2cm 0cm 2cm  0cm},clip]{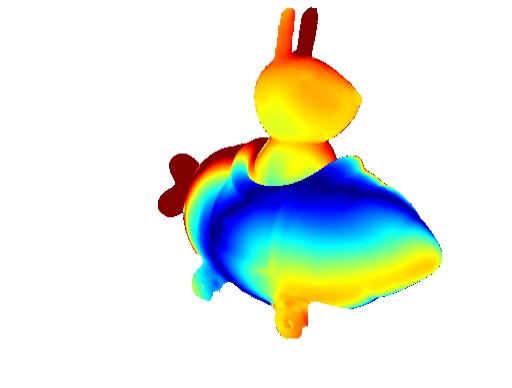} &
\includegraphics[width=0.15\textwidth,trim={2cm 0cm 2cm  0cm},clip]{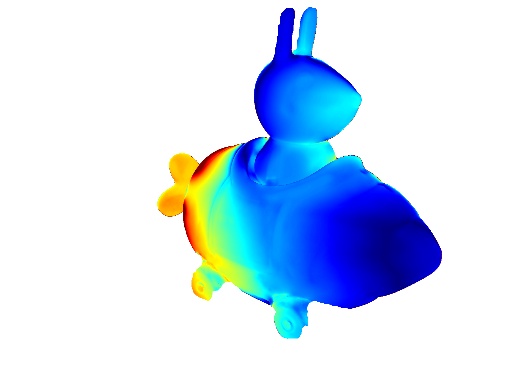} \\
%
\end{tabular}
\caption{Evaluations 1-4/14 }
\label{fig:eval1}
\end{figure*}

\begin{figure*}[t]
\begin{tabular}{c c c c c c}
~ & L17 & Q18 & I18 & S20 & L20 \\
\begin{sideways} {Die-3D Shape} \end{sideways} &
\includegraphics[width=0.15\textwidth,trim={12cm 0cm 8cm  0cm},clip]{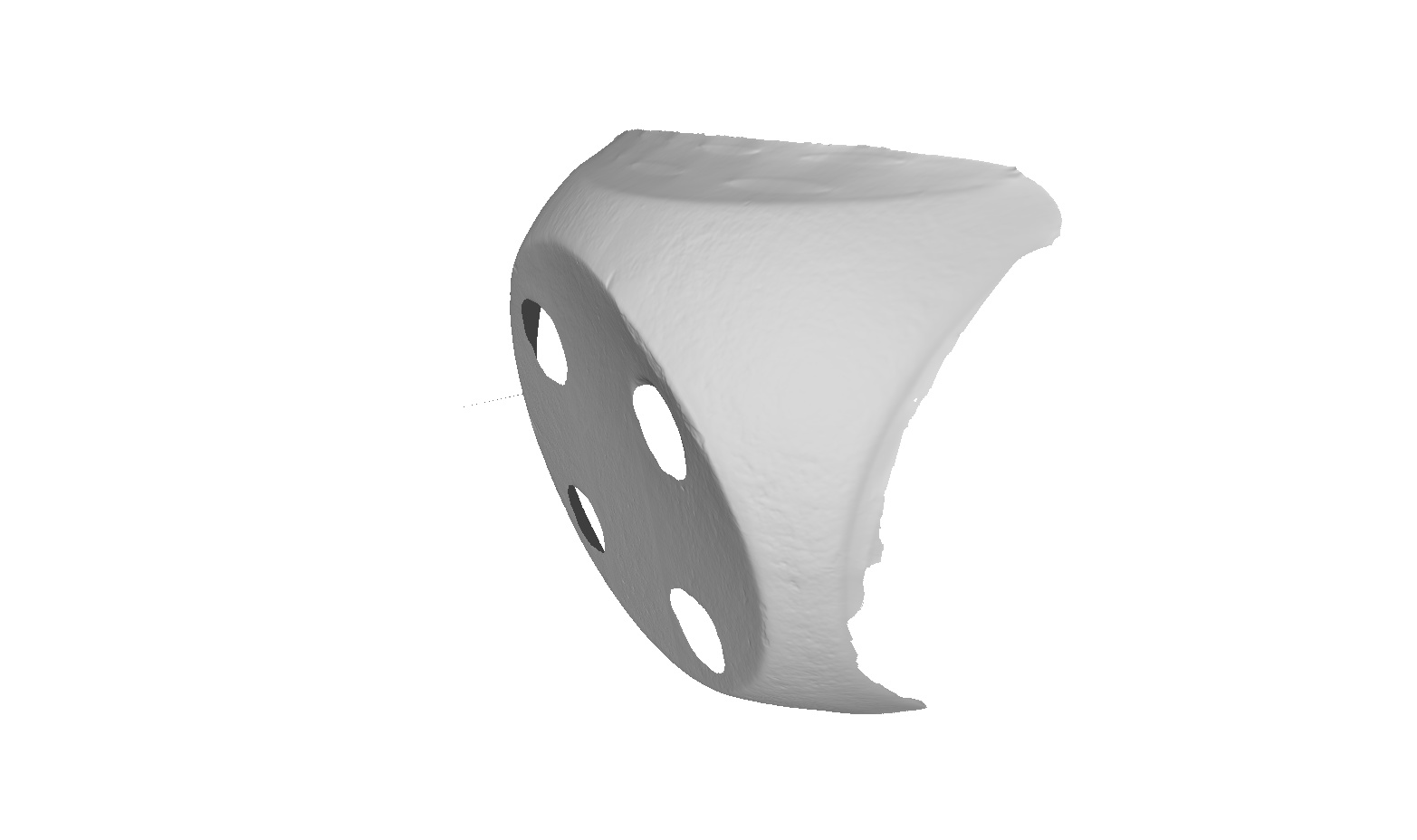} &
\includegraphics[width=0.15\textwidth,trim={12cm 0cm 8cm  0cm},clip]{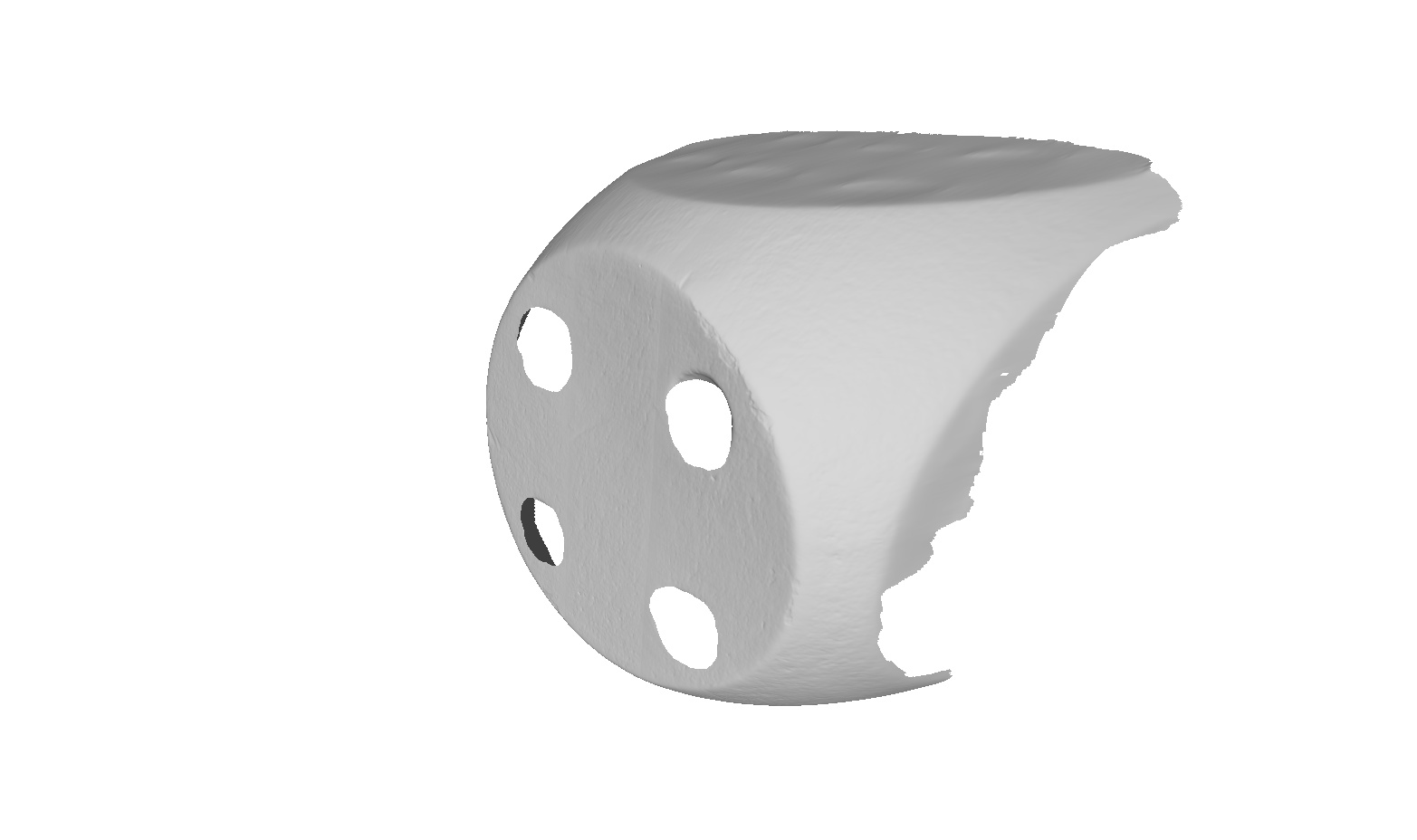} &
\includegraphics[width=0.15\textwidth,trim={12cm 0cm 8cm  0cm},clip]{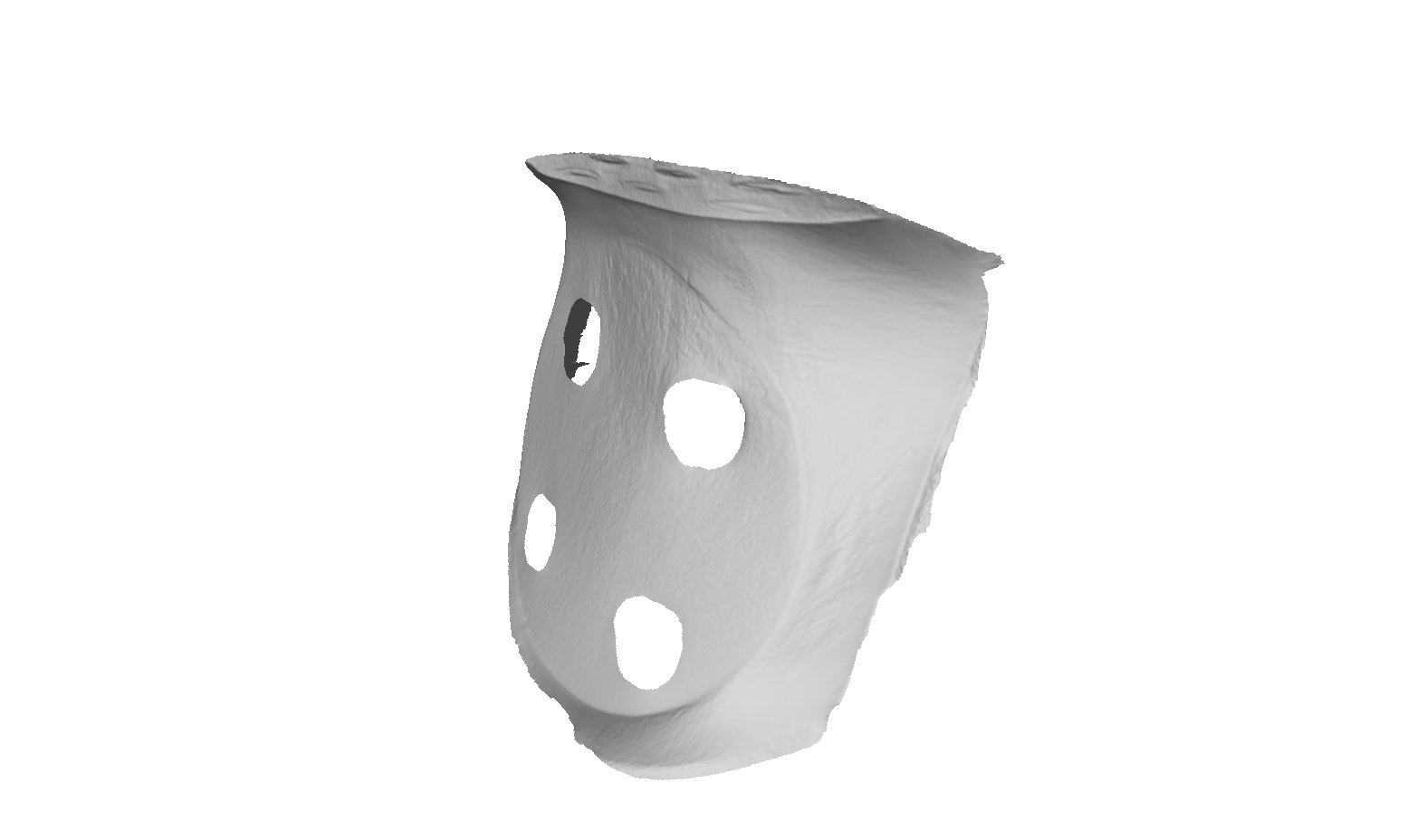} &
\includegraphics[width=0.15\textwidth,trim={12cm 0cm 8cm  0cm},clip]{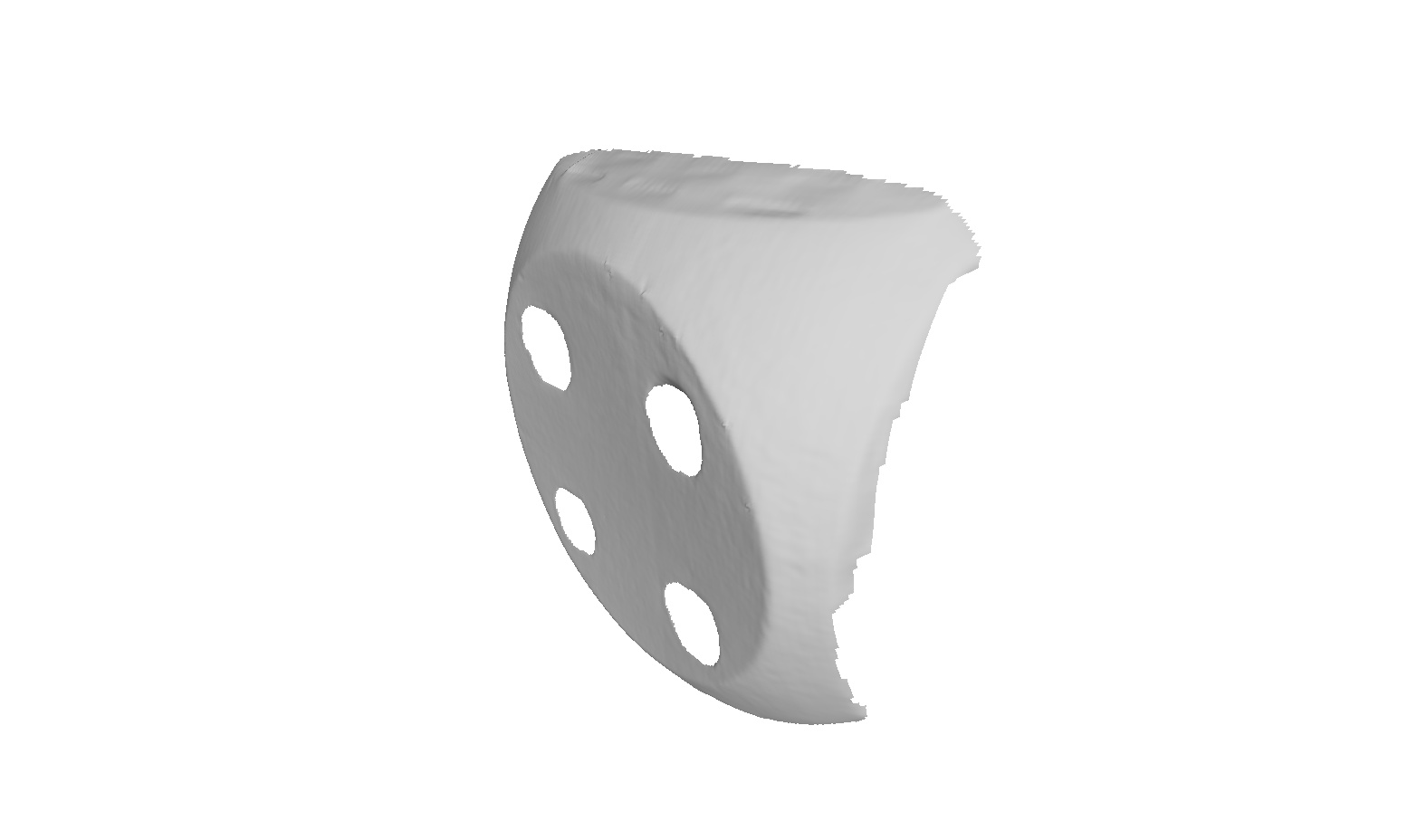} &
\includegraphics[width=0.15\textwidth,trim={12cm 0cm 8cm  0cm},clip]{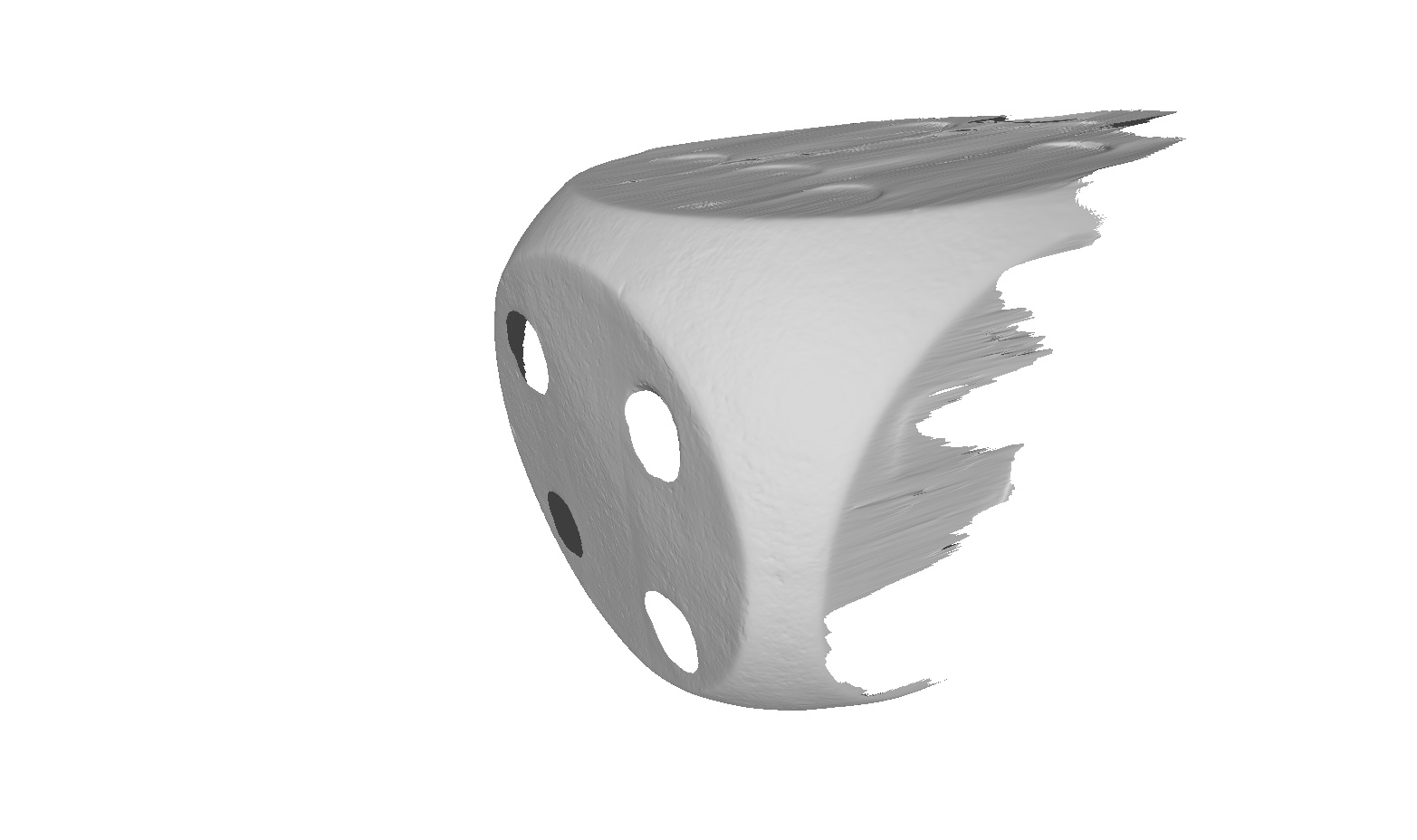} \\
\begin{sideways} {$Z$ Error (mm)} \end{sideways} &
\includegraphics[width=0.15\textwidth,trim={2cm 0cm 2cm  0cm},clip]{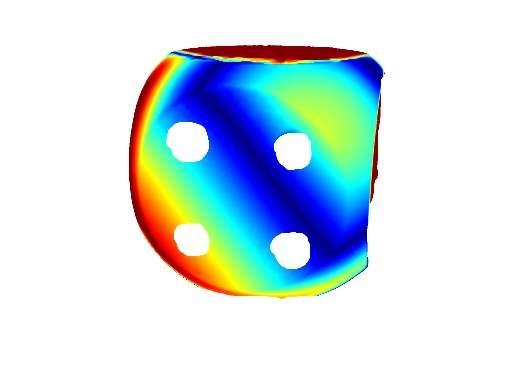} &
\includegraphics[width=0.15\textwidth,trim={2cm 0cm 2cm  0cm},clip]{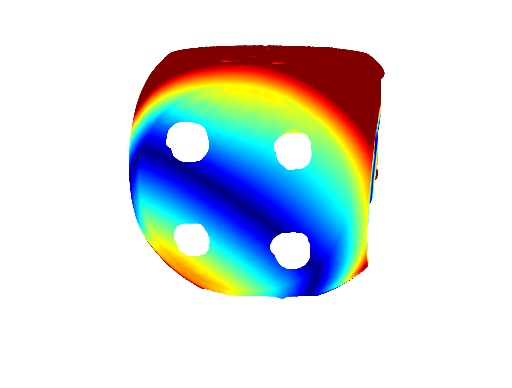} &
\includegraphics[width=0.15\textwidth,trim={2cm 0cm 2cm  0cm},clip]{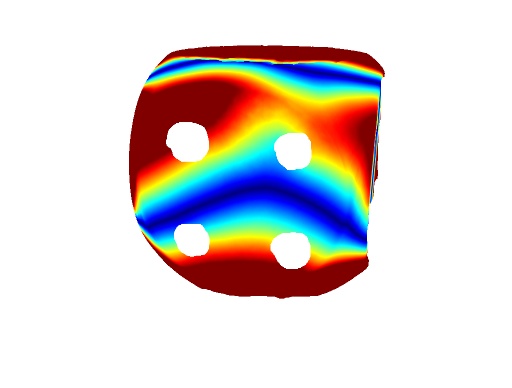} &
\includegraphics[width=0.15\textwidth,trim={2cm 0cm 2cm  0cm},clip]{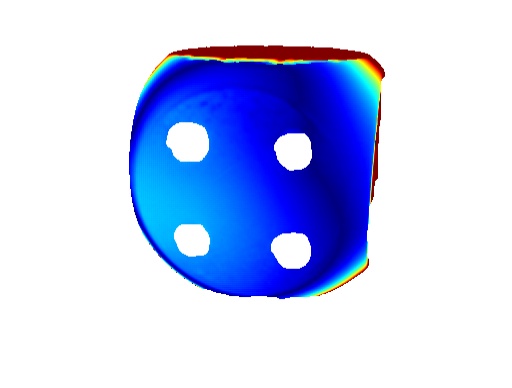} &
\includegraphics[width=0.15\textwidth,trim={2cm 0cm 2cm  0cm},clip]{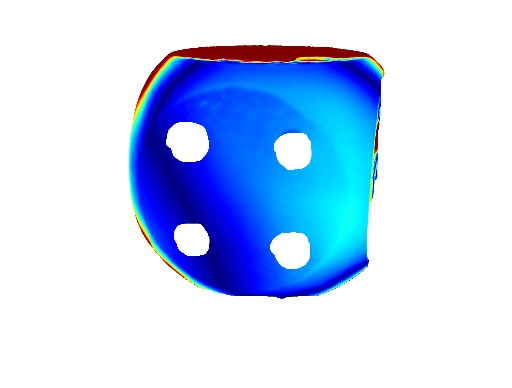} \\
%

\begin{sideways} {Hippo-3D Shape} \end{sideways} &
\includegraphics[width=0.15\textwidth,trim={5cm 0cm 4cm  0cm},clip]{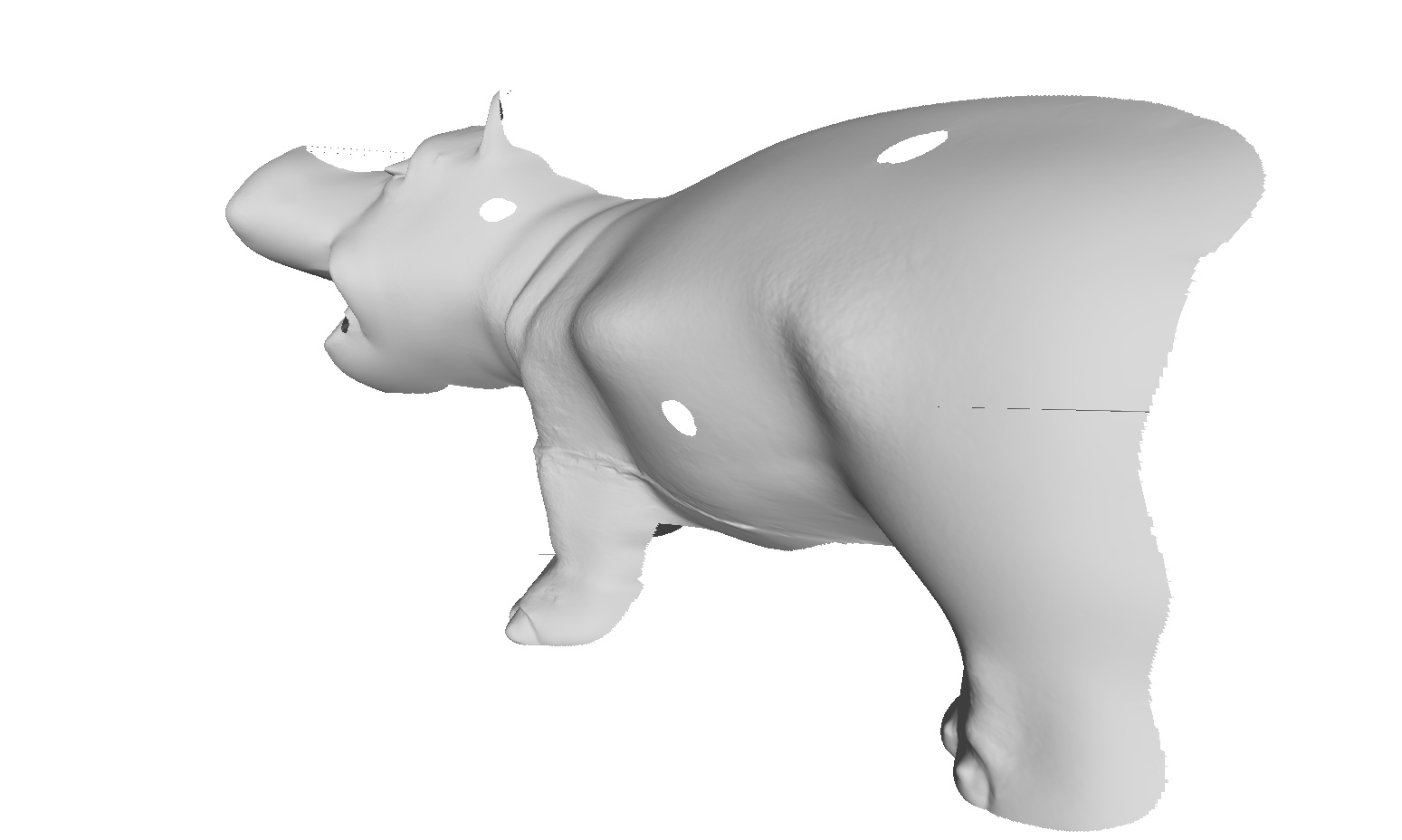} &
\includegraphics[width=0.15\textwidth,trim={5cm 0cm 4cm  0cm},clip]{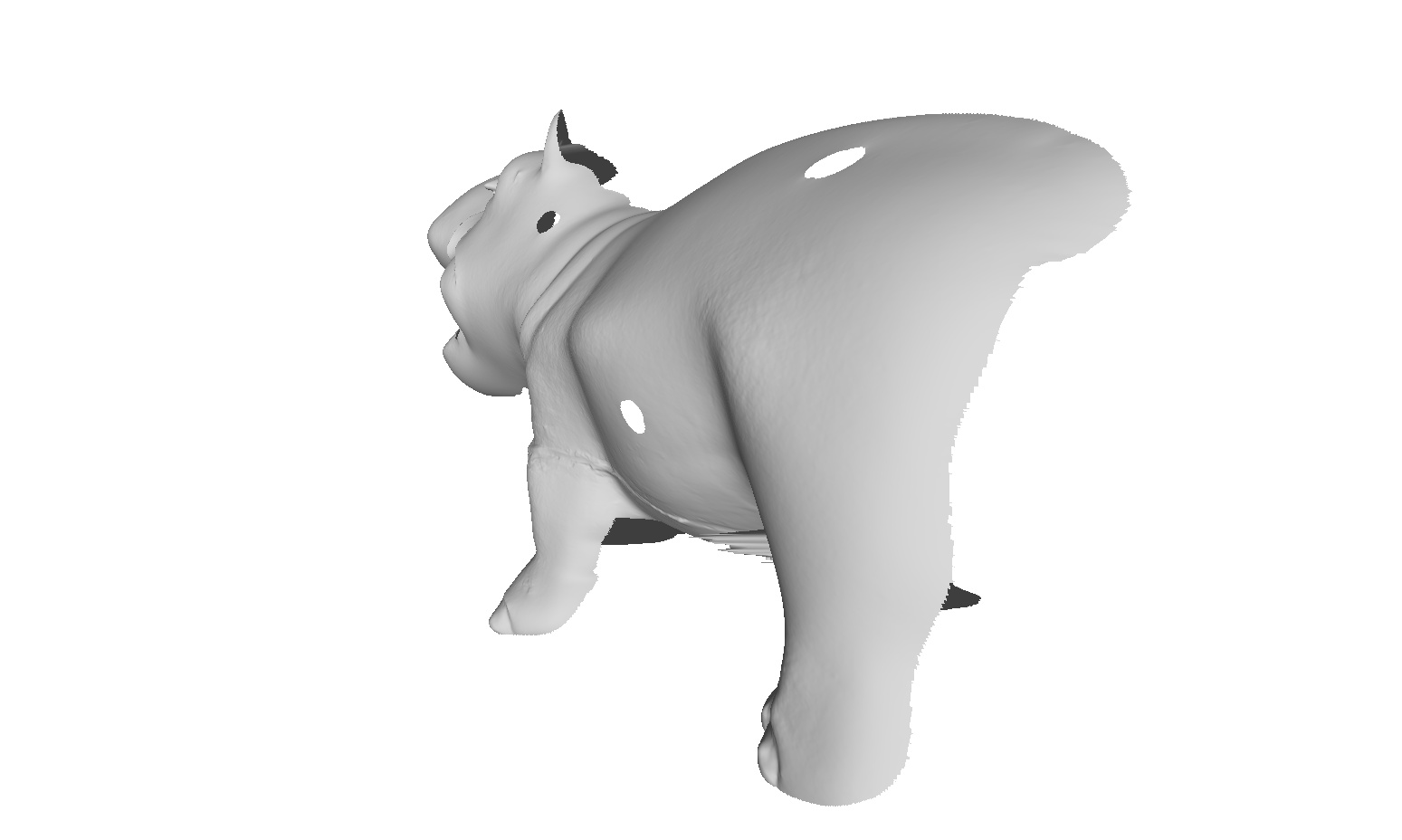} &
\includegraphics[width=0.15\textwidth,trim={5cm 0cm 4cm  0cm},clip]{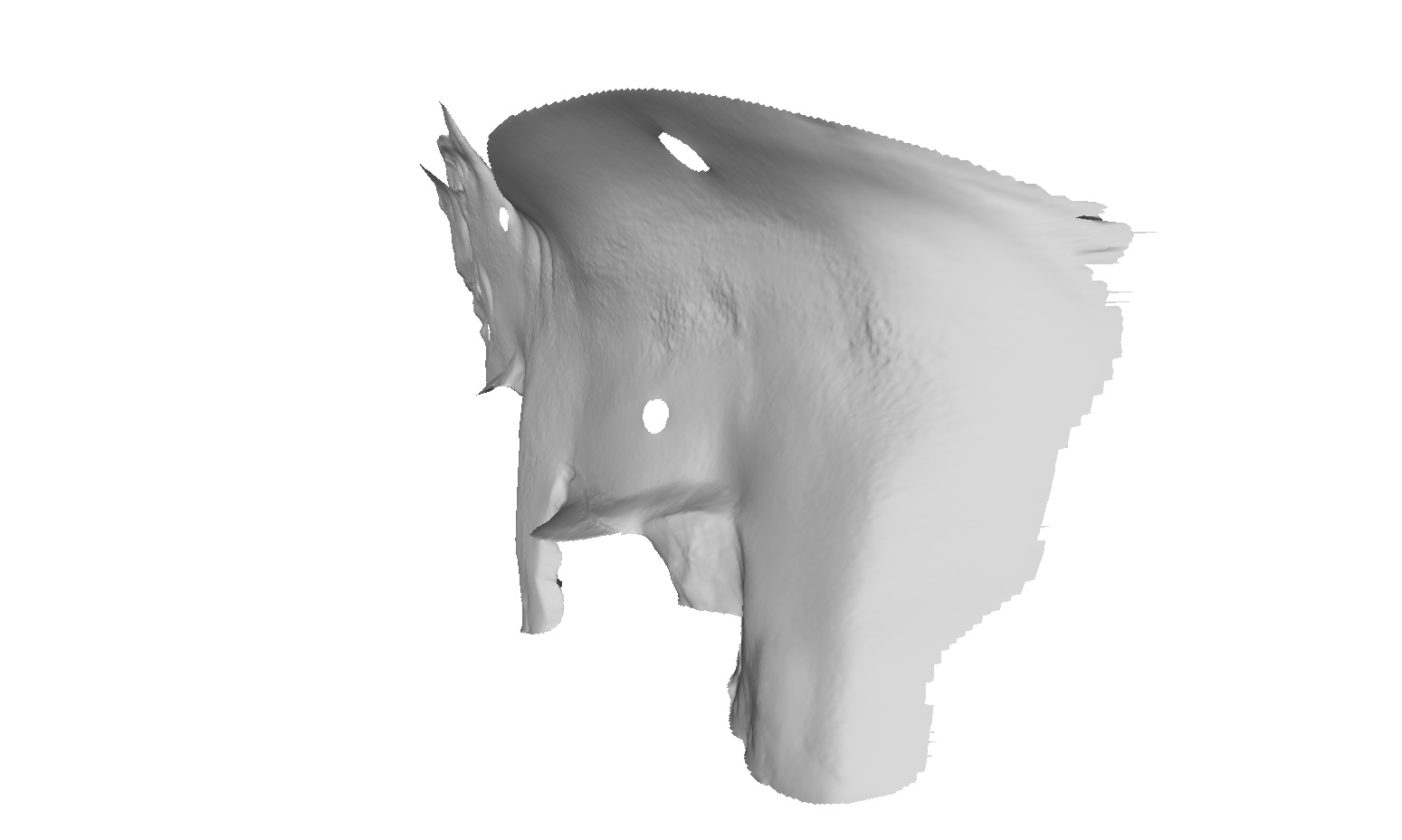} &
\includegraphics[width=0.15\textwidth,trim={5cm 0cm 4cm  0cm},clip]{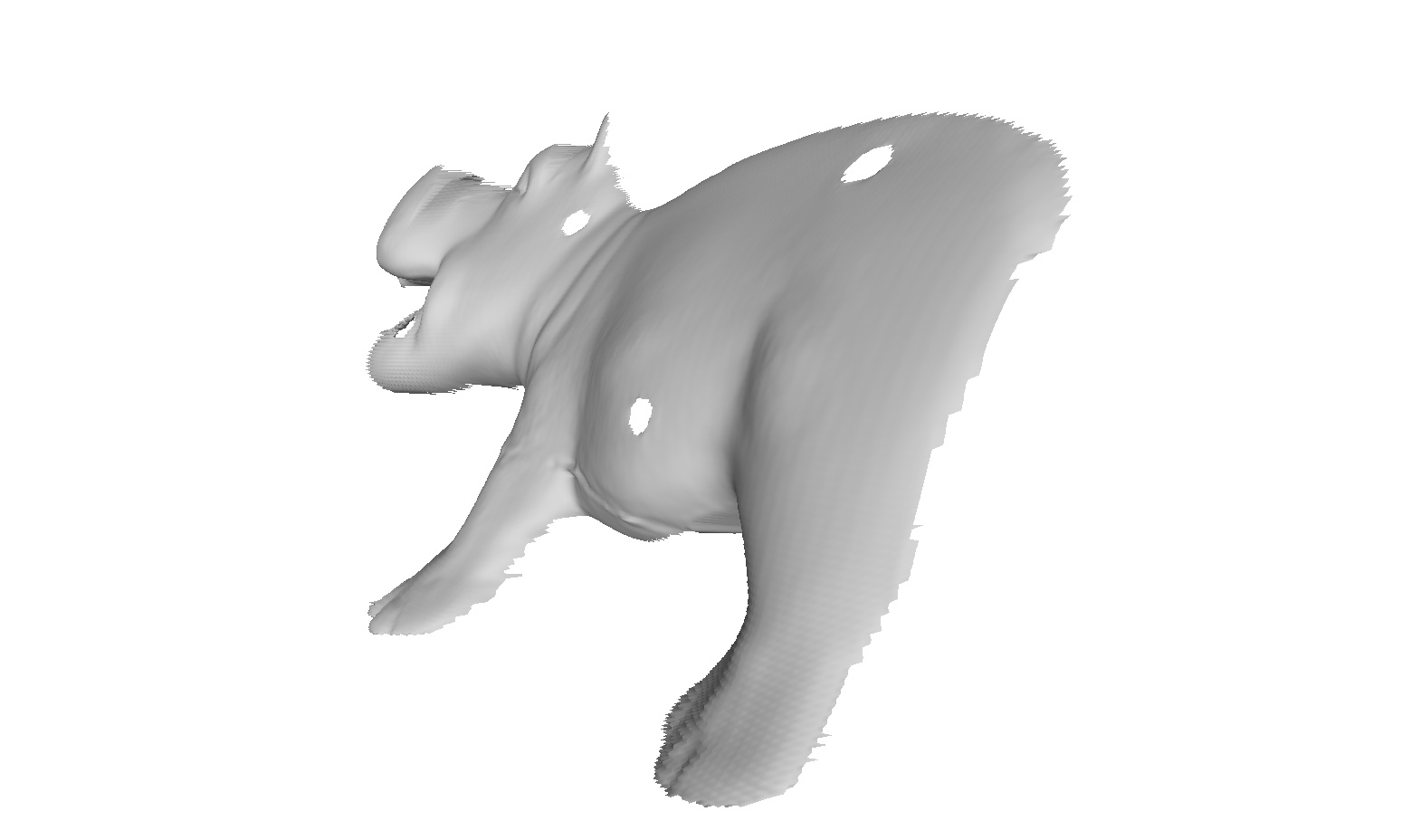} &
\includegraphics[width=0.15\textwidth,trim={5cm 0cm 4cm  0cm},clip]{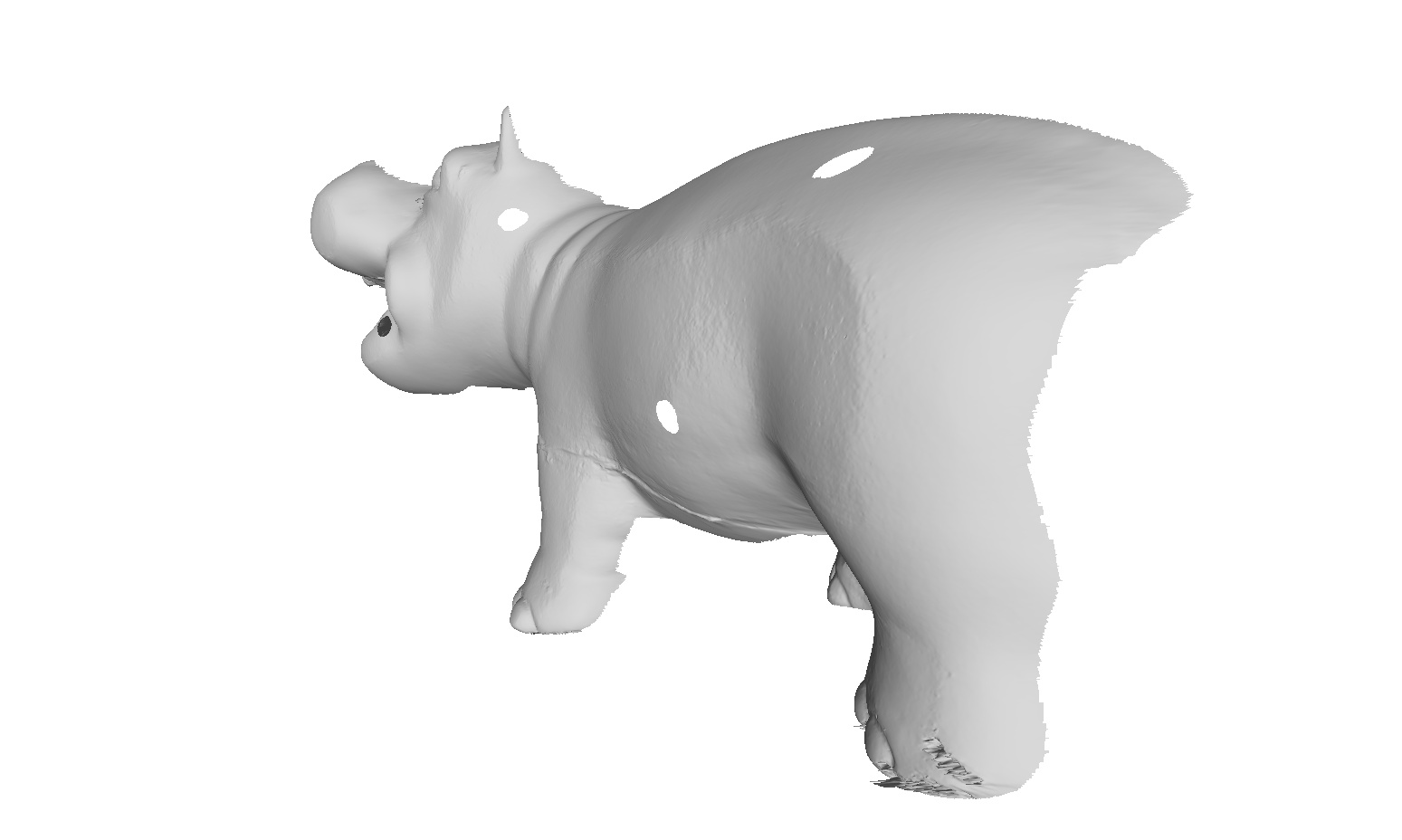} \\
\begin{sideways} {$Z$ Error (mm)} \end{sideways} &
\includegraphics[width=0.15\textwidth,trim={1cm 0cm 2cm  0cm},clip]{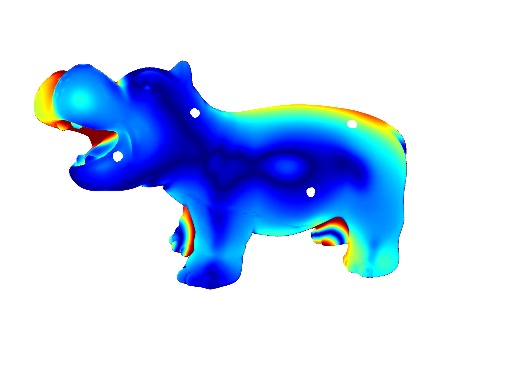} &
\includegraphics[width=0.15\textwidth,trim={1cm 0cm 2cm  0cm},clip]{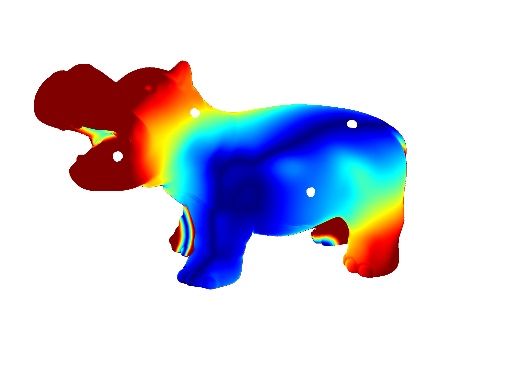} &
\includegraphics[width=0.15\textwidth,trim={1cm 0cm 2cm  0cm},clip]{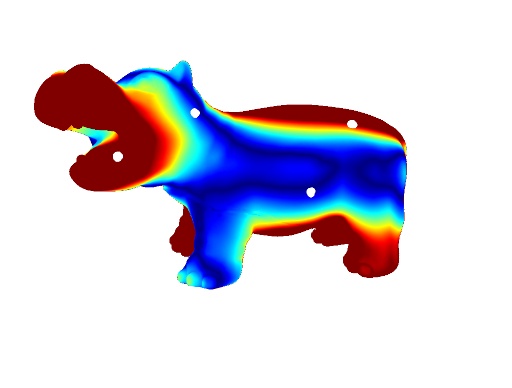} &
\includegraphics[width=0.15\textwidth,trim={1cm 0cm 2cm  0cm},clip]{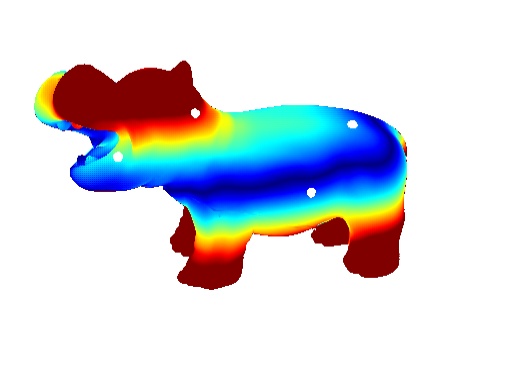} &
\includegraphics[width=0.15\textwidth,trim={1cm 0cm 2cm  0cm},clip]{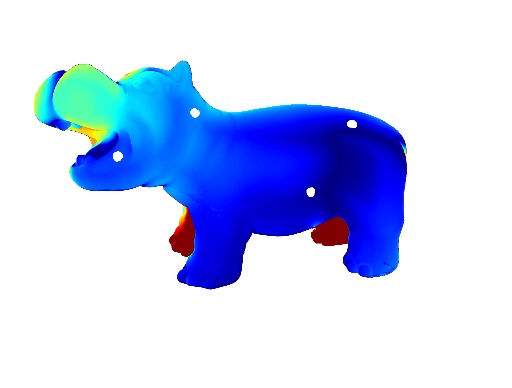} \\

\begin{sideways} {House-3D Shape} \end{sideways} &
\includegraphics[width=0.15\textwidth,trim={12cm 0cm 12cm  0cm},clip]{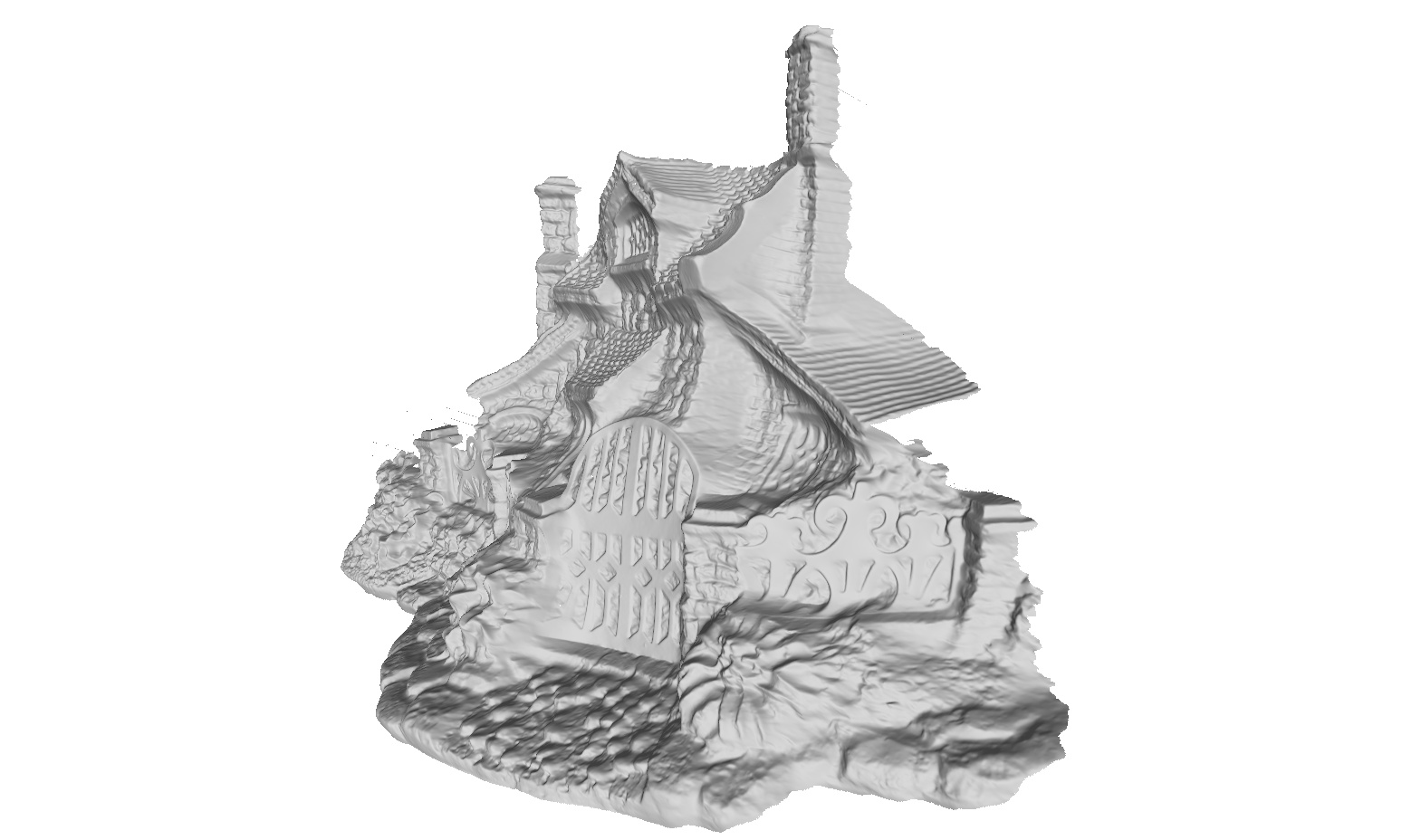} &
\includegraphics[width=0.15\textwidth,trim={12cm 0cm 12cm  0cm},clip]{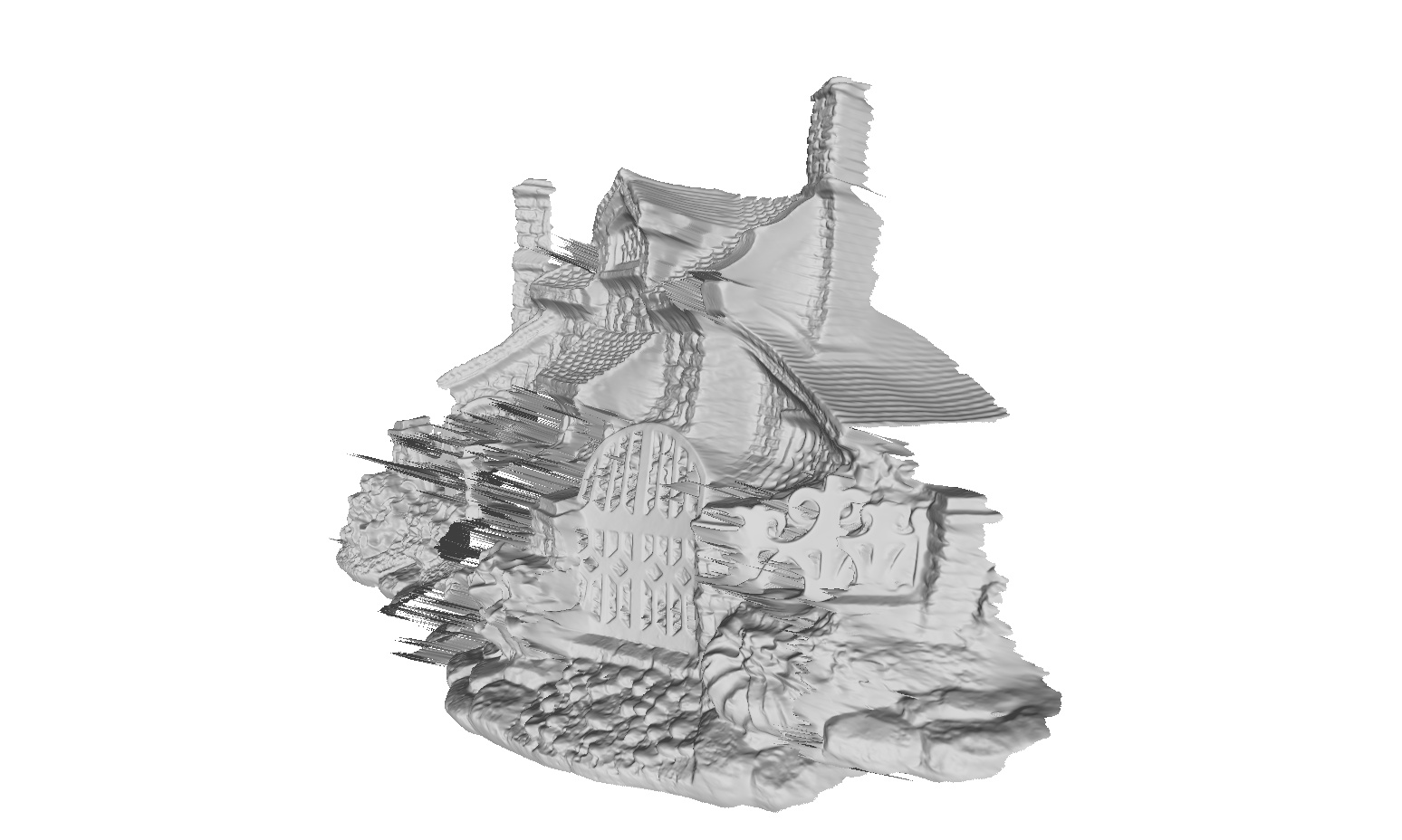} &
\includegraphics[width=0.15\textwidth,trim={12cm 0cm 12cm  0cm},clip]{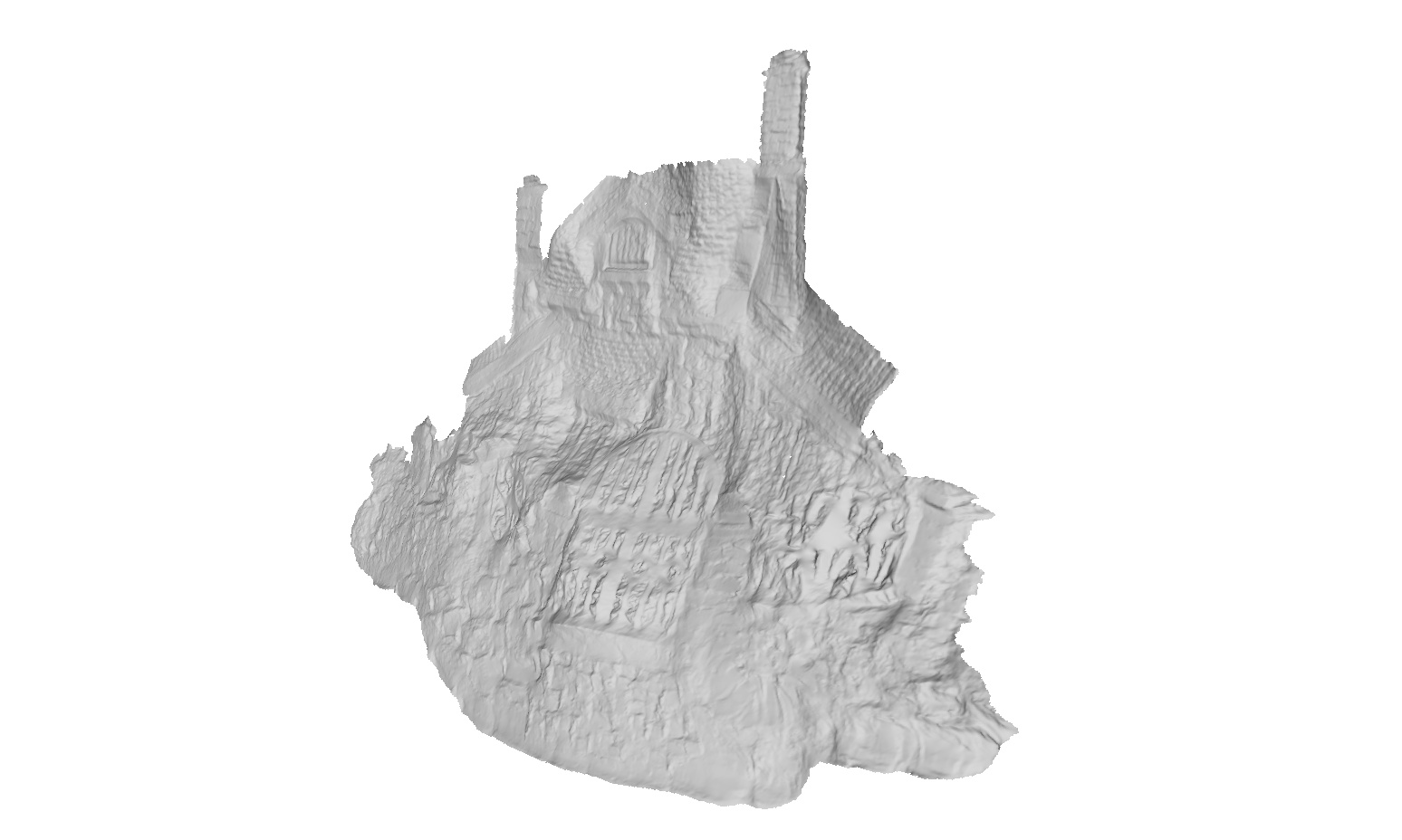} &
\includegraphics[width=0.15\textwidth,trim={12cm 0cm 12cm  0cm},clip]{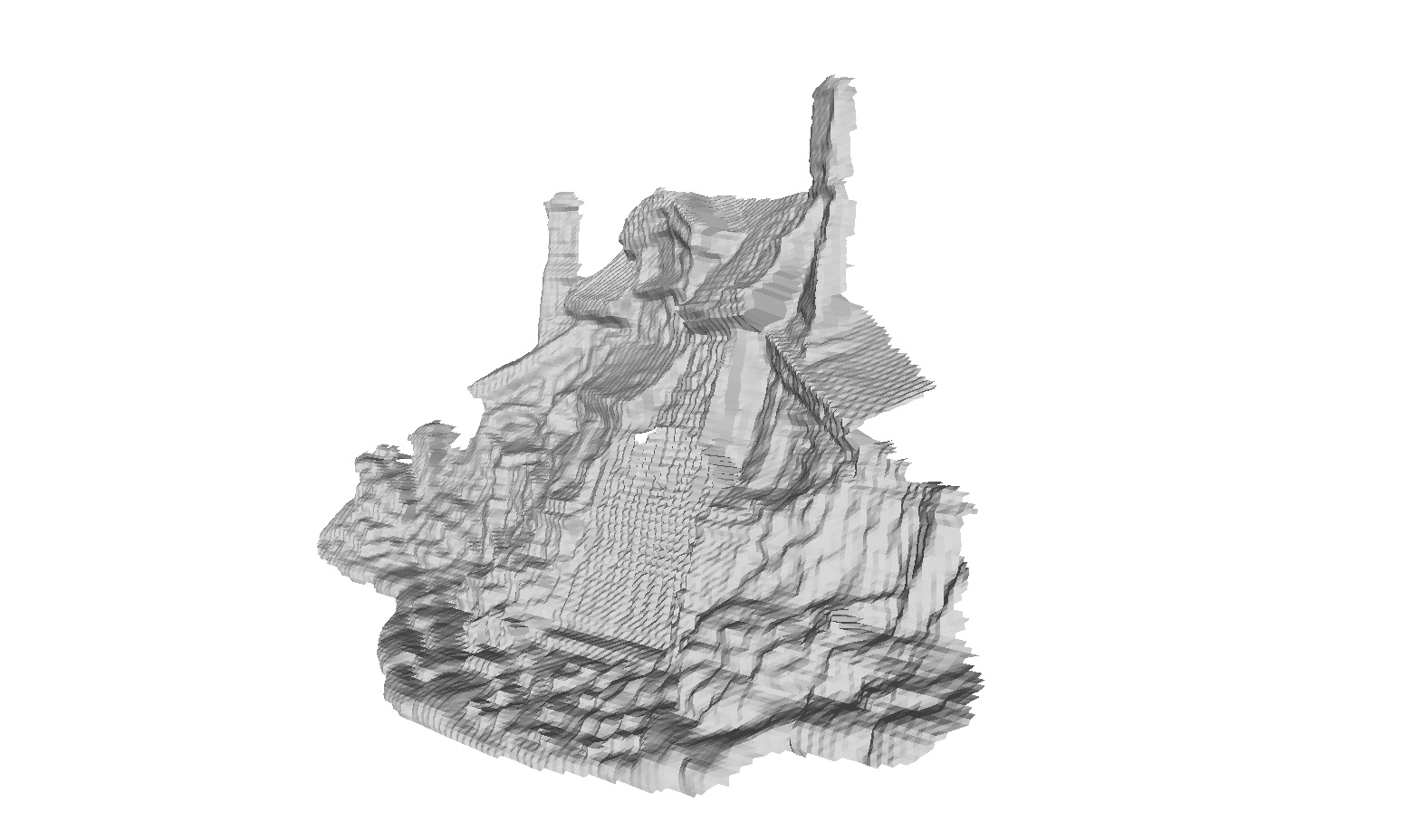} &
\includegraphics[width=0.15\textwidth,trim={12cm 0cm 12cm  0cm},clip]{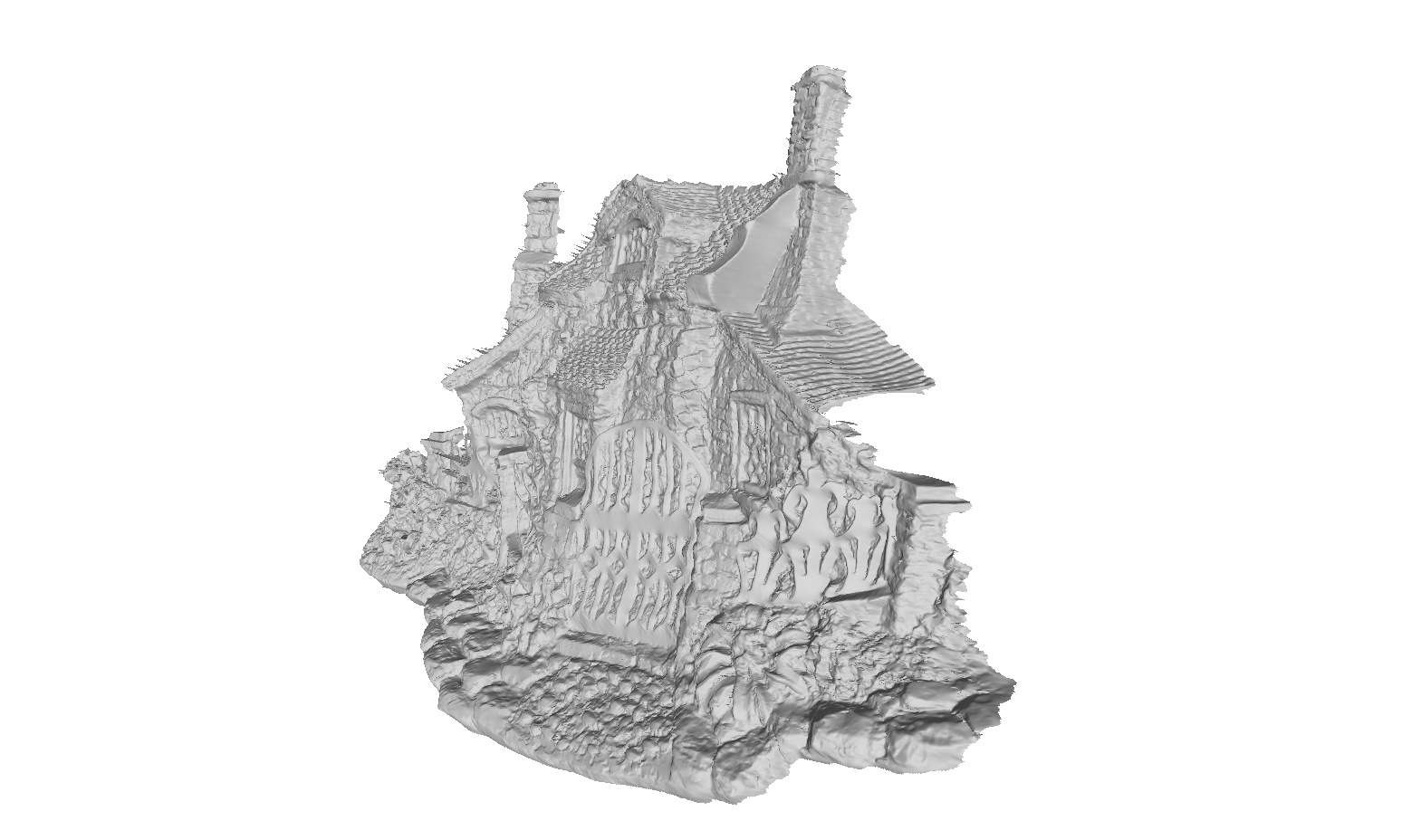} \\
\begin{sideways} {$Z$ Error (mm)} \end{sideways} &
\includegraphics[width=0.15\textwidth,trim={1cm 0cm 1cm  0cm},clip]{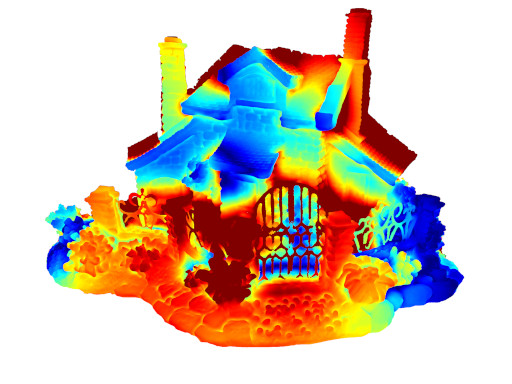} &
\includegraphics[width=0.15\textwidth,trim={1cm 0cm 1cm  0cm},clip]{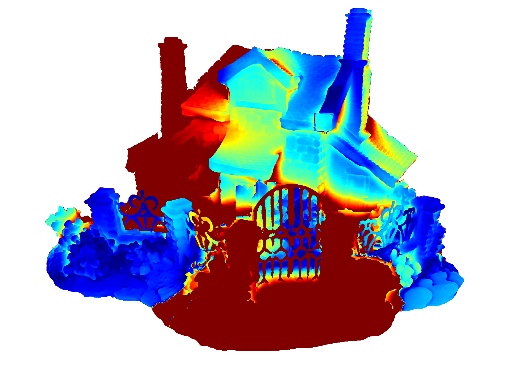} &
\includegraphics[width=0.15\textwidth,trim={1cm 0cm 1cm  0cm},clip]{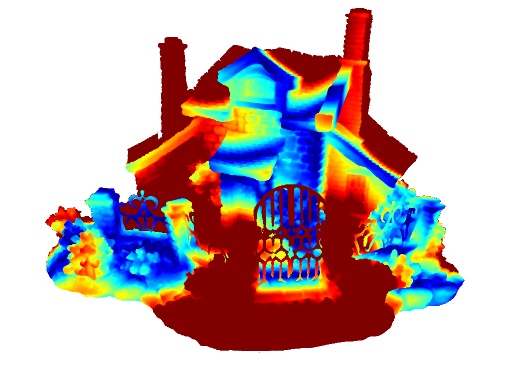} &
\includegraphics[width=0.15\textwidth,trim={1cm 0cm 1cm  0cm},clip]{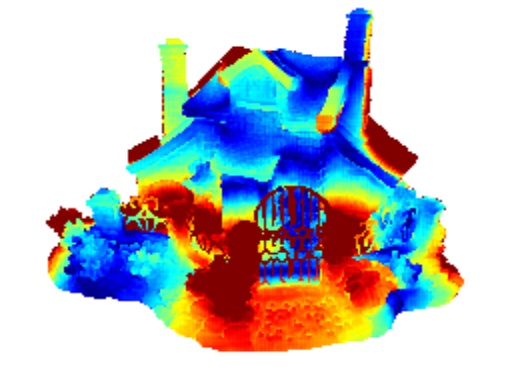} &
\includegraphics[width=0.15\textwidth,trim={1cm 0cm 1cm  0cm},clip]{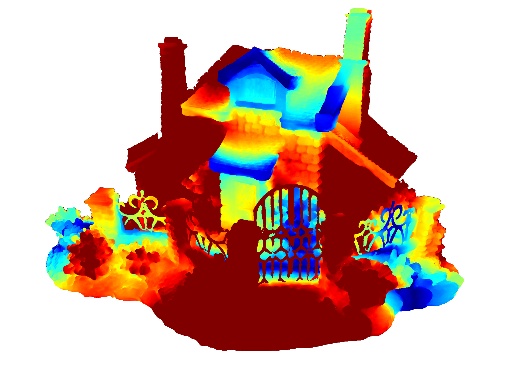} \\
%

\begin{sideways} {Cup-3D Shape} \end{sideways} &
\includegraphics[width=0.15\textwidth,trim={13cm 0cm 13cm  0cm},clip]{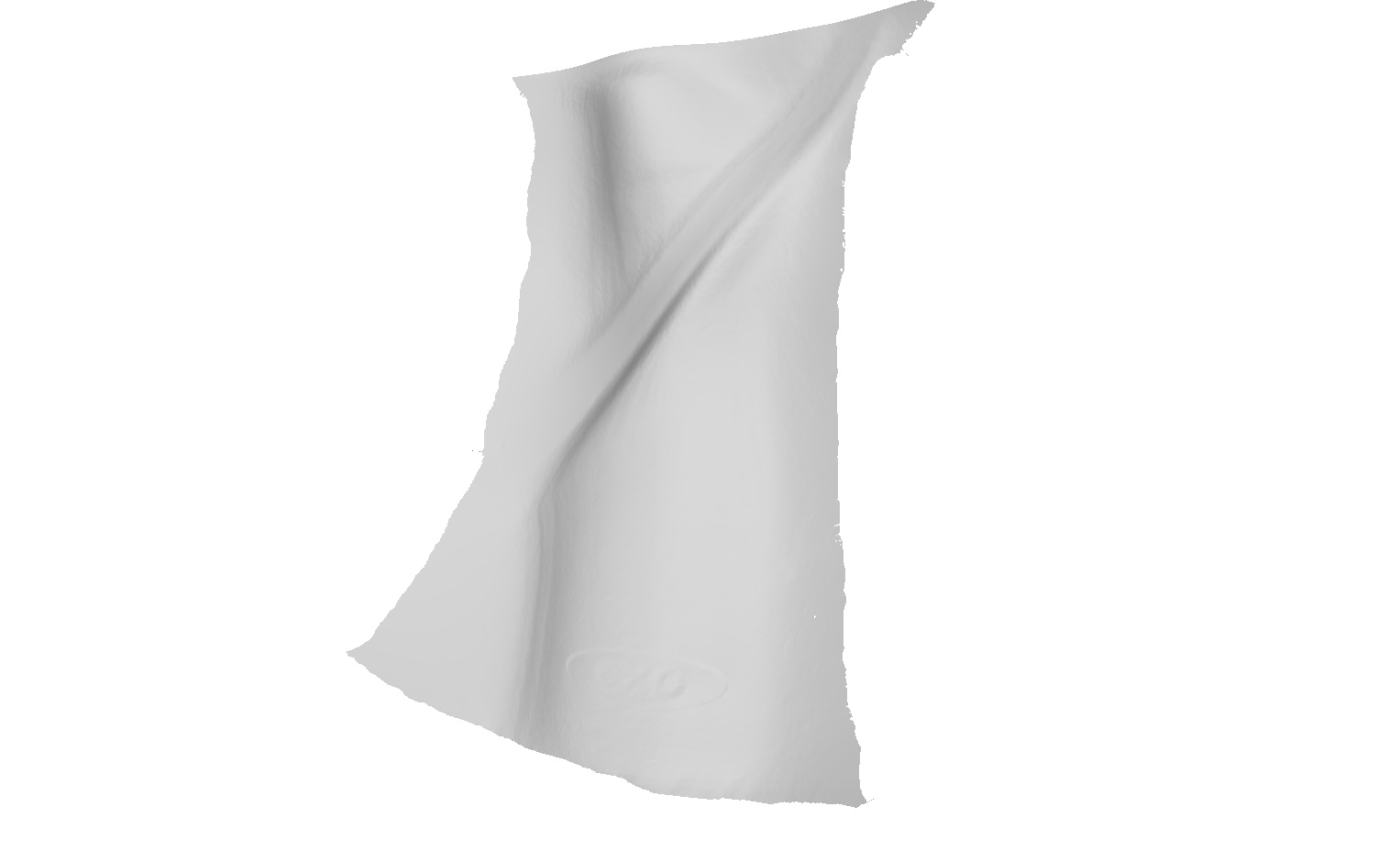} &
\includegraphics[width=0.15\textwidth,trim={13cm 0cm 13cm  0cm},clip]{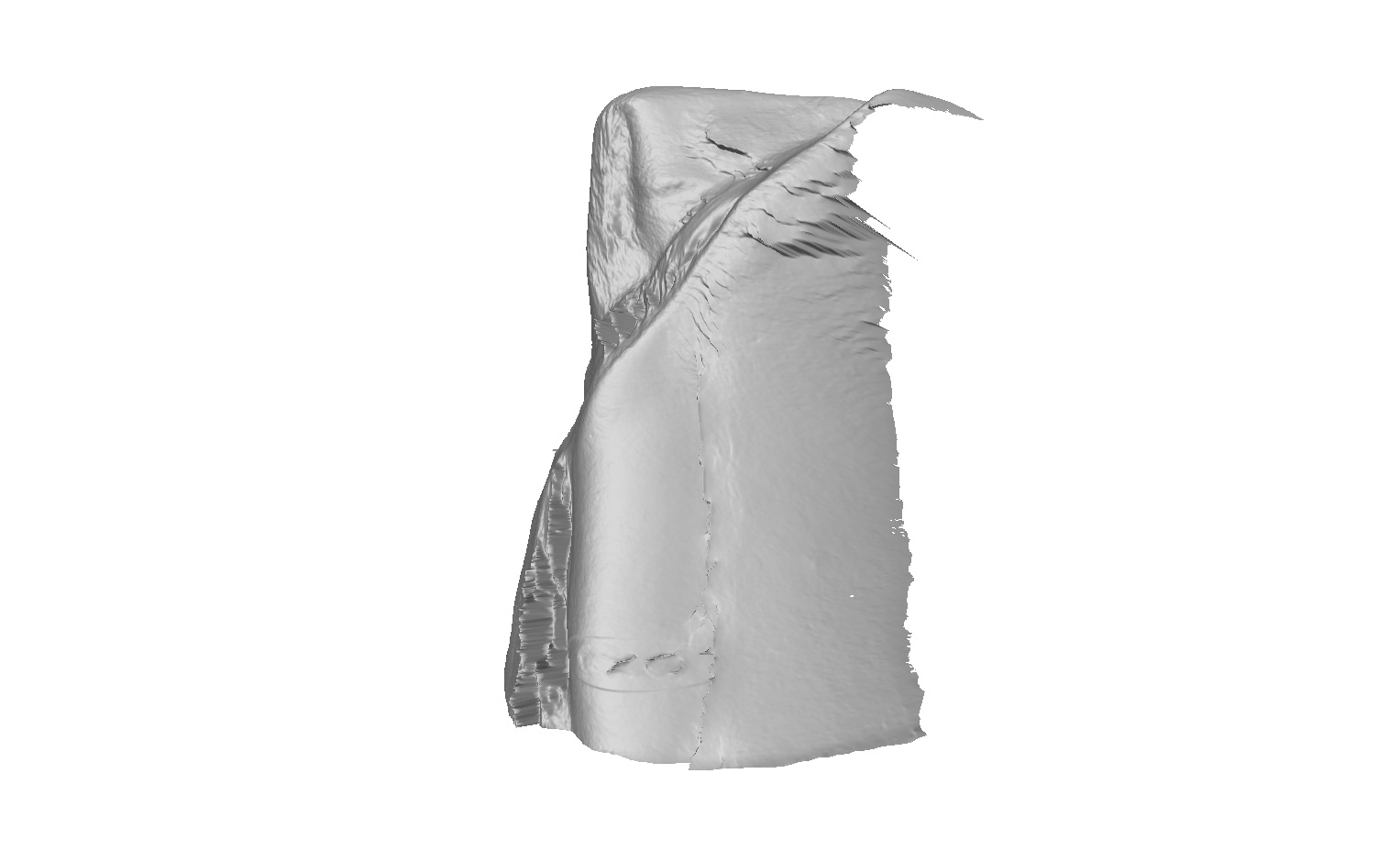} &
\includegraphics[width=0.15\textwidth,trim={13cm 0cm 13cm  0cm},clip]{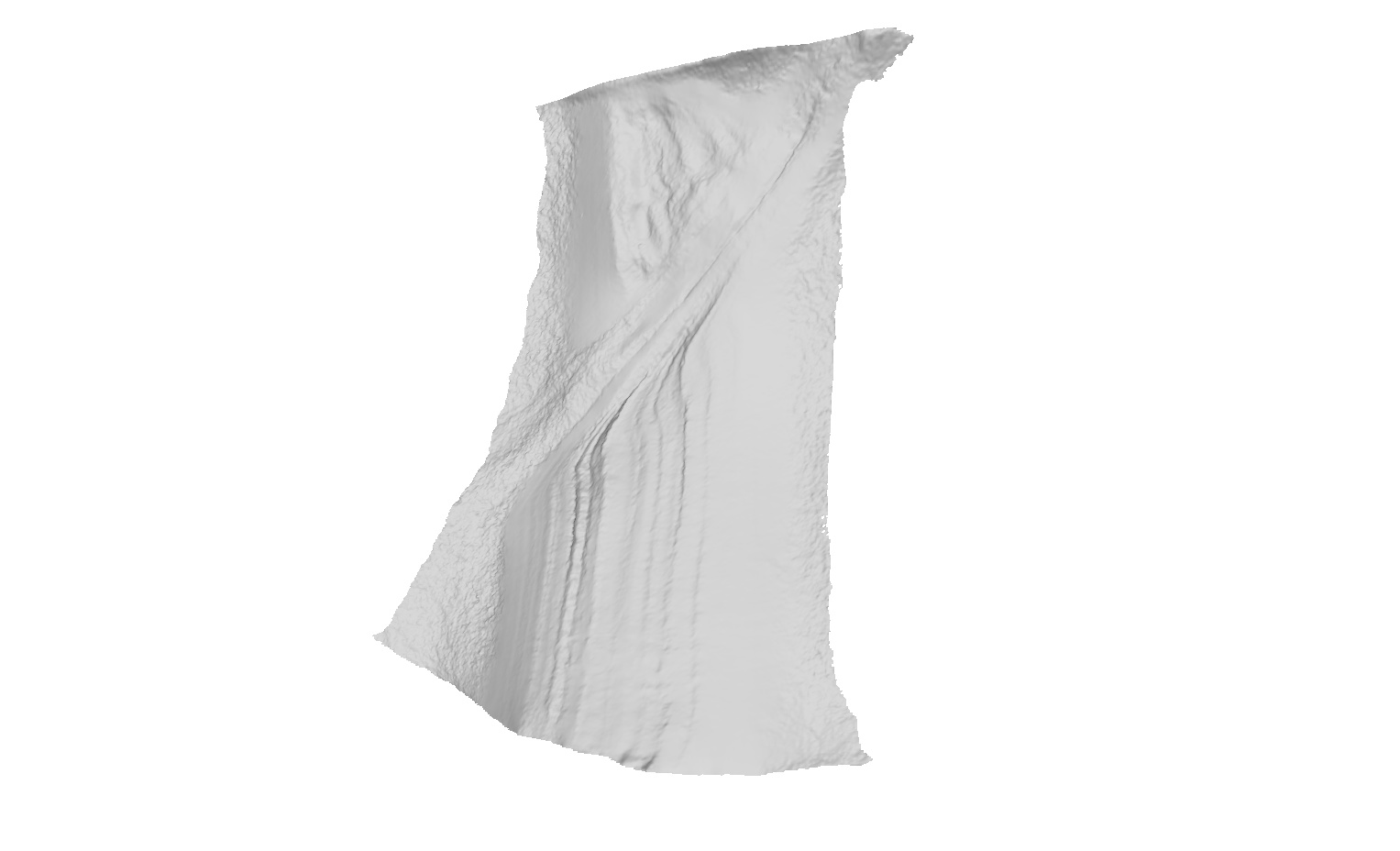} &
\includegraphics[width=0.15\textwidth,trim={13cm 0cm 13cm  0cm},clip]{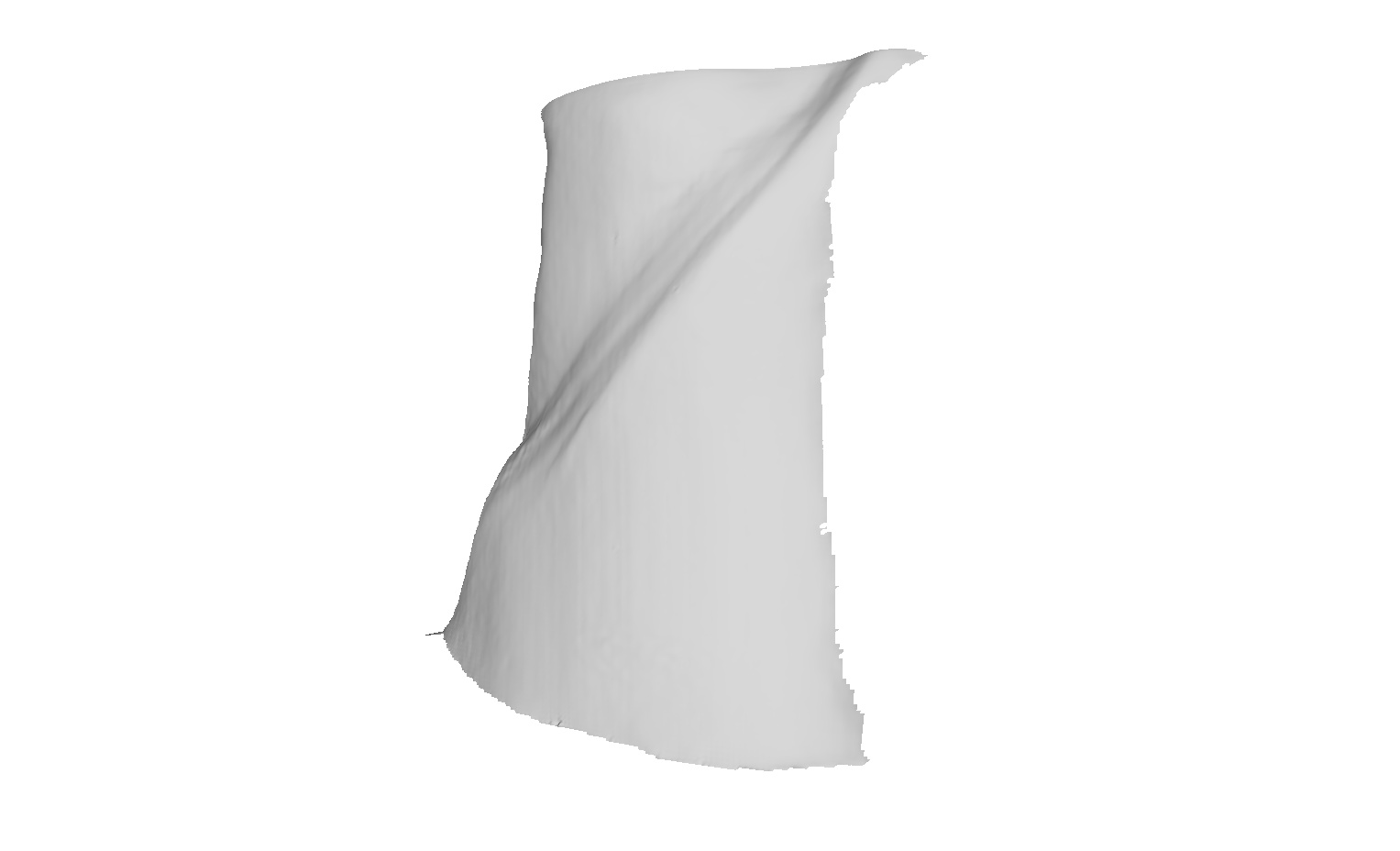} &
\includegraphics[width=0.15\textwidth,trim={13cm 0cm 13cm  0cm},clip]{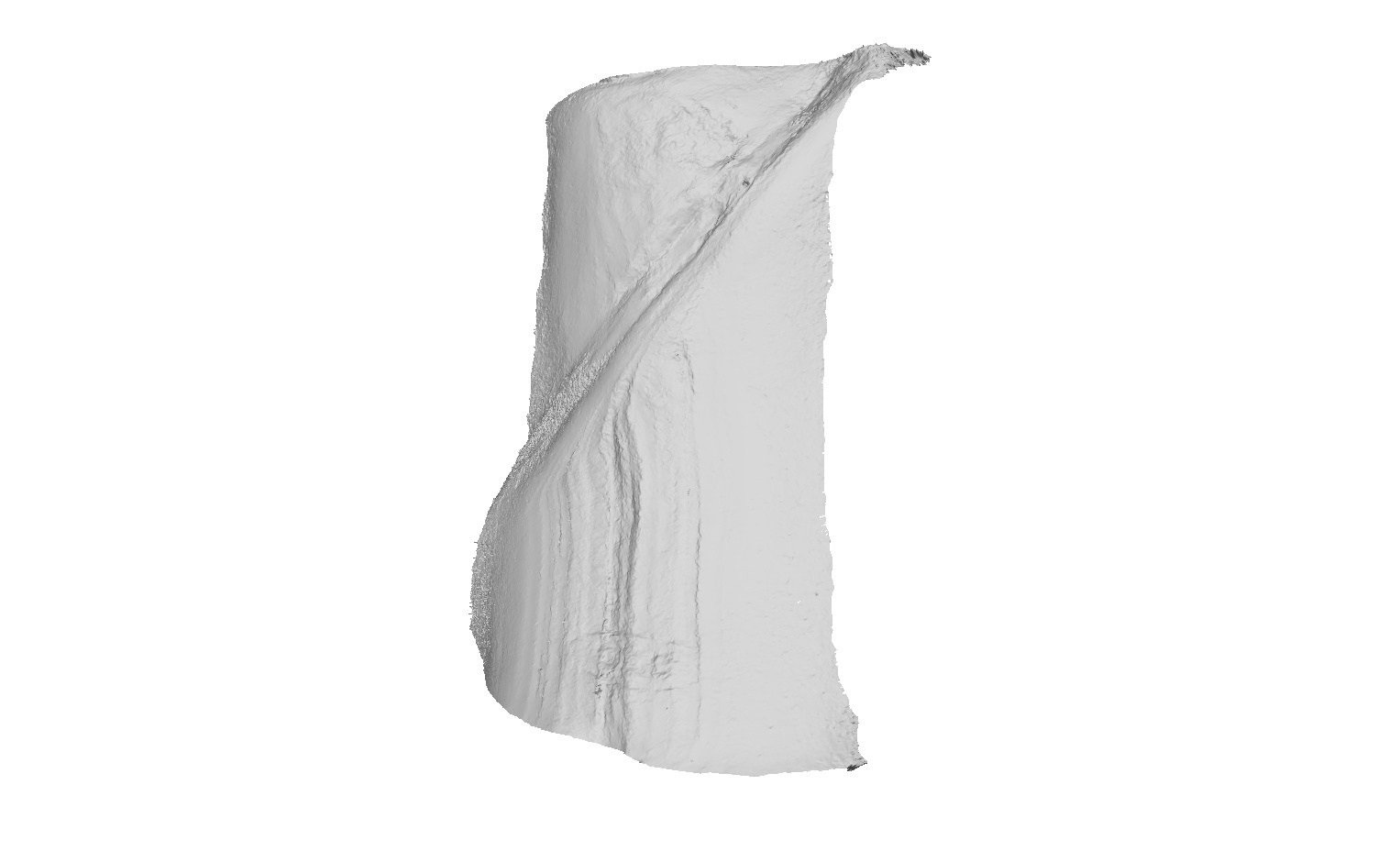} \\
\begin{sideways} {$Z$ Error (mm)} \end{sideways} &
\includegraphics[width=0.15\textwidth,trim={3cm 0cm 3cm  0cm},clip]{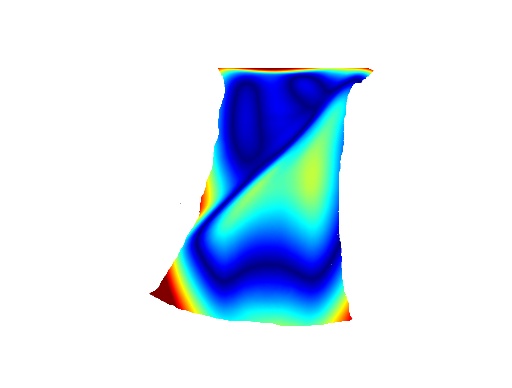} &
\includegraphics[width=0.15\textwidth,trim={3cm 0cm 3cm  0cm},clip]{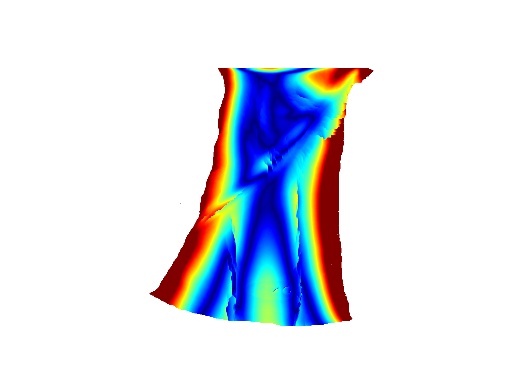} &
\includegraphics[width=0.15\textwidth,trim={3cm 0cm 3cm  0cm},clip]{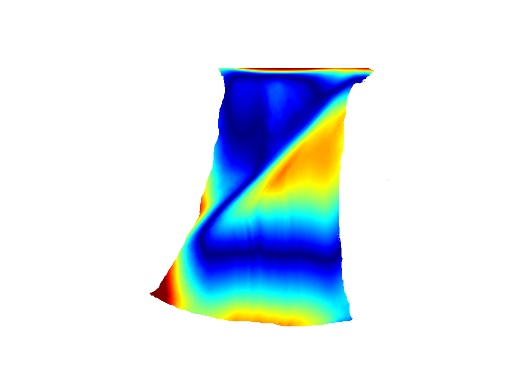} &
\includegraphics[width=0.15\textwidth,trim={3cm 0cm 3cm  0cm},clip]{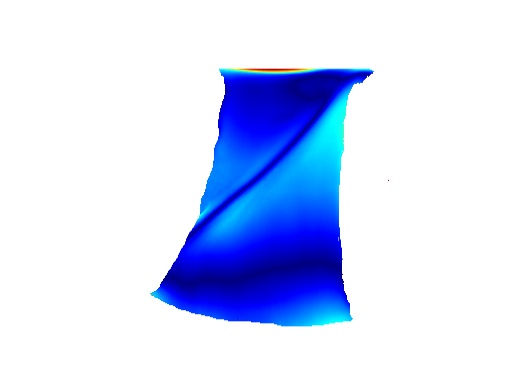} &
\includegraphics[width=0.15\textwidth,trim={3cm 0cm 3cm  0cm},clip]{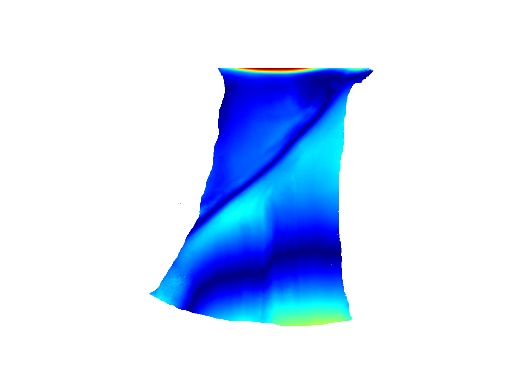} \\
\end{tabular}
\caption{Evaluations 5-8/14 }
\label{fig:eval3}
\end{figure*}

\begin{figure*}[t]
\begin{tabular}{c c c c c c}
~ & L17 & Q18 & I18 & S20 & L20 \\
\begin{sideways} {Owl-3D Shape} \end{sideways} &
\includegraphics[width=0.15\textwidth,trim={5cm 0cm 6cm  0cm},clip]{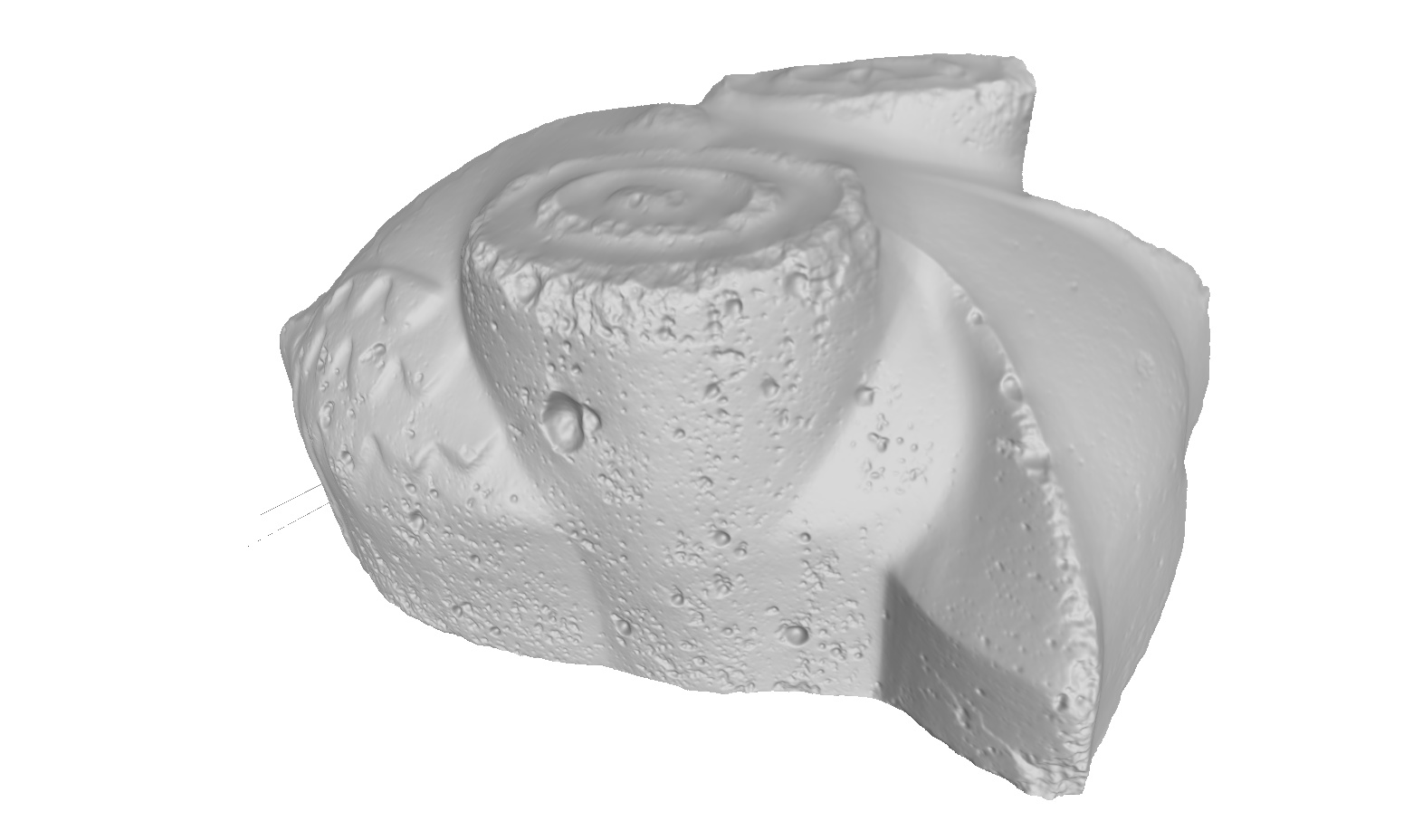} &
\includegraphics[width=0.15\textwidth,trim={5cm 0cm 6cm  0cm},clip]{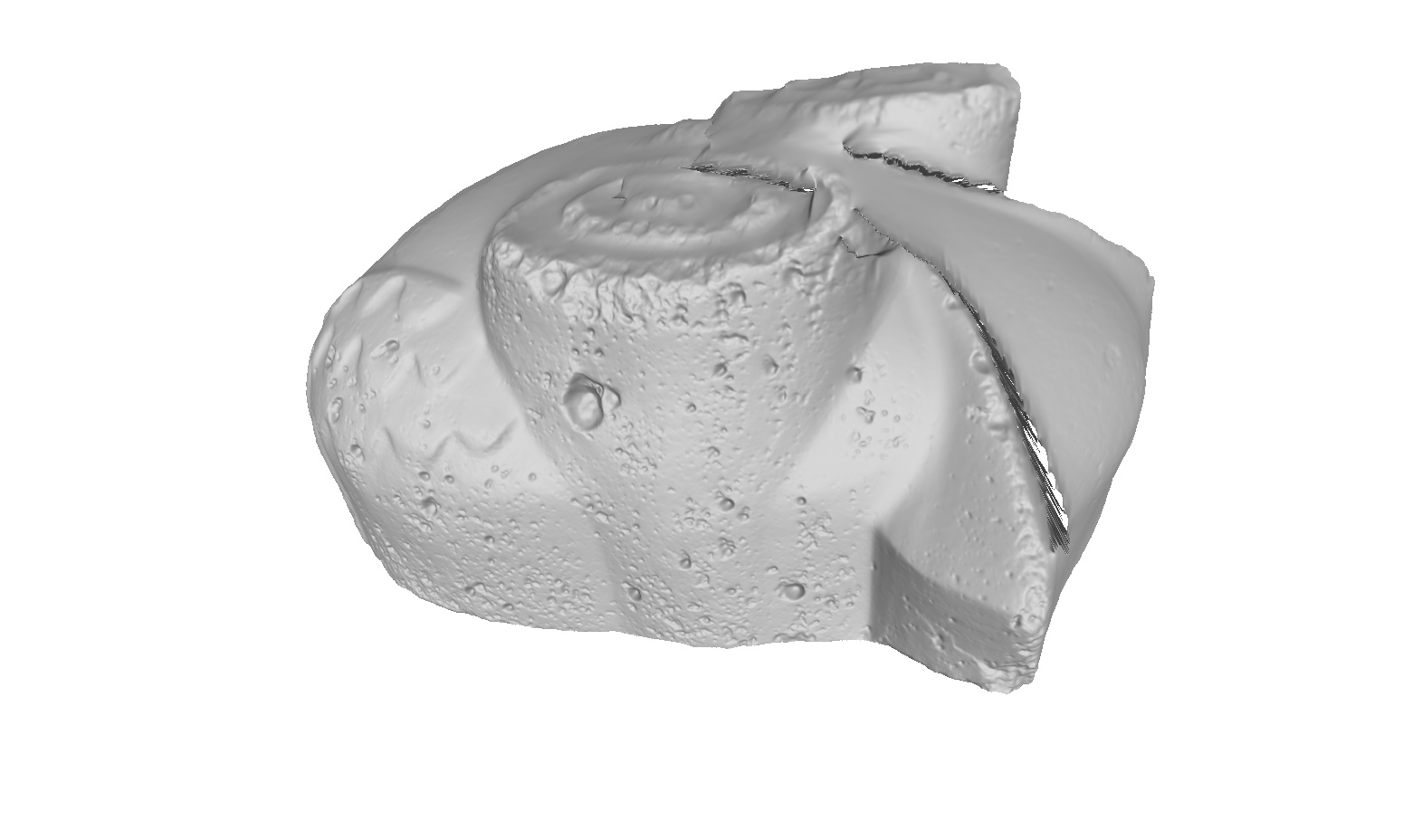} &
\includegraphics[width=0.15\textwidth,trim={5cm 0cm 6cm  0cm},clip]{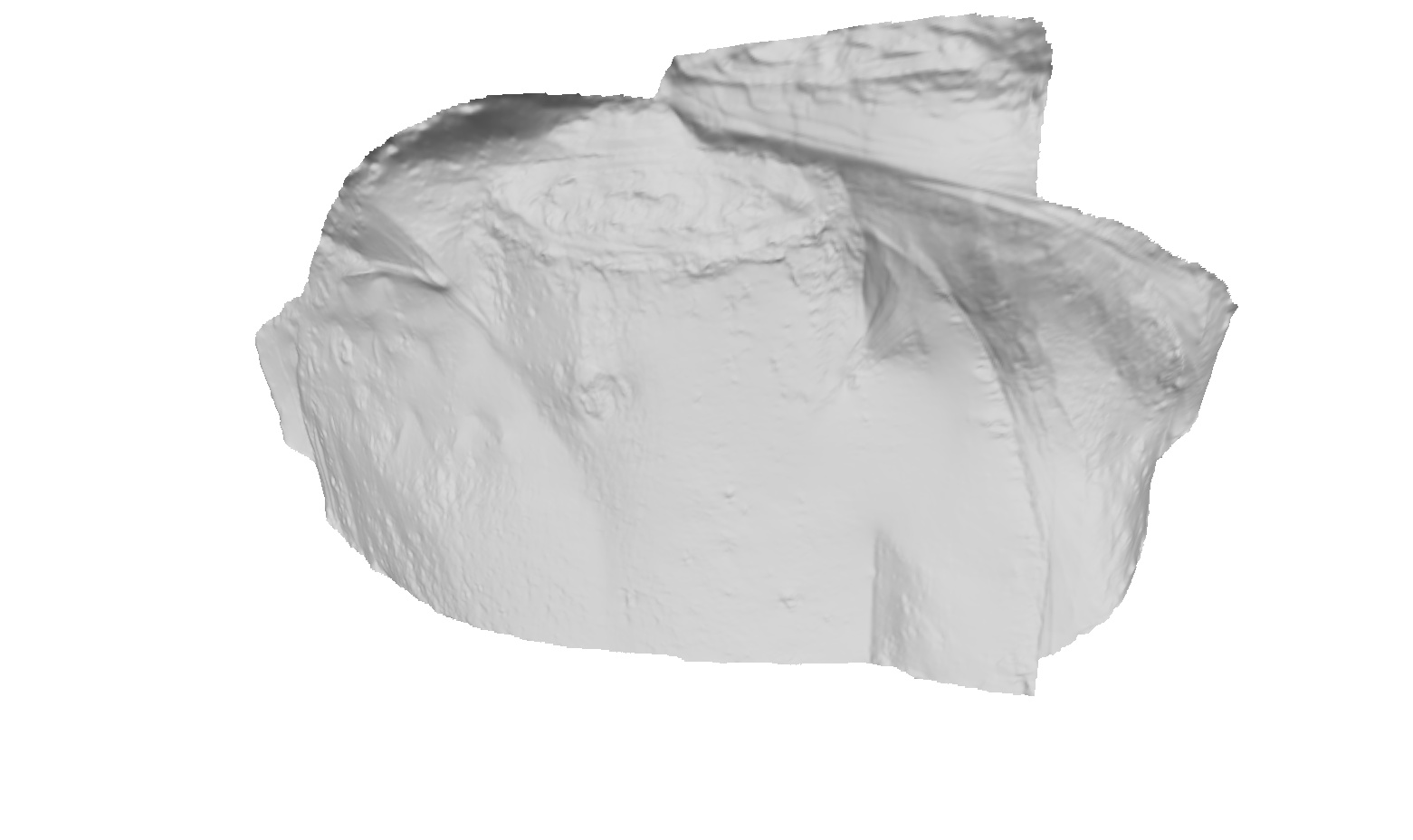} &
\includegraphics[width=0.15\textwidth,trim={5cm 0cm 6cm  0cm},clip]{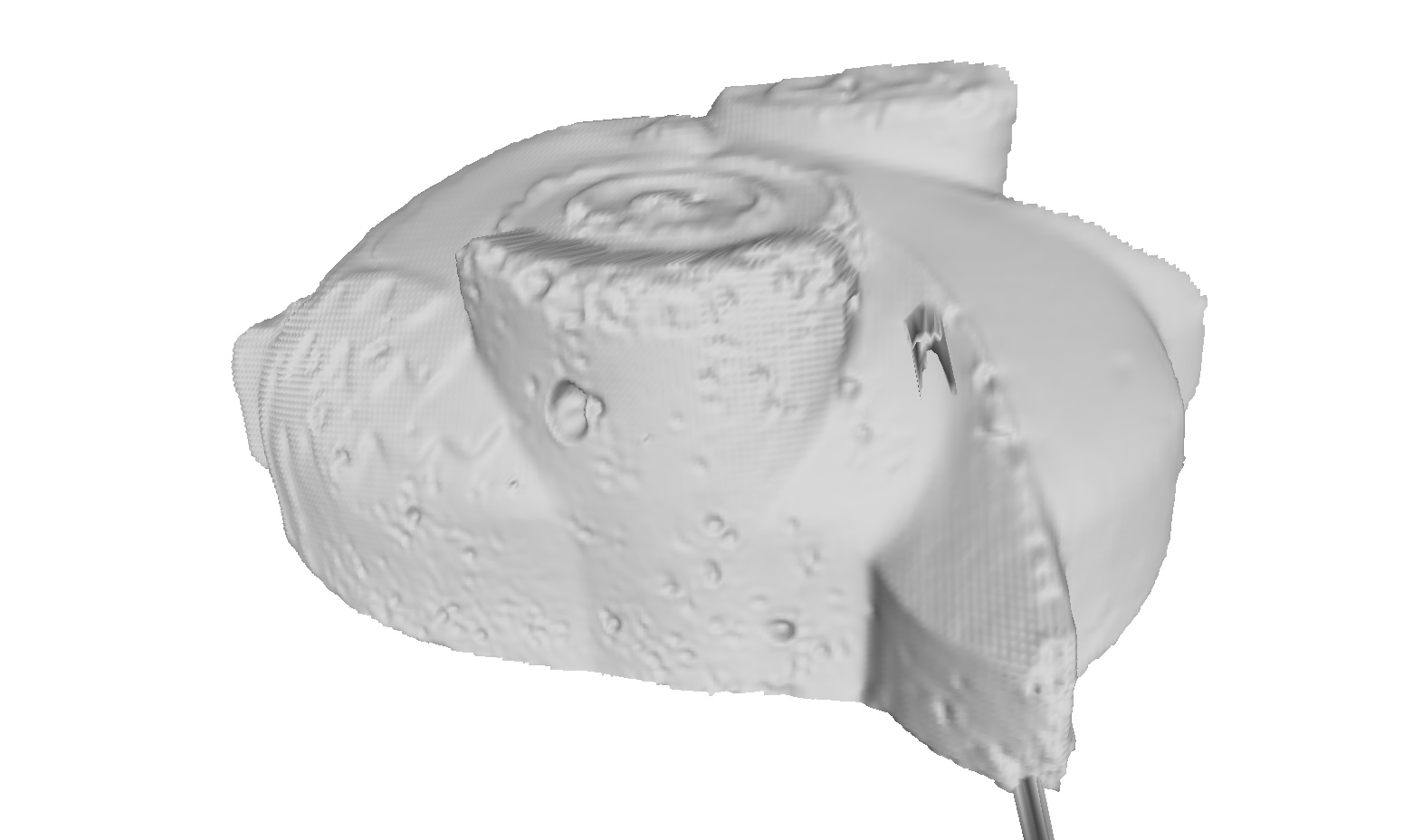} &
\includegraphics[width=0.15\textwidth,trim={5cm 0cm 6cm  0cm},clip]{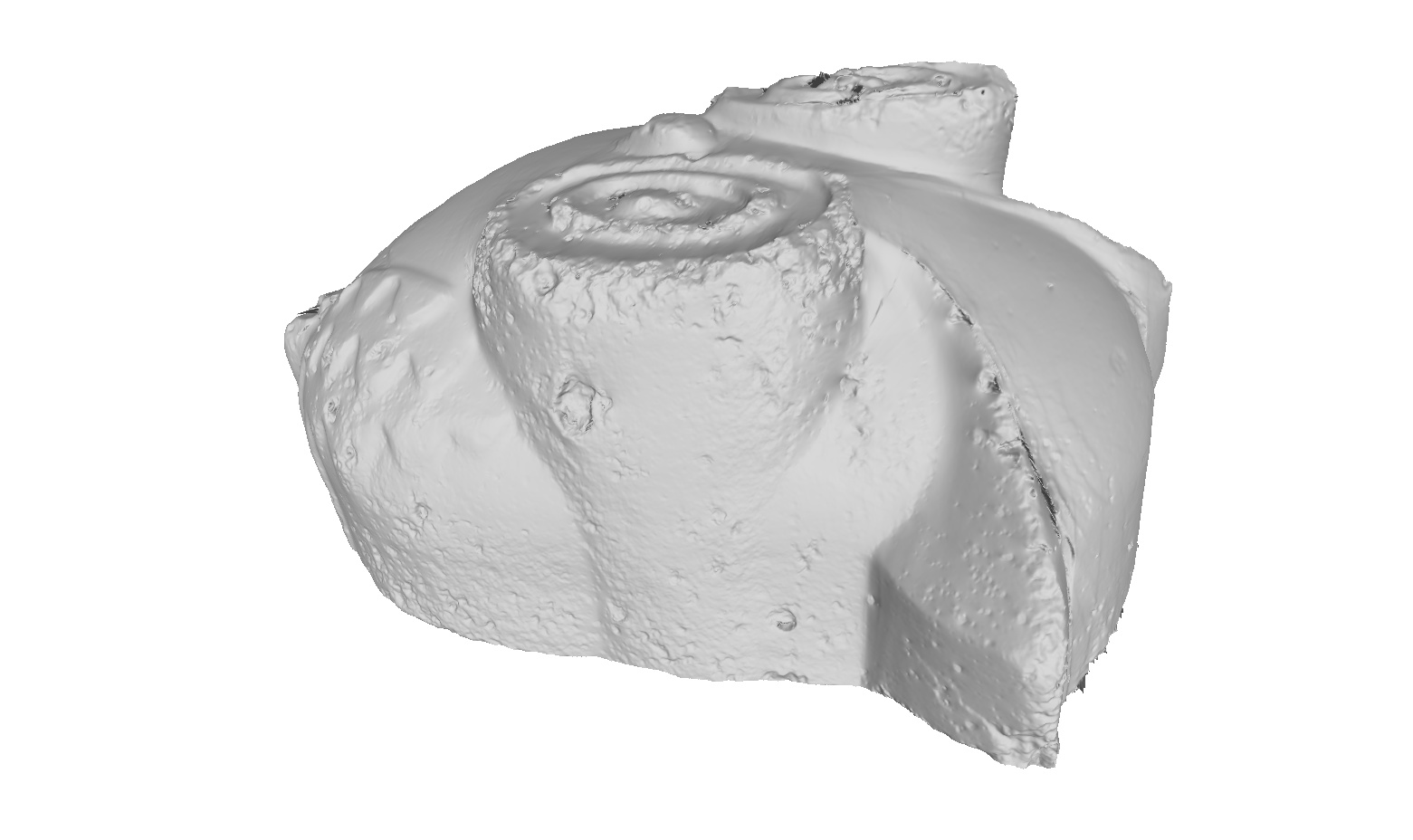} \\
\begin{sideways} {$Z$ Error (mm)} \end{sideways} &
\includegraphics[width=0.15\textwidth,trim={1cm 0cm 1cm  0cm},clip]{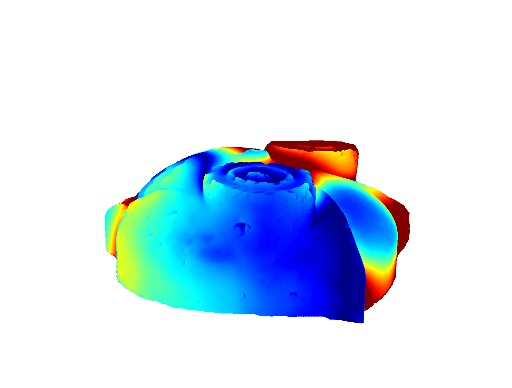} &
\includegraphics[width=0.15\textwidth,trim={1cm 0cm 1cm  0cm},clip]{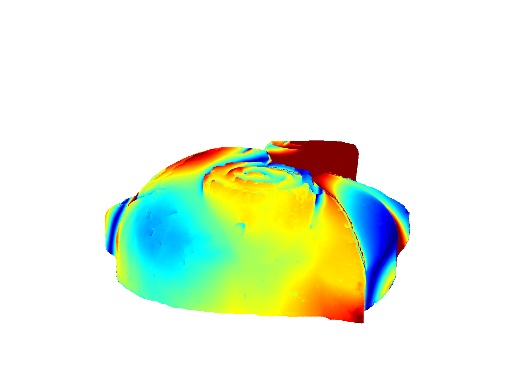} &
\includegraphics[width=0.15\textwidth,trim={1cm 0cm 1cm  0cm},clip]{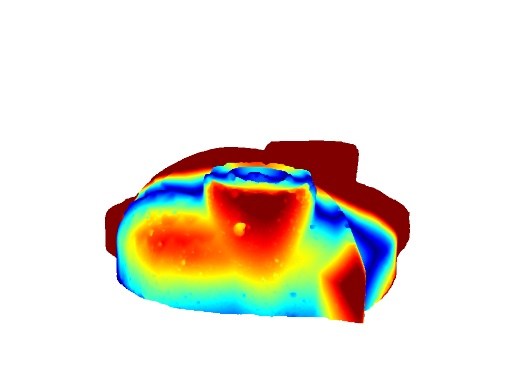} &
\includegraphics[width=0.15\textwidth,trim={1cm 0cm 1cm  0cm},clip]{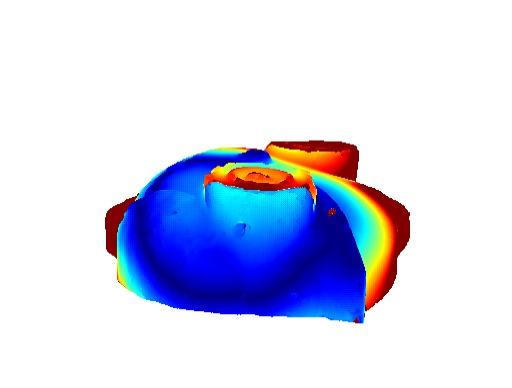} &
\includegraphics[width=0.15\textwidth,trim={1cm 0cm 1cm  0cm},clip]{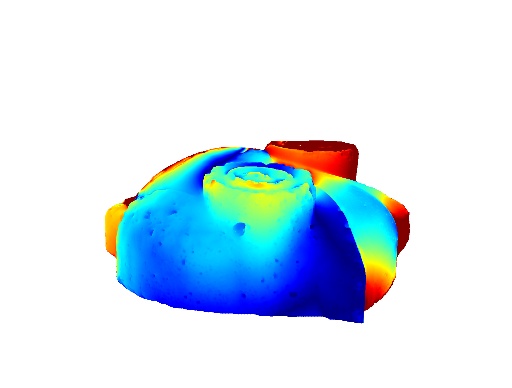} \\
%

\begin{sideways} {Jar-3D Shape} \end{sideways} &
\includegraphics[width=0.15\textwidth,trim={7cm 0cm 6cm  0cm},clip]{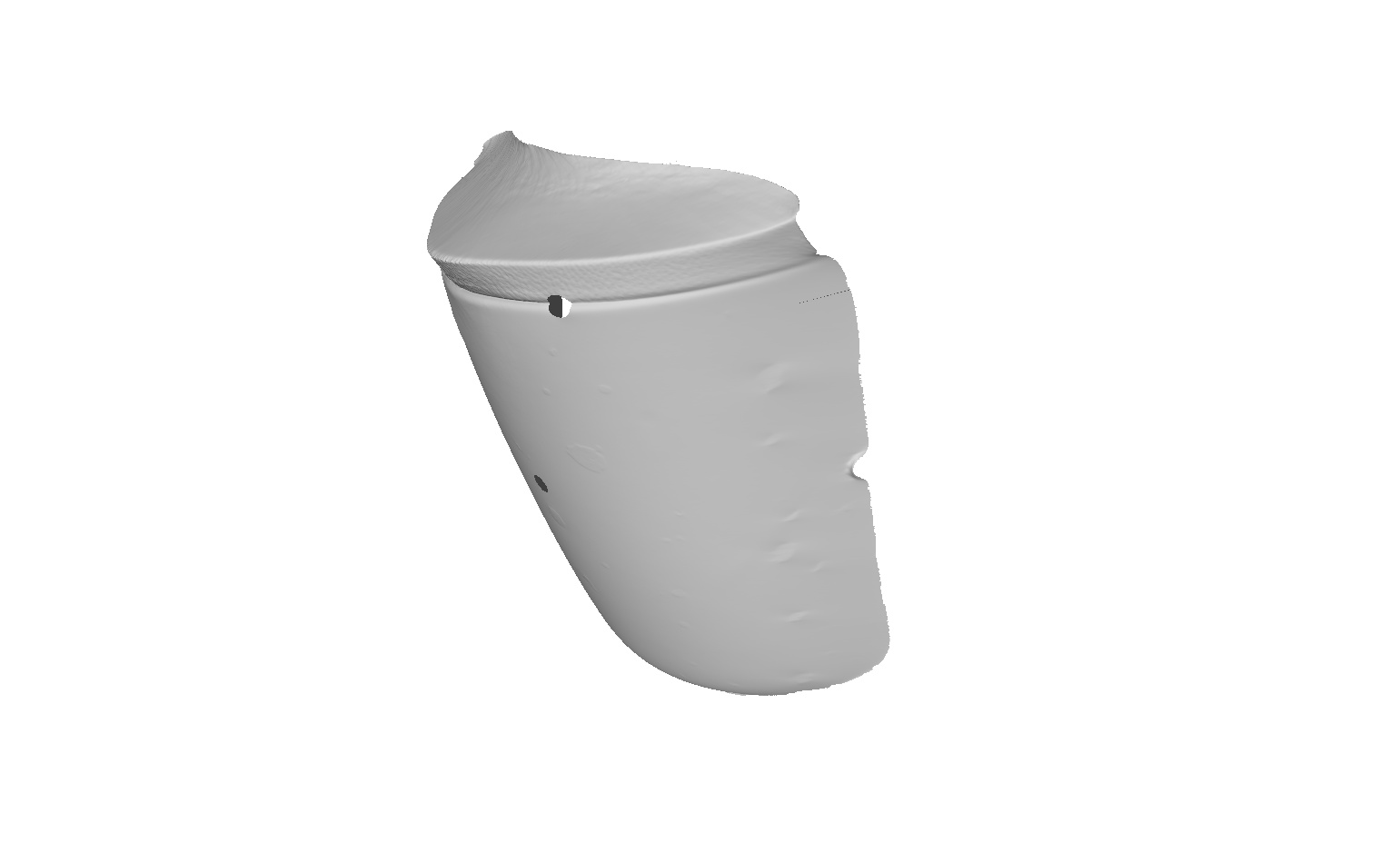} &
\includegraphics[width=0.15\textwidth,trim={7cm 0cm 6cm  0cm},clip]{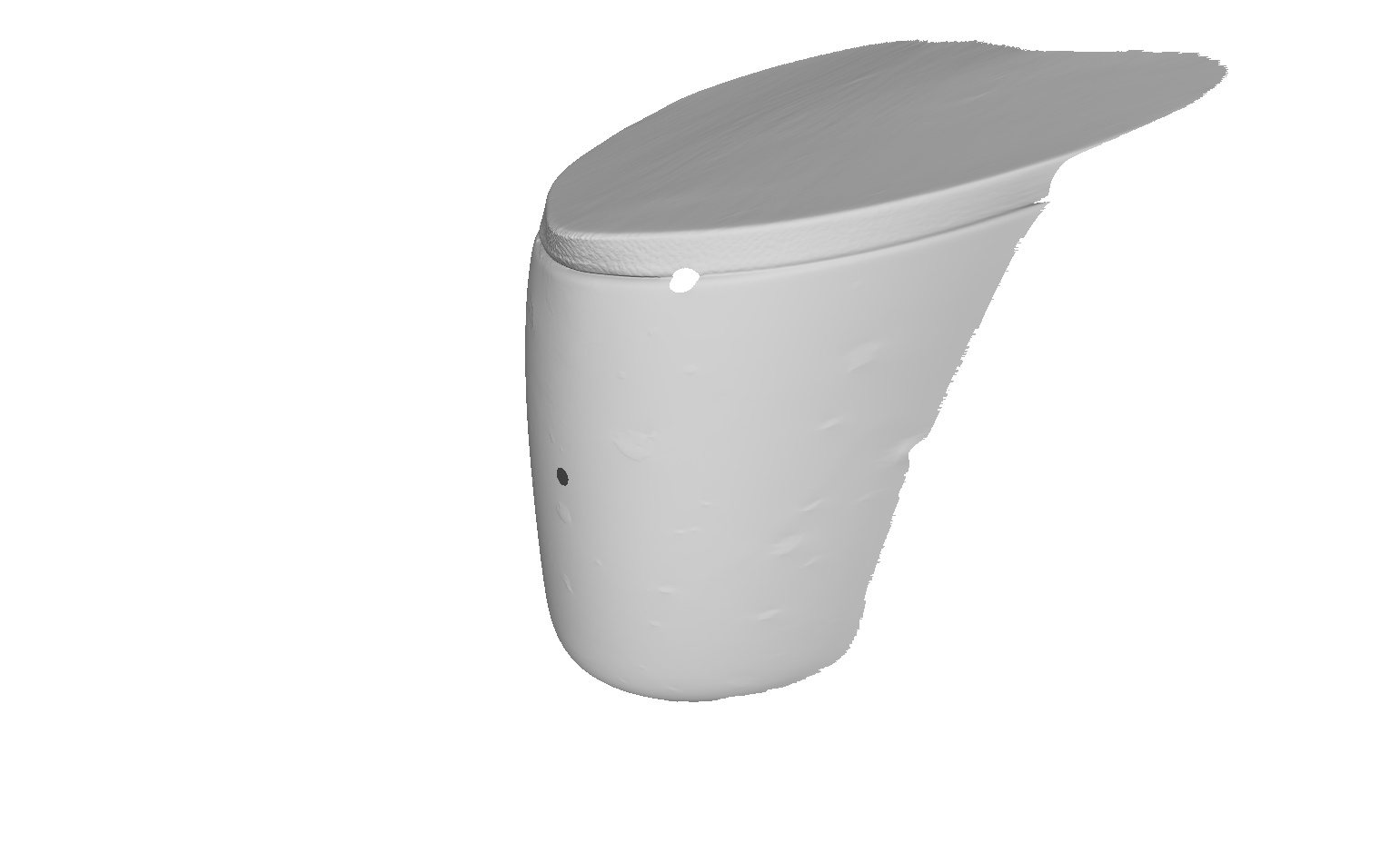} &
\includegraphics[width=0.15\textwidth,trim={7cm 0cm 6cm  0cm},clip]{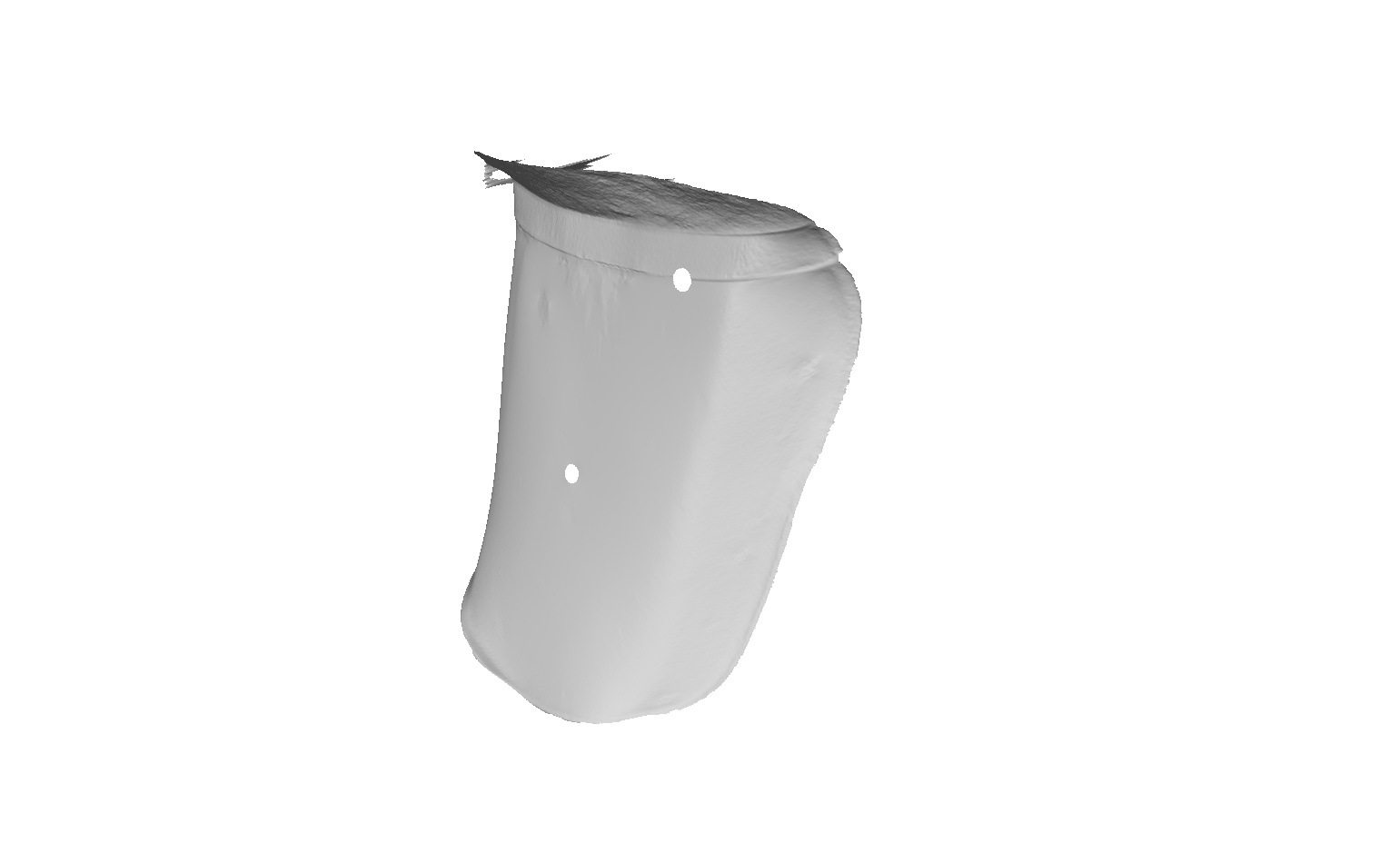} &
\includegraphics[width=0.15\textwidth,trim={7cm 0cm 6cm  0cm},clip]{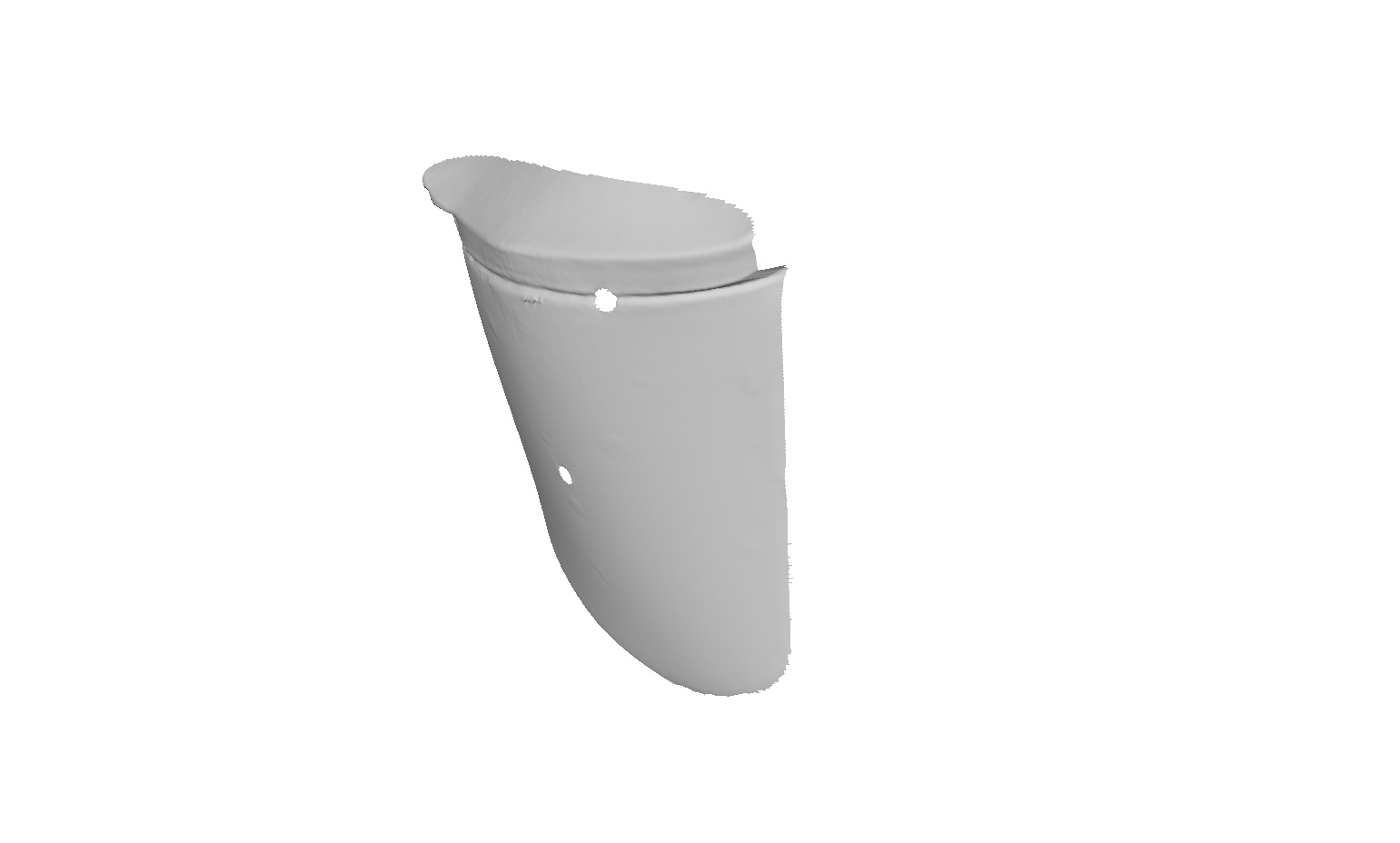} &
\includegraphics[width=0.07\textwidth,trim={5cm 0cm 3cm  0cm},clip]{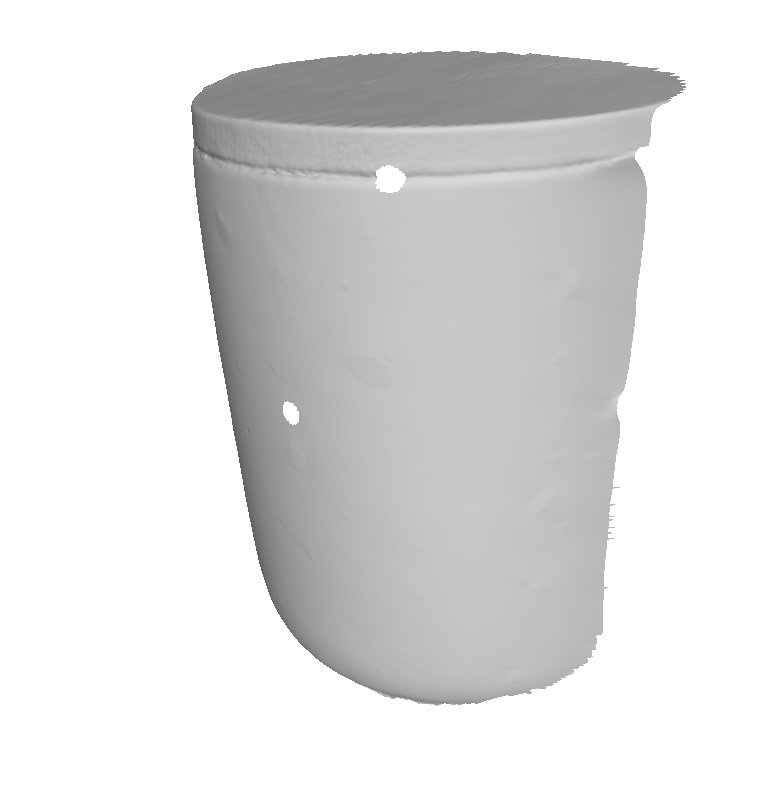} \\

\begin{sideways} {$Z$ Error (mm)} \end{sideways} &
\includegraphics[width=0.15\textwidth,trim={1cm 0cm 2cm  0cm},clip]{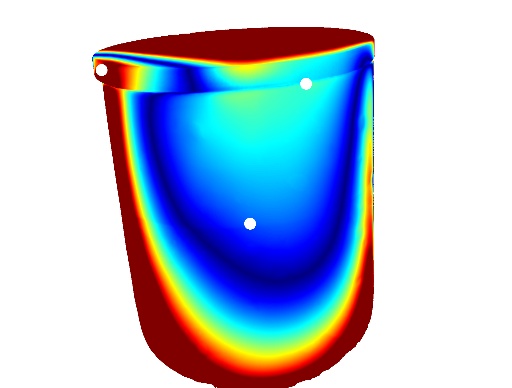} &
\includegraphics[width=0.15\textwidth,trim={1cm 0cm 2cm  0cm},clip]{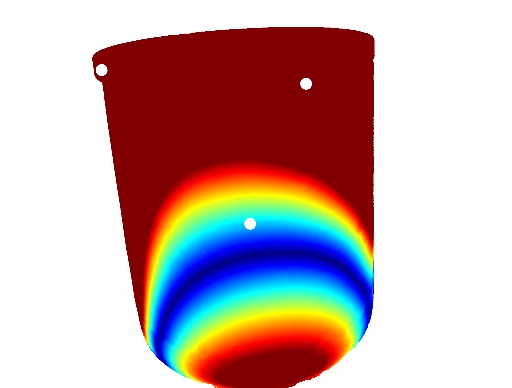} &
\includegraphics[width=0.15\textwidth,trim={1cm 0cm 2cm  0cm},clip]{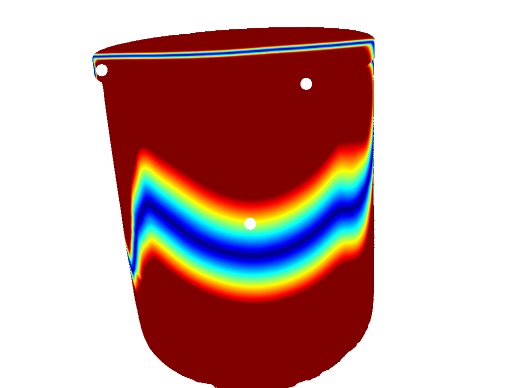} &
\includegraphics[width=0.15\textwidth,trim={1cm 0cm 2cm  0cm},clip]{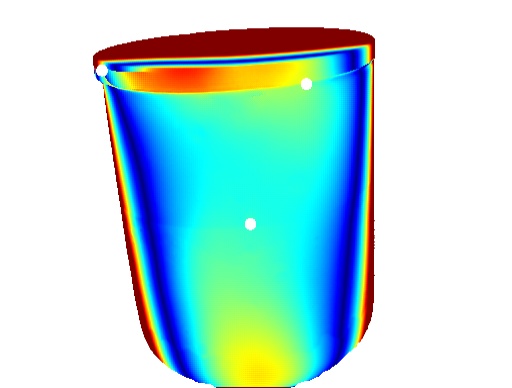} &
\includegraphics[width=0.15\textwidth,trim={1cm 0cm 2cm  0cm},clip]{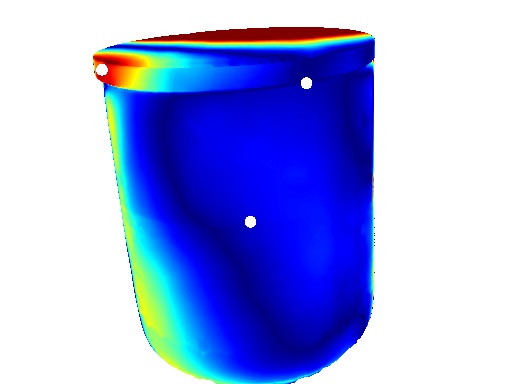} \\

\begin{sideways} {Queen-3D Shape} \end{sideways} &
\includegraphics[width=0.15\textwidth,trim={14cm 0cm 14cm  0cm},clip]{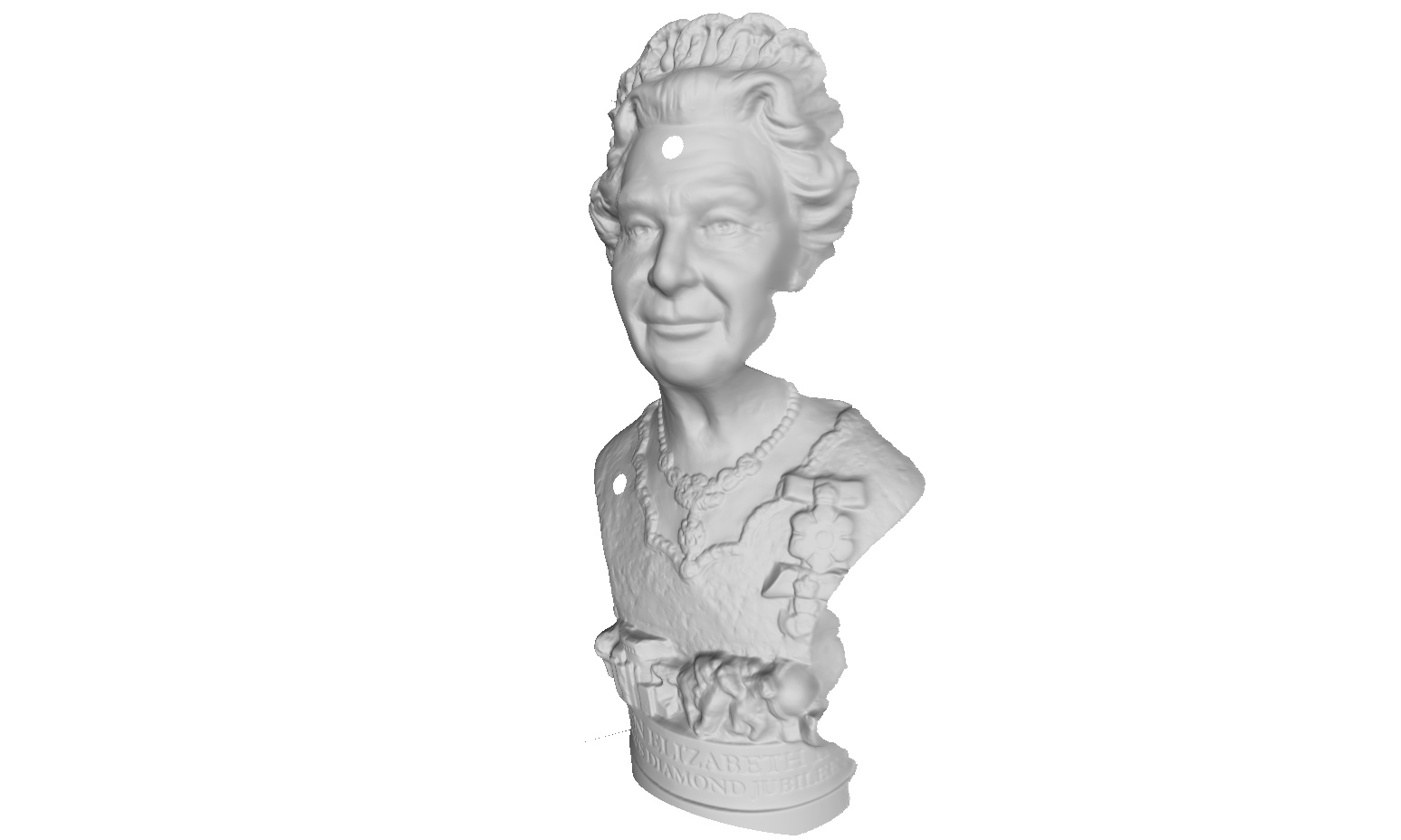} &
\includegraphics[width=0.15\textwidth,trim={14cm 0cm 14cm  0cm},clip]{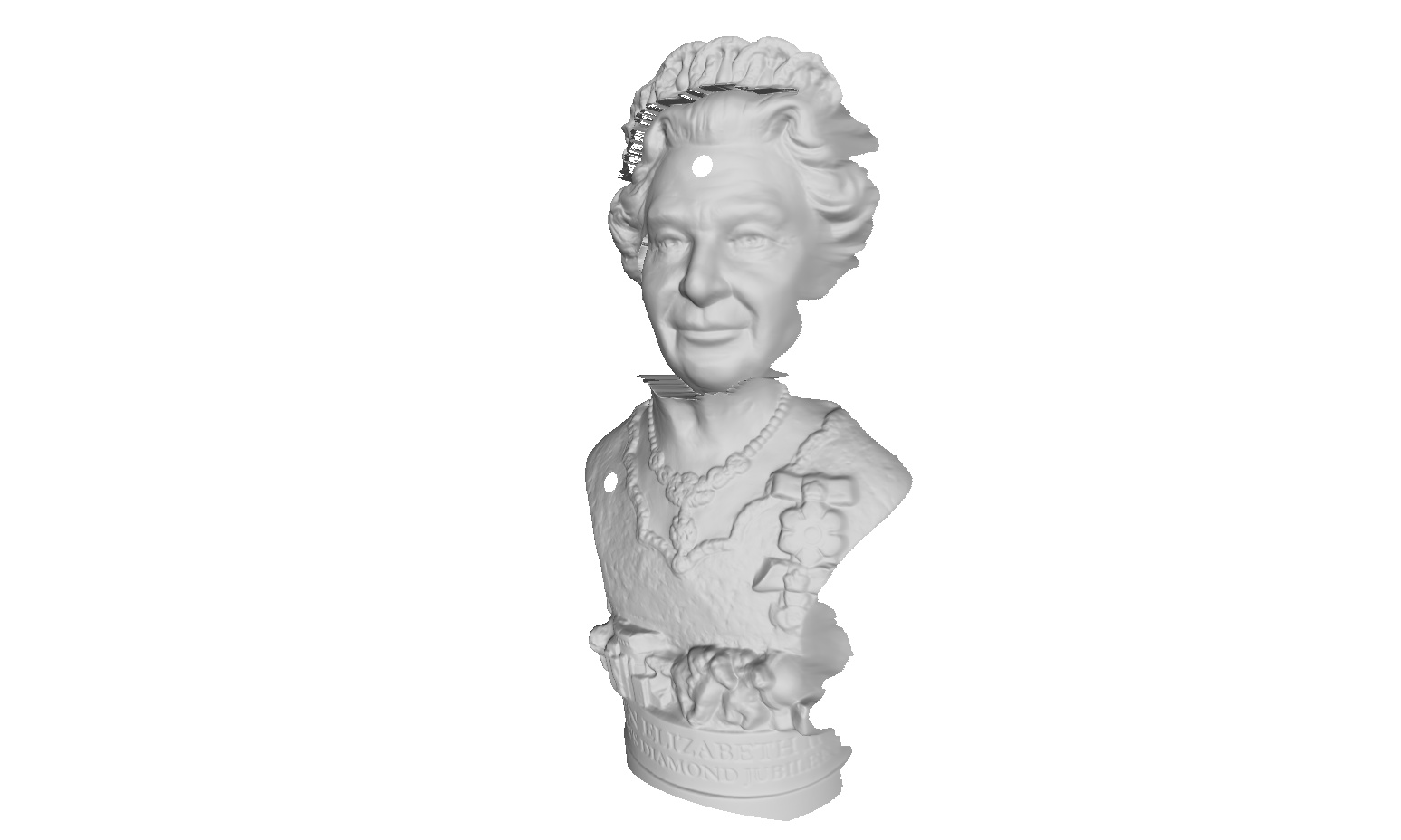} &
\includegraphics[width=0.15\textwidth,trim={14cm 0cm 14cm  0cm},clip]{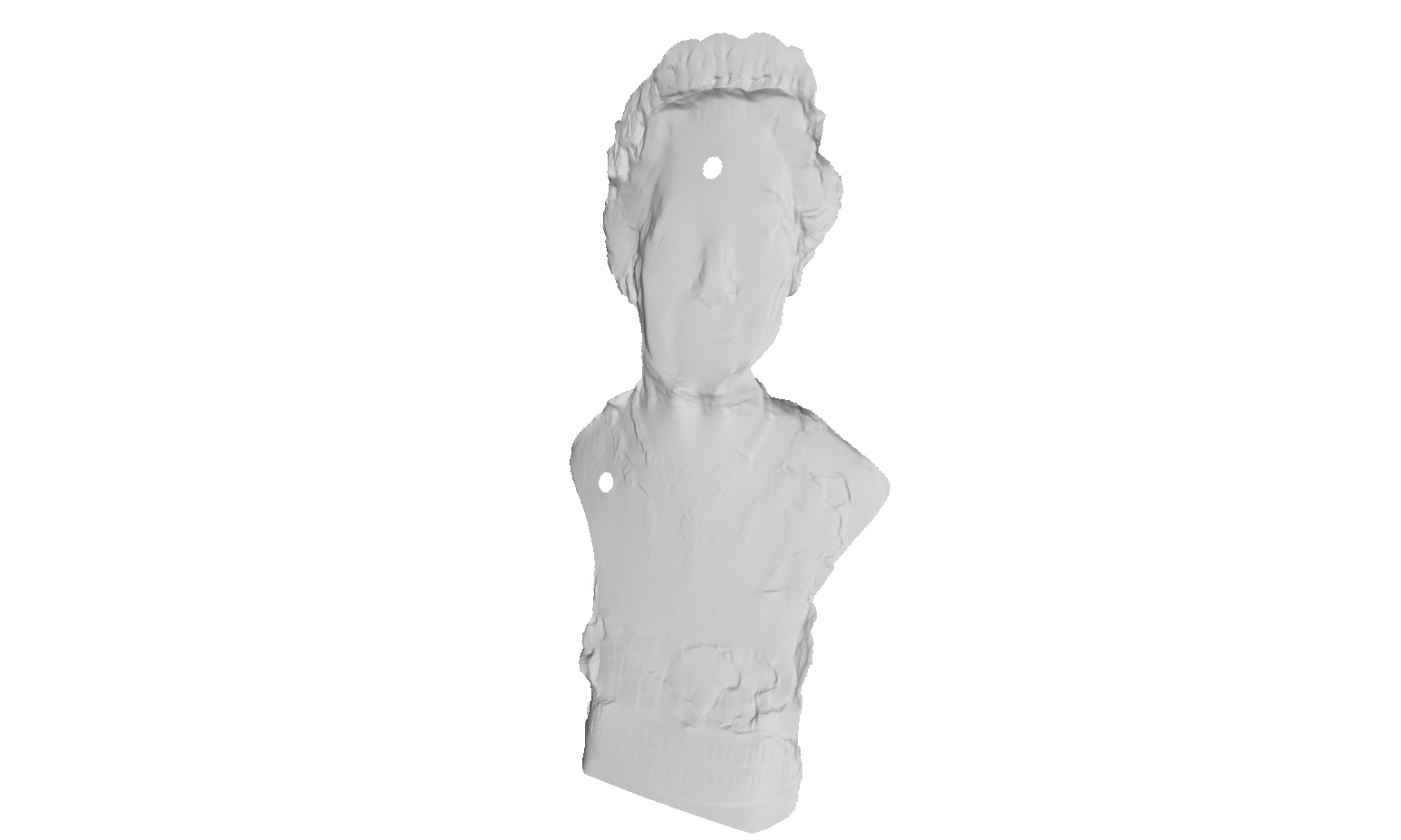} &
\includegraphics[width=0.15\textwidth,trim={14cm 0cm 14cm  0cm},clip]{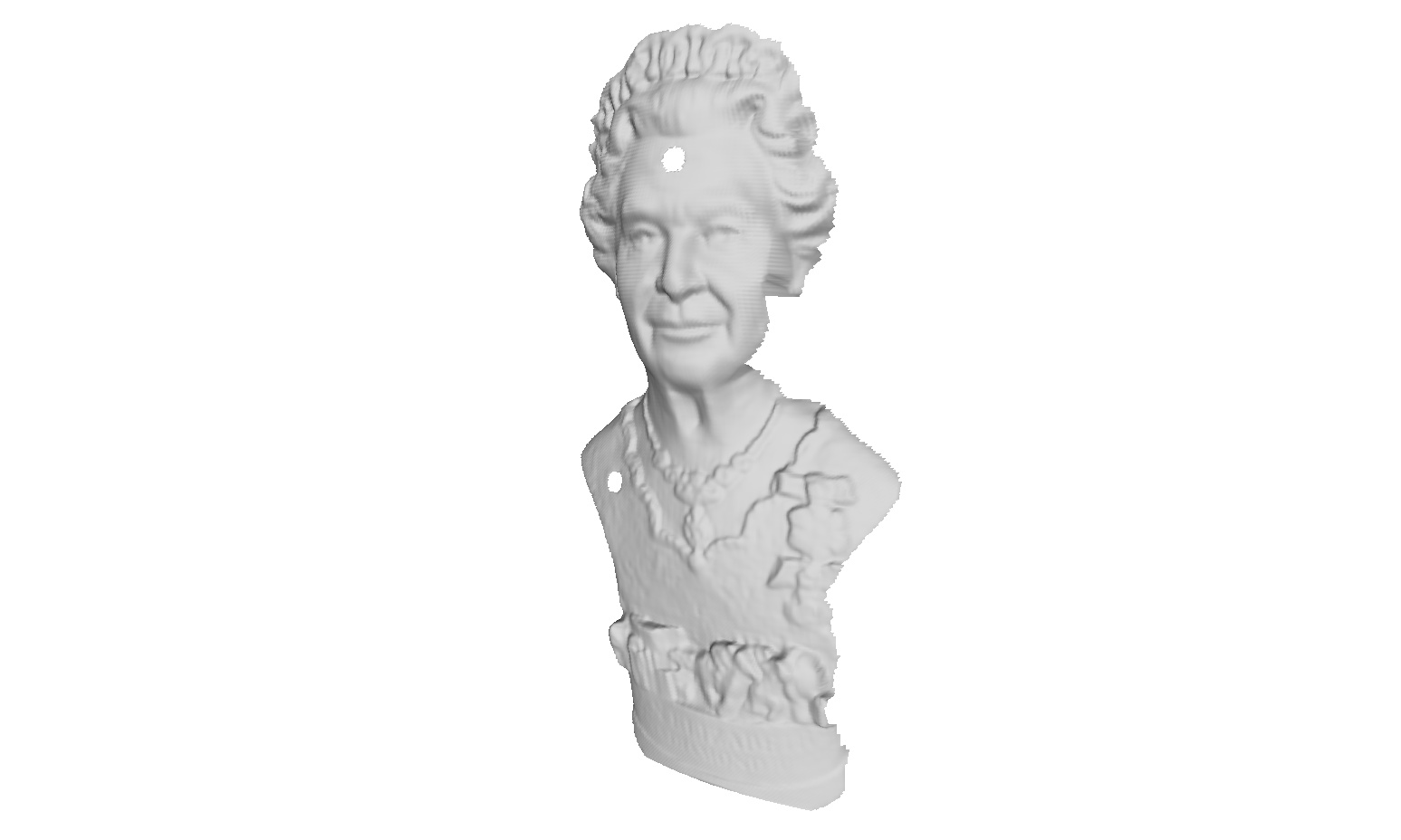} &
\includegraphics[width=0.15\textwidth,trim={14cm 0cm 14cm  0cm},clip]{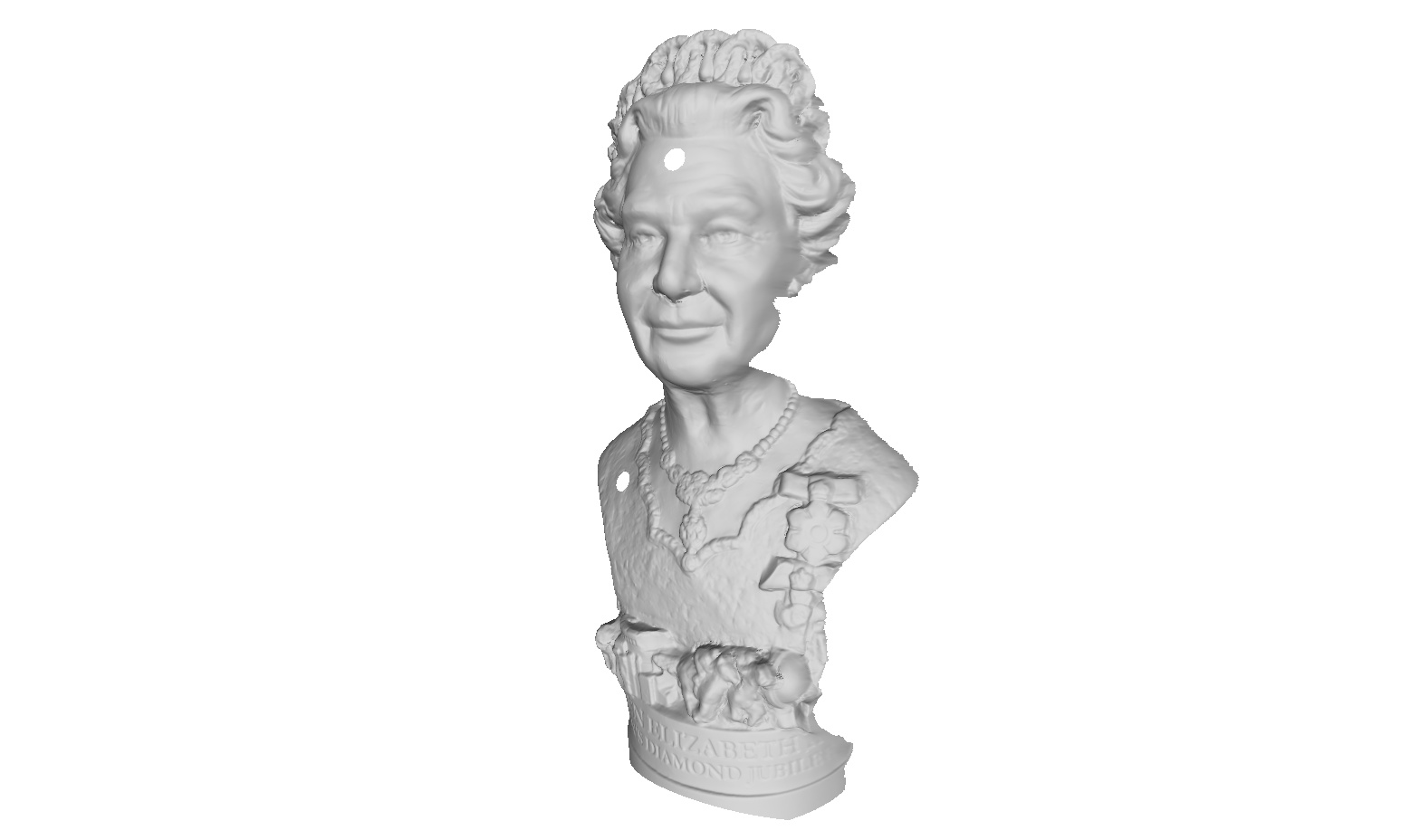} \\
\begin{sideways} {$Z$ Error (mm)} \end{sideways} &
\includegraphics[width=0.15\textwidth,trim={2cm 0cm 2cm  0cm},clip]{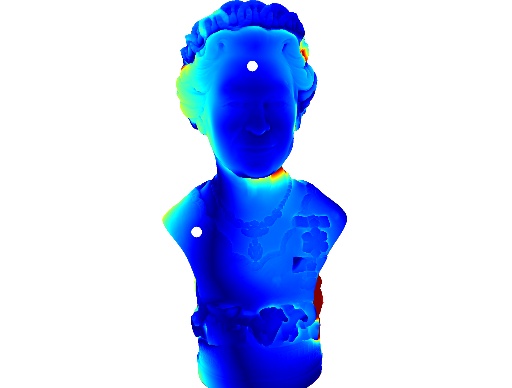} &
\includegraphics[width=0.15\textwidth,trim={2cm 0cm 2cm  0cm},clip]{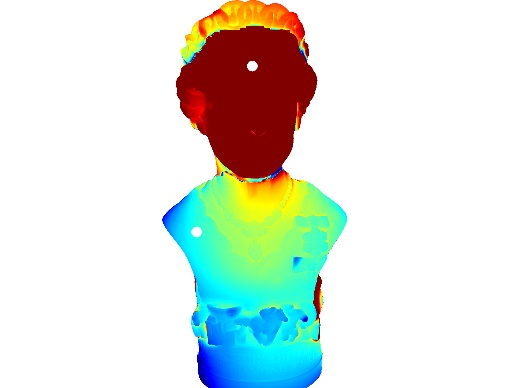} &
\includegraphics[width=0.15\textwidth,trim={2cm 0cm 2cm  0cm},clip]{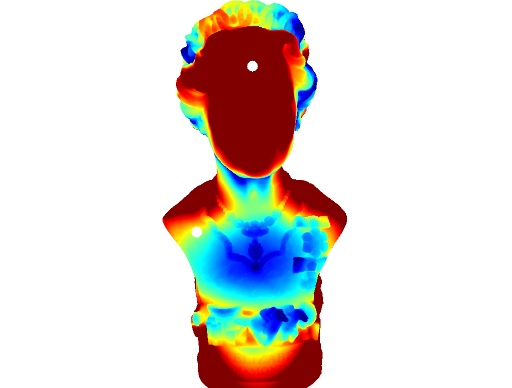} &
\includegraphics[width=0.15\textwidth,trim={2cm 0cm 2cm  0cm},clip]{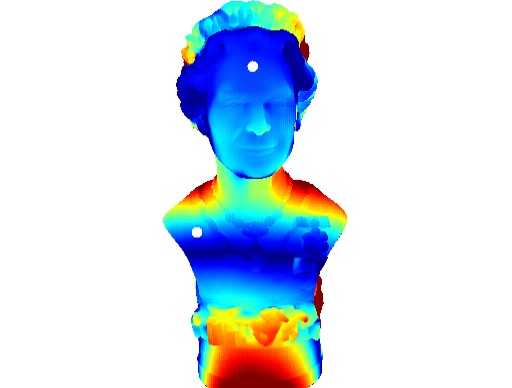} &
\includegraphics[width=0.15\textwidth,trim={2cm 0cm 2cm  0cm},clip]{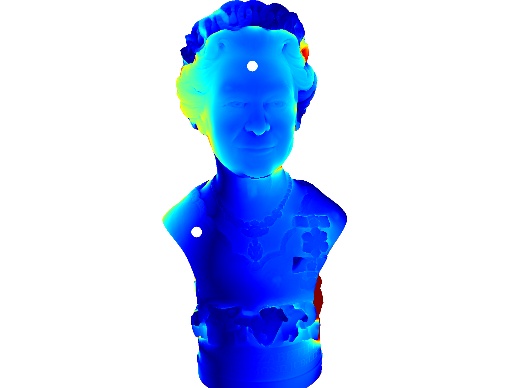} \\
%
%
\begin{sideways} {Squirrel-3D Shape} \end{sideways} &
\includegraphics[width=0.15\textwidth,trim={15cm 0cm 12cm  0cm},clip]{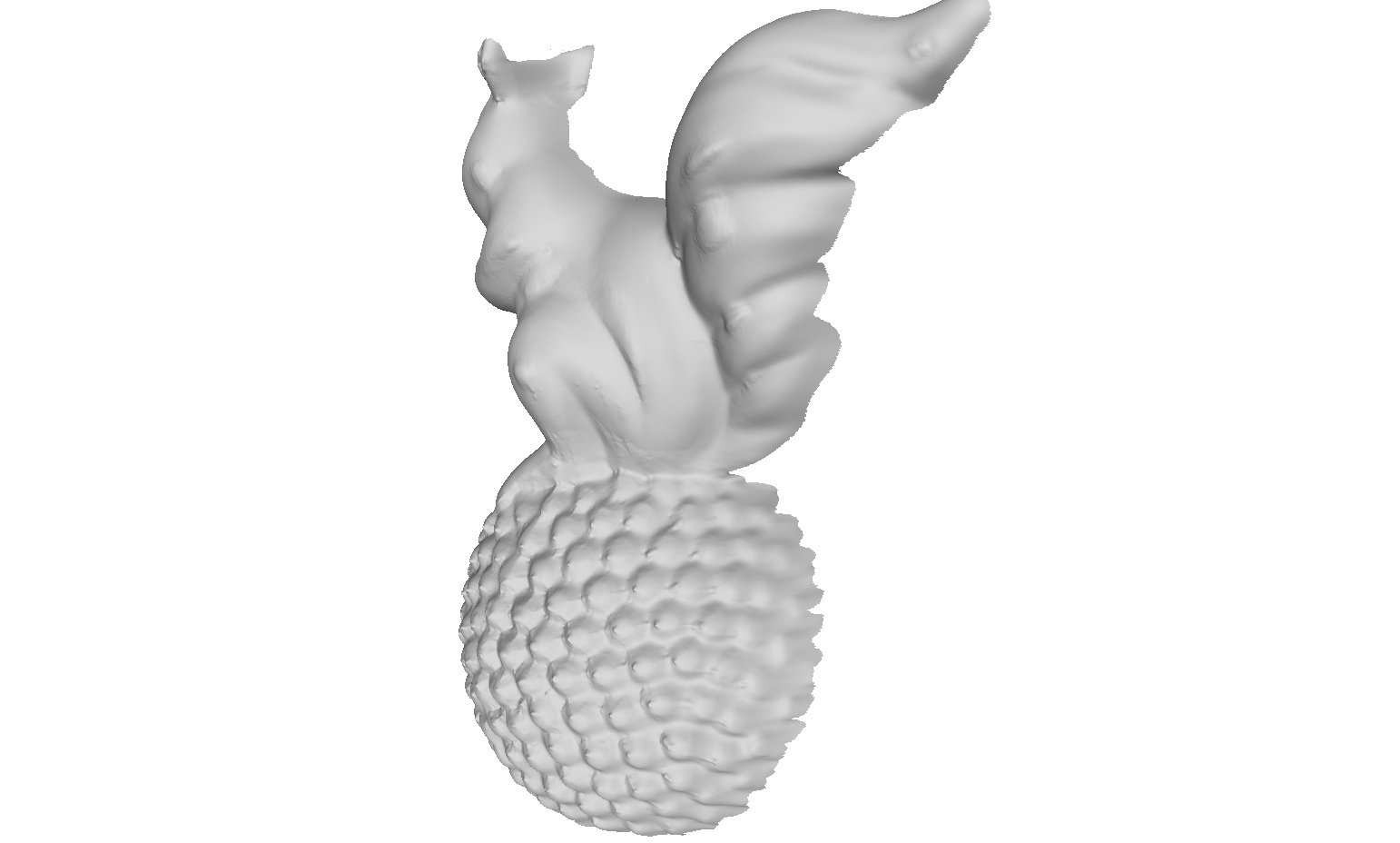} &
\includegraphics[width=0.15\textwidth,trim={15cm 0cm 12cm  0cm},clip]{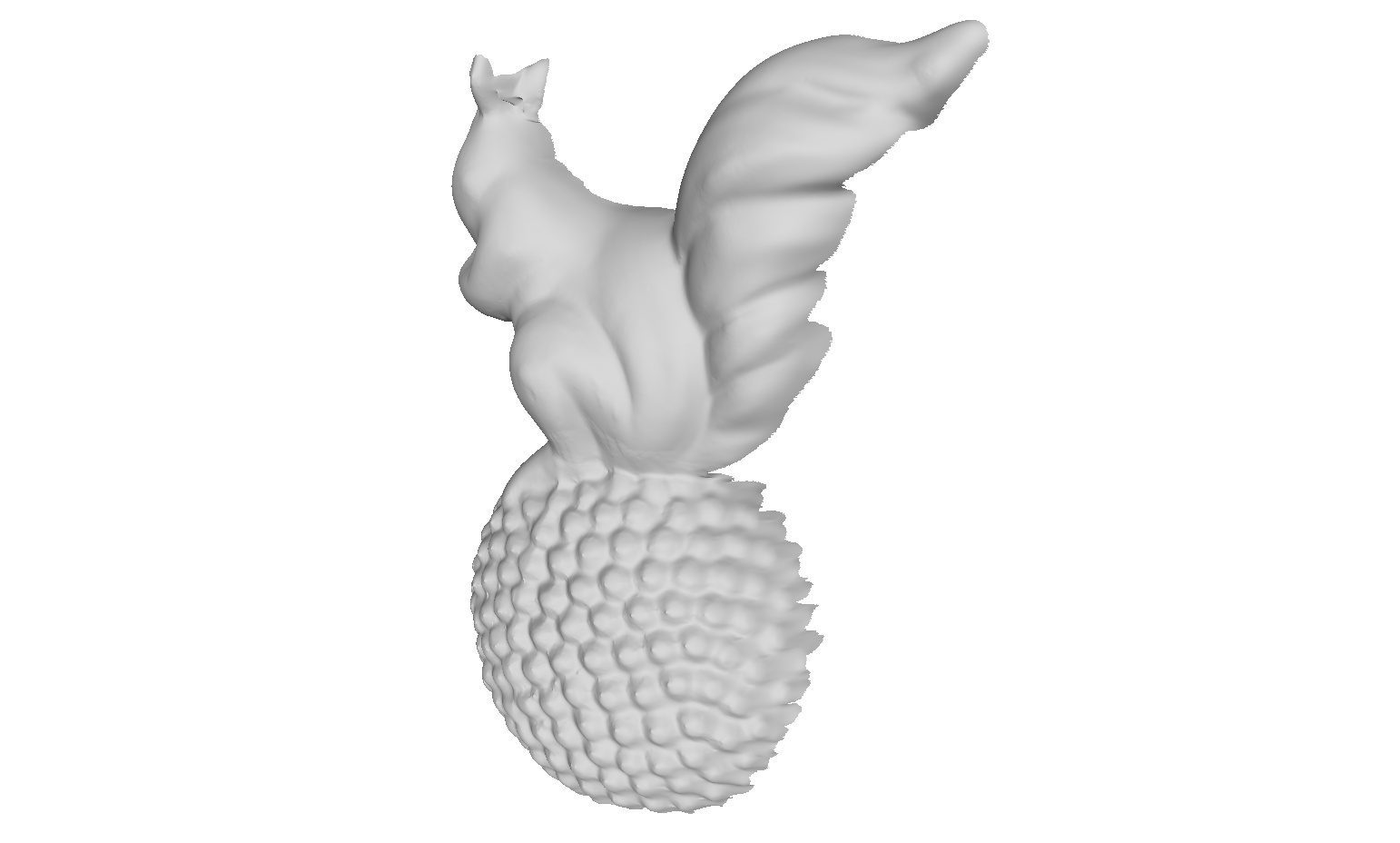} &
\includegraphics[width=0.15\textwidth,trim={15cm 0cm 12cm  0cm},clip]{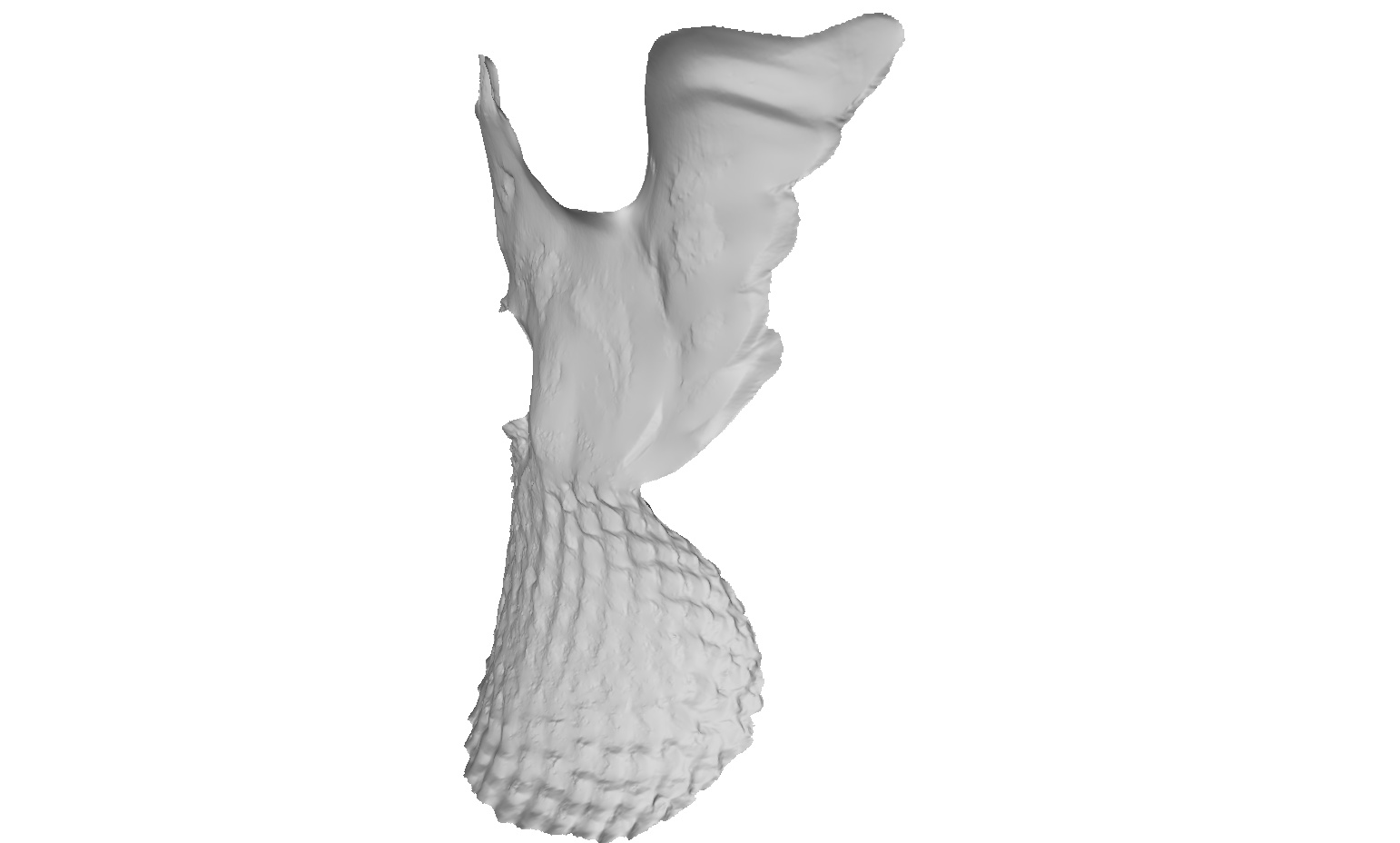} &
\includegraphics[width=0.15\textwidth,trim={15cm 0cm 12cm  0cm},clip]{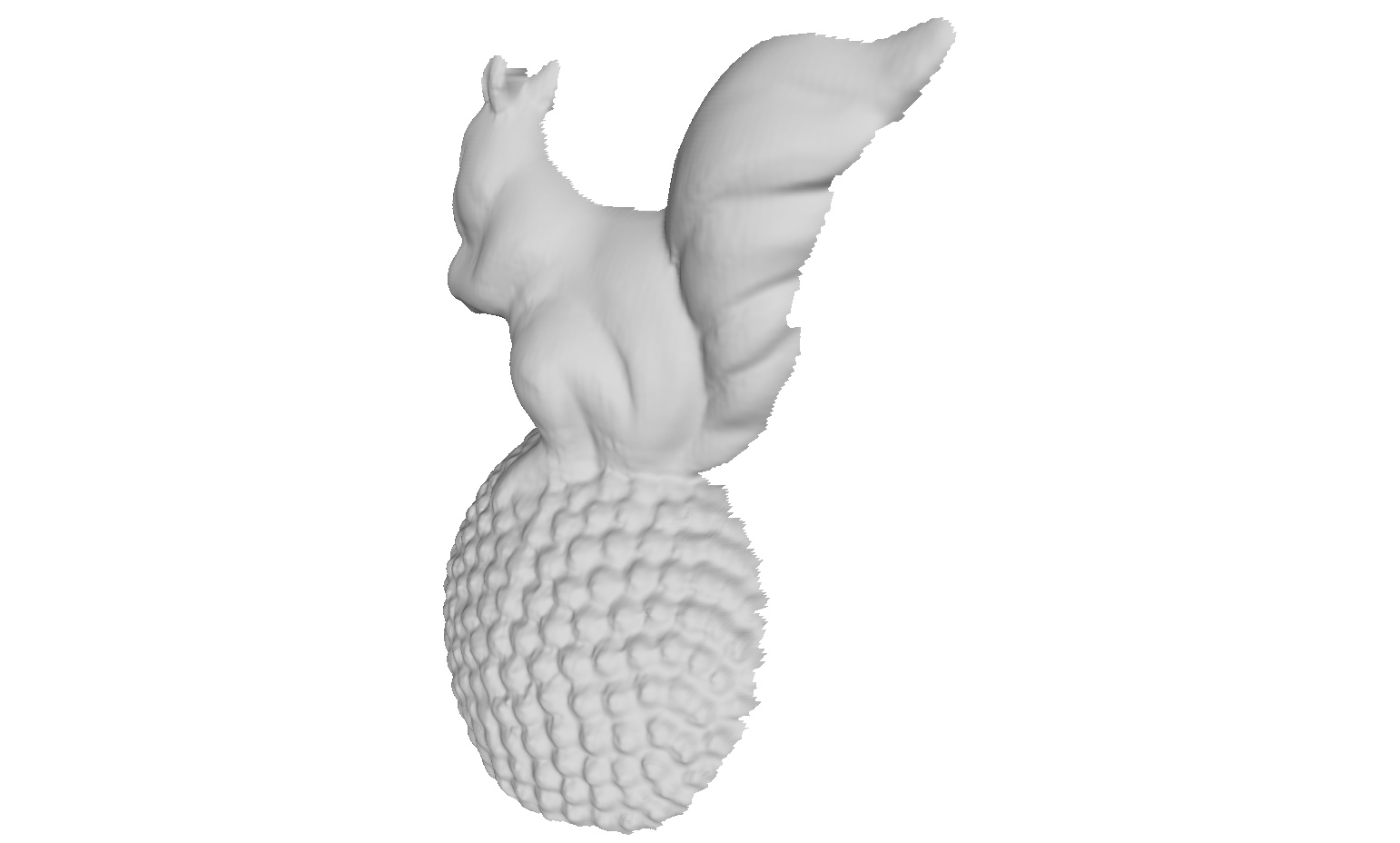} &
\includegraphics[width=0.15\textwidth,trim={15cm 0cm 12cm  0cm},clip]{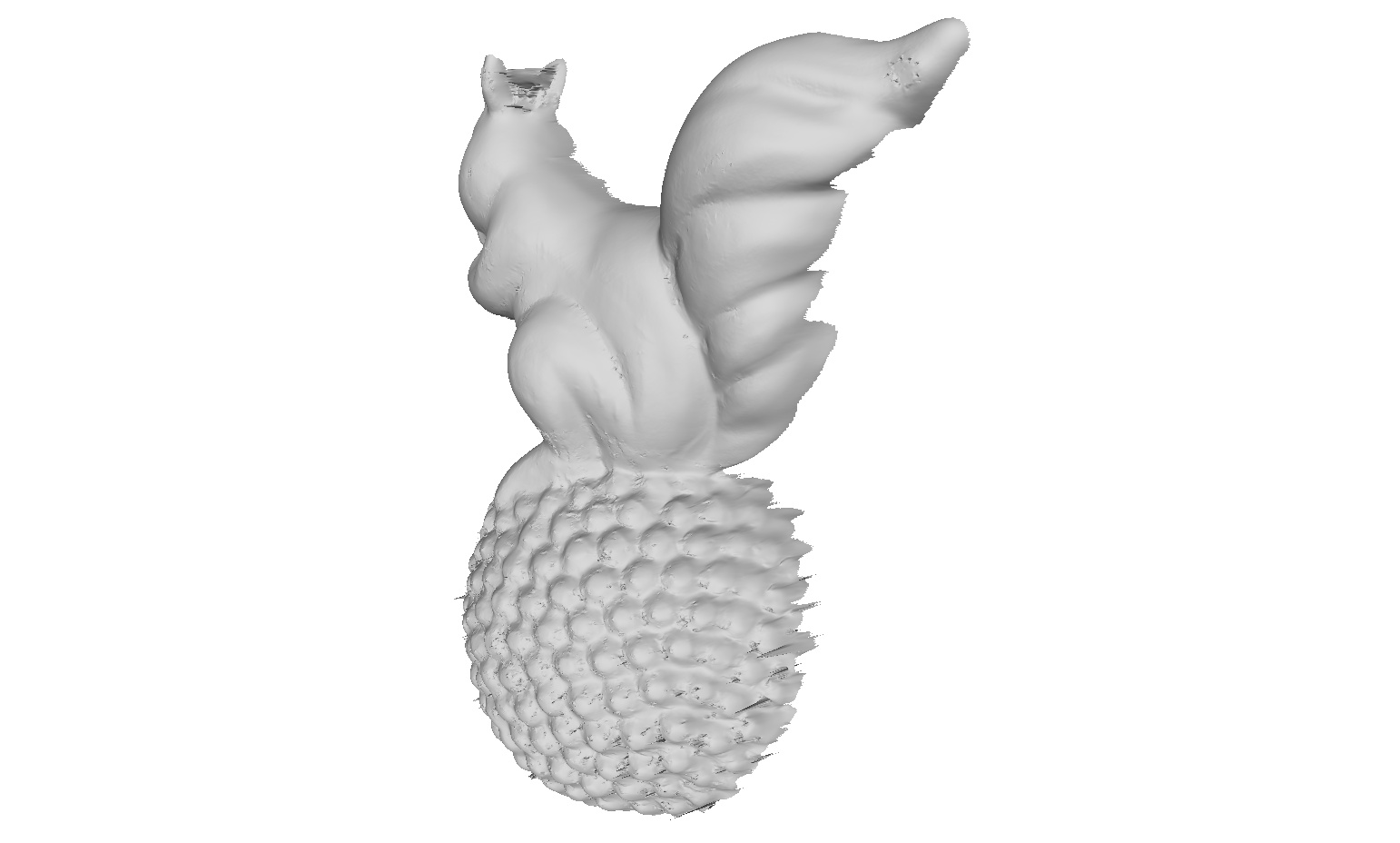} \\
\begin{sideways} {$Z$ Error (mm)} \end{sideways} &
\includegraphics[width=0.15\textwidth,trim={2cm 0cm 2cm  0cm},clip]{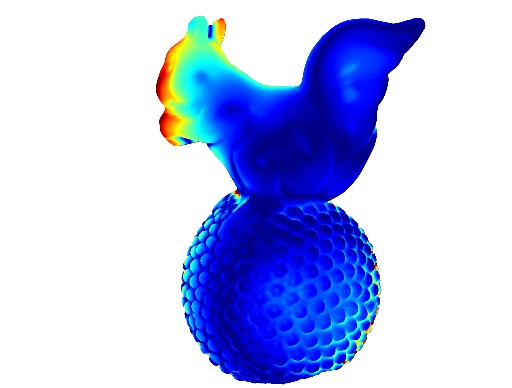} &
\includegraphics[width=0.15\textwidth,trim={2cm 0cm 2cm  0cm},clip]{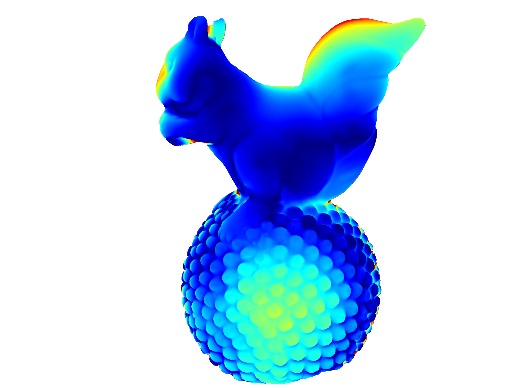} &
\includegraphics[width=0.15\textwidth,trim={2cm 0cm 2cm  0cm},clip]{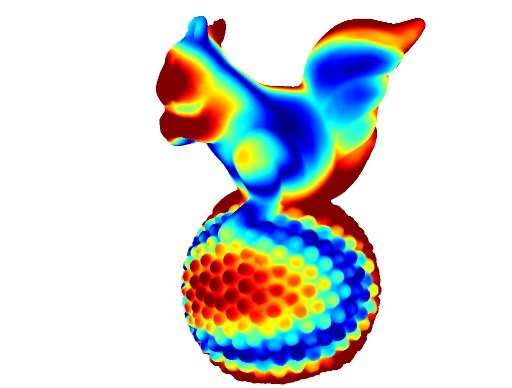} &
\includegraphics[width=0.15\textwidth,trim={2cm 0cm 2cm  0cm},clip]{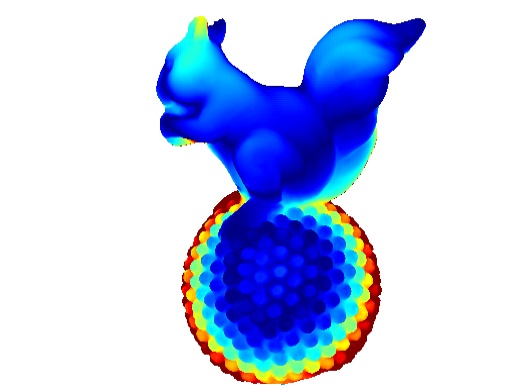} &
\includegraphics[width=0.15\textwidth,trim={2cm 0cm 2cm  0cm},clip]{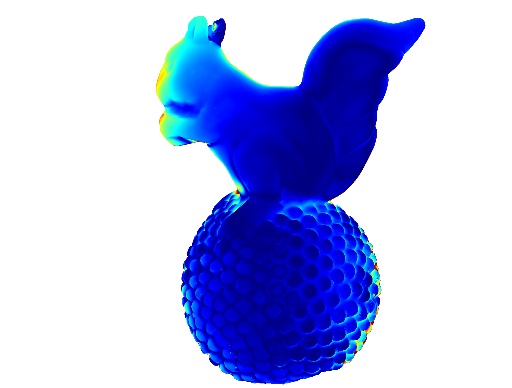} \\
\end{tabular}
\caption{Evaluations 9-12/14 }
\label{fig:eva5}
\end{figure*}

\begin{figure*}[!t]
\begin{tabular}{c c c c c c}
~ & L17 & Q18 & I18 & S20 & L20 \\
\begin{sideways} {Bowl-3D Shape} \end{sideways} &
\includegraphics[width=0.15\textwidth,trim={10cm 0cm 10cm  0cm},clip]{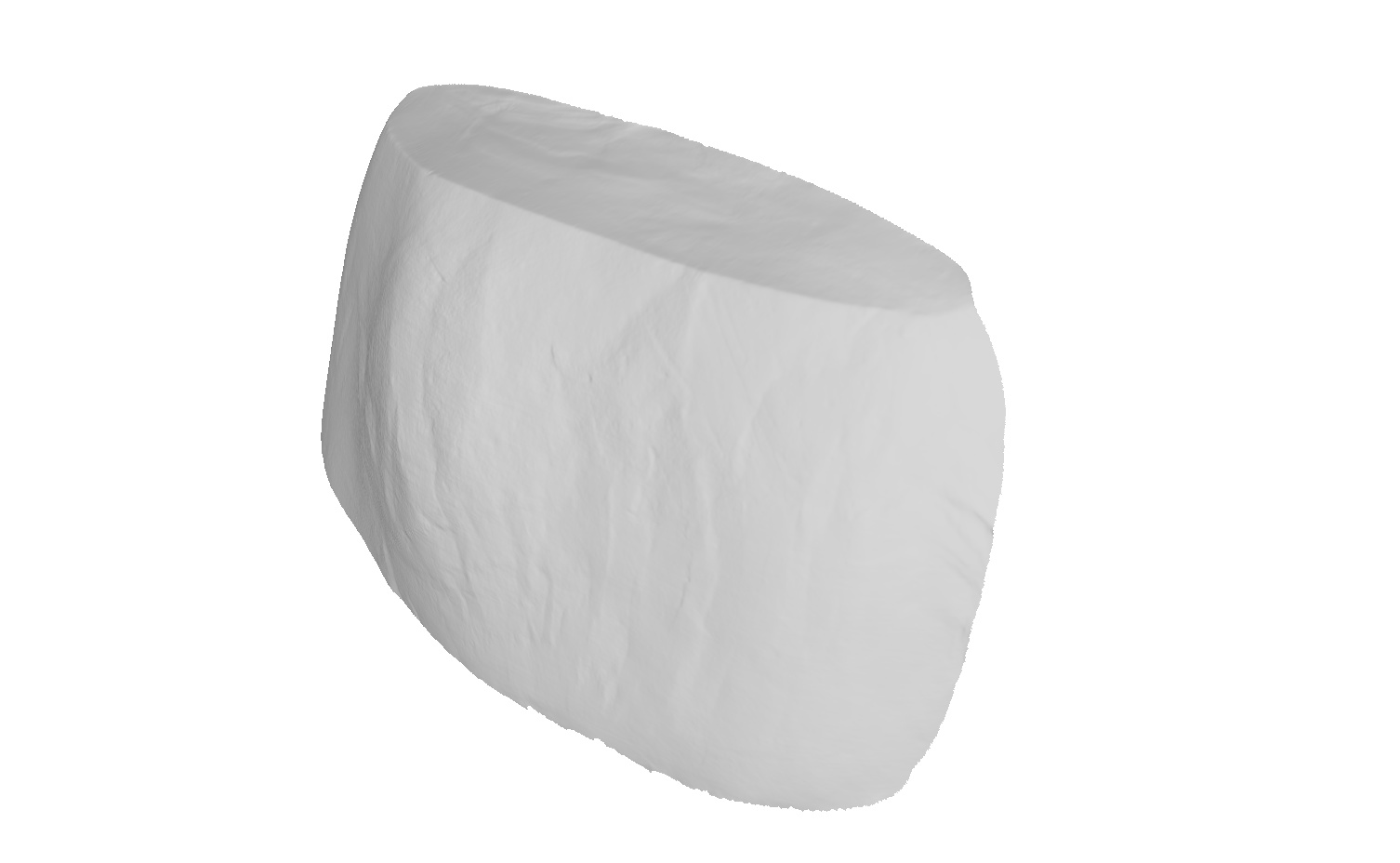} &
\includegraphics[width=0.15\textwidth,trim={10cm 0cm 10cm  0cm},clip]{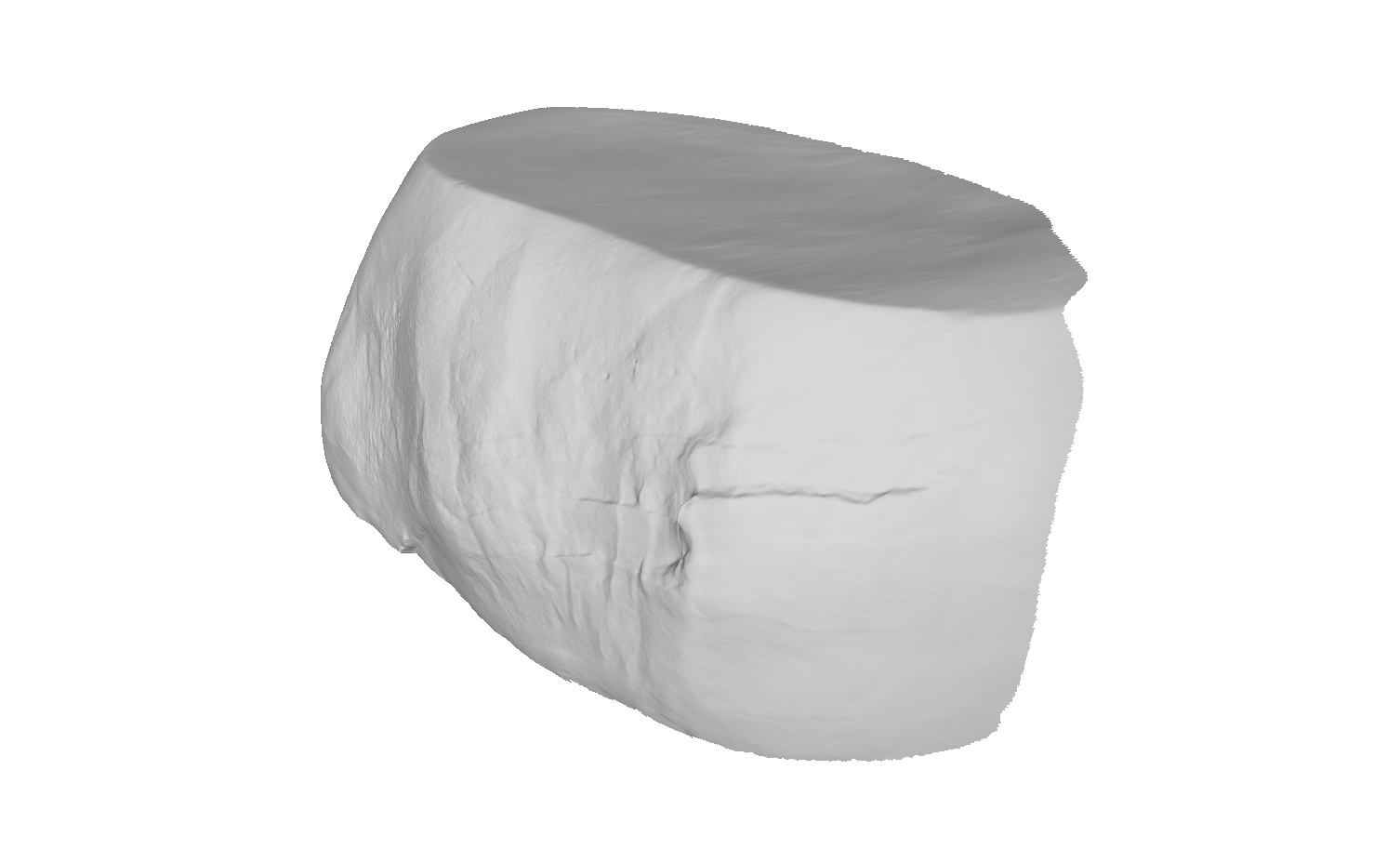} &
\includegraphics[width=0.15\textwidth,trim={10cm 0cm 10cm  0cm},clip]{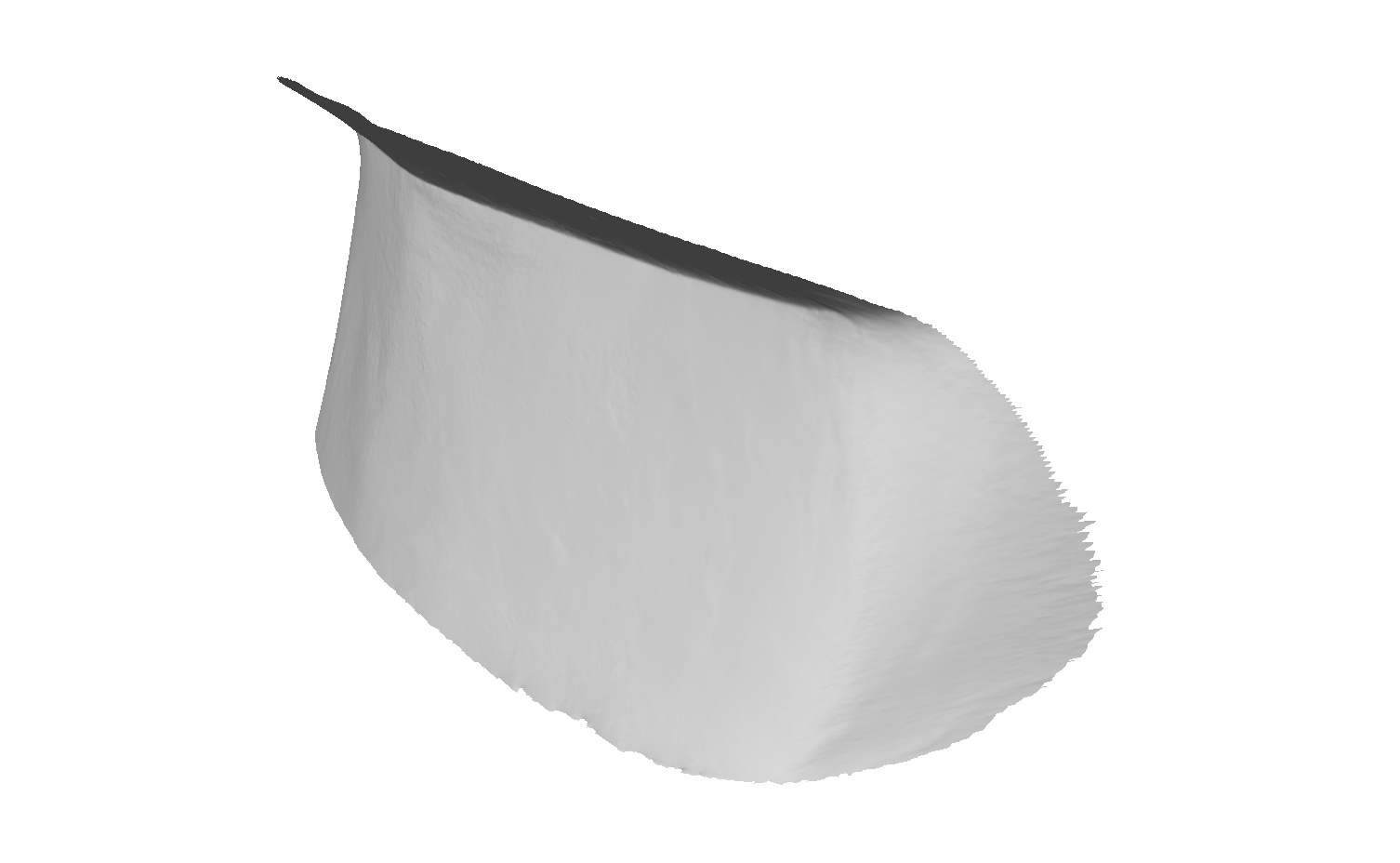} &
\includegraphics[width=0.15\textwidth,trim={10cm 0cm 10cm  0cm},clip]{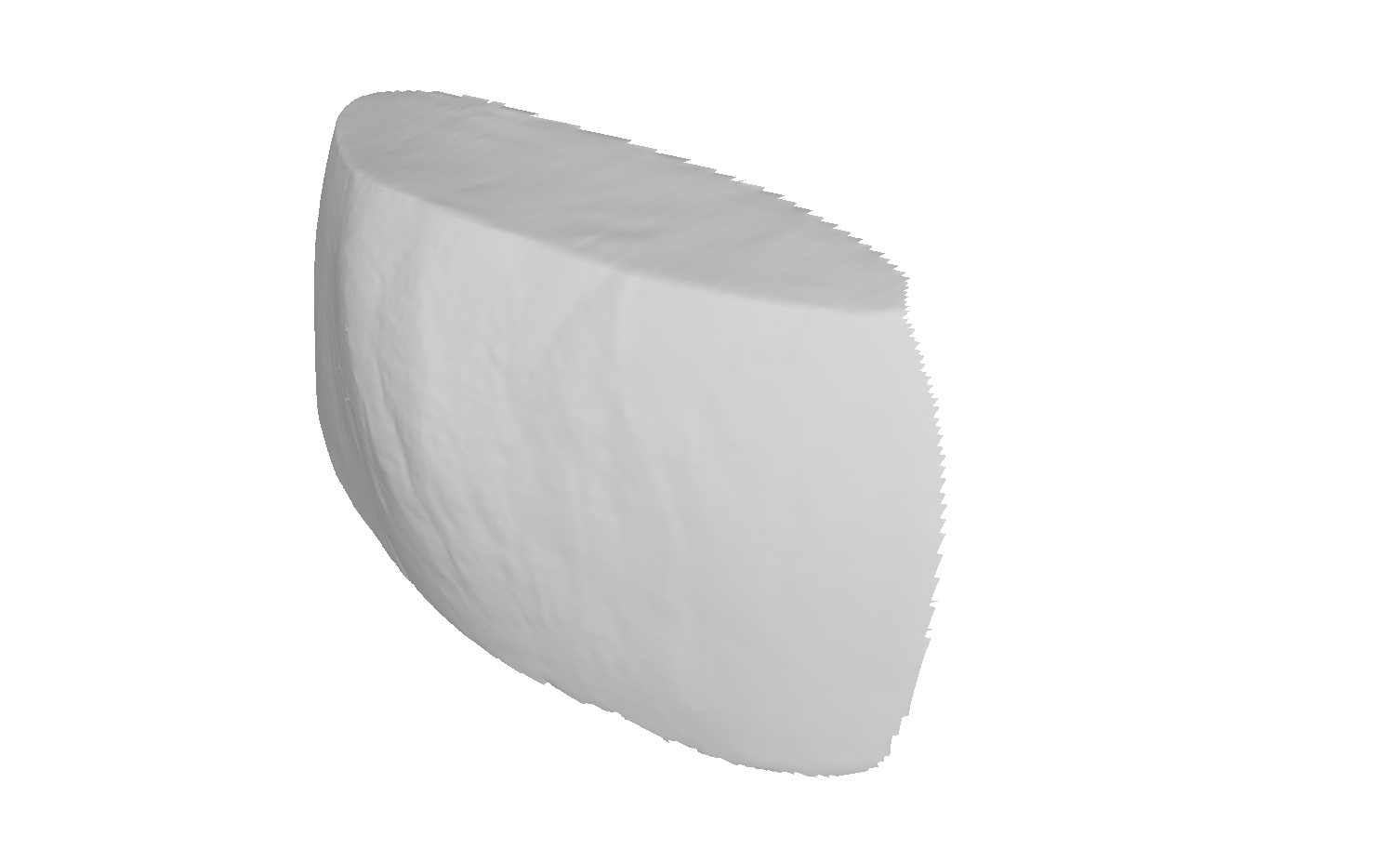} &
\includegraphics[width=0.15\textwidth,trim={10cm 0cm 10cm  0cm},clip]{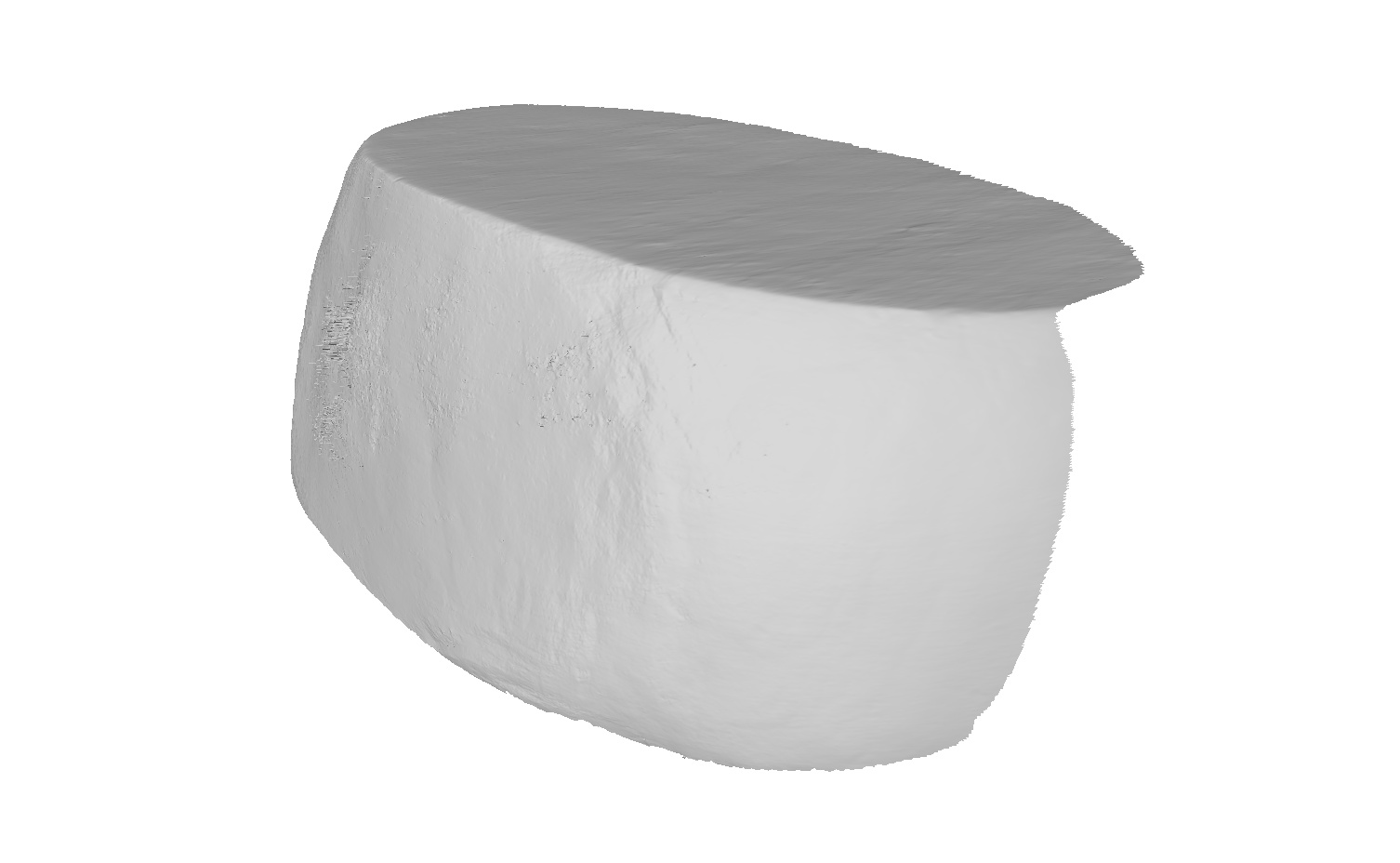} \\
\begin{sideways} {$Z$ Error (mm)} \end{sideways} &
\includegraphics[width=0.15\textwidth,trim={2cm 0cm 2cm  0cm},clip]{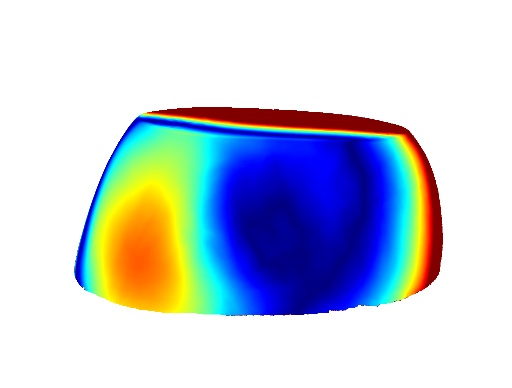} &
\includegraphics[width=0.15\textwidth,trim={2cm 0cm 2cm  0cm},clip]{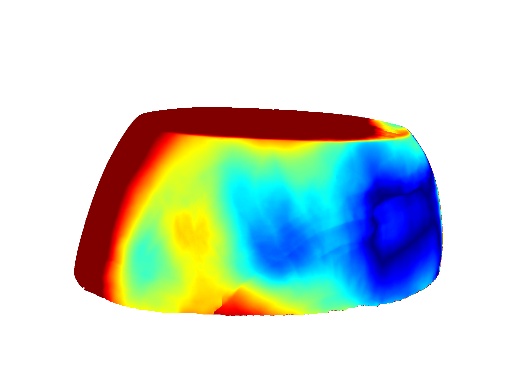} &
\includegraphics[width=0.15\textwidth,trim={2cm 0cm 2cm  0cm},clip]{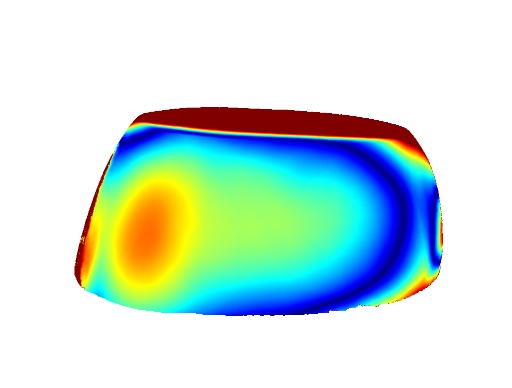} &
\includegraphics[width=0.15\textwidth,trim={2cm 0cm 2cm  0cm},clip]{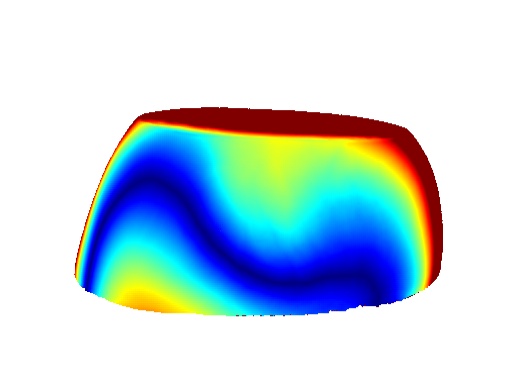} &
\includegraphics[width=0.15\textwidth,trim={2cm 0cm 2cm  0cm},clip]{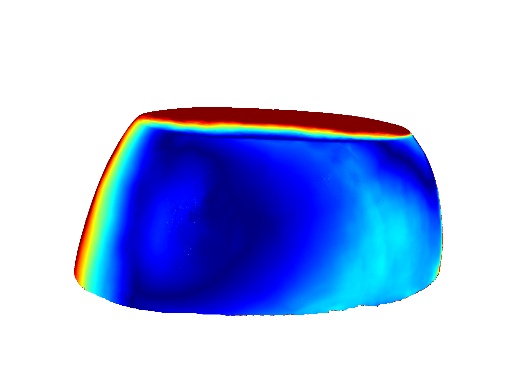} \\
%
\begin{sideways} {Tool-3D Shape} \end{sideways} &
\includegraphics[width=0.15\textwidth,trim={10cm 0cm 11cm  0cm},clip]{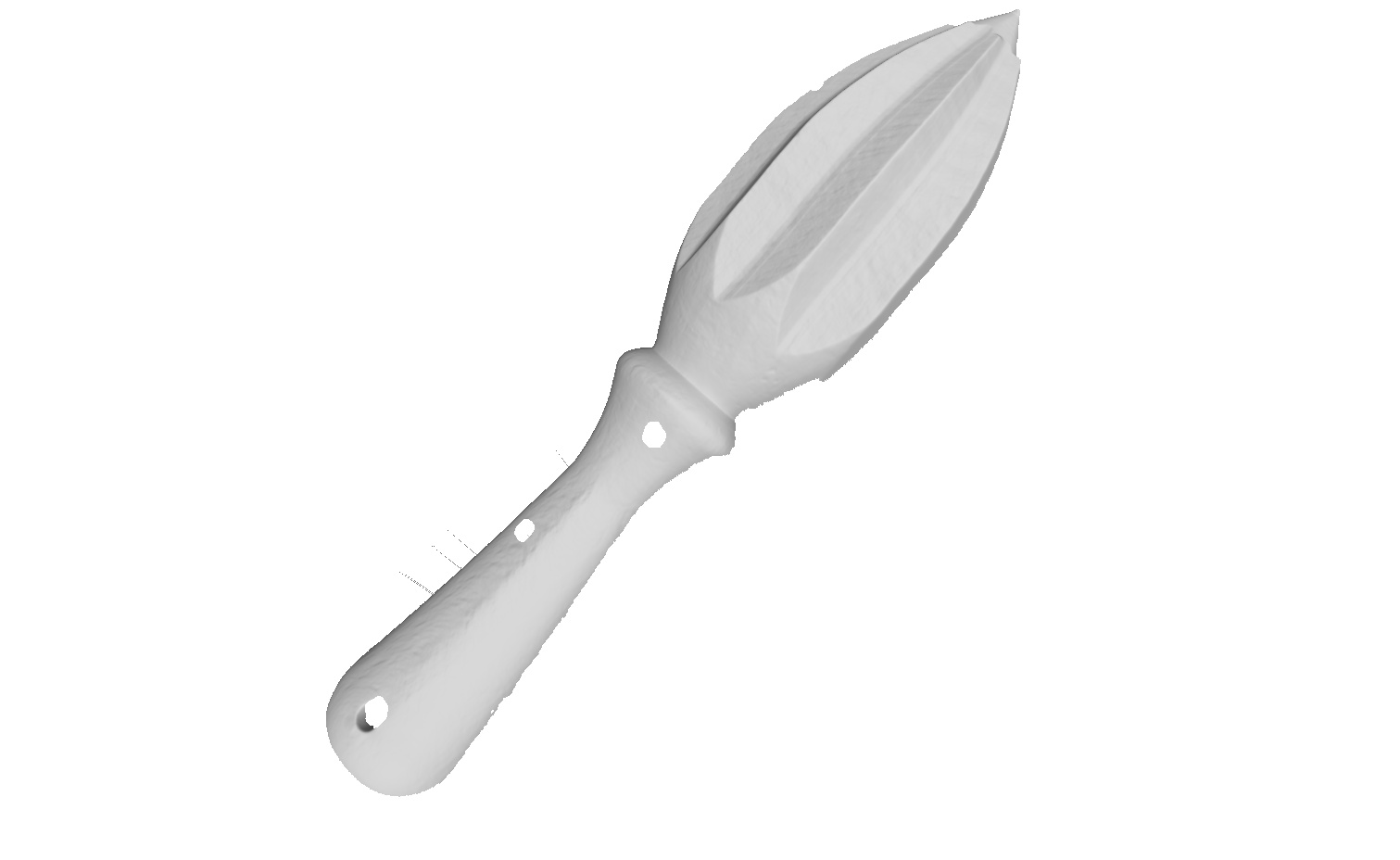} &
\includegraphics[width=0.15\textwidth,trim={10cm 0cm 11cm  0cm},clip]{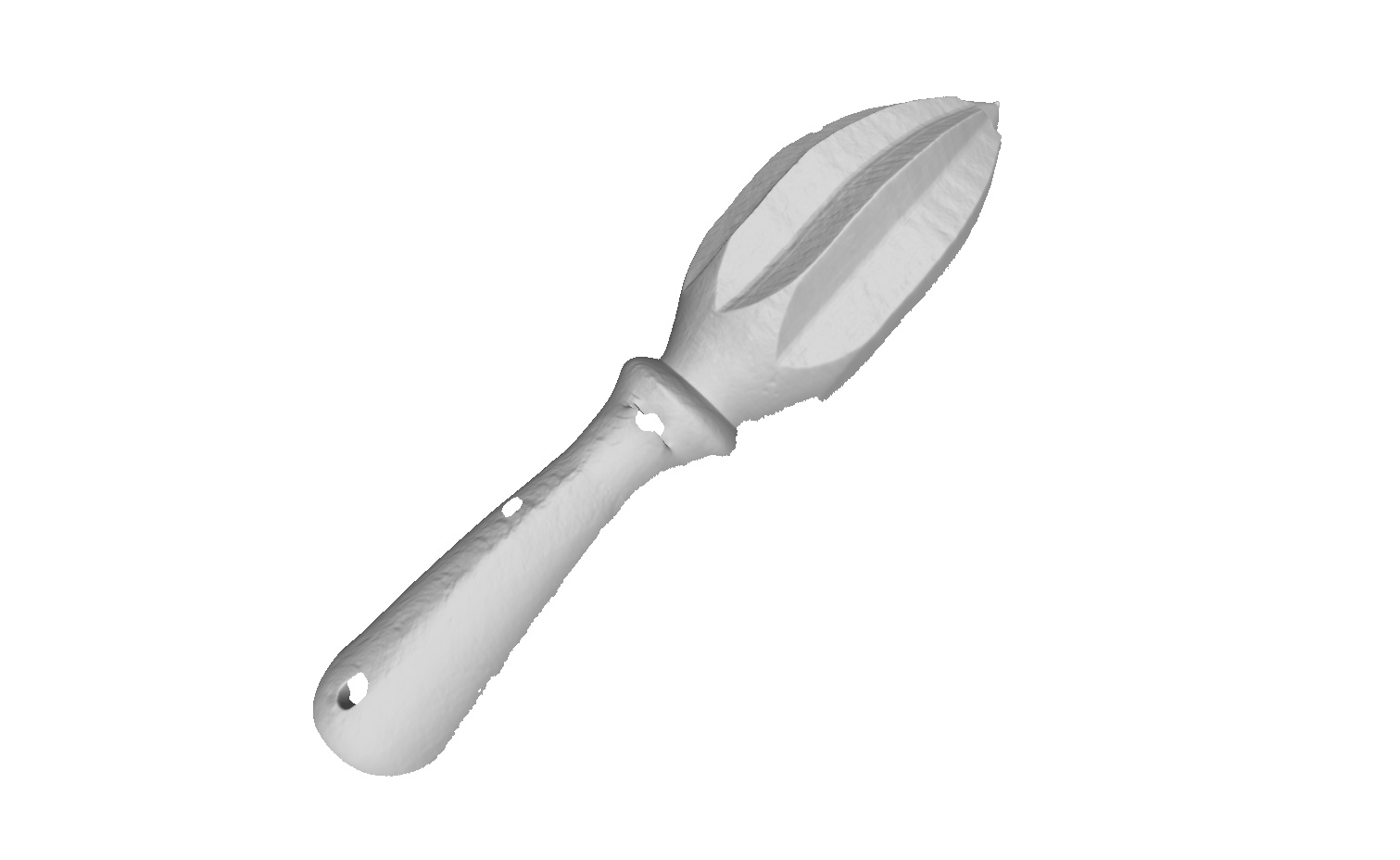} &
\includegraphics[width=0.15\textwidth,trim={10cm 0cm 11cm  0cm},clip]{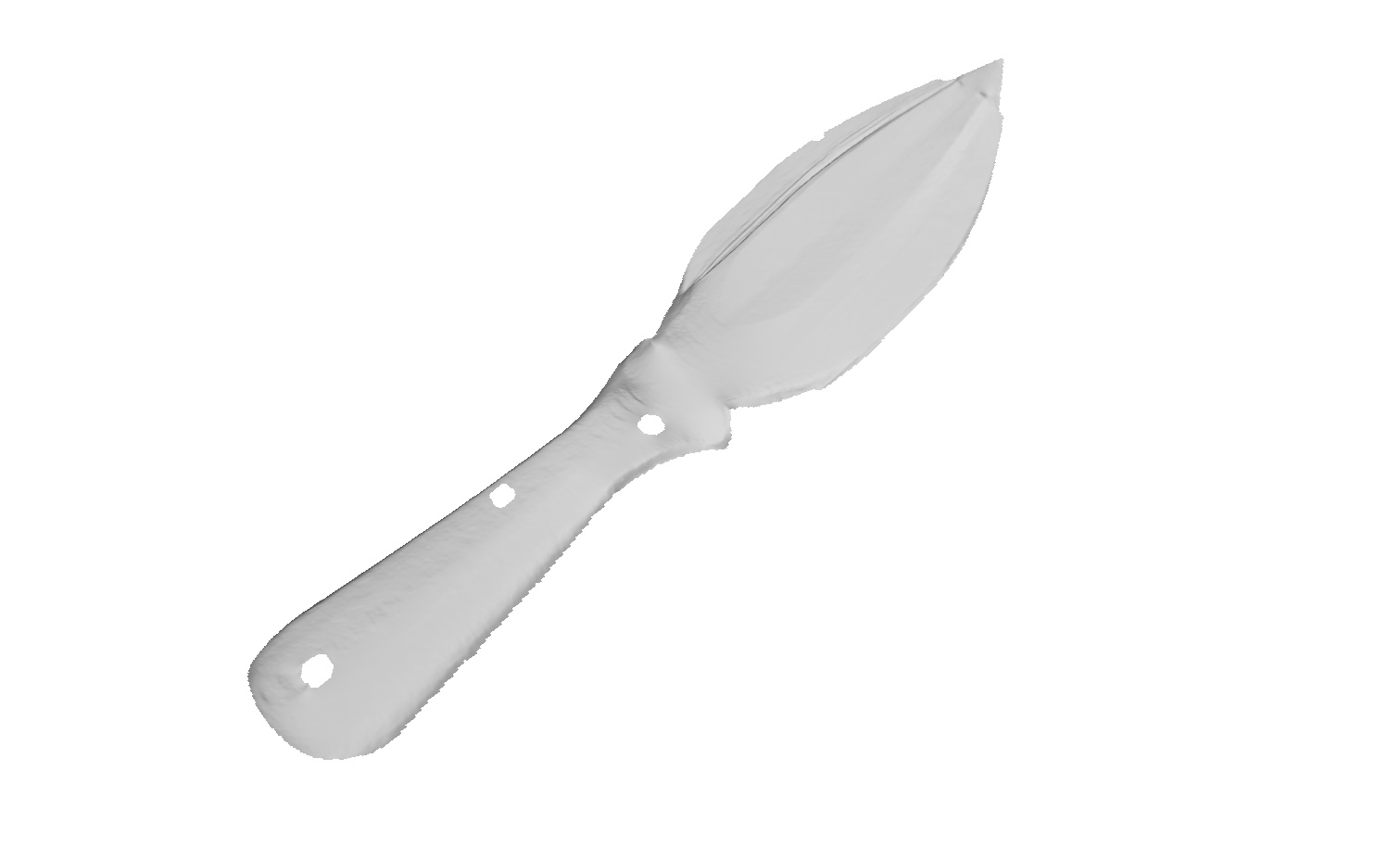} &
\includegraphics[width=0.15\textwidth,trim={10cm 0cm 11cm  0cm},clip]{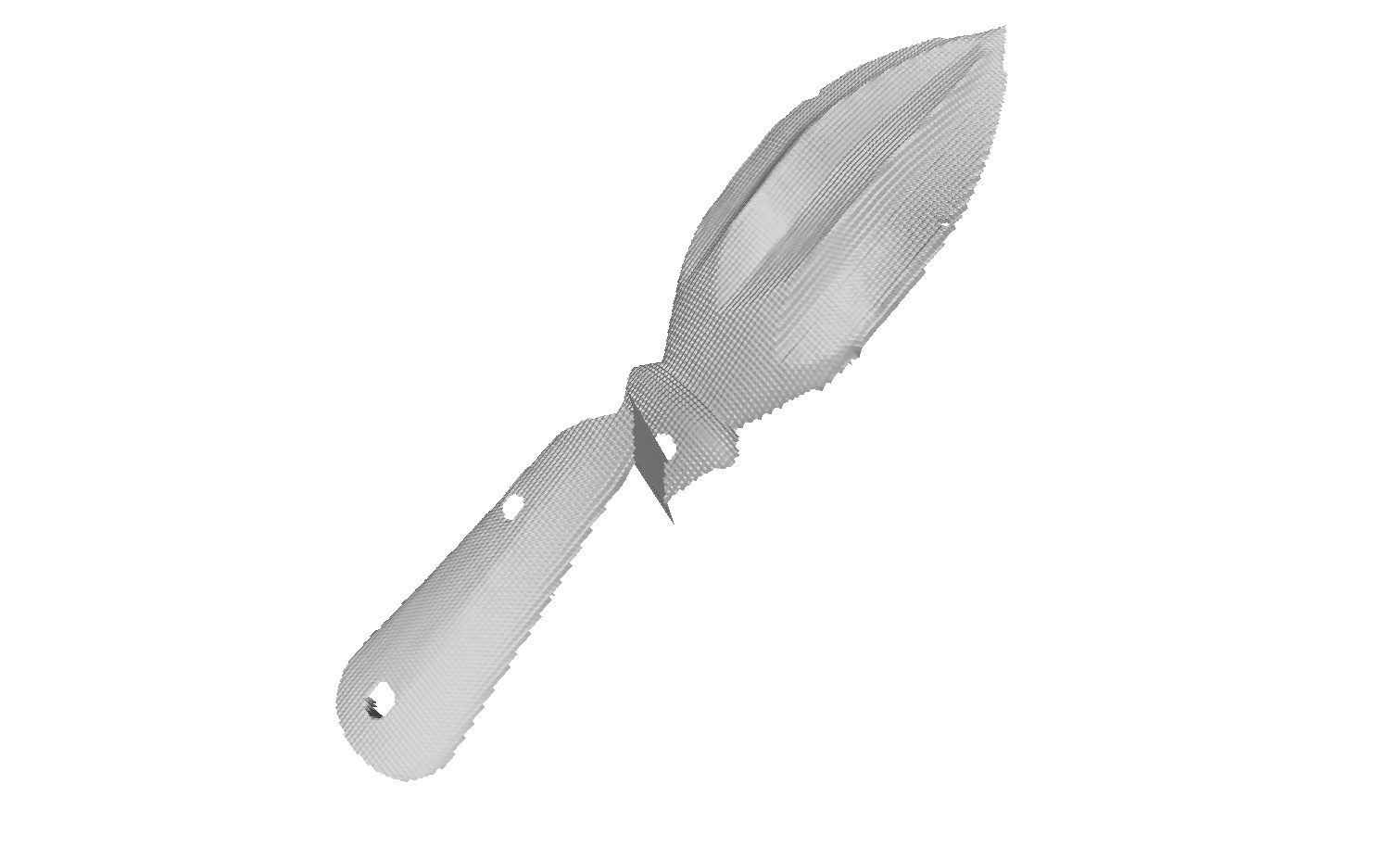} &
\includegraphics[width=0.15\textwidth,trim={10cm 0cm 11cm  0cm},clip]{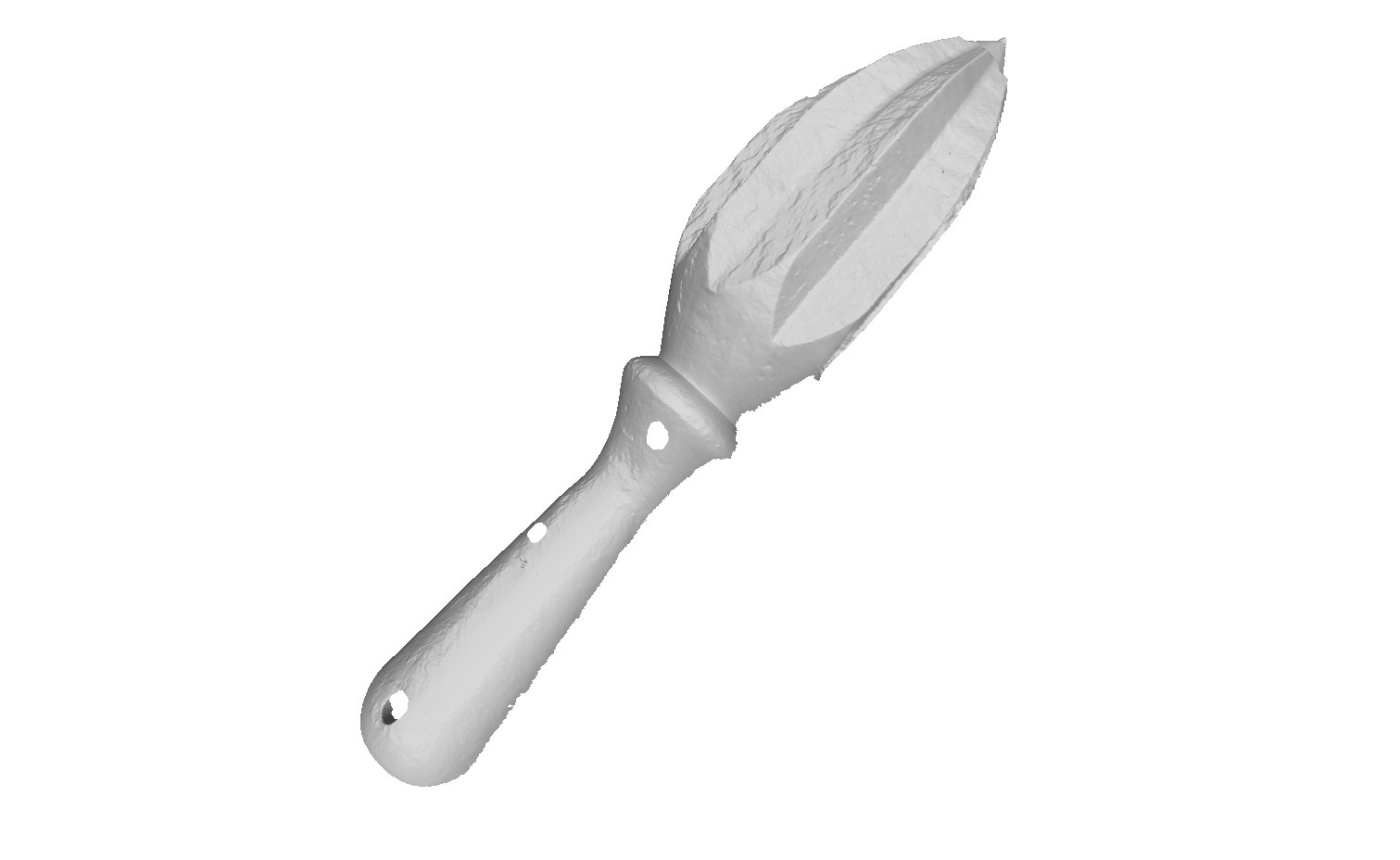} \\
\begin{sideways} {$Z$ Error (mm)} \end{sideways} &
\includegraphics[width=0.15\textwidth,trim={2cm 0cm 2cm  0cm},clip]{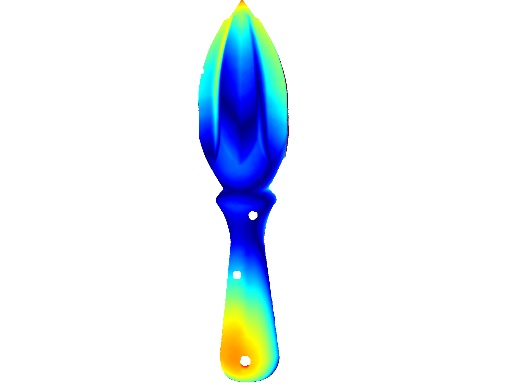} &
\includegraphics[width=0.15\textwidth,trim={2cm 0cm 2cm  0cm},clip]{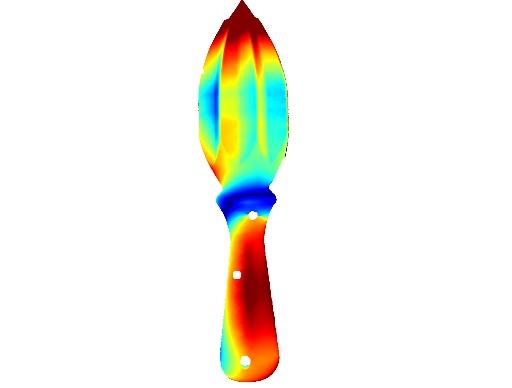} &
\includegraphics[width=0.15\textwidth,trim={2cm 0cm 2cm  0cm},clip]{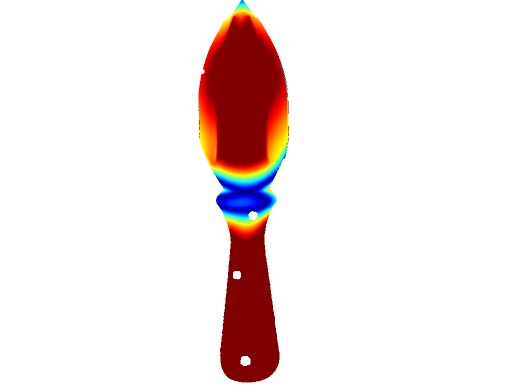} &
\includegraphics[width=0.15\textwidth,trim={2cm 0cm 2cm  0cm},clip]{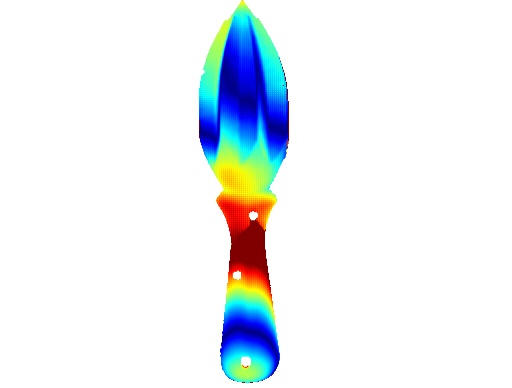} &
\includegraphics[width=0.15\textwidth,trim={2cm 0cm 2cm  0cm},clip]{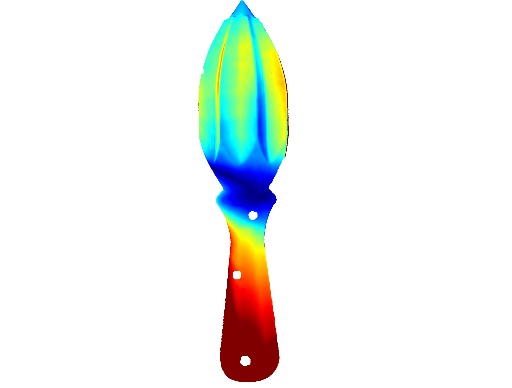} \\
%
\end{tabular}
\caption{Evaluations 13-14/14 }
\label{fig:eval7}
\end{figure*}

This section contains a complete qualitative comparison of all 5 methods L17~\cite{logothetis2017semi},  Q18~\cite{queau2018led}  I18~~\cite{ikehata2018cnn},  S20~\cite{santo2020deep} and  L20~\cite{logothetis2020cnn}. Estimated 3D surface view as well as depth $Z$  
error maps are provided in Figures~\ref{fig:eval1} to~\ref{fig:eval7}. Note that errors of predicted normals are provided in the main publication.

